%% file: main.tex
\documentclass[10pt,twocolumn,letterpaper]{article}
\usepackage[dvipsnames,table]{xcolor}
\usepackage[framemethod=tikz]{mdframed}

\usepackage{cvpr}              %
\usepackage{amsmath,amsfonts,bm,mathtools}
\usepackage{amsthm,amssymb}

\PassOptionsToPackage{sort&compress,numbers}{natbib}
\usepackage{natbib}
\usepackage{siunitx}
\usepackage{lipsum}
\usepackage{symbols}
\usepackage{tabularx}
\usepackage{algorithm}
\usepackage{algorithmic}
\newcommand*{\ShowNotes}{} %
\input{macros.tex}

\input{math_commands}

\usepackage{array,colortbl,booktabs}
\newcolumntype{S}{p{5pt}}

\usepackage[absolute,overlay]{textpos}

\usepackage{tikz}
\usetikzlibrary{matrix}
\usepackage{multirow}
\usepackage{graphicx}

\usepackage{mathtools}
\usepackage[utf8]{inputenc} %
\usepackage[T1]{fontenc}    %
\usepackage{url}            %
\usepackage{booktabs}       %
\usepackage{amsfonts}       %
\usepackage{nicefrac}       %
\usepackage{microtype}      %

\usepackage{thm-restate}
\definecolor{cvprblue}{rgb}{0.21,0.49,0.74}
\usepackage[pagebackref,breaklinks,colorlinks,allcolors=cvprblue]{hyperref}
\usepackage[capitalize,noabbrev]{cleveref}

\def\paperID{ } %
\def\confName{CVPR}
\def\confYear{2026}

\title{Tunable Soft Equivariance with Guarantees}

\author{
Md~Ashiqur~Rahman$^{1}$ \quad
Lim~Jun~Hao$^{2}$ \quad
Jeremiah~Jiang$^{2}$ \quad
Teck{-}Yian~Lim$^{2}$\quad
Raymond~A.~Yeh$^{1}$\\
$^{1}$Purdue University \quad
$^{2}$DSO National Laboratories\\
}

\begin{document}
\twocolumn[{%
\renewcommand\twocolumn[1][]{#1}%
\maketitle
\vspace{-0.65cm}
\input{figs/teaser}
\vspace{.45cm}
}]

\begin{abstract}
    \input{src/abs}

\end{abstract}

\input{src/intro}

\input{src/rel}

\input{src/prelim}

\input{src/method}

\input{src/exp}

\input{src/conc}

\clearpage

\clearpage
{
    \small
    \bibliographystyle{ieeenat_fullname}
    \bibliography{ref}
}

\input{supp/appendix}

\end{document}

%% file: macros.tex
\definecolor{darkred}{rgb}{0.7,0.1,0.1}
\definecolor{darkgreen}{rgb}{0.1,0.7,0.1}
\definecolor{cyan}{rgb}{0.7,0.0,0.7}
\definecolor{dblue}{rgb}{0.2,0.2,0.8}
\definecolor{maroon}{rgb}{0.76,.13,.28}
\definecolor{burntorange}{rgb}{0.81,.33,0}
\definecolor{tealblue}{rgb}{0.212,0.459, 0.533}

\definecolor{deepolive}{rgb}{0.333,0.420,0.282}      %
\definecolor{mutedolive}{rgb}{0.502,0.502,0.000}    %
\definecolor{deepsage}{rgb}{0.270,0.360,0.240}      %

\definecolor{palesage}{rgb}{0.957,0.969,0.929}      %
\definecolor{warmparchment}{rgb}{0.973,0.965,0.941} %
\definecolor{creamstone}{rgb}{0.980,0.960,0.910}    %
\definecolor{lighttaupe}{rgb}{0.950,0.940,0.900}    %

\definecolor{mypink}{rgb}{0.93359375, 0.62109375, 0.83984375}

\definecolor{pp}{rgb}{0.43921569, 0.18823529, 0.62745098}
\definecolor{rr}{rgb}{0.5254902 , 0.00784314, 0.12941176}
\definecolor{bb}{rgb}{0.09019608, 0.23529412, 0.37647059}
\definecolor{yy}{rgb}{0.49803922, 0.3372549 , 0.0}
\definecolor{gg}{rgb}{0.02352941, 0.3372549 , 0.17647059}
\definecolor{myblue}{rgb}{0.22352941, 0.64705882, 0.87058824}

\newcommand{\myparagraph}[1]{\vspace*{2pt}{\bf\noindent #1}}
\ifdefined\ShowNotes
  \newcommand{\colornote}[3]{{\color{#1}\bf{#2: #3}\normalfont}}
\else
  \newcommand{\colornote}[3]{}
\fi

\newcommand{\eat}[1]{} %

\mdfdefinestyle{MyFrame}{%
    linecolor=Black,
    outerlinewidth=0.1pt,
    roundcorner=2pt,
    innertopmargin=0.3\baselineskip,
    innerbottommargin=0.3\baselineskip,
    innermargin = 0.2cm,
    outermargin = 0.cm,
    innerrightmargin=5pt,
    innerleftmargin=5pt,
    backgroundcolor=mybb!3!white}

\usepackage[many]{tcolorbox}
\newtheorem{definition}{Definition}
\tcolorboxenvironment{definition}{
  boxrule=0pt,
  boxsep=0pt,
  colback={palesage!50!White},
  enhanced jigsaw, 
  borderline west={1pt}{0pt}{mutedolive},%
  sharp corners,
  before skip=10pt,
  after skip=10pt,
  left=5pt,
  right=5pt,
}

\tcolorboxenvironment{myexp}{
  boxrule=0pt,
  boxsep=0pt,
  colback={White!90!White},
  enhanced jigsaw, 
  borderline west={1pt}{0pt}{BurntOrange},
  sharp corners,
  before skip=3pt,
  after skip=6pt,
  left=5pt,
  right=5pt,
  bottom=1pt,
  top=3pt,
  width=\linewidth
}

\tcolorboxenvironment{myclaim}{
  boxrule=0pt,
  boxsep=0pt,
  colback={White!90!White},
  enhanced jigsaw, 
  borderline west={1pt}{0pt}{deepsage},%
  sharp corners,
  before skip=10pt,
  after skip=10pt,
  left=5pt,
  right=5pt,
}

\newtcolorbox{myclaime}[1][]{
  notitle, %
  #1, %
  boxrule=0pt,
  boxsep=0pt,
  colback={palesage!50!White},
  enhanced jigsaw, 
  borderline west={1pt}{0pt}{deepsage},%
  sharp corners,
  before skip=2pt,
  after skip=3pt,
  left=5pt,
  right=5pt,
  bottom=1pt,
  top=3pt,
}

\tcolorboxenvironment{example}{
  boxrule=0pt,
  boxsep=0pt,
  blanker,
  borderline west={1pt}{0pt}{Black},
  sharp corners,
  before skip=10pt,
  after skip=10pt,
  left=5pt,
  right=5pt,
}

\tcolorboxenvironment{mylemma}{
  boxrule=0pt,
  boxsep=0pt,
    colback={palesage!50!White},
  enhanced jigsaw, 
  borderline west={1pt}{0pt}{Gray},
  sharp corners,
  before skip=10pt,
  after skip=10pt,
  left=5pt,
  right=5pt,
}

\newtcolorbox{mylemmae}[1][]{
  notitle, %
  #1, %
  boxrule=0pt,
  boxsep=0pt,
  colback={palesage!50!White},
  enhanced jigsaw, 
  borderline west={1pt}{0pt}{Gray},
  sharp corners,
  before skip=1pt,
  after skip=3pt,
  left=5pt,
  right=5pt,
  bottom=1pt,
  top=2pt,
}

\usepackage[multiple]{footmisc}

%% file: math_commands.tex
\usepackage{amsmath,amsfonts,bm}
\usepackage{amsthm,amssymb}

\def\1{\bm{1}}

\def\sp{space}

\def\rmB{{\mathbf{B}}}

\def\rmI{{\mathbf{I}}}
\def\rmJ{{\mathbf{J}}}

\def\rmU{{\mathbf{U}}}
\def\rmV{{\mathbf{V}}}
\def\rmW{{\mathbf{W}}}

\def\vzero{{\bm{0}}}

\def\vu{{\bm{u}}}
\def\vv{{\bm{v}}}
\def\vw{{\bm{w}}}
\def\vx{{\bm{x}}}
\def\vy{{\bm{y}}}
\def\vz{{\bm{z}}}

\def\evalpha{{\alpha}}
\def\evbeta{{\beta}}

\def\evlambda{{\lambda}}

\def\eva{{a}}
\def\evb{{b}}
\def\evc{{c}}
\def\evd{{d}}

\def\mA{{\bm{A}}}
\def\mB{{\bm{B}}}
\def\mC{{\bm{C}}}

\def\mI{{\bm{I}}}
\def\mJ{{\bm{J}}}

\def\mL{{\bm{L}}}
\def\mM{{\bm{M}}}

\def\mS{{\bm{S}}}
\def\mT{{\bm{T}}}
\def\mU{{\bm{U}}}
\def\mV{{\bm{V}}}
\def\mW{{\bm{W}}}

\def\mSigma{{\bm{\Sigma}}}

\DeclareMathAlphabet{\mathsfit}{\encodingdefault}{\sfdefault}{m}{sl}
\SetMathAlphabet{\mathsfit}{bold}{\encodingdefault}{\sfdefault}{bx}{n}

\def\gD{{\mathcal{D}}}

\def\gX{{\mathcal{X}}}
\def\gY{{\mathcal{Y}}}

\def\sR{{\mathbb{R}}}
\def\sS{{\mathbb{S}}}

\newcommand{\E}{\mathbb{E}}

\newcommand{\R}{\mathbb{R}}

\newcommand{\Sym}{\operatorname{Sym}}
\newcommand{\diag}{\operatorname{diag}}

%% file: figs/teaser.tex
\setlength{\tabcolsep}{1.8pt}%
\renewcommand{\arraystretch}{0.8}
\begin{minipage}{\textwidth}
\newcommand{\onebyfour}[4]{%
\begin{minipage}{0.20\textwidth}\centering
    \includegraphics[width=0.30\linewidth]{#1}\hspace{0.0095\linewidth}%
    \includegraphics[width=0.30\linewidth]{#2}\hspace{0.0095\linewidth}%
    \includegraphics[width=0.30\linewidth]{#3}\hspace{0.0095\linewidth}%
\end{minipage}}

\begin{minipage}{0.073\textwidth}
\hspace{1cm}
\end{minipage}
\begin{minipage}{0.927\textwidth}
    \noindent\resizebox{\textwidth}{!}{%
    \begin{tikzpicture}[x=1\textwidth/10]
        \draw[thick, <->] (0.,0) -- (10.,0);
        \node[anchor=north] at (0.42,+0.6) {\small Equivariant};
        \node[anchor=north] at (5,+0.6) {\small Soft equivariant};
        \node[anchor=north] at (9.45,+0.6) {\small Non-equivariant};
    \end{tikzpicture}
    }%
\vspace{6pt}
\end{minipage}

\resizebox{\textwidth}{!}{%
\begin{tabular}{cccccc}
    Weights & \onebyfour{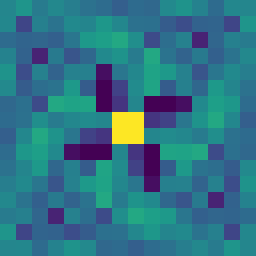}{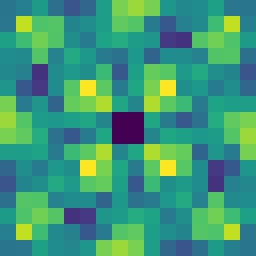}{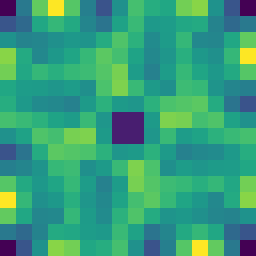}{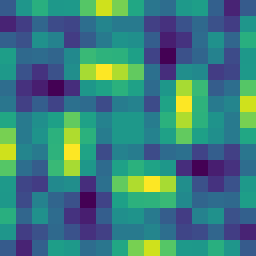} &
    \onebyfour{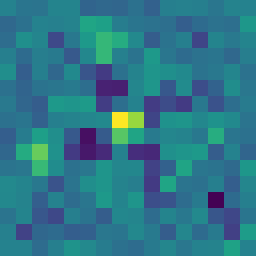}{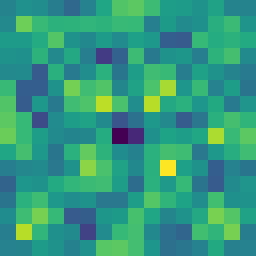}{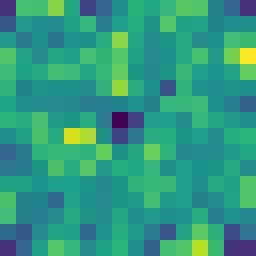}{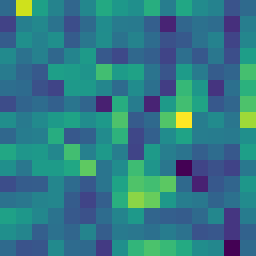} &
    \onebyfour{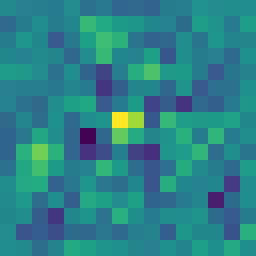}{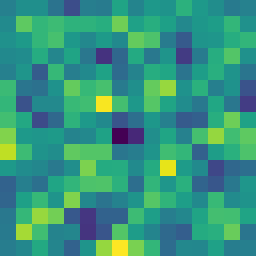}{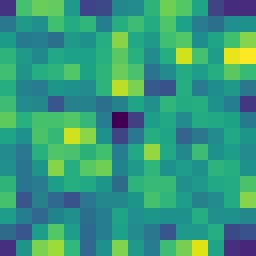}{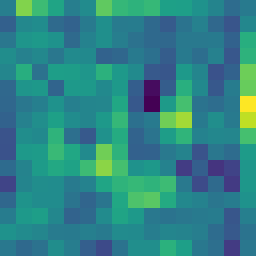} &
    \onebyfour{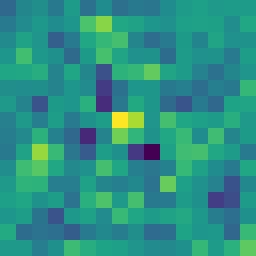}{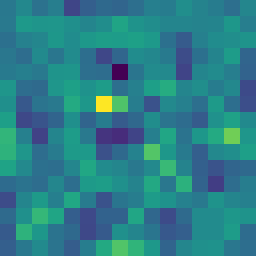}{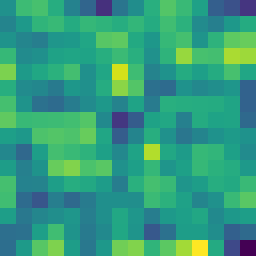}{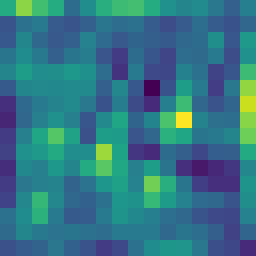} &
    \onebyfour{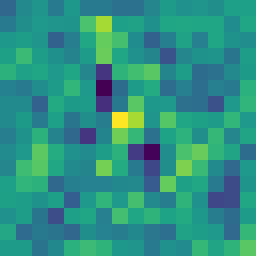}{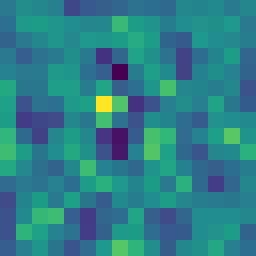}{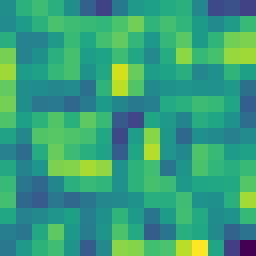}{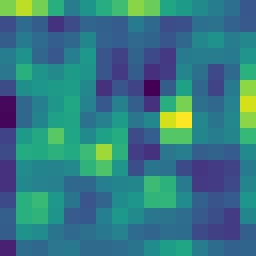} \\[0.42cm]
    \begin{minipage}{0.065\textwidth}\centering\includegraphics[width=0.91\linewidth,angle=90,origin=c]{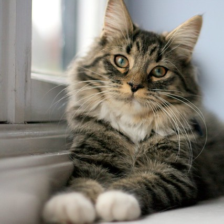}\end{minipage} & \onebyfour{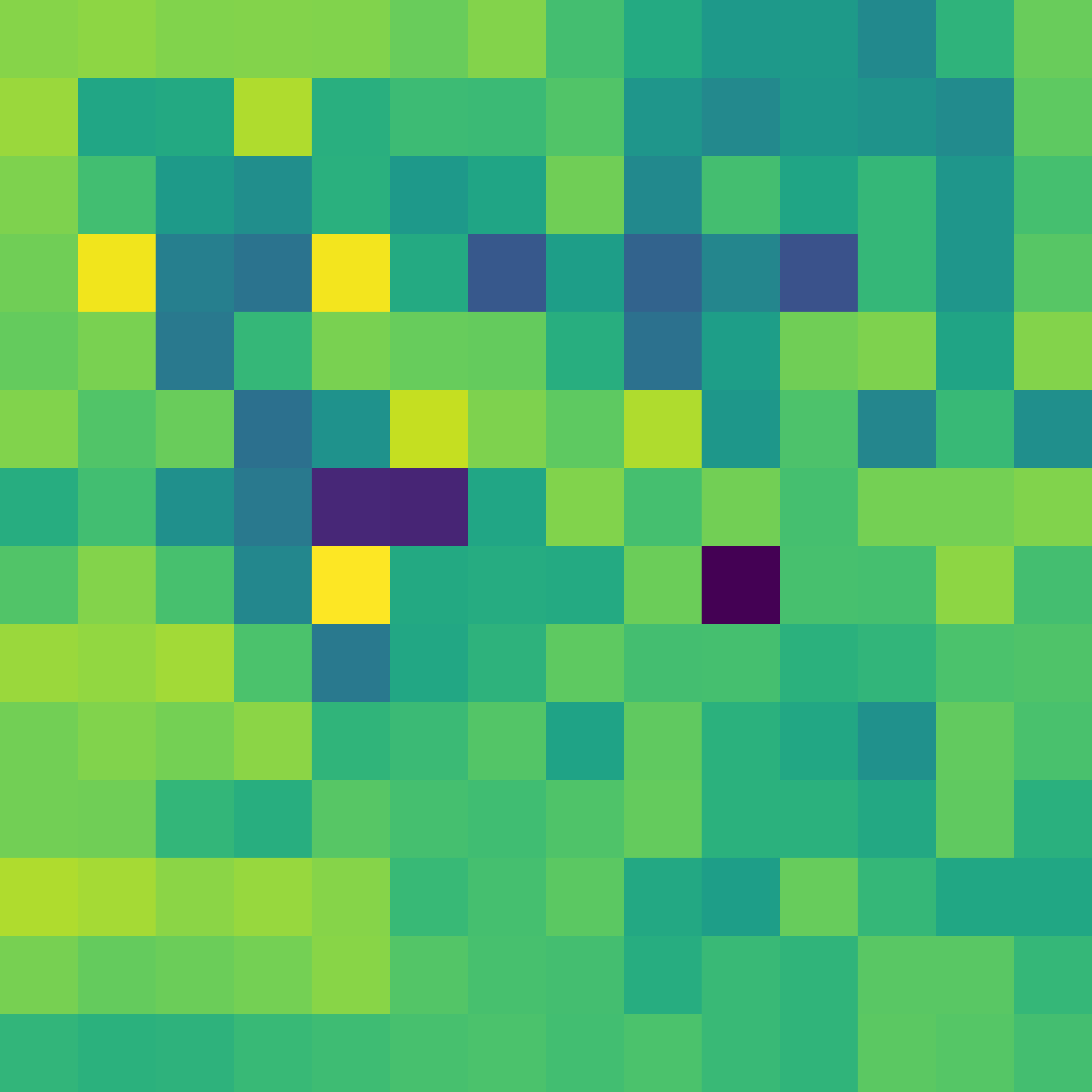}{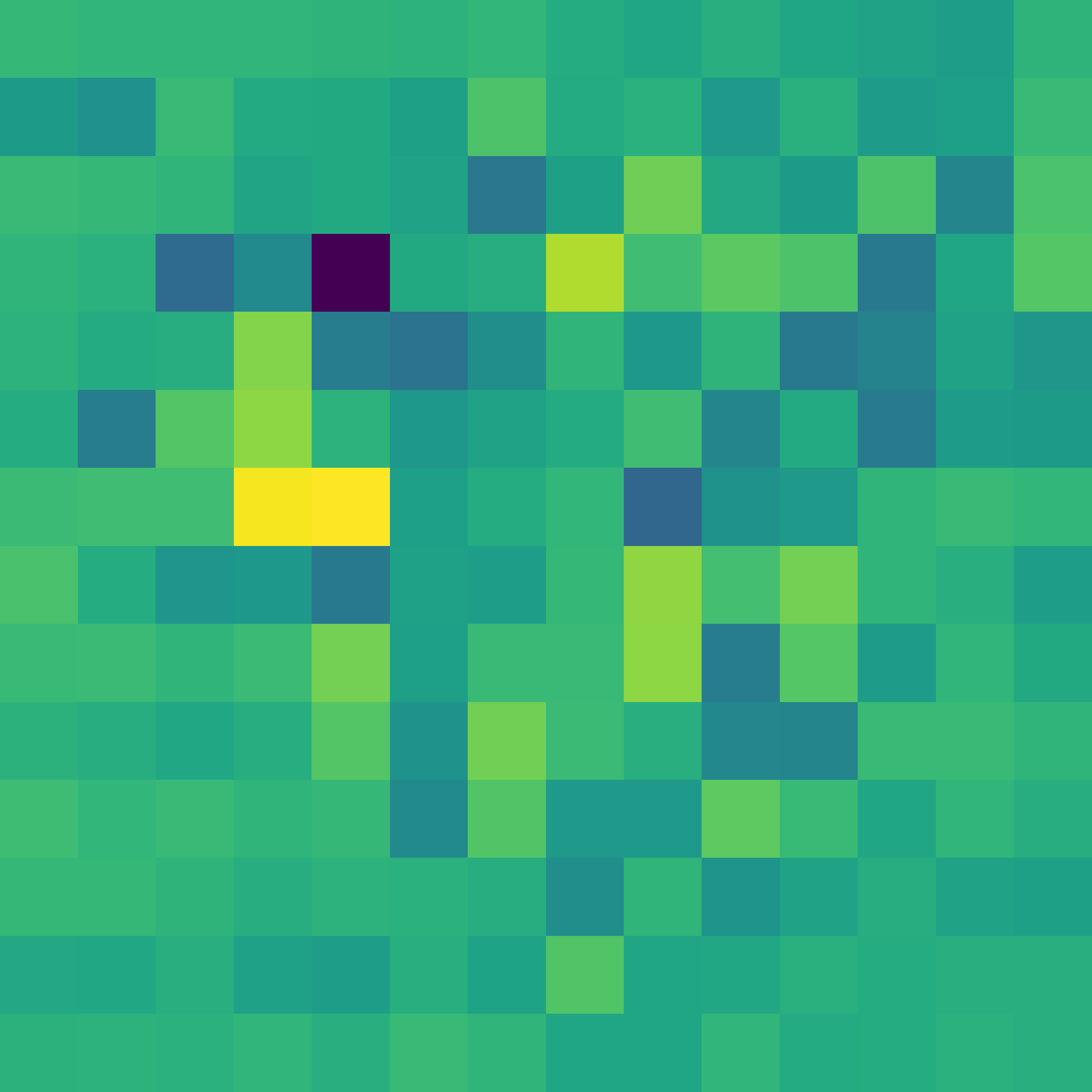}{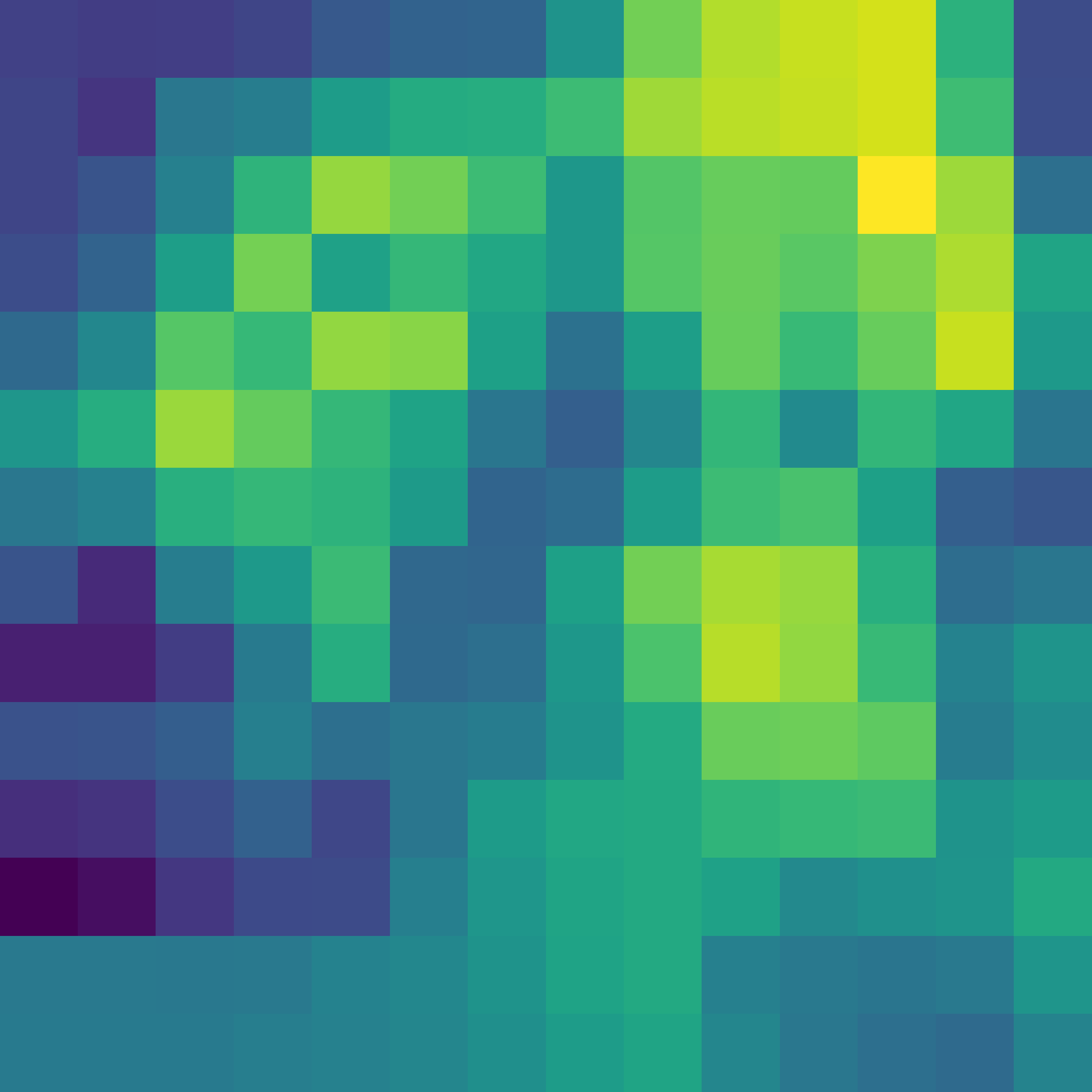}{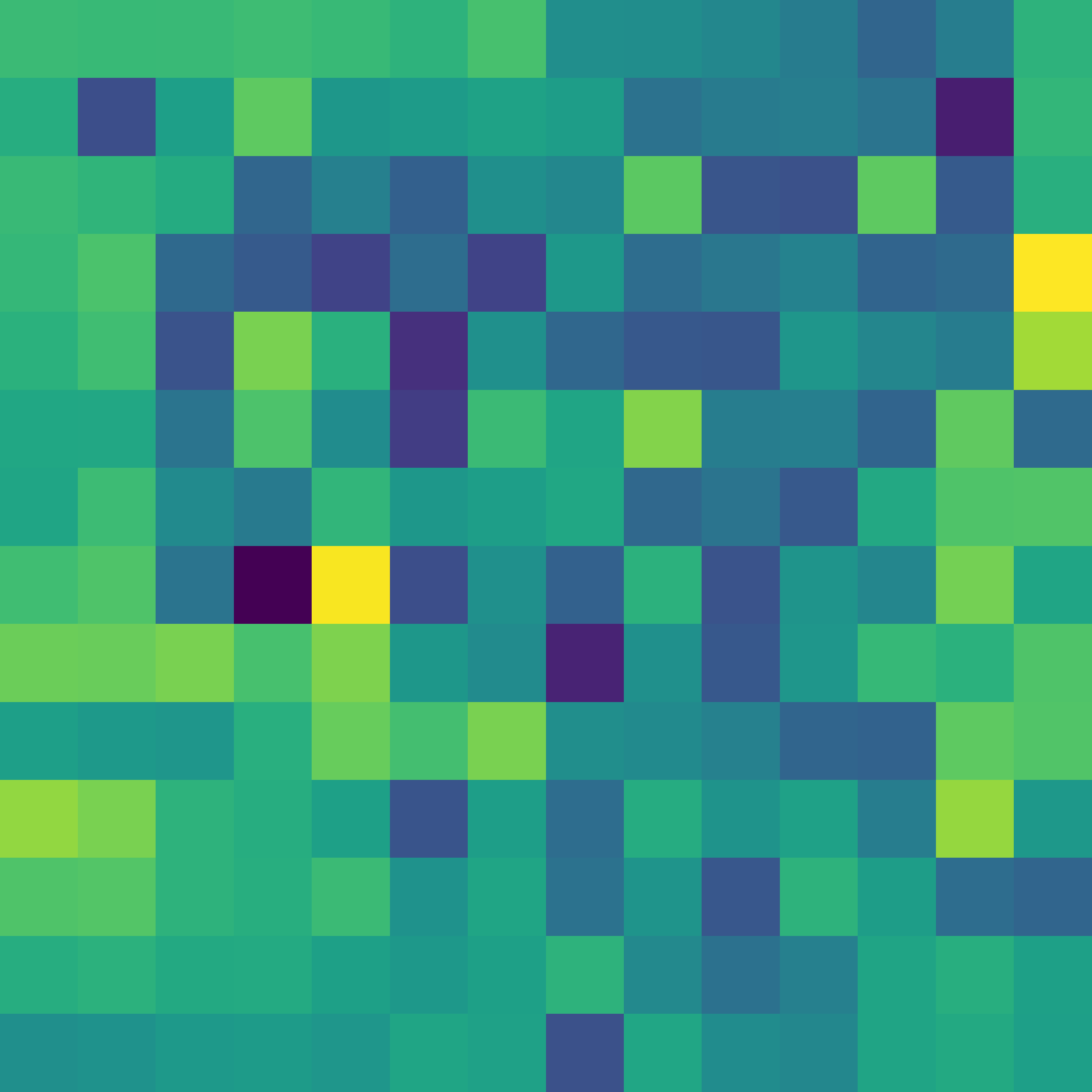} &
    \onebyfour{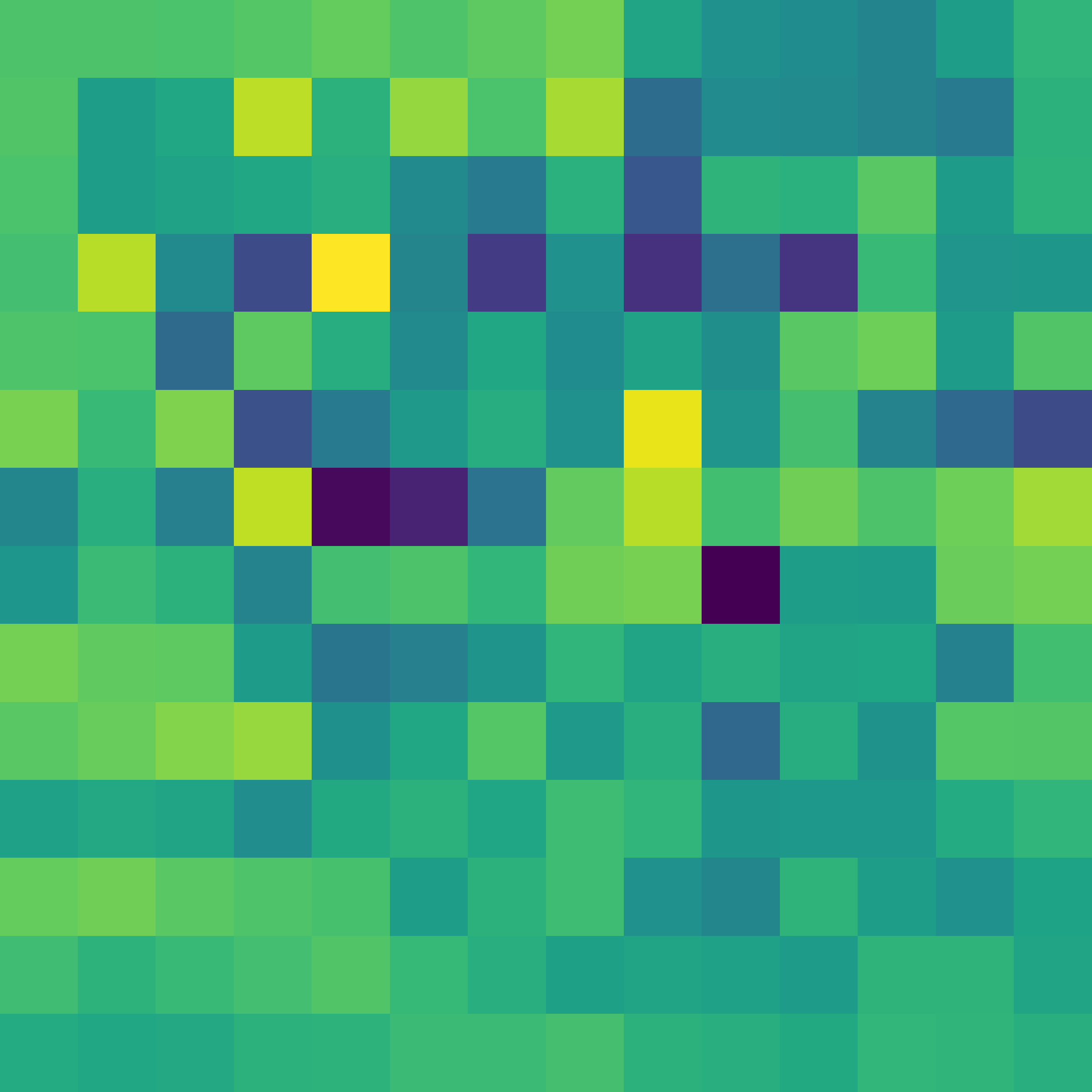}{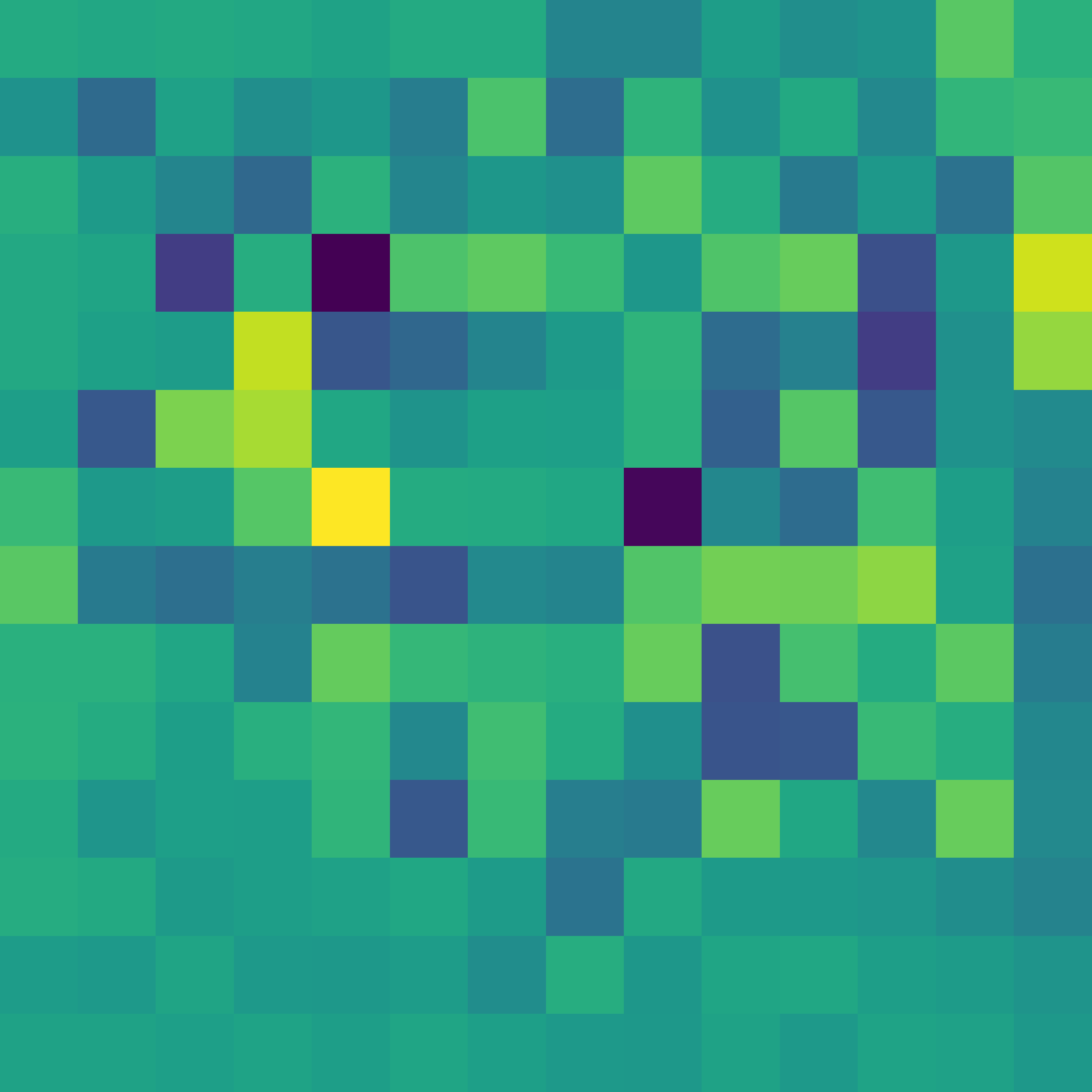}{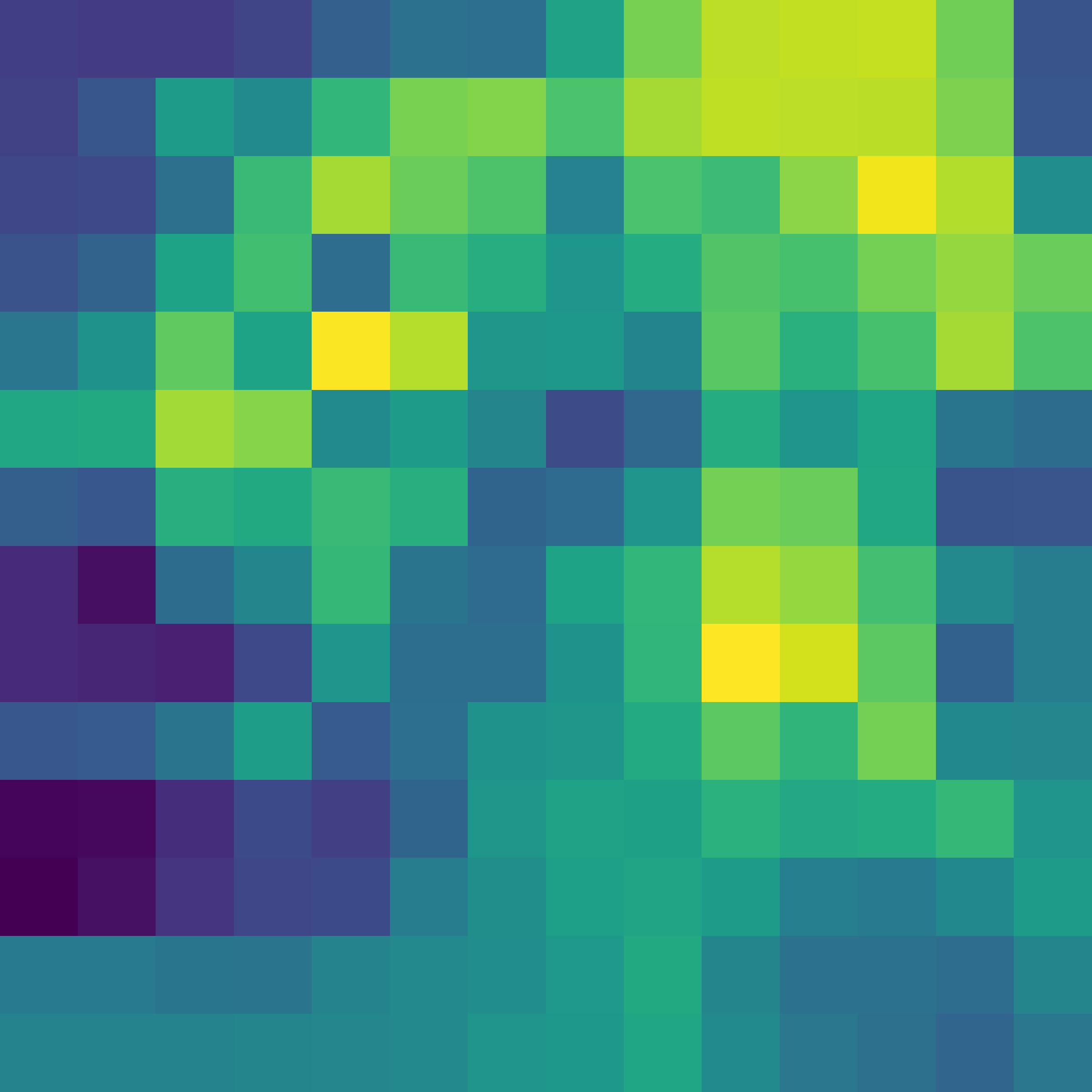}{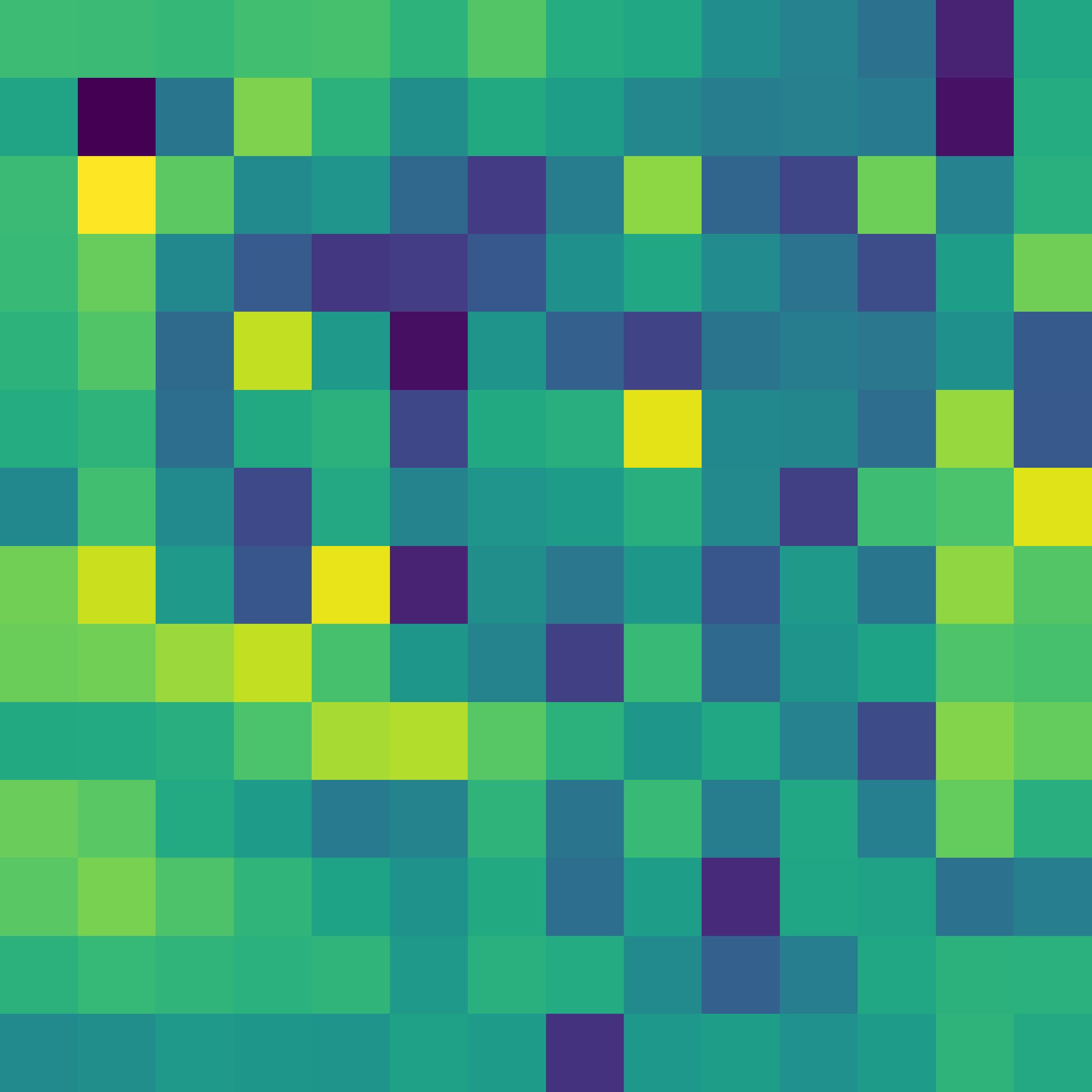} &
    \onebyfour{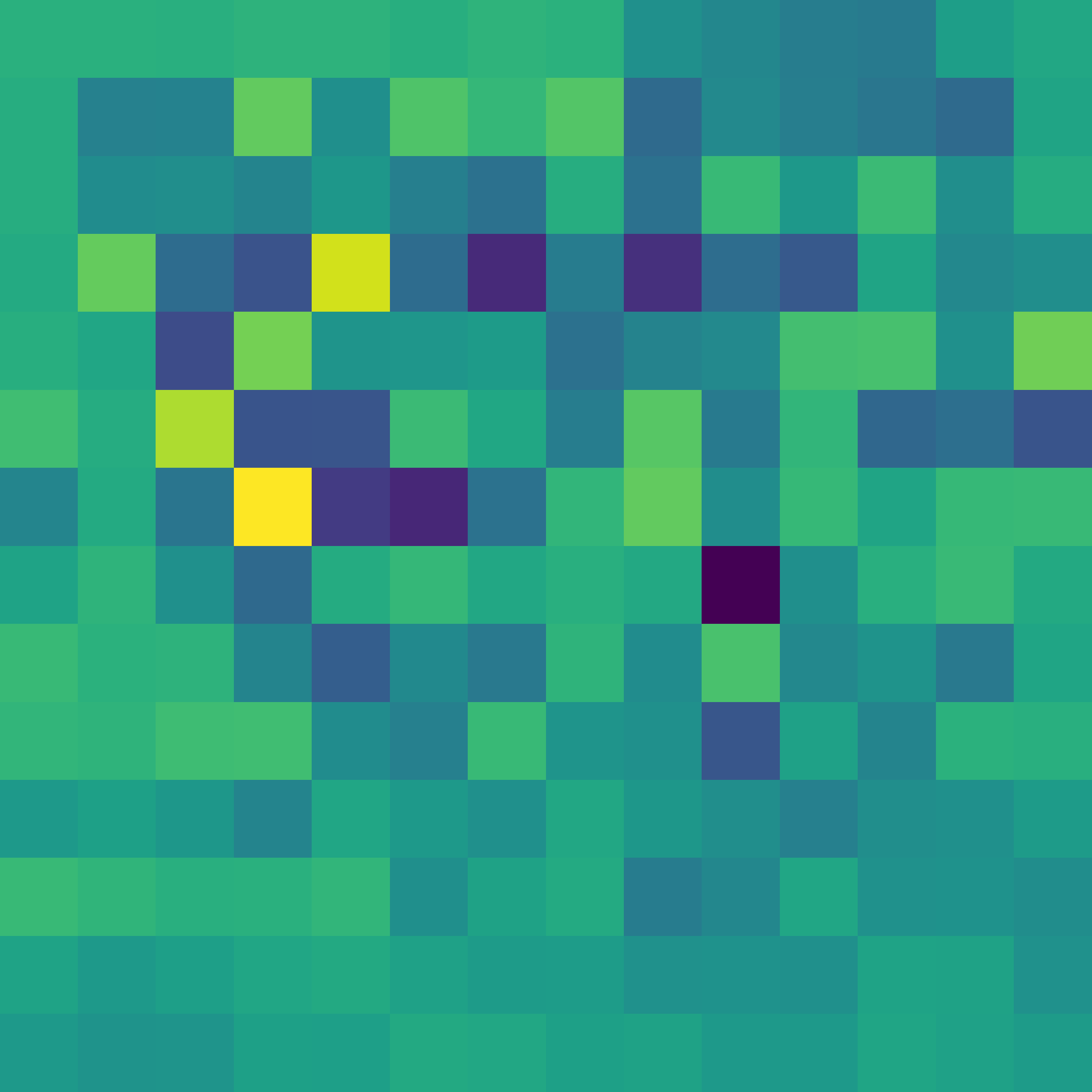}{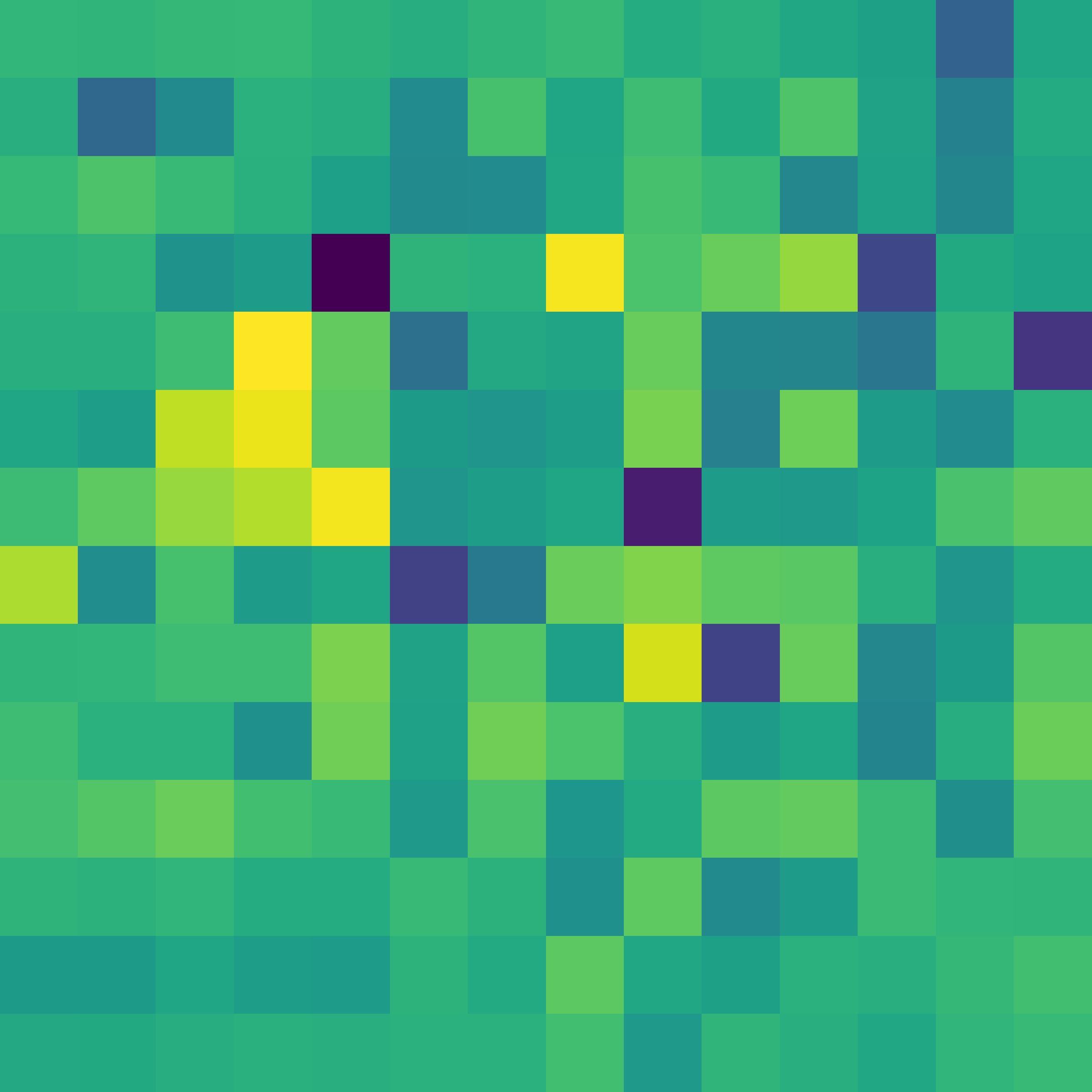}{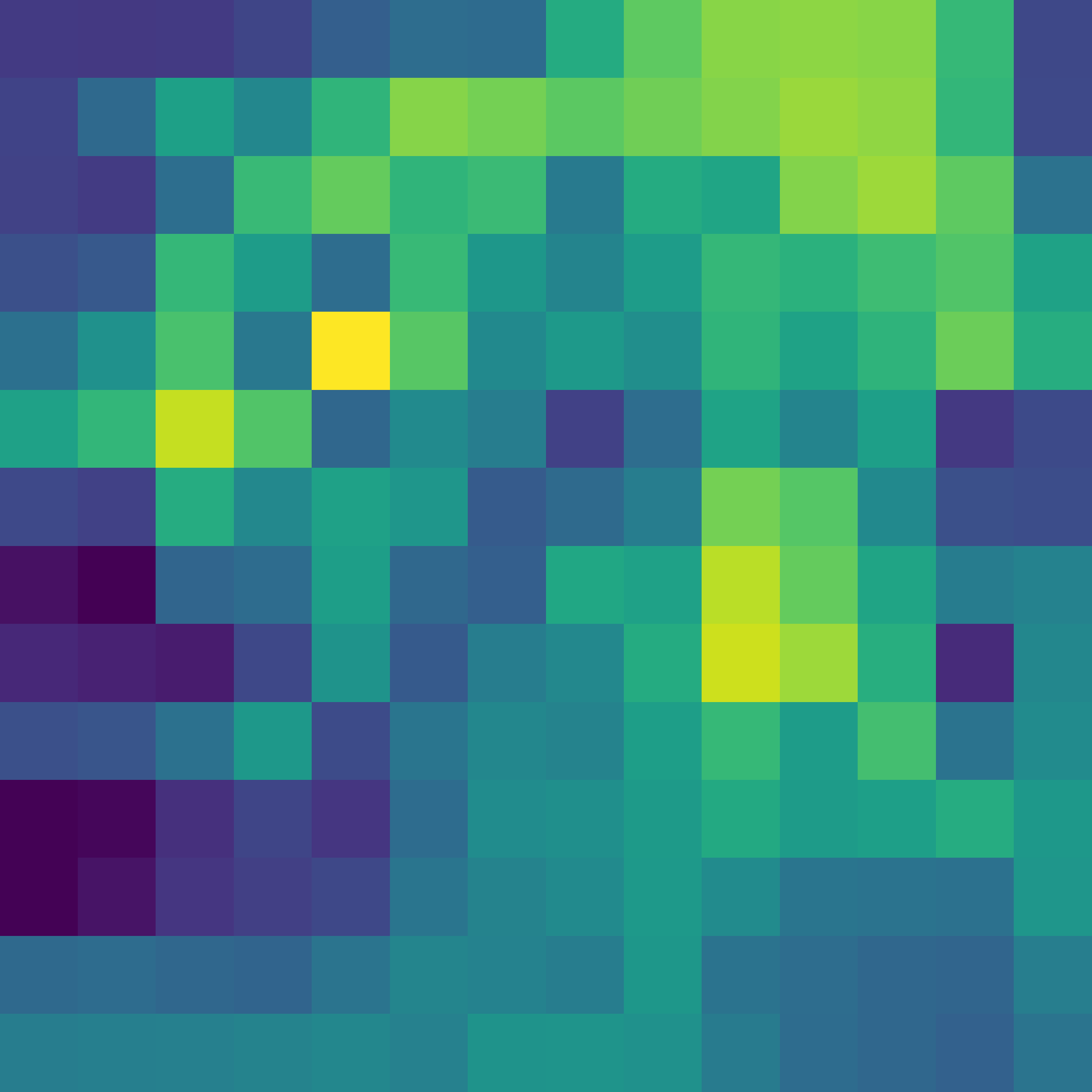}{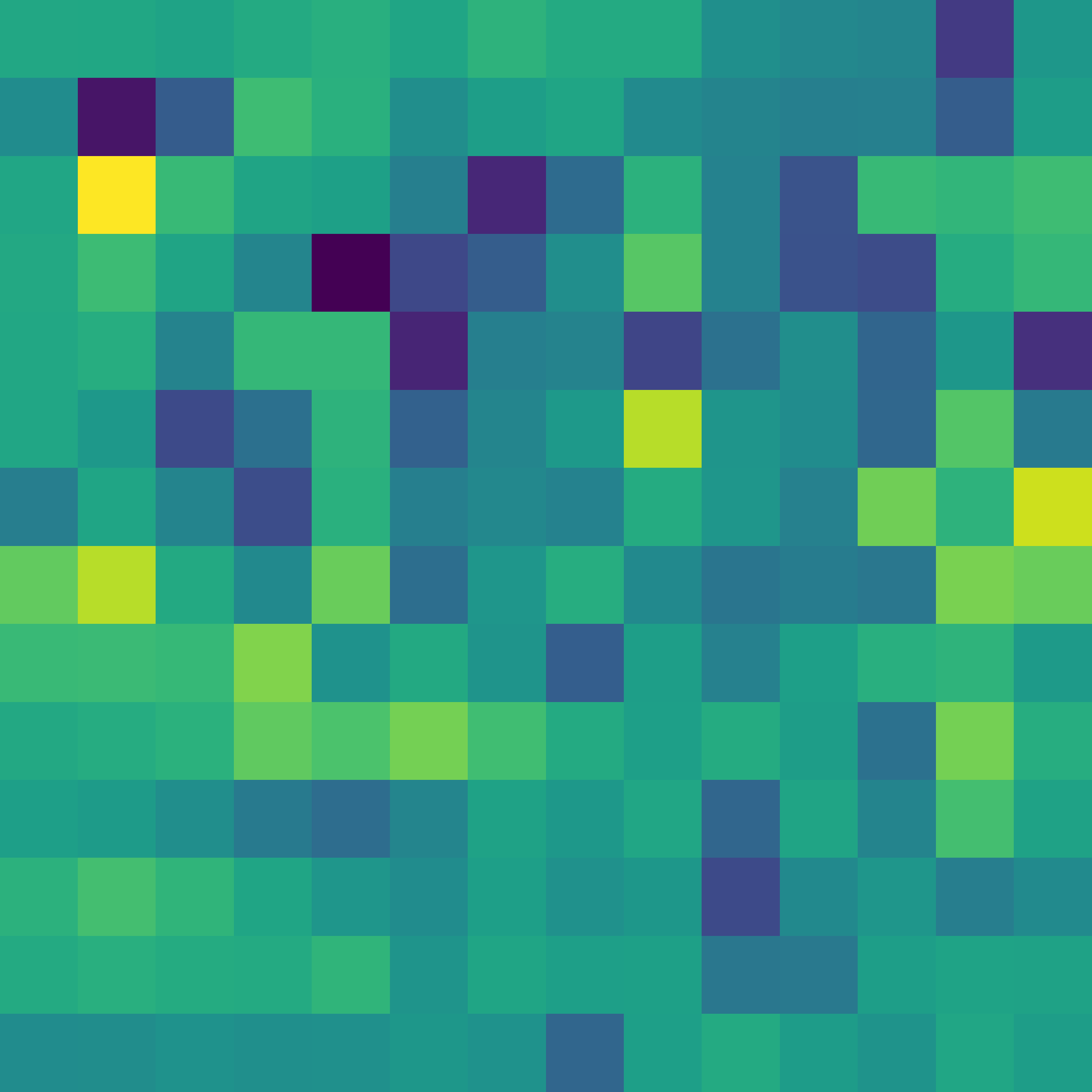} &
    \onebyfour{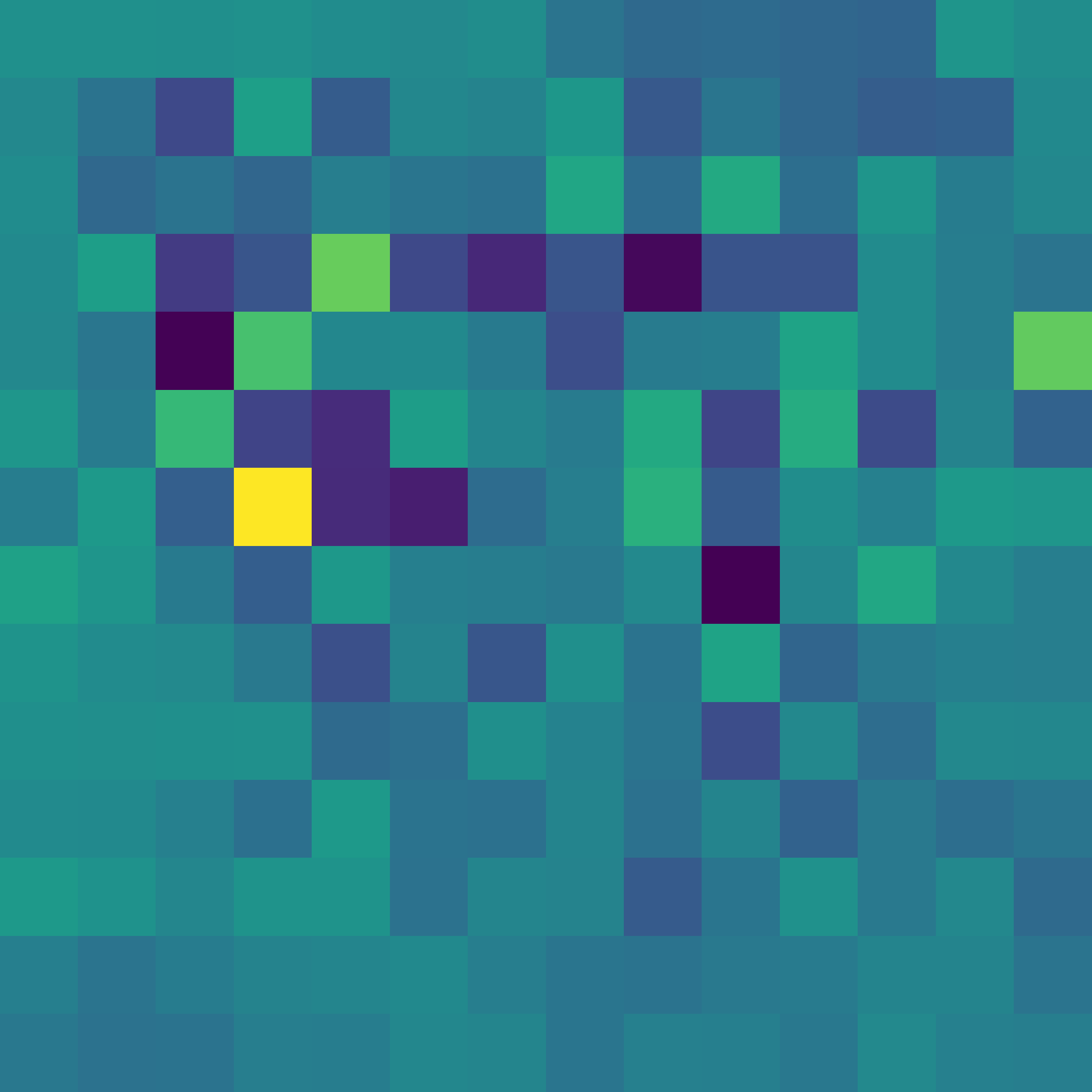}{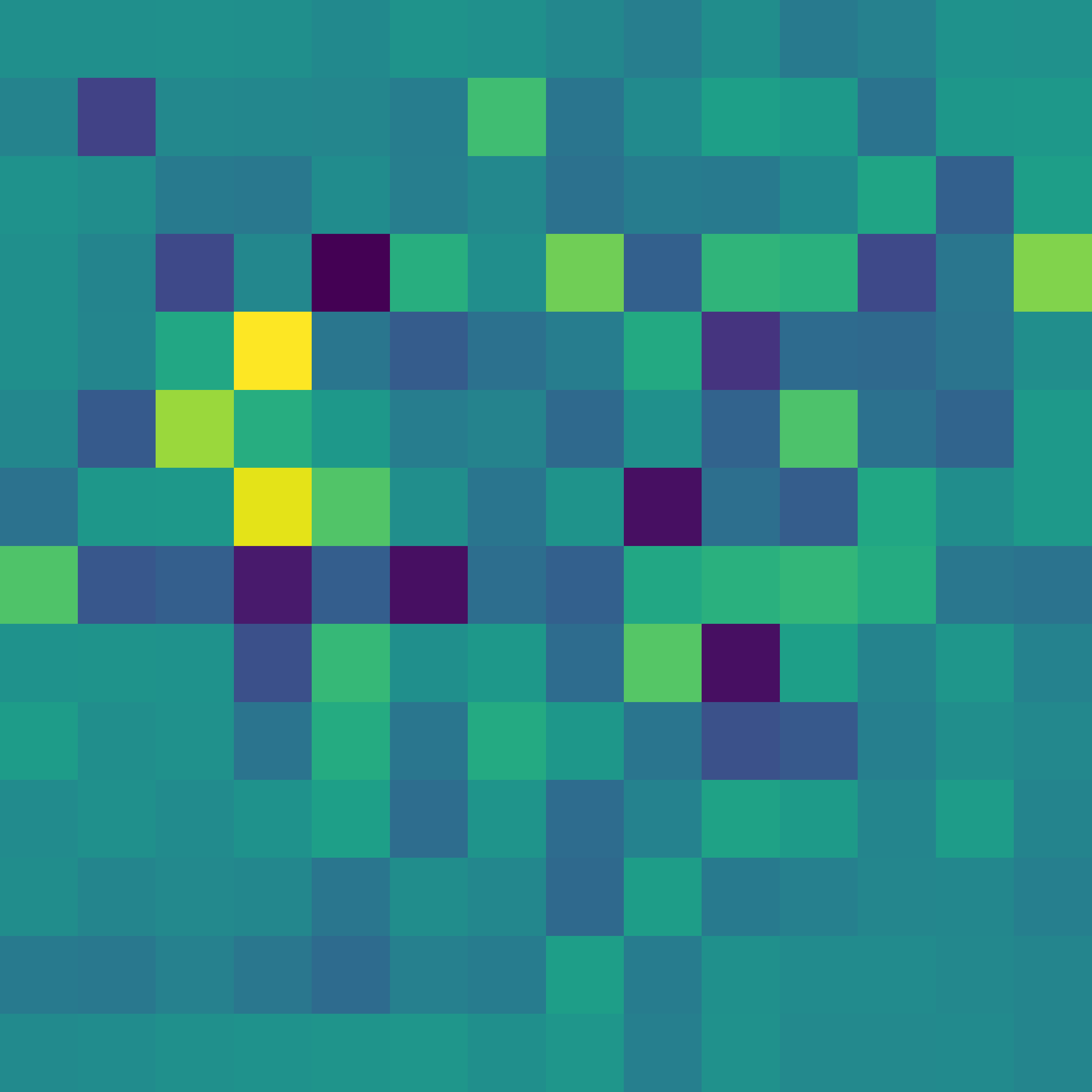}{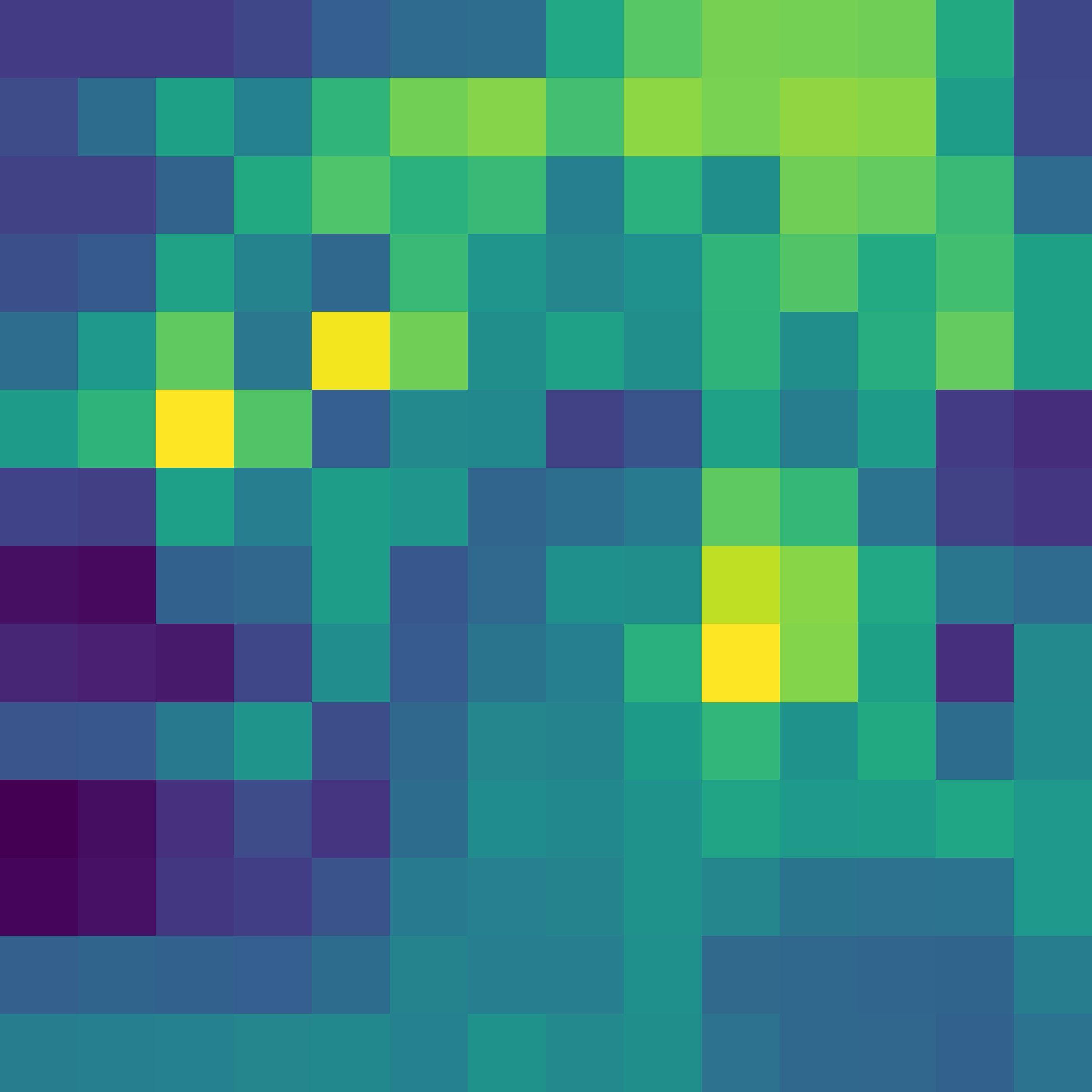}{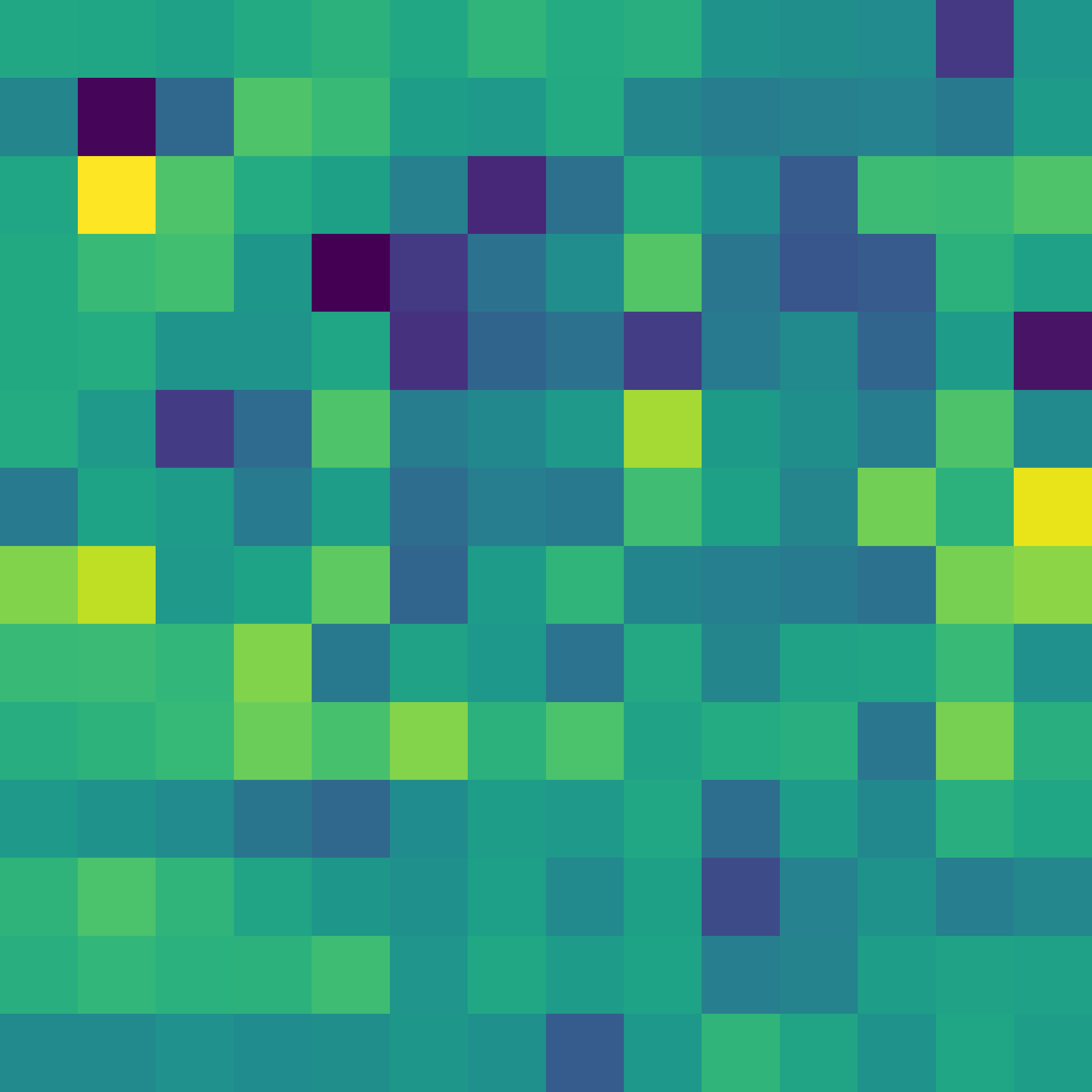} &
    \onebyfour{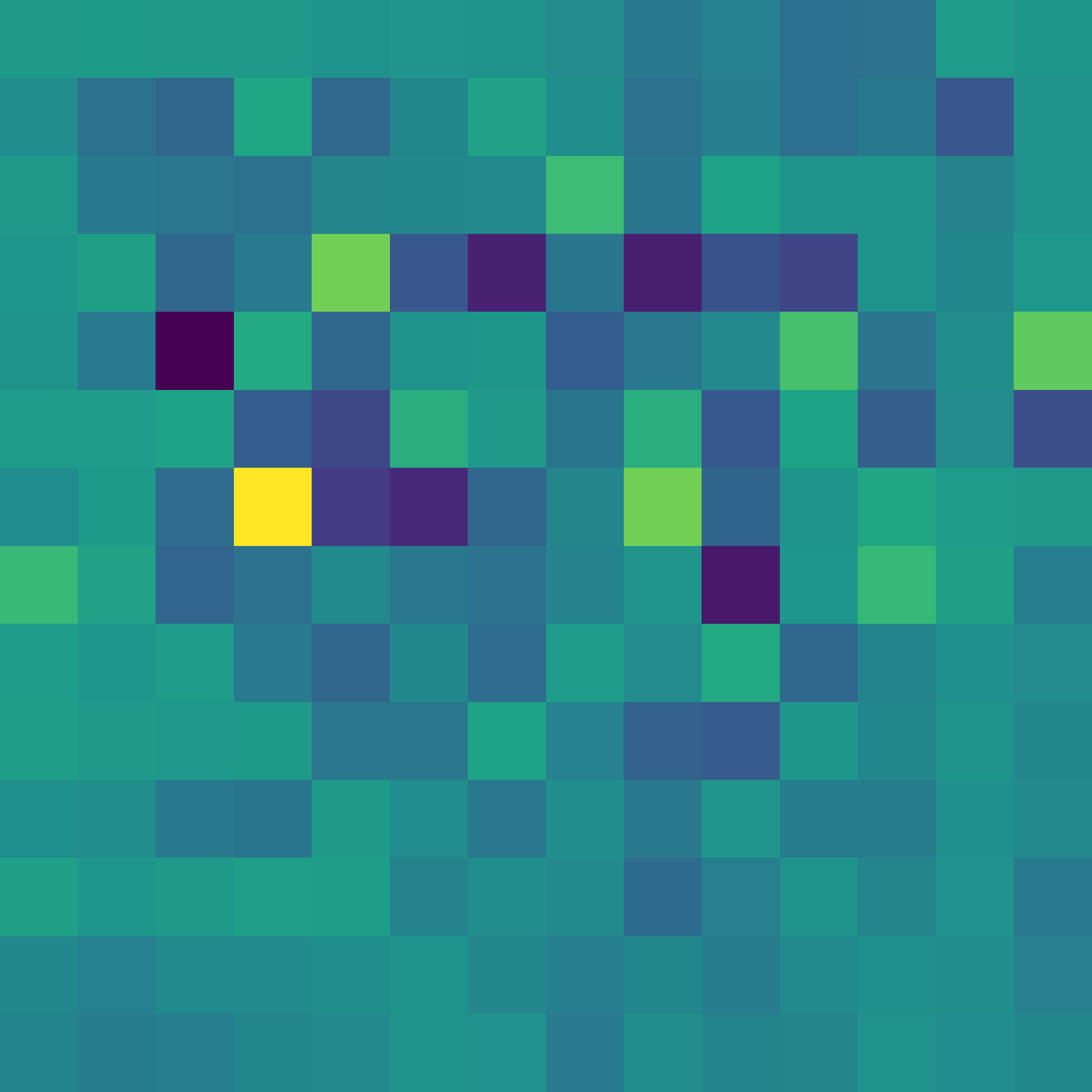}{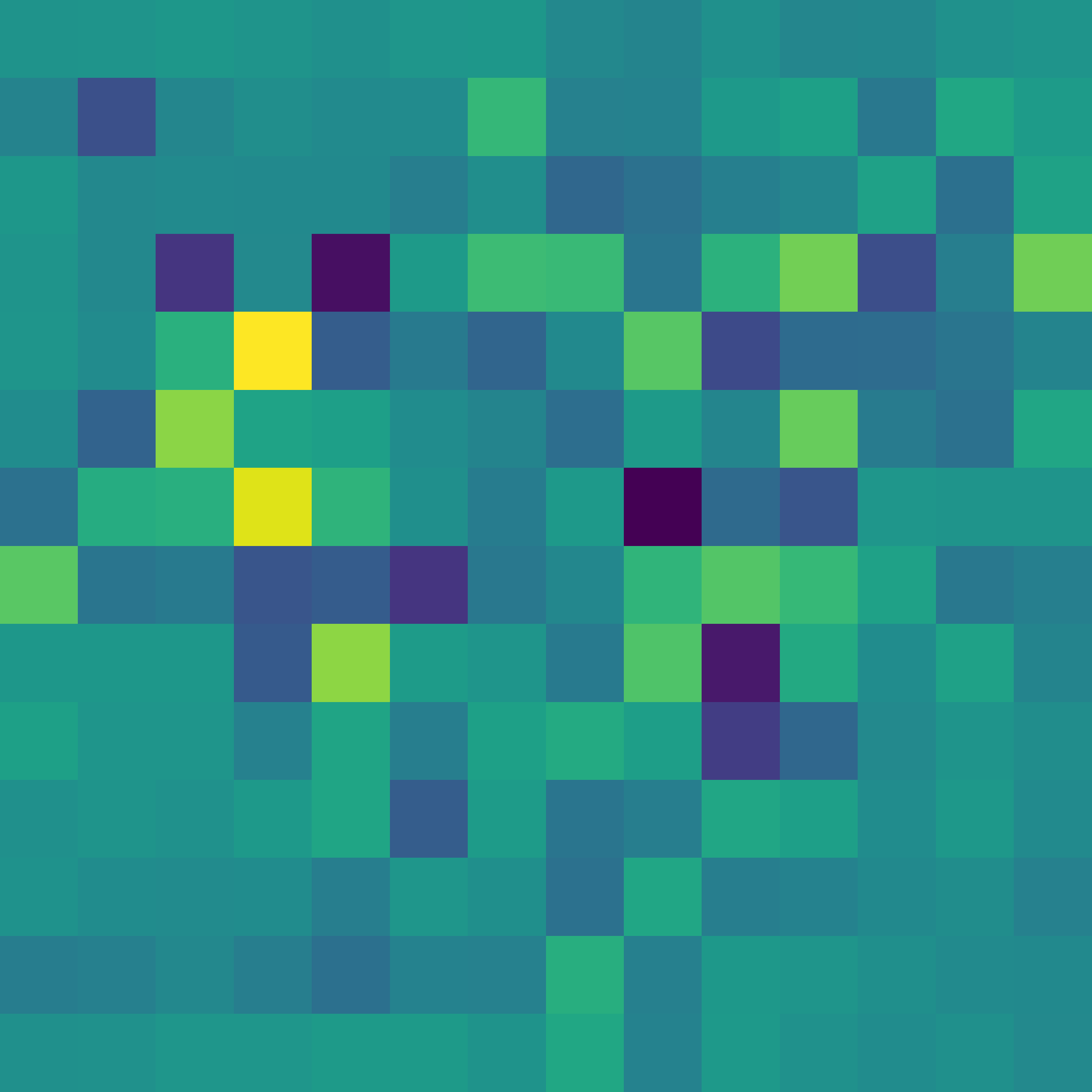}{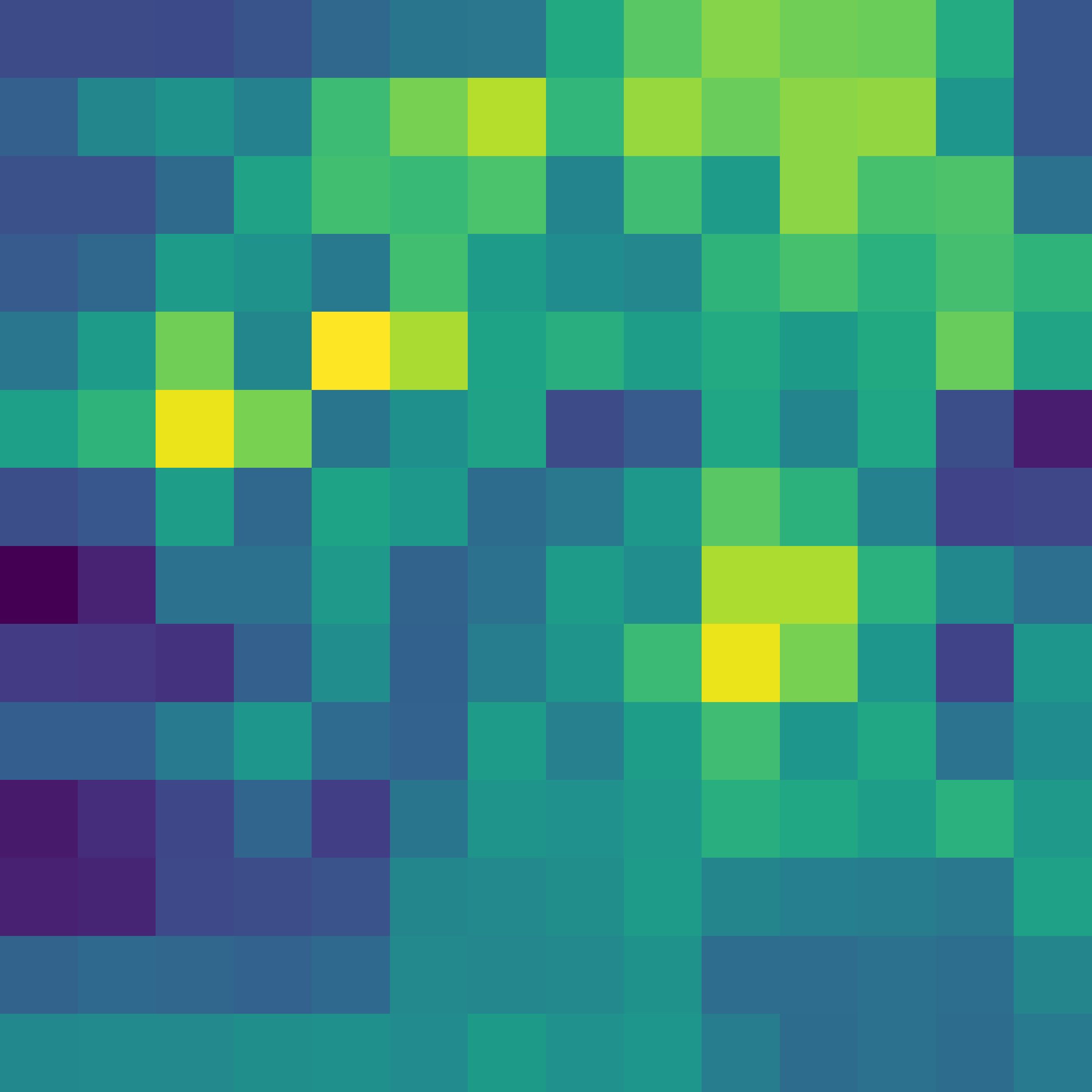}{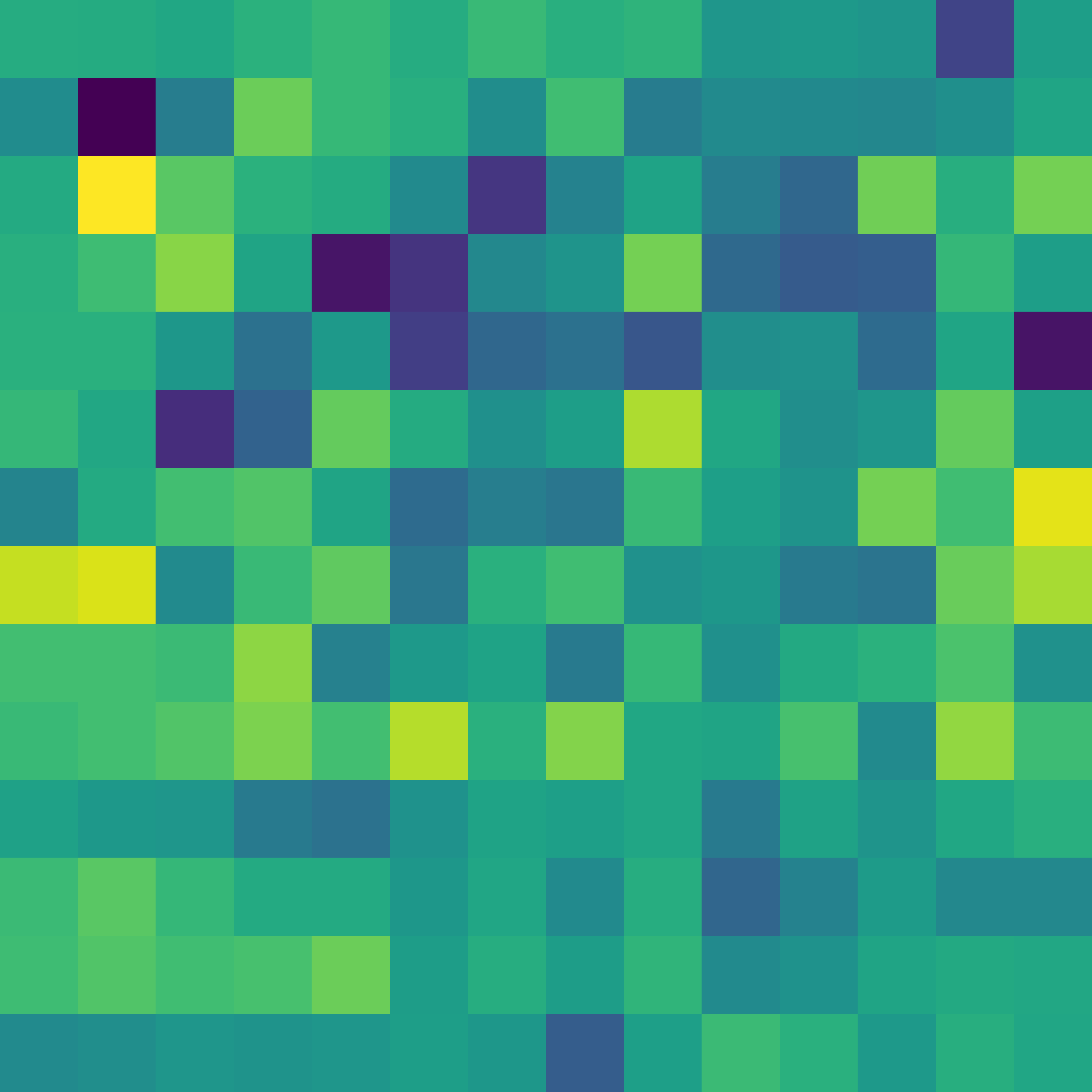} \\[0.4cm]
    \begin{minipage}{0.065\textwidth}\centering\includegraphics[width=0.91\linewidth]{figs/teaser_figs/reference_images/4824411109_9124f1d397_z.png}\end{minipage} & \onebyfour{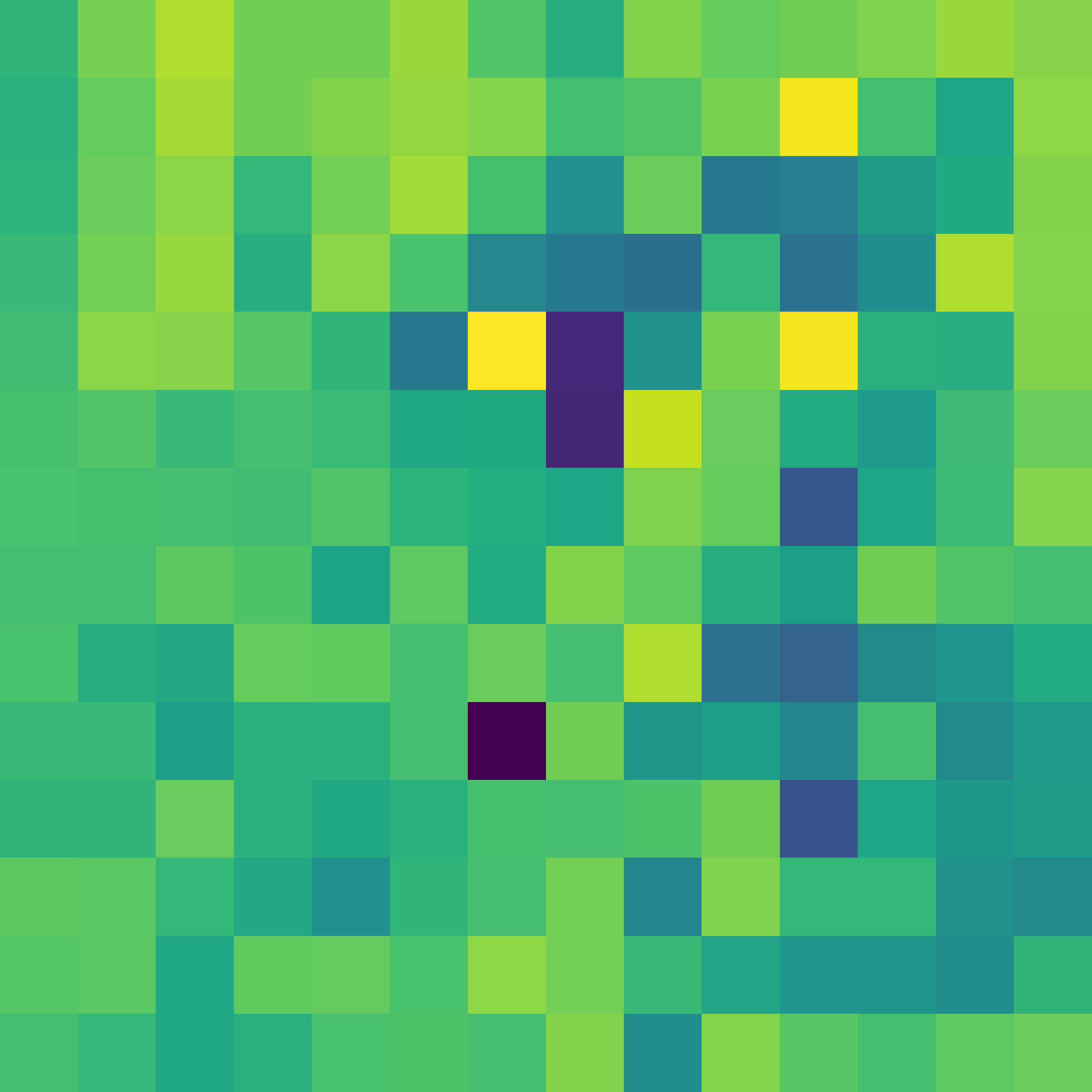}{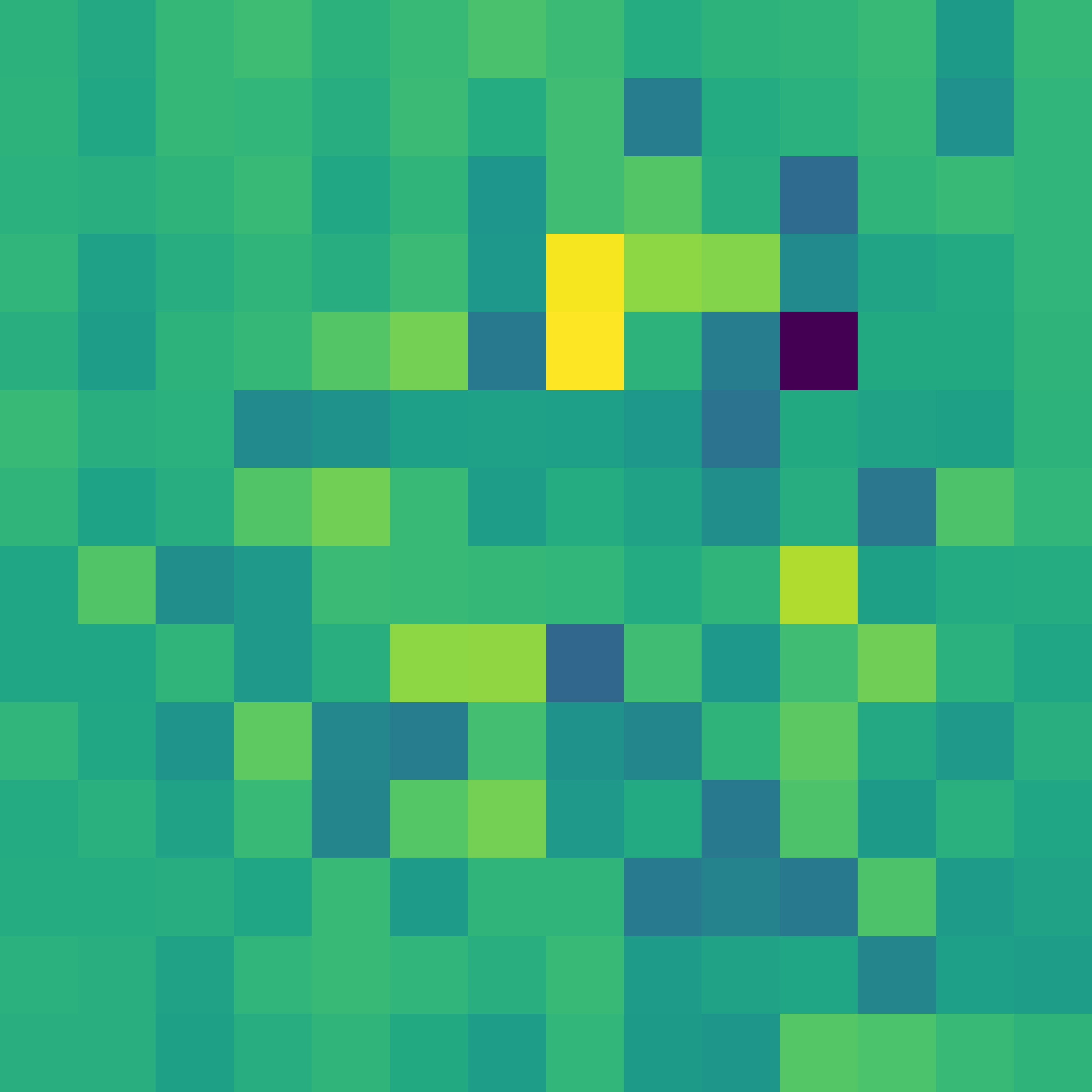}{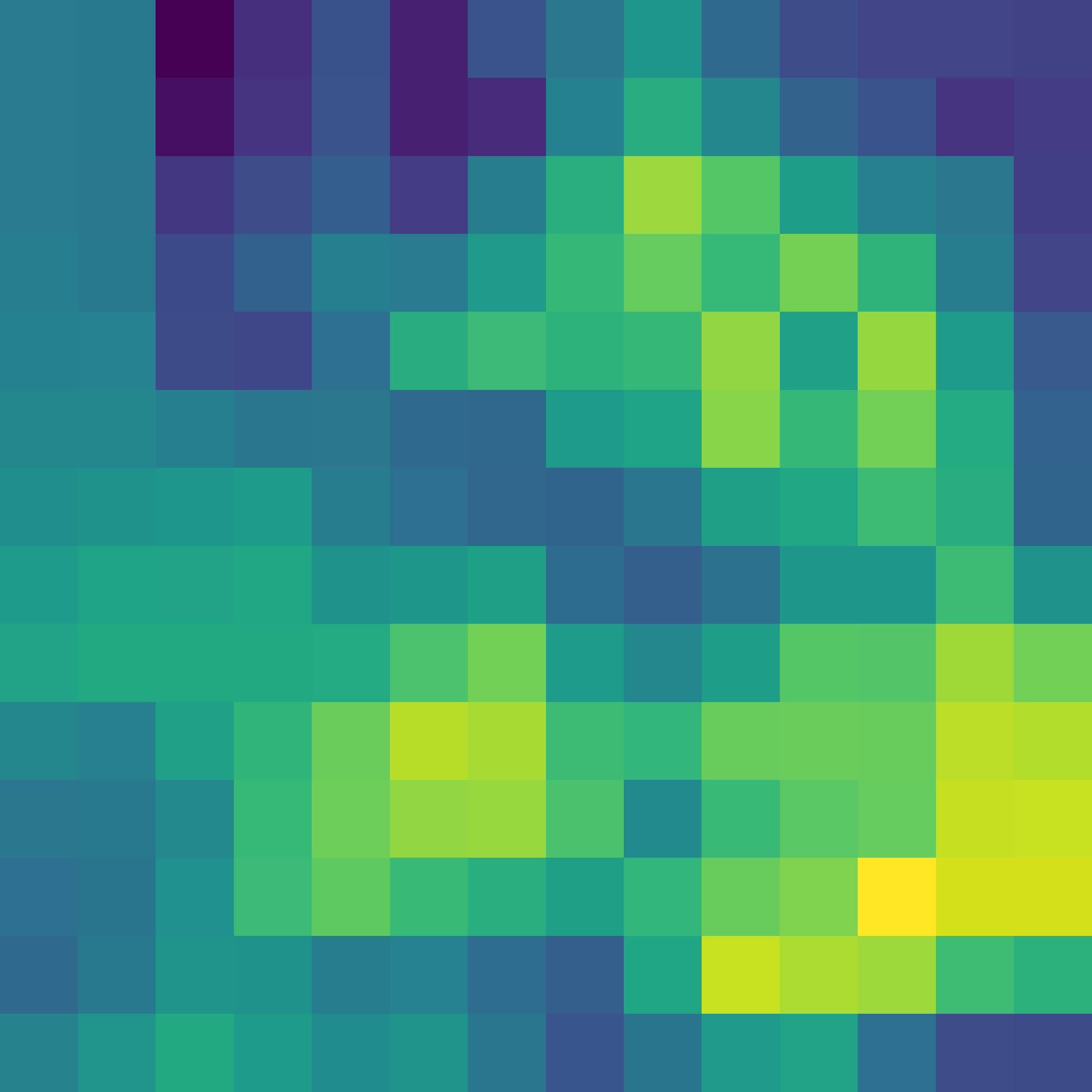}{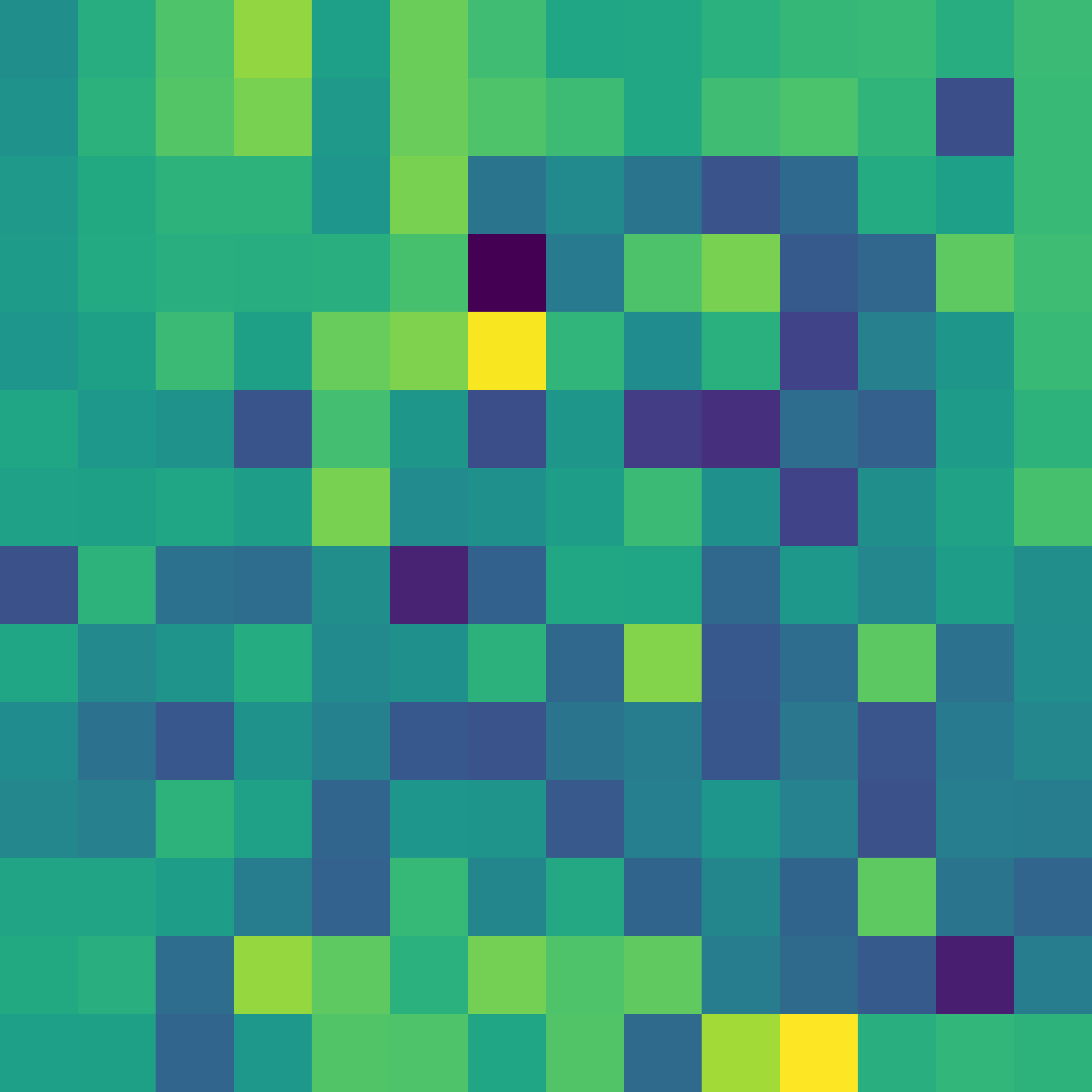} &
    \onebyfour{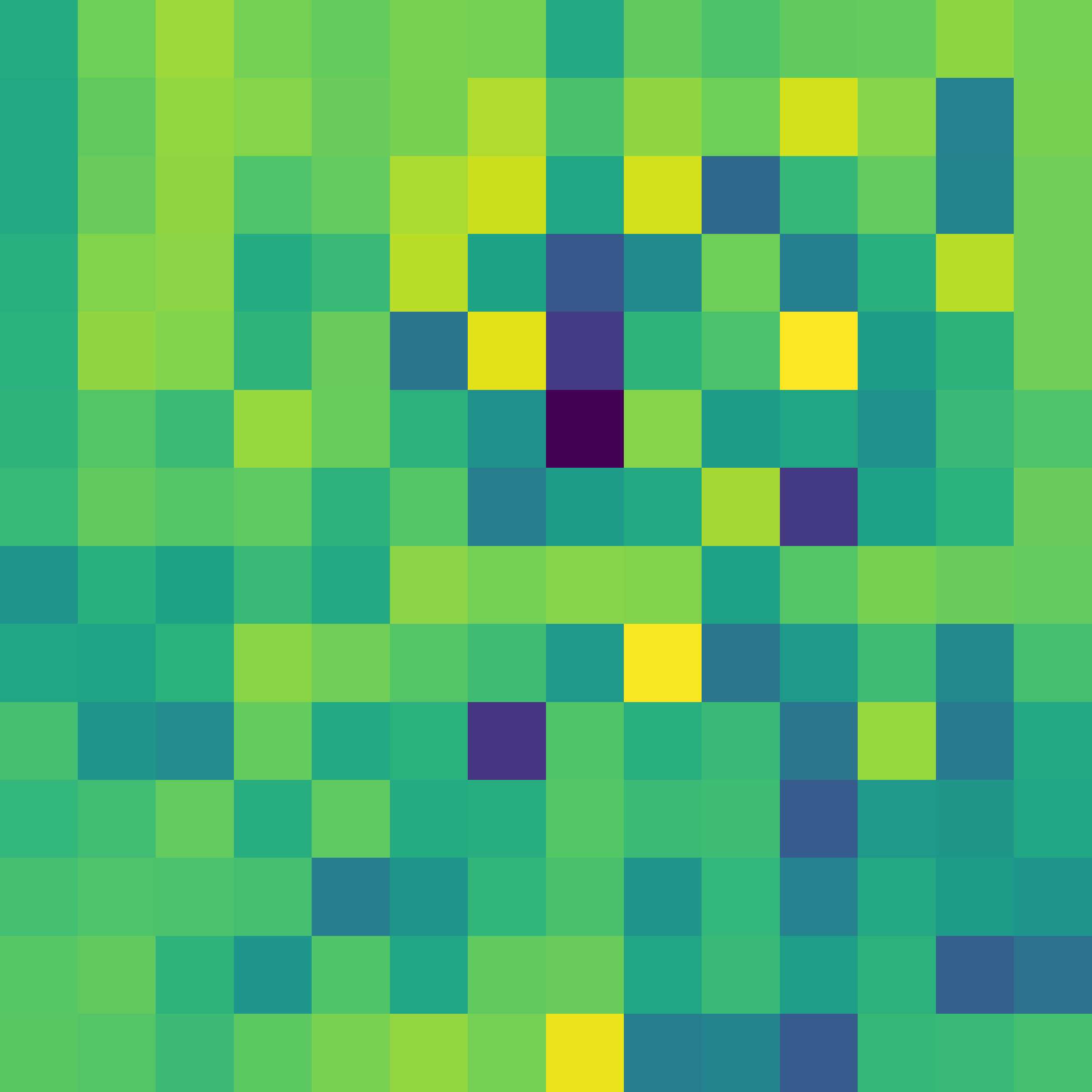}{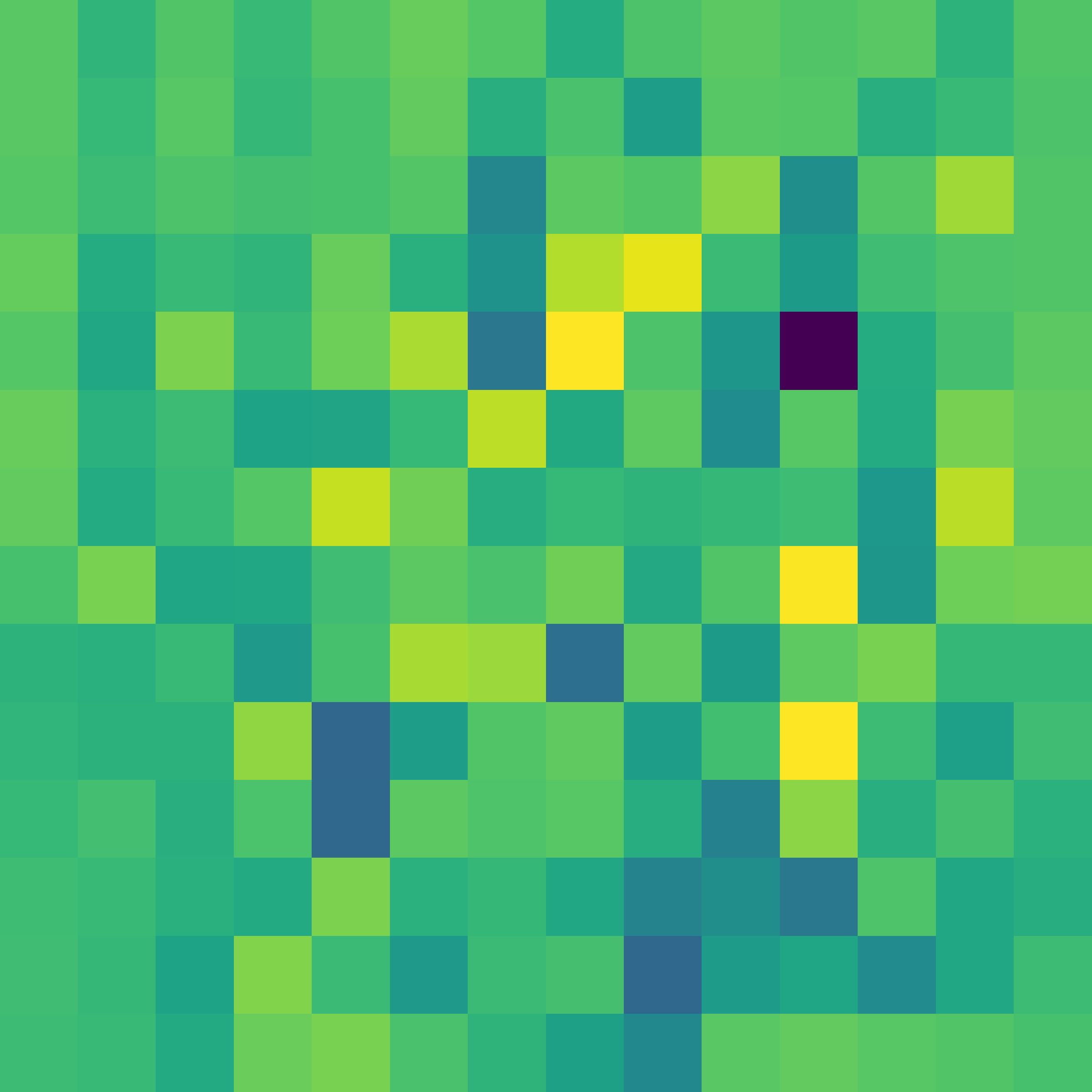}{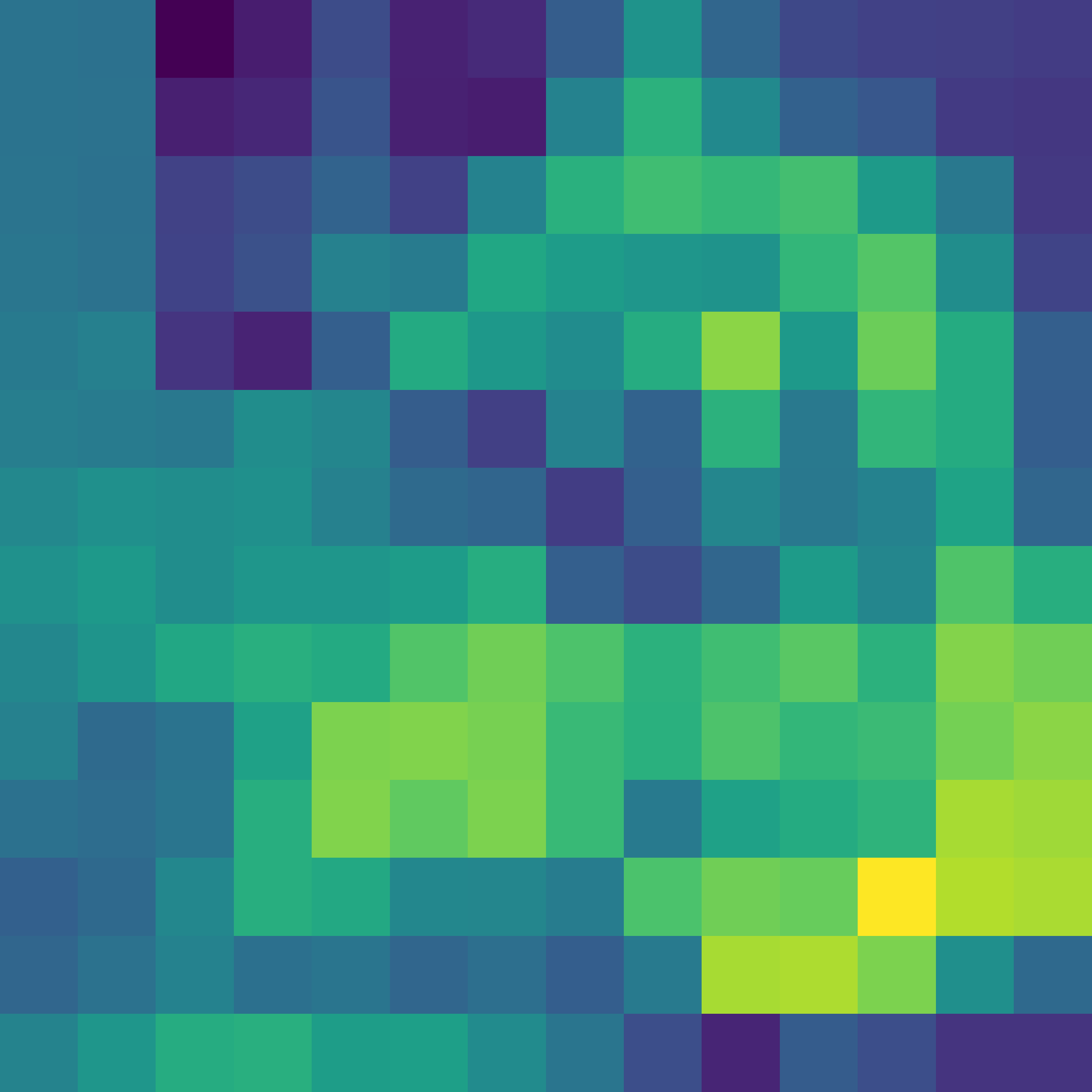}{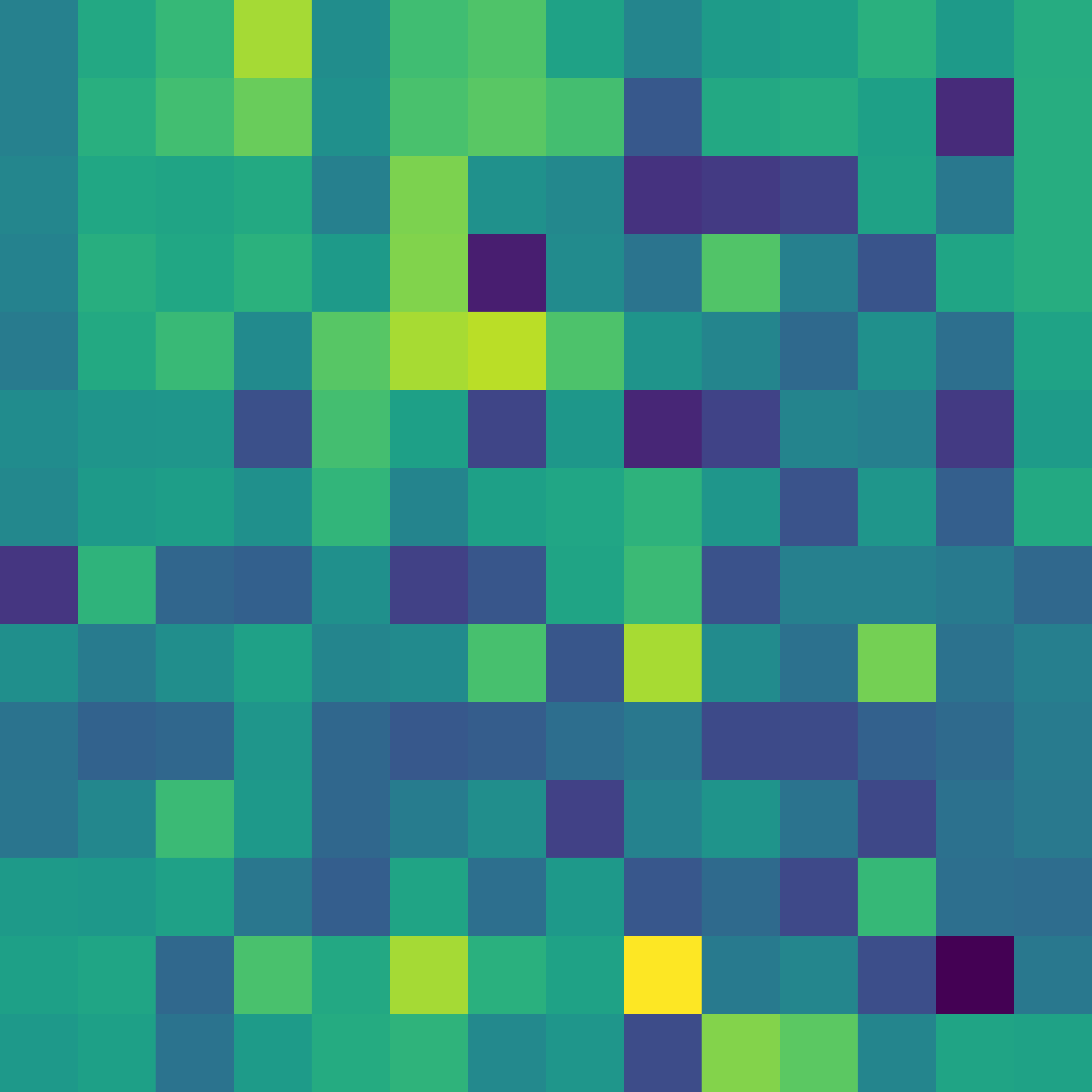} &
    \onebyfour{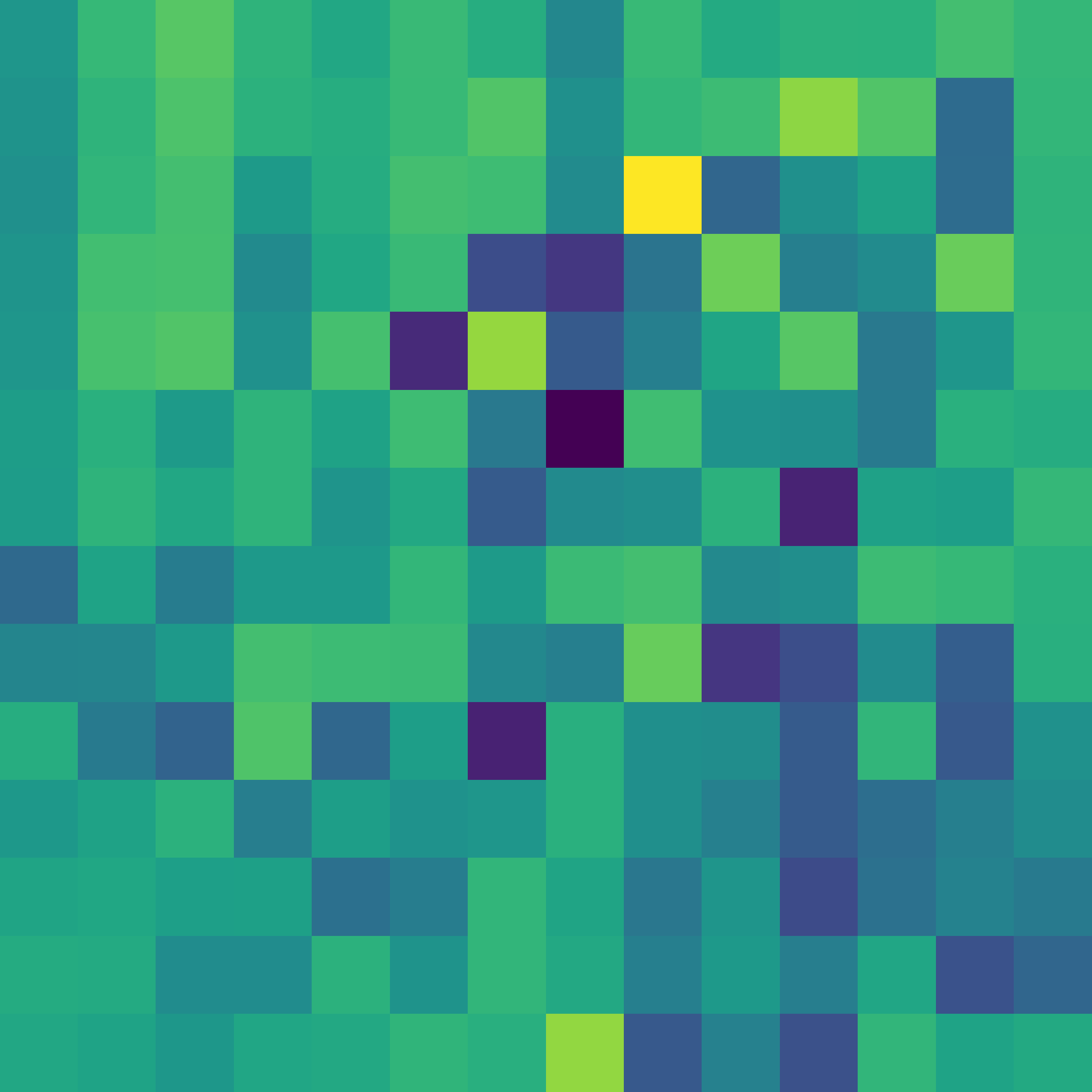}{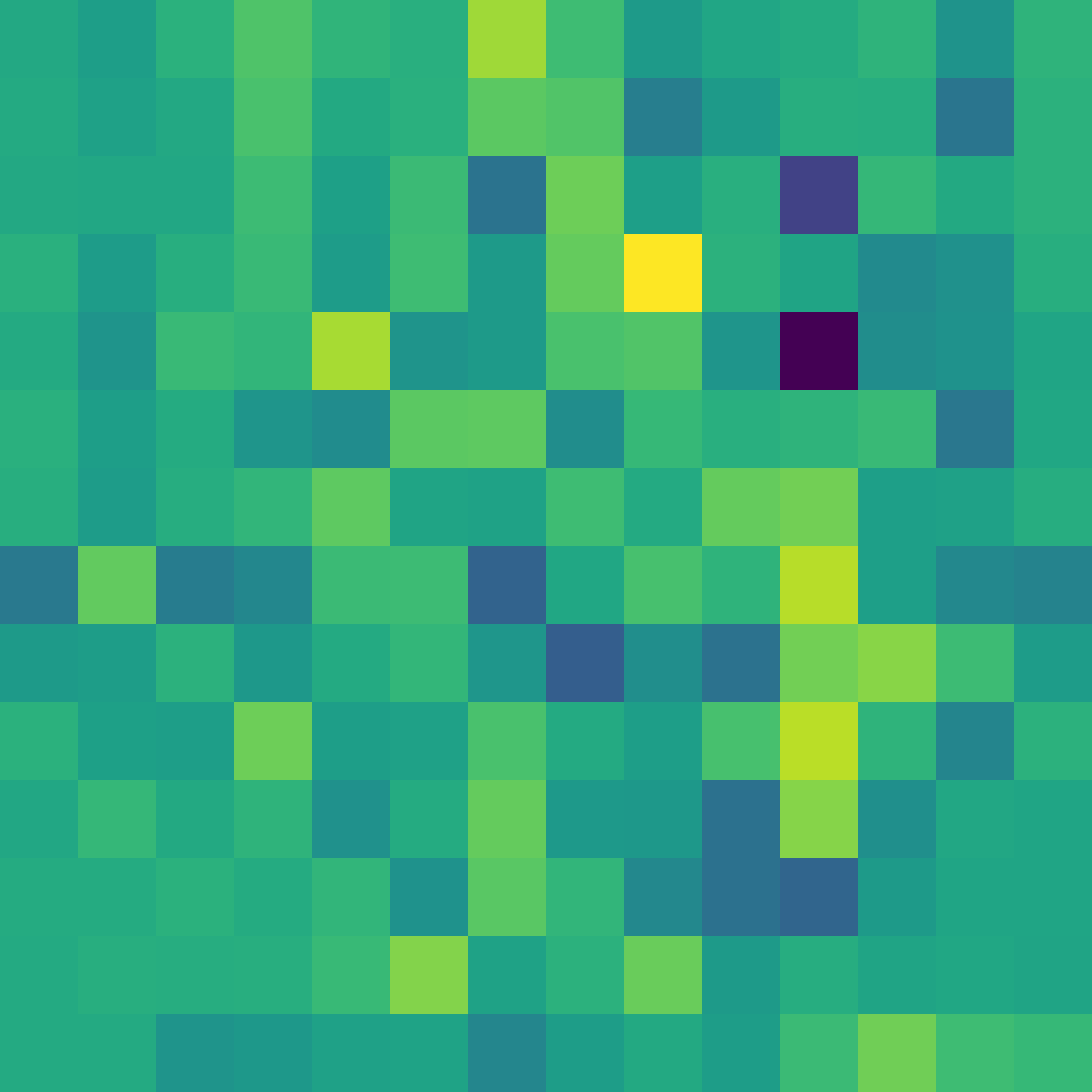}{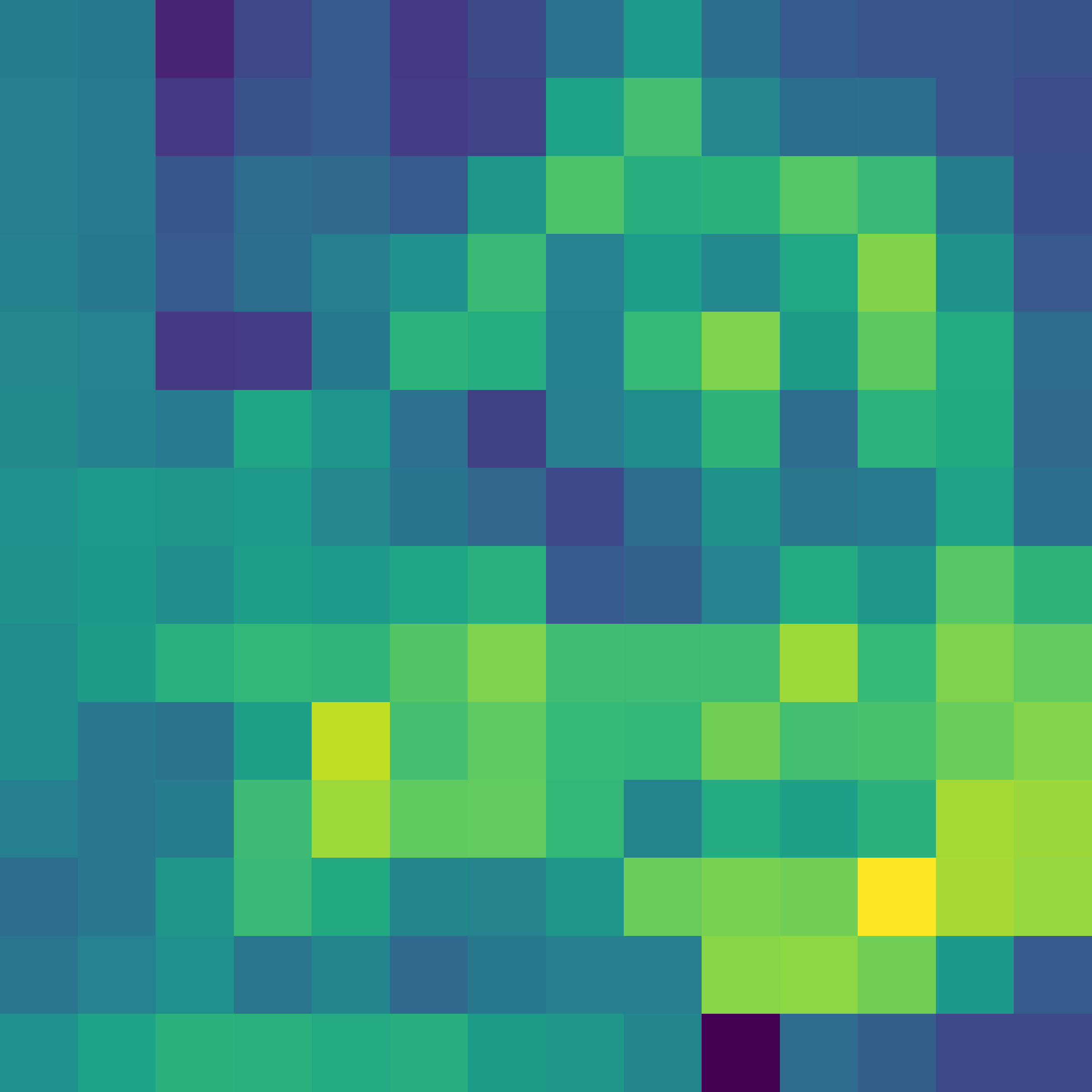}{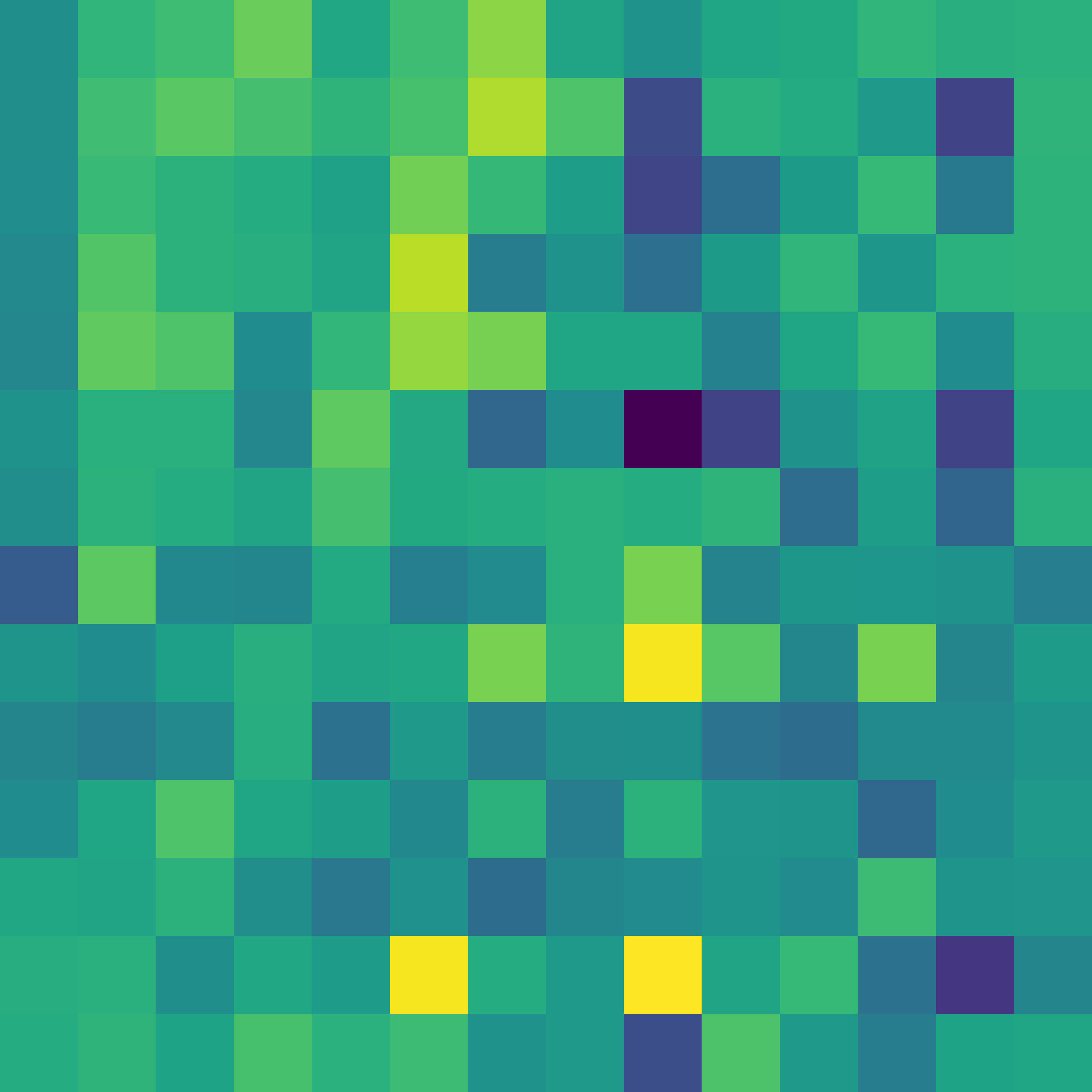} &
    \onebyfour{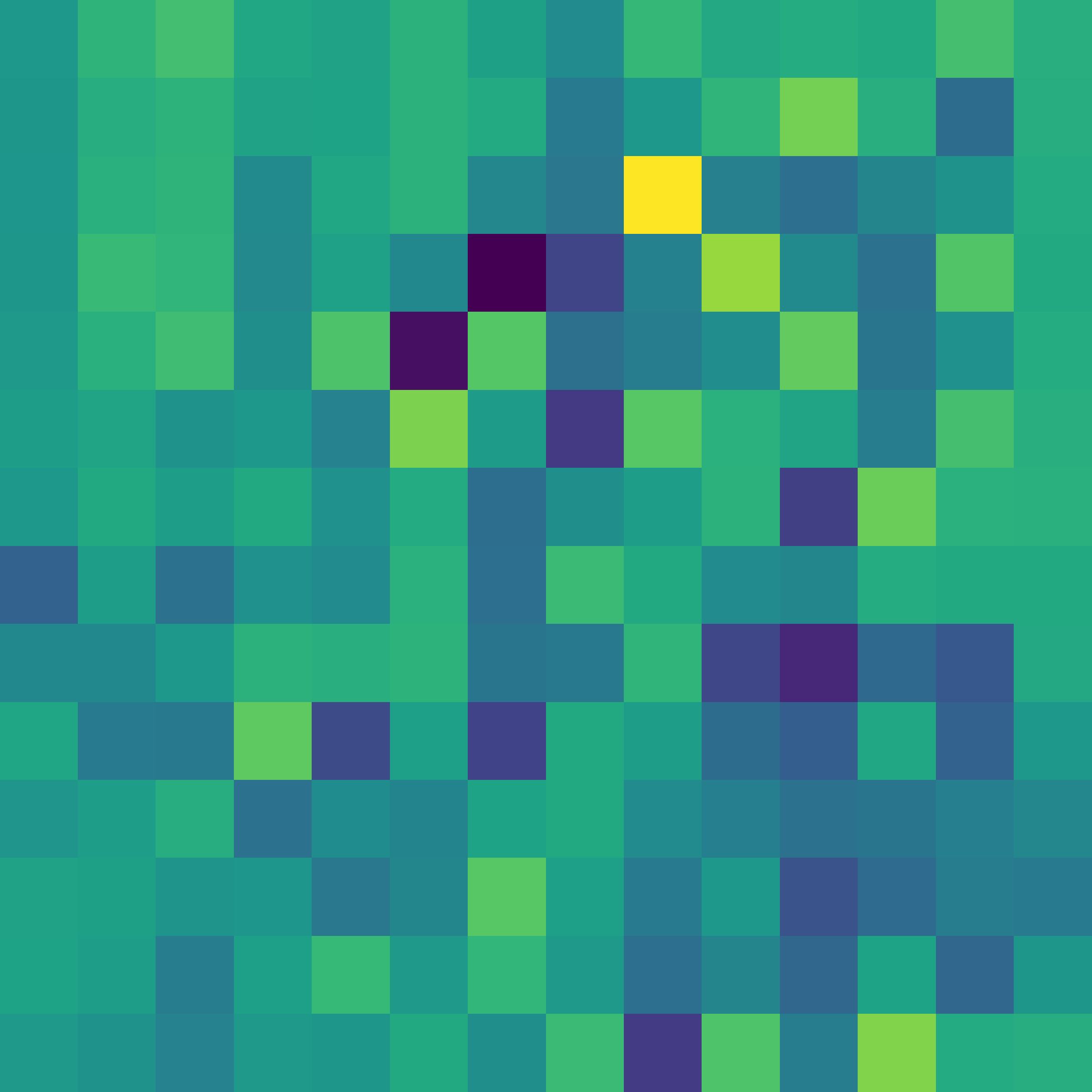}{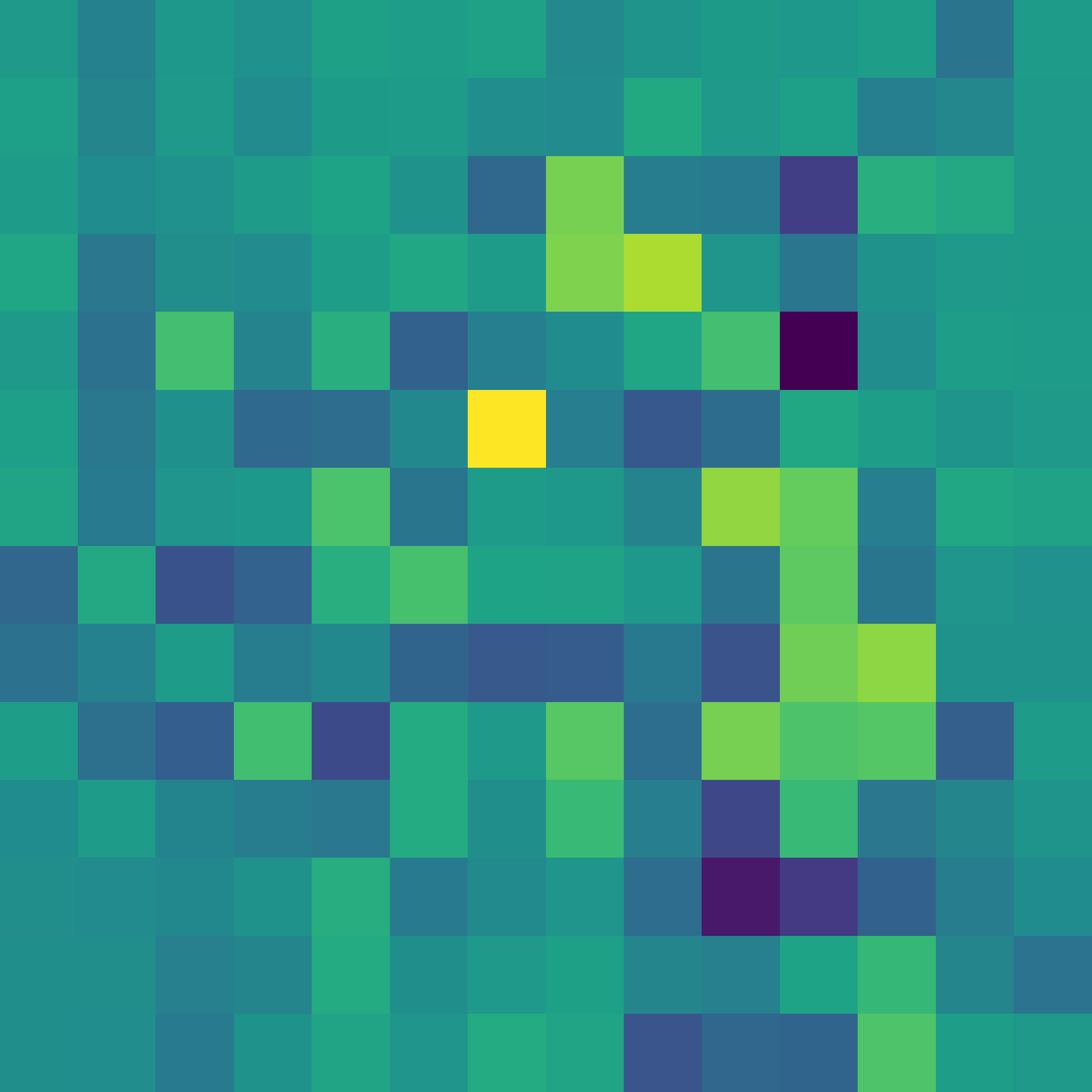}{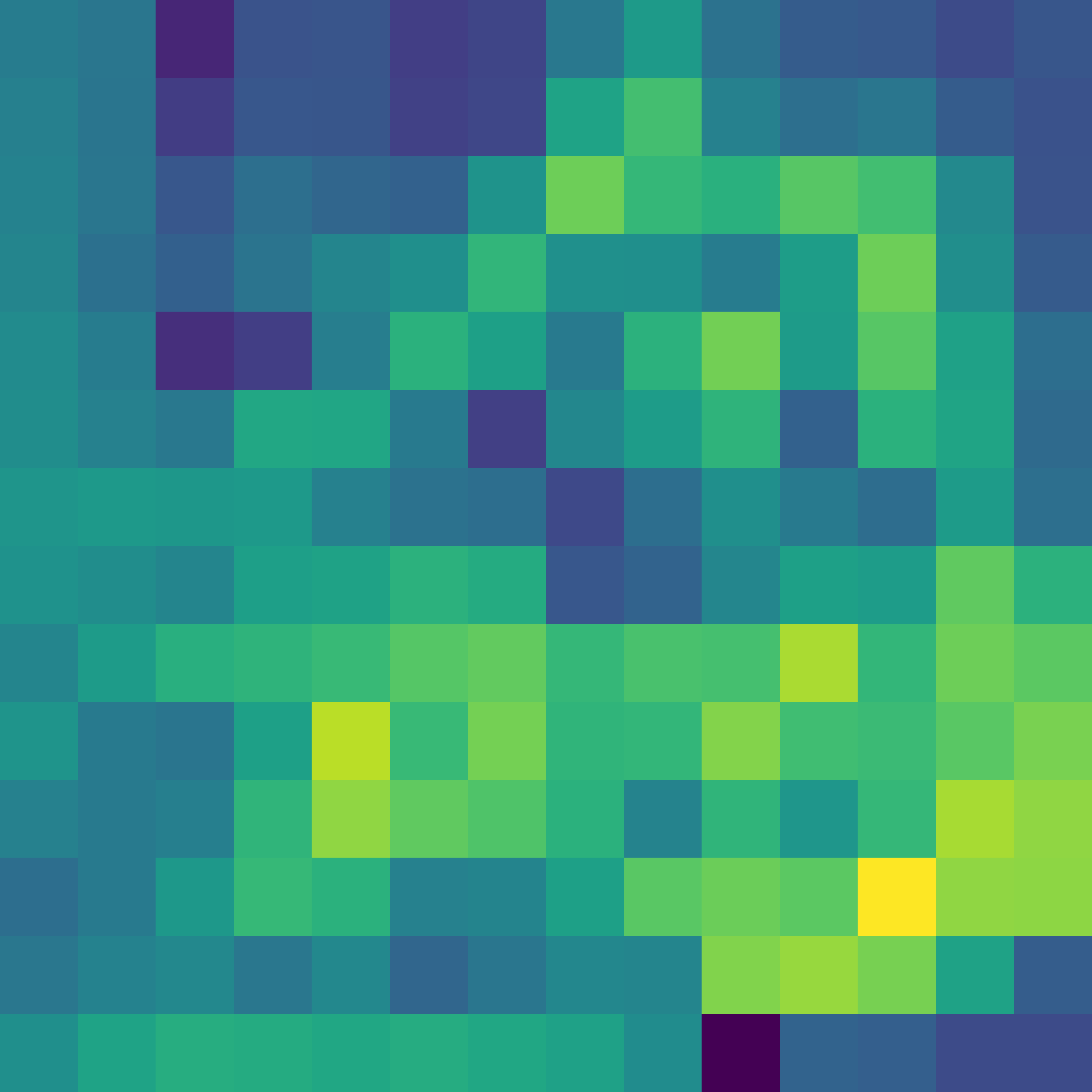}{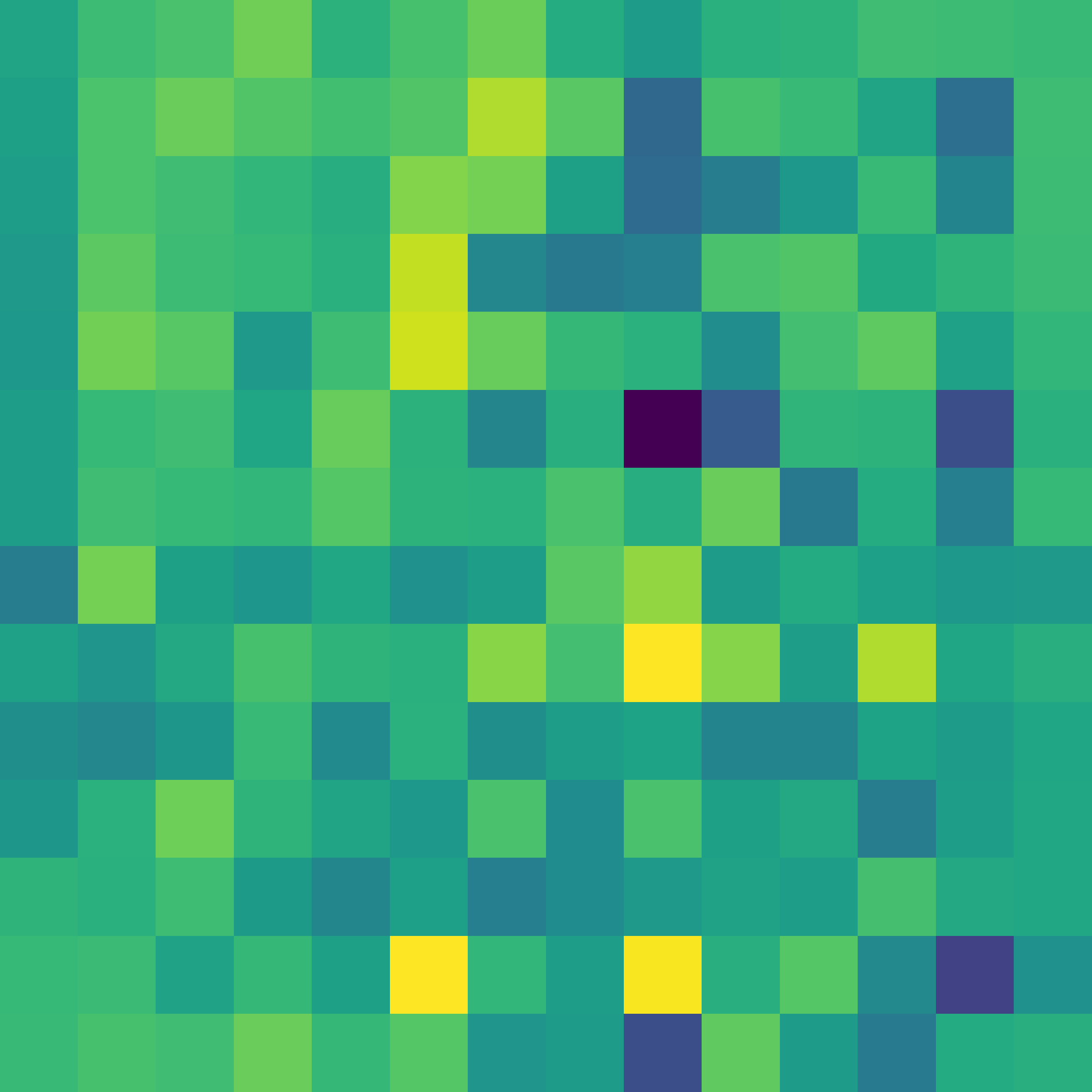} &
    \onebyfour{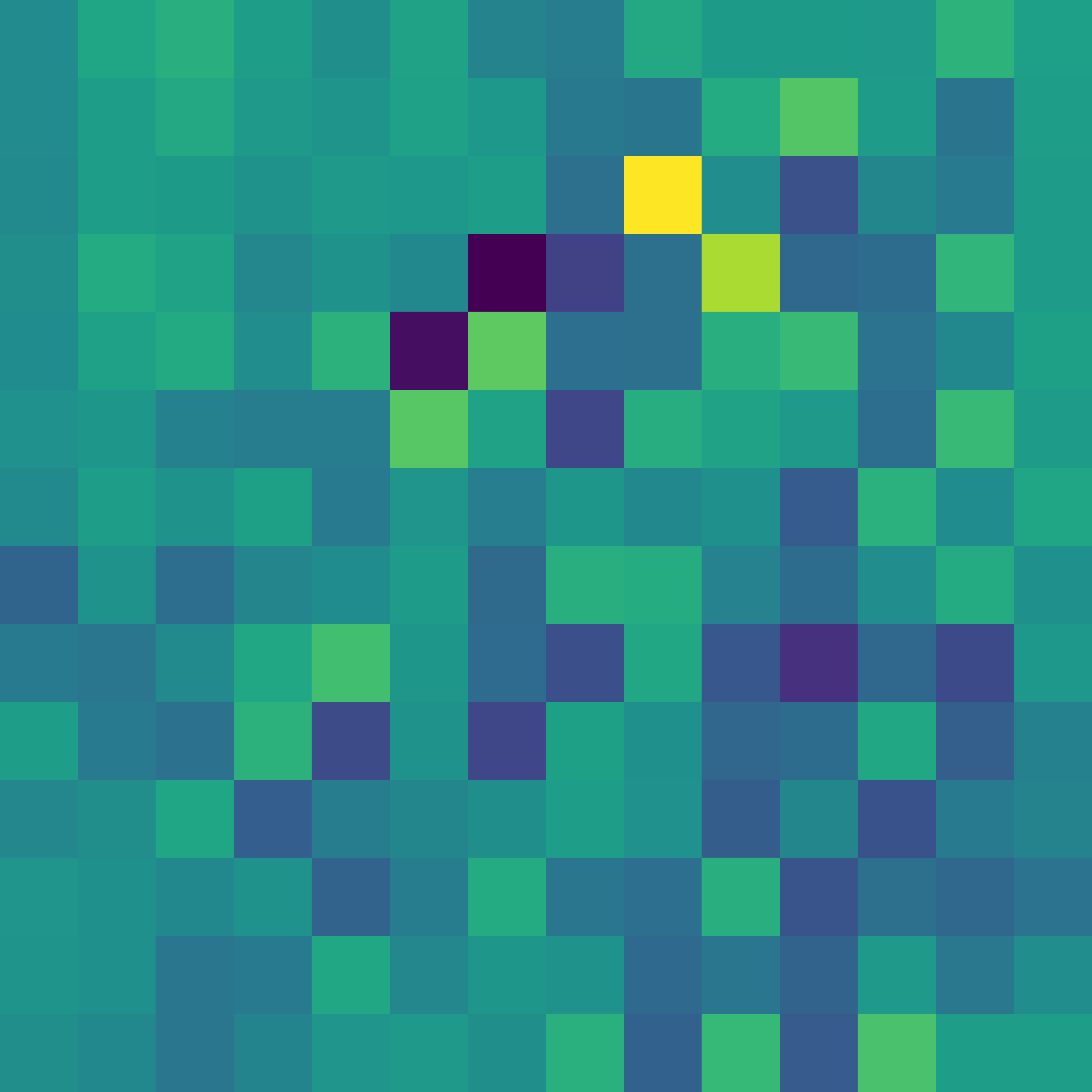}{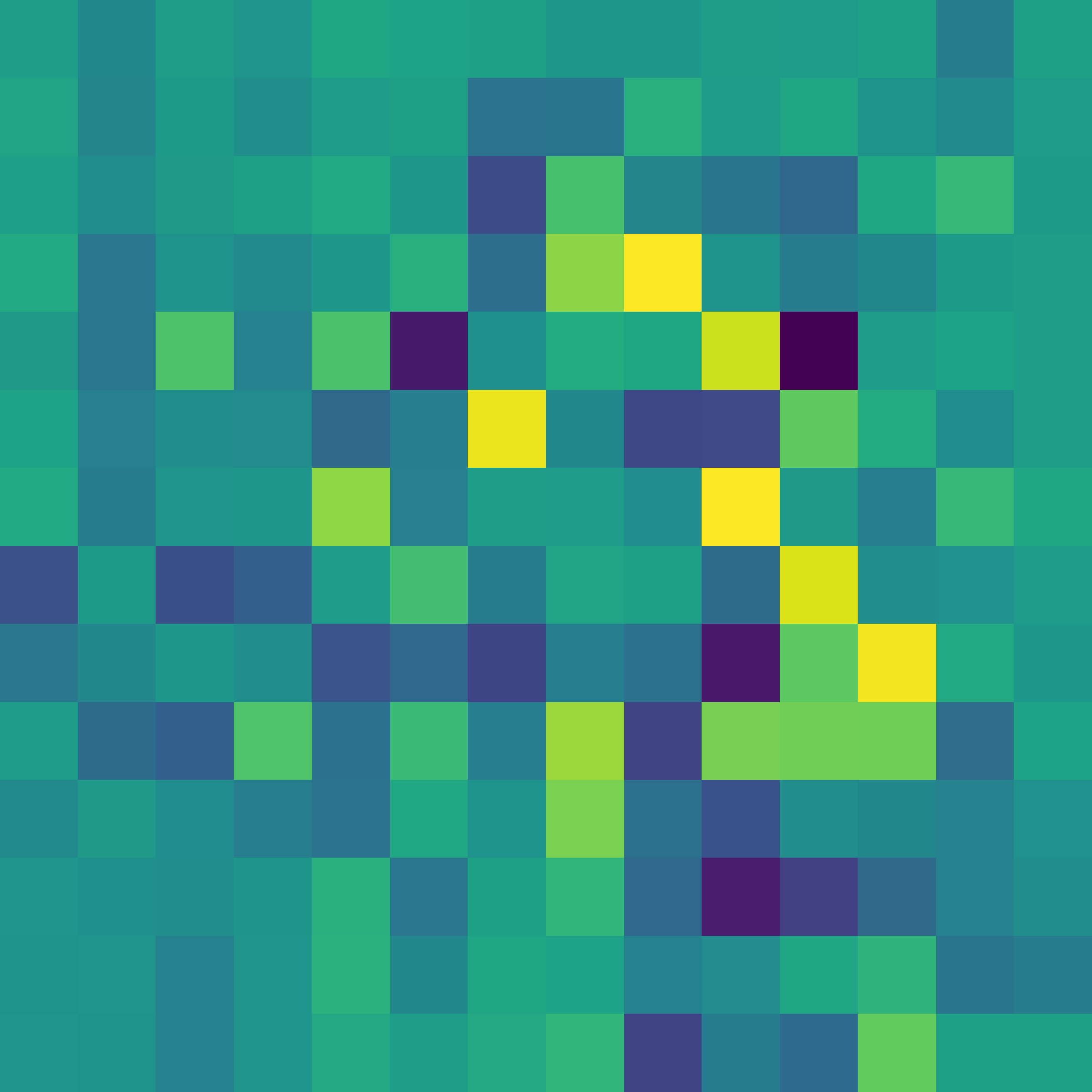}{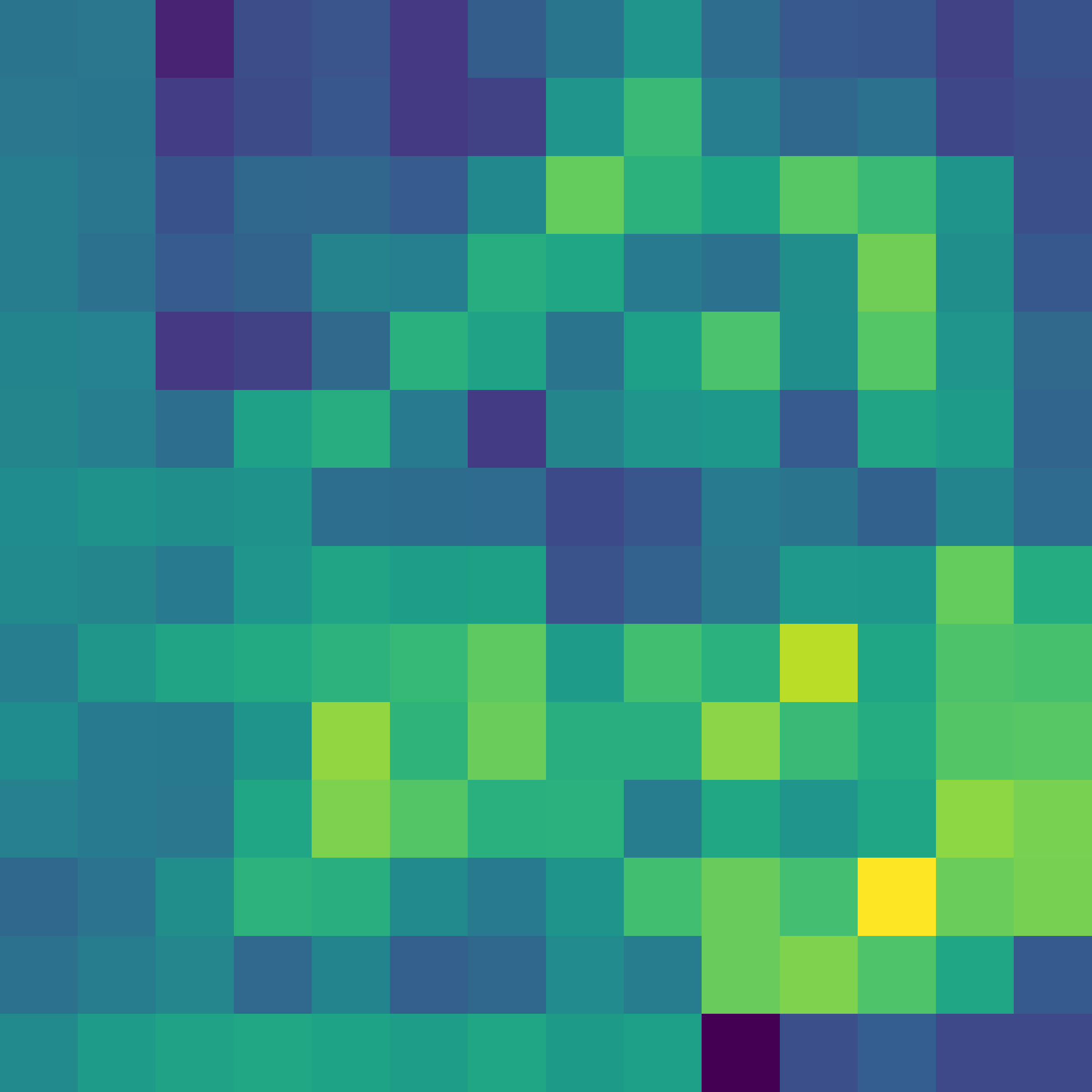}{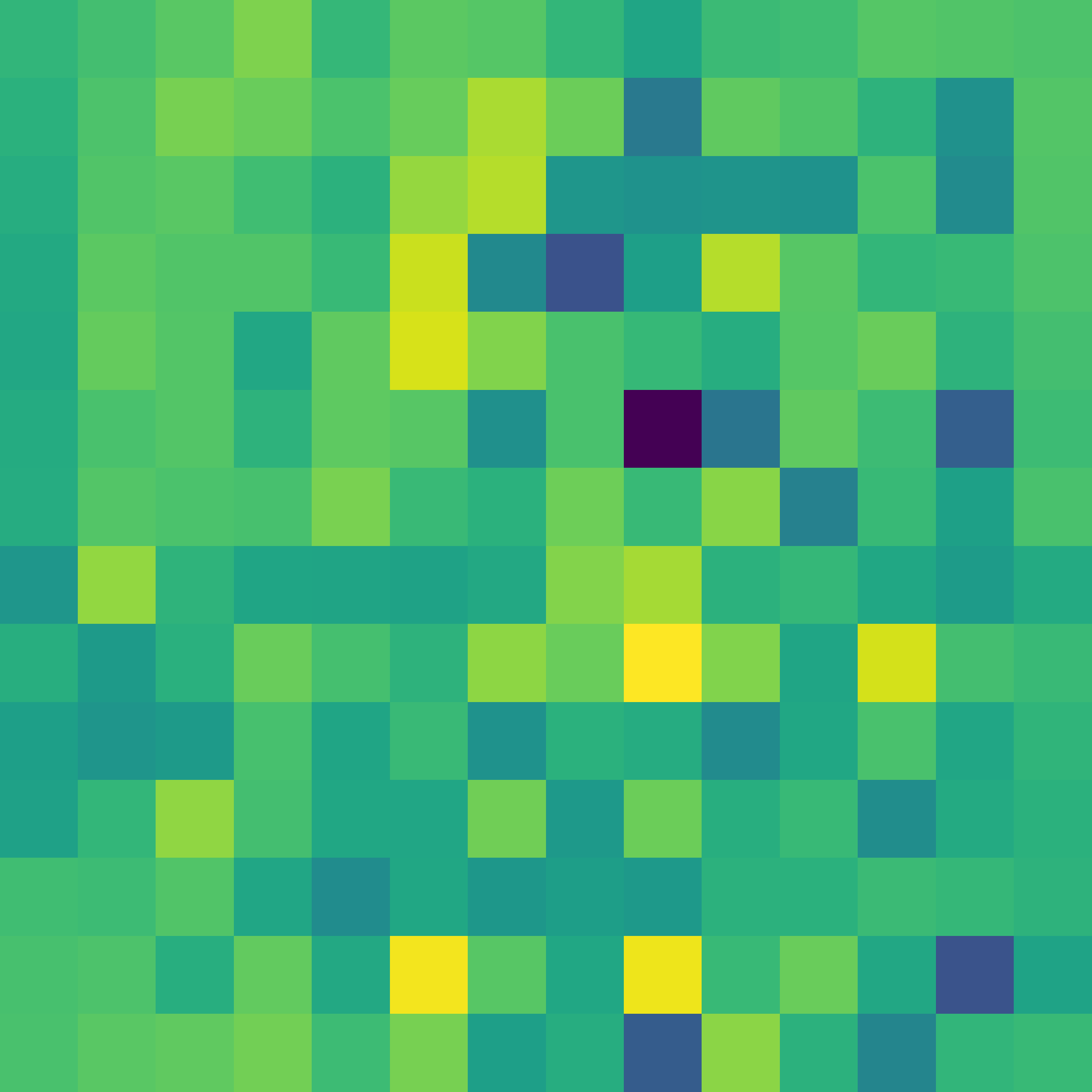} \\[0.45cm]
    Eq. Error & \onebyfour{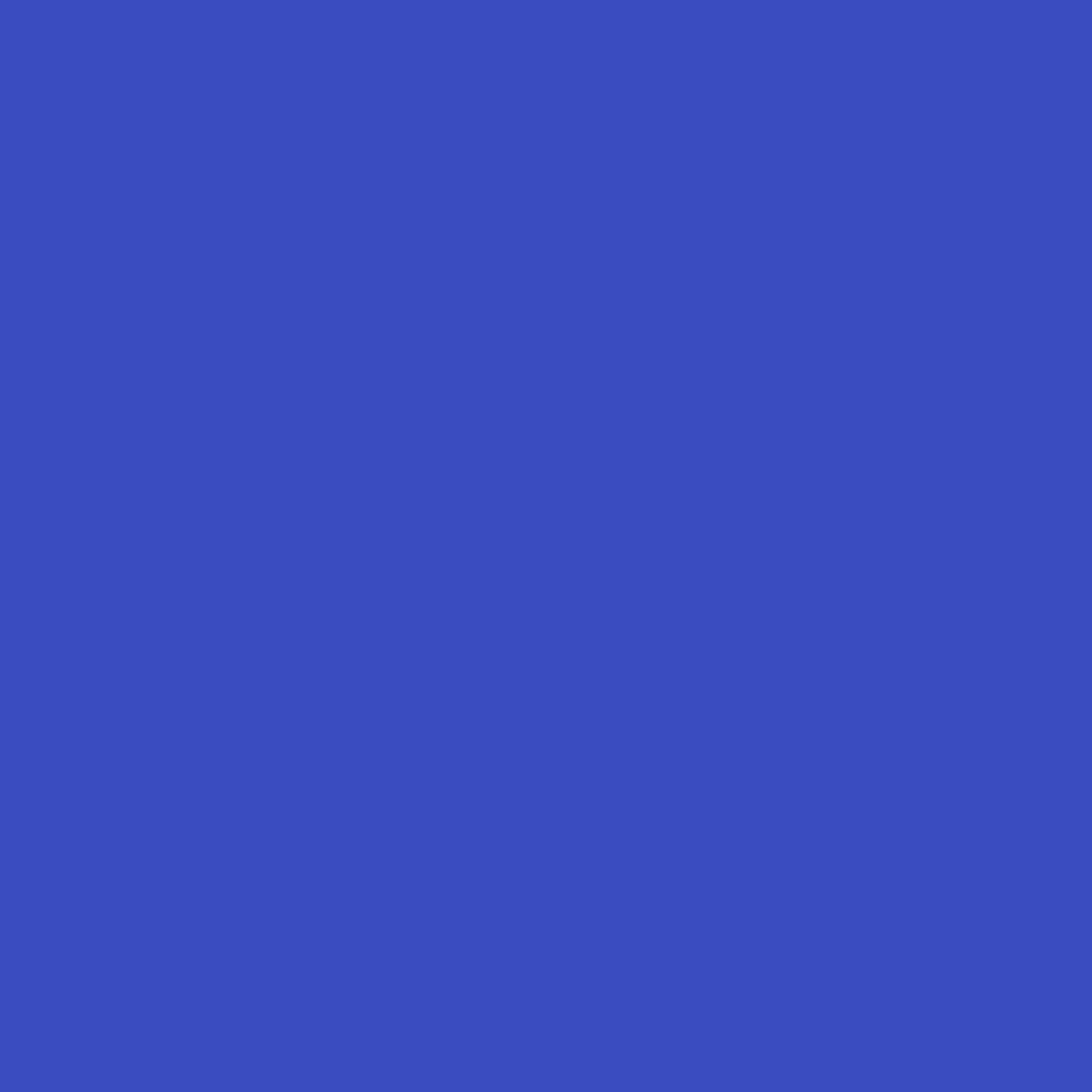}{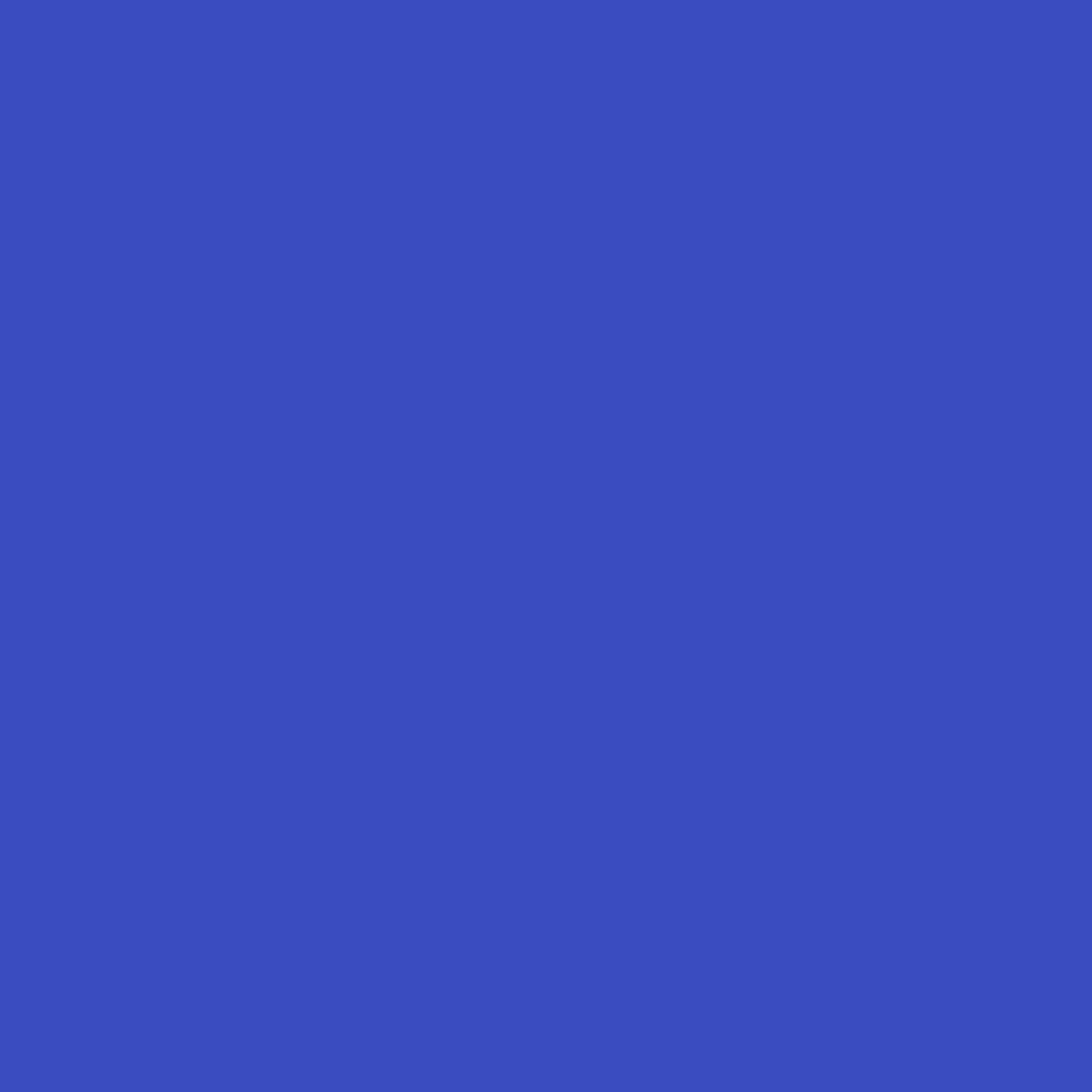}{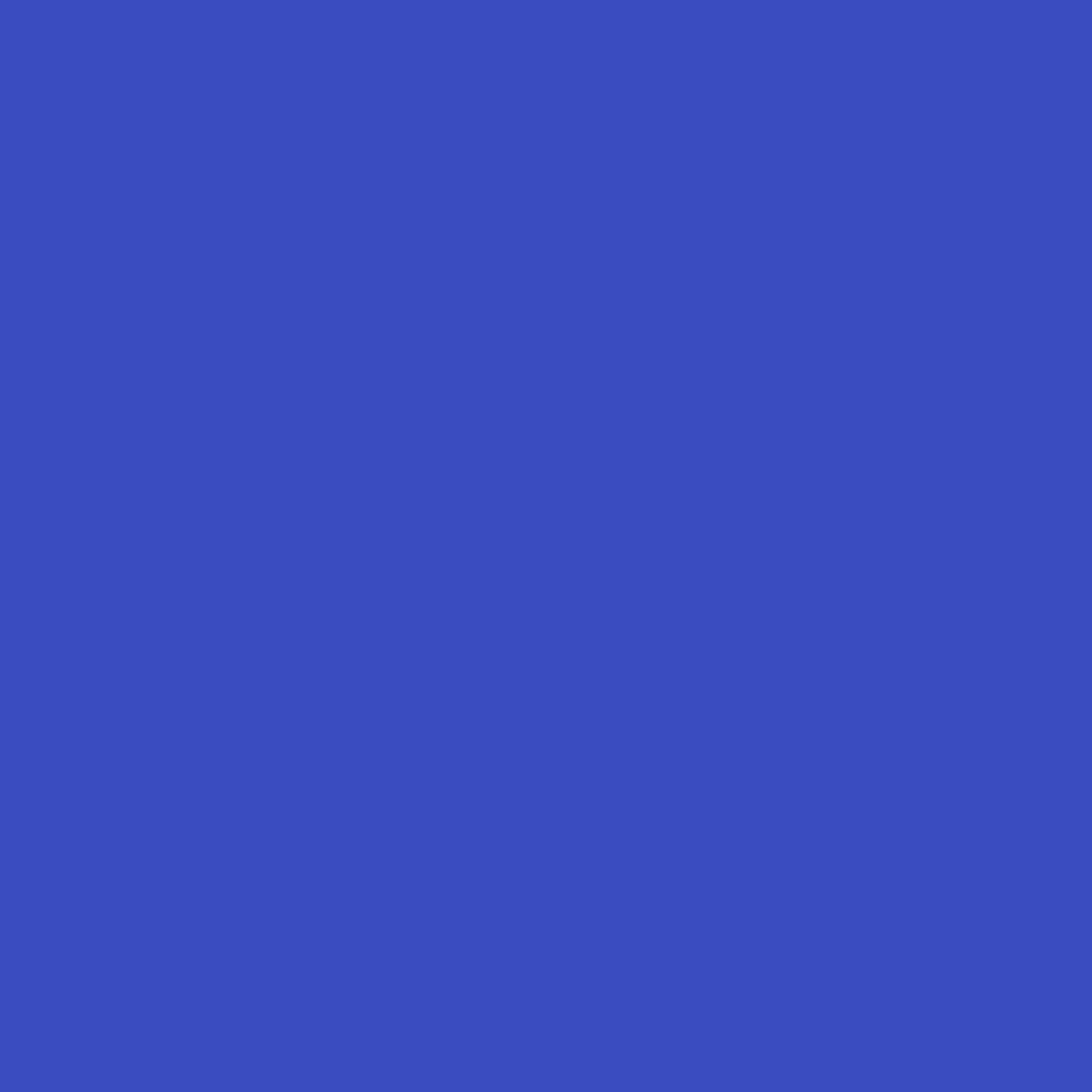}{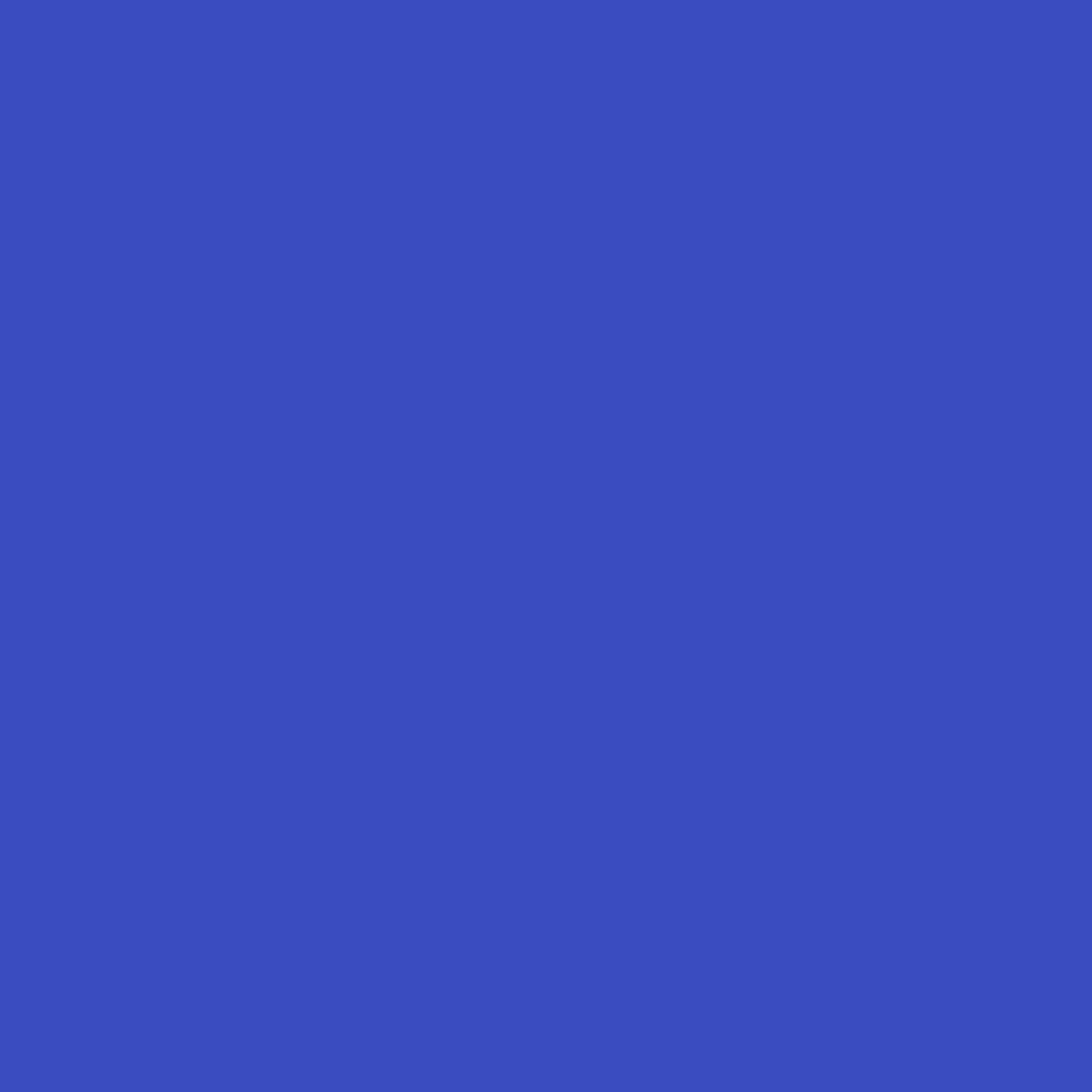} &
    \onebyfour{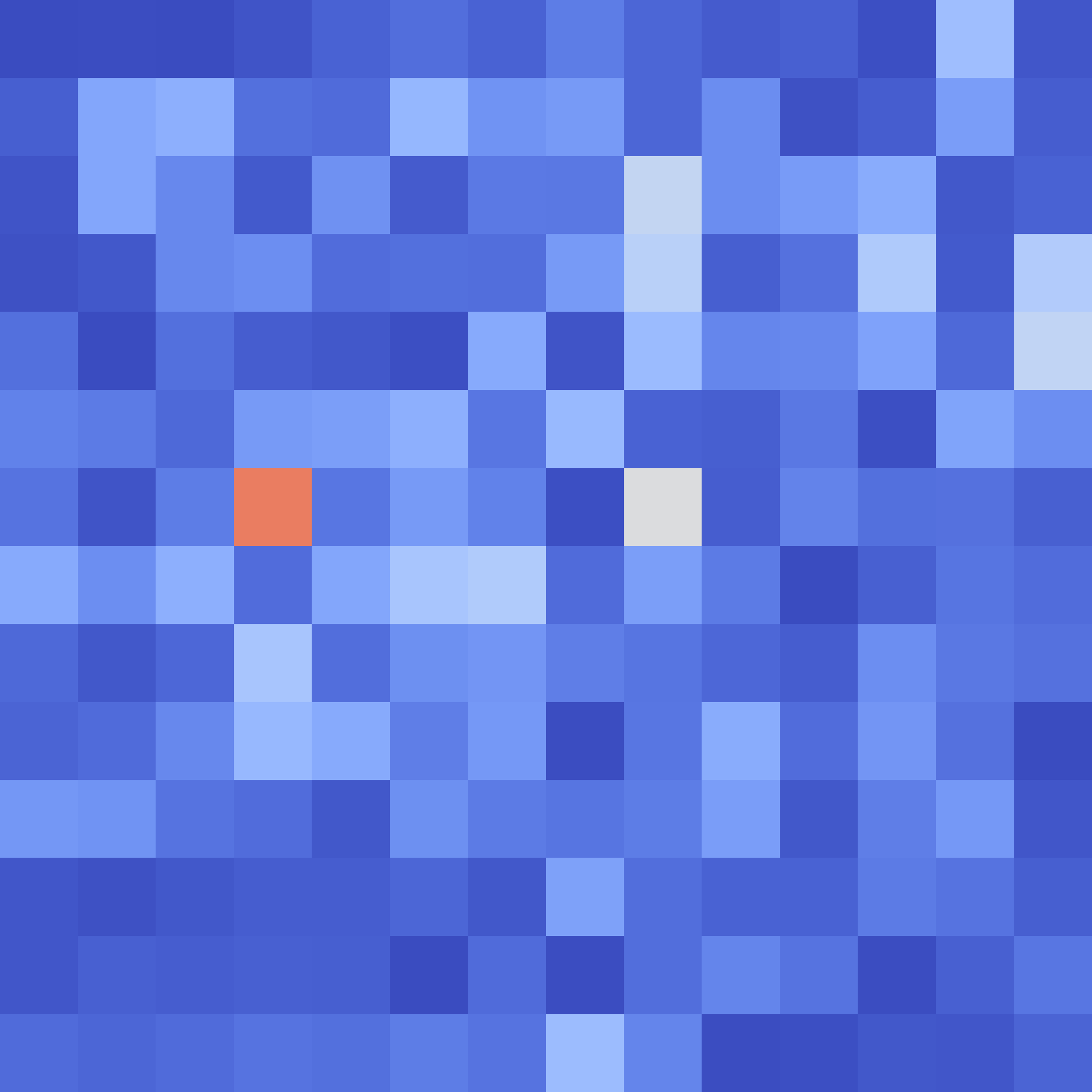}{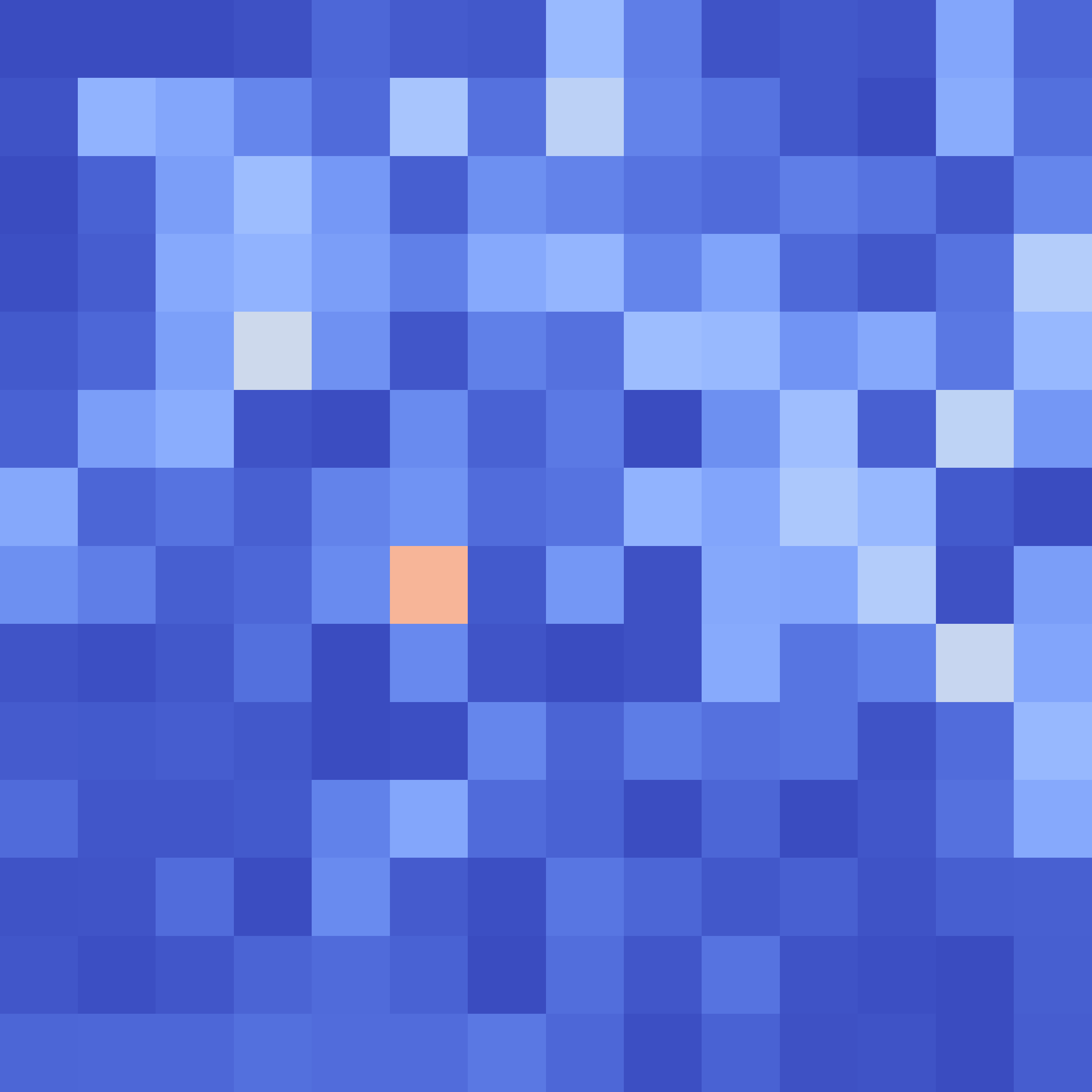}{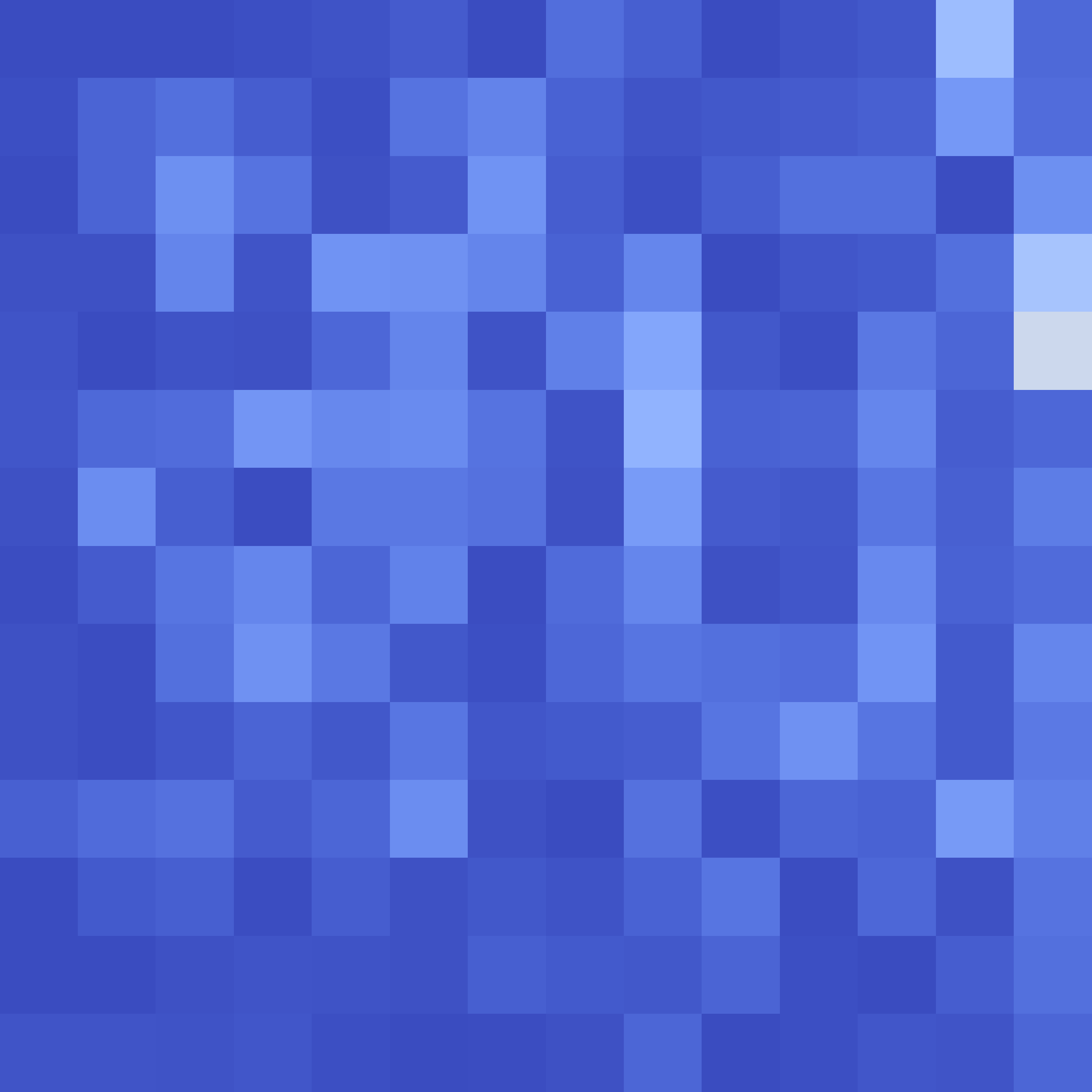}{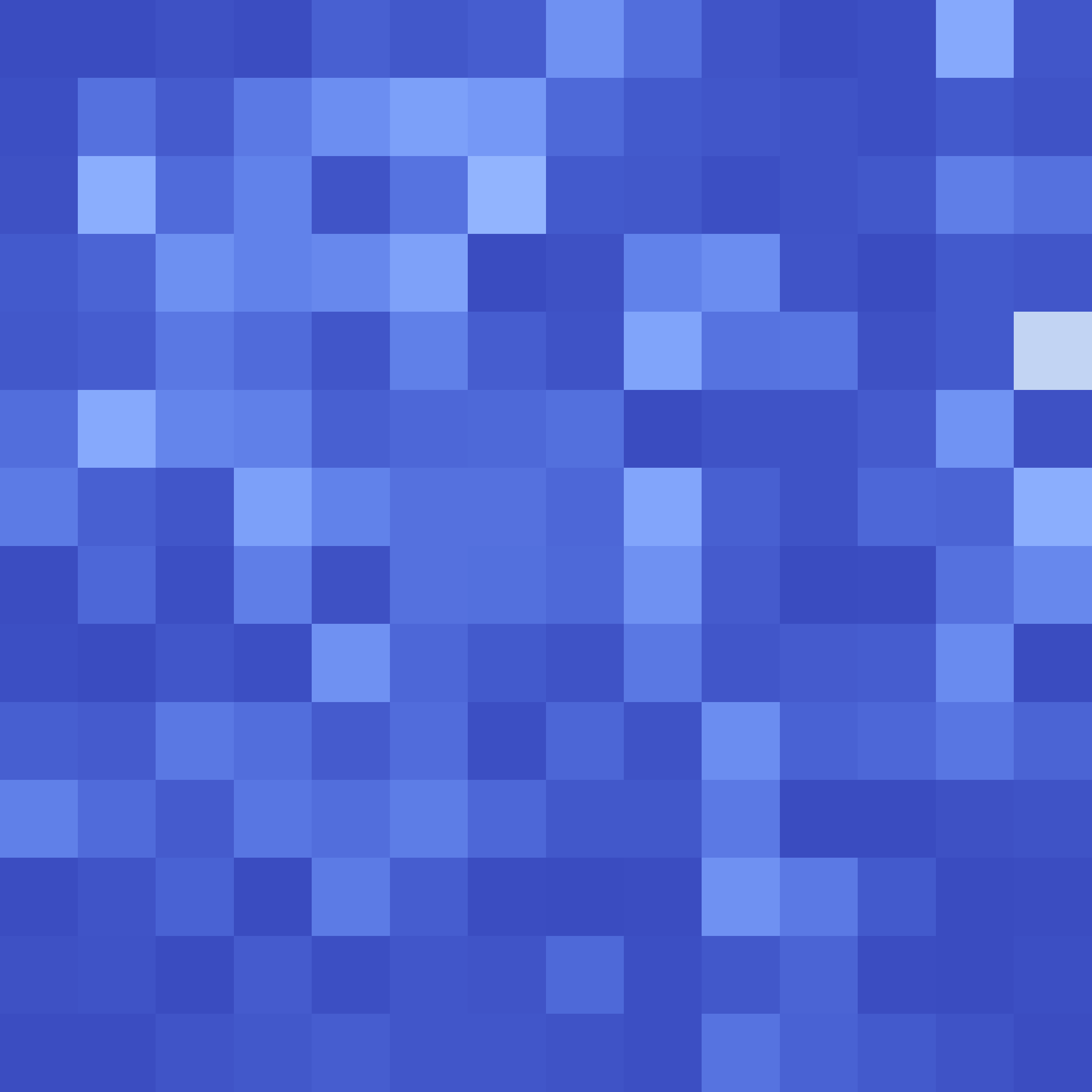} &
    \onebyfour{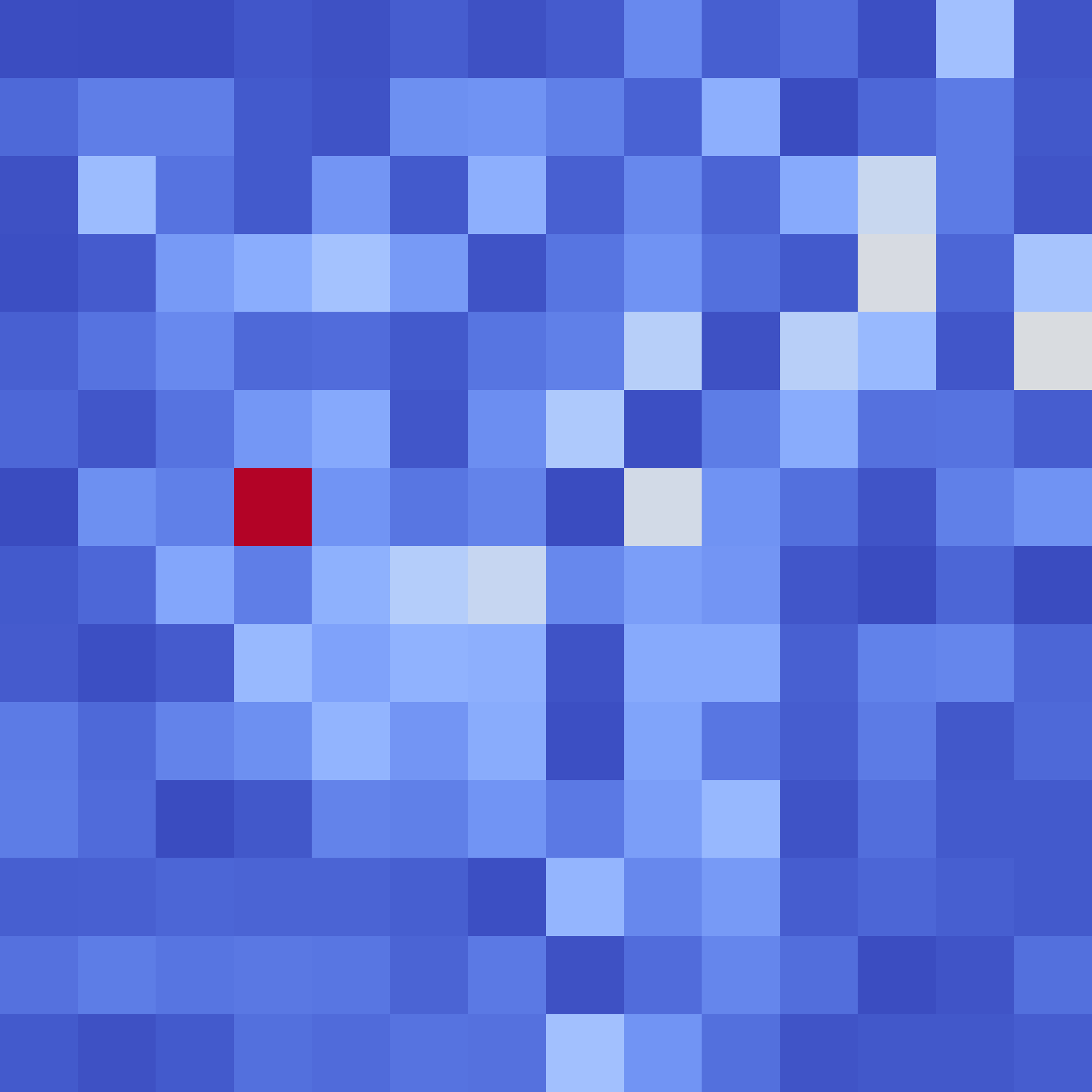}{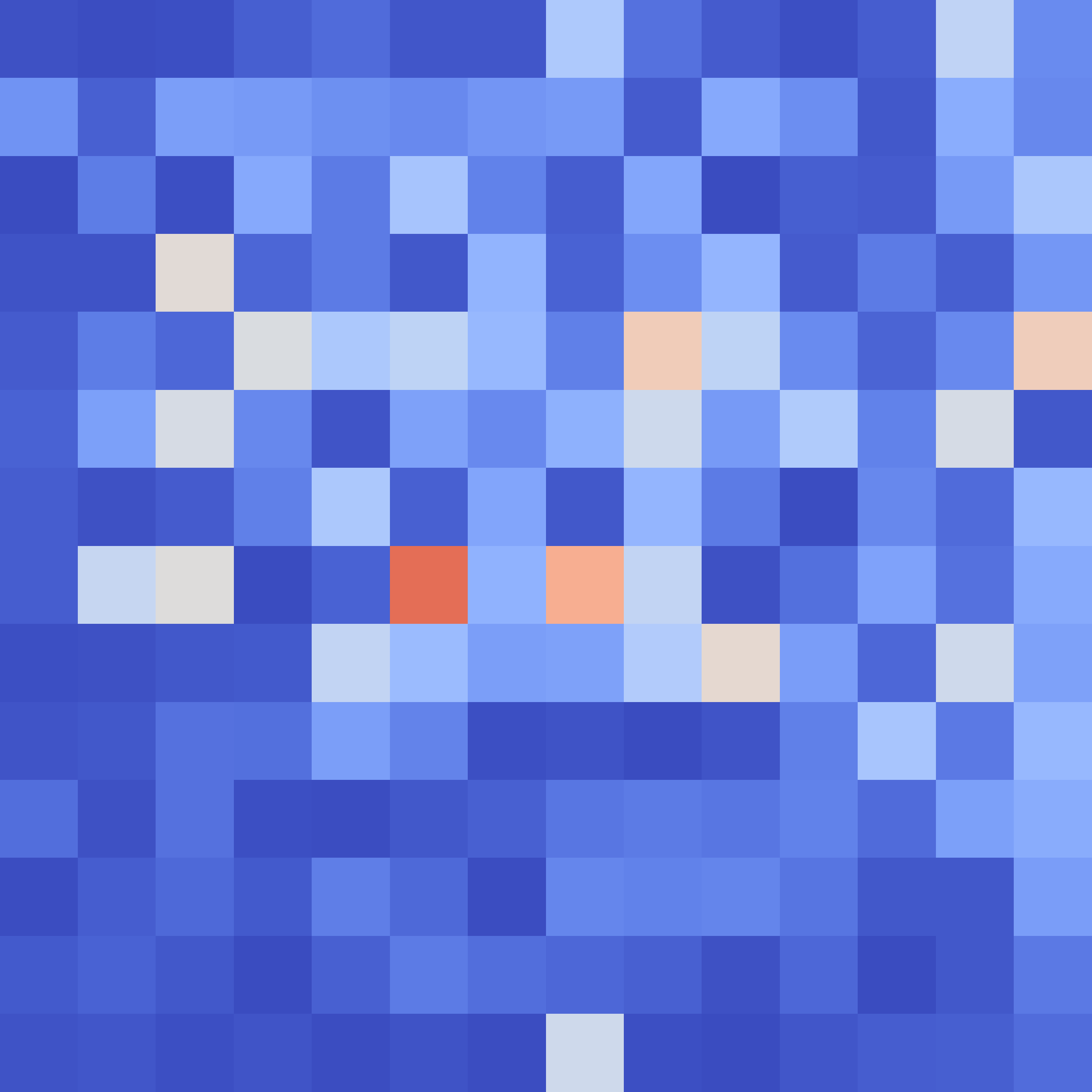}{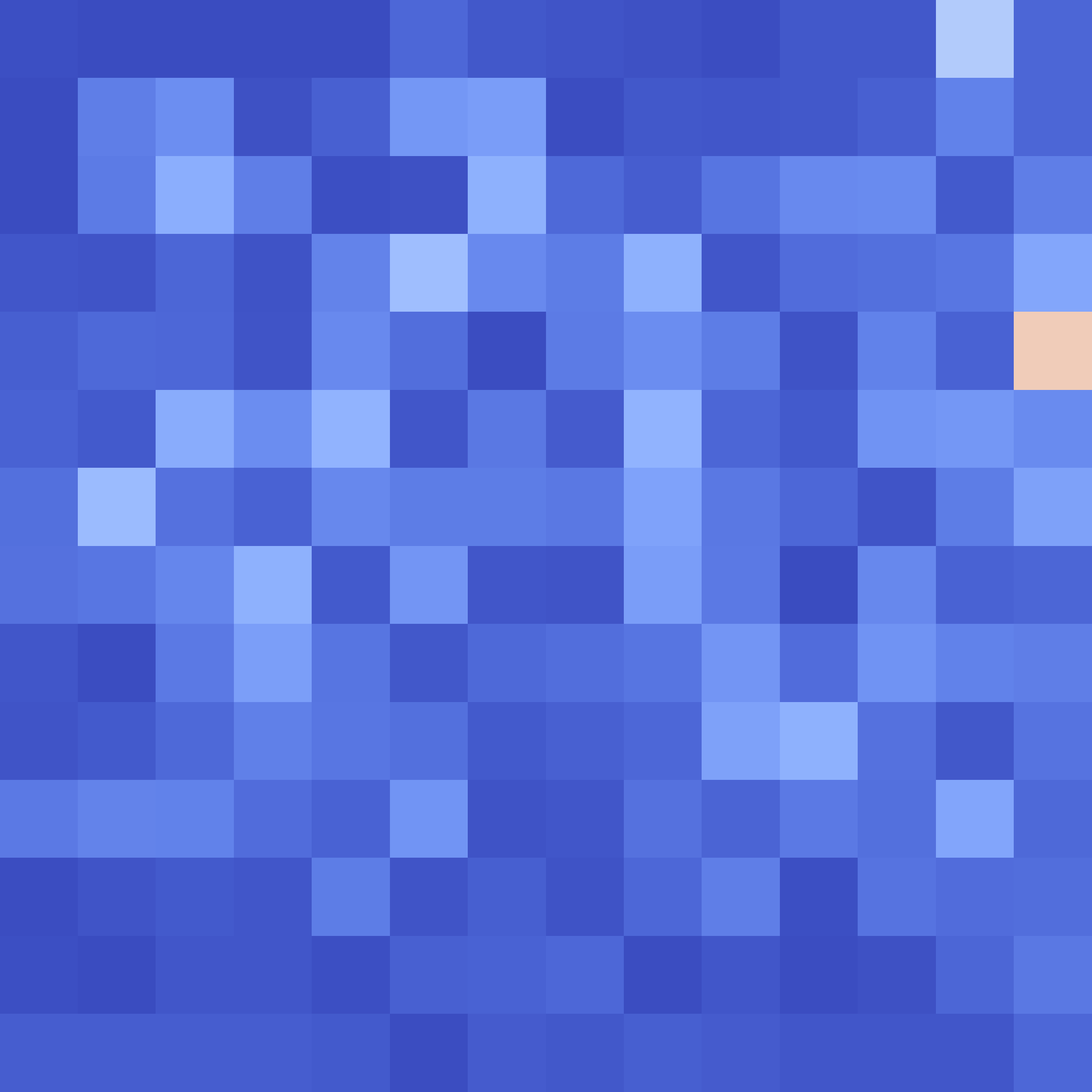}{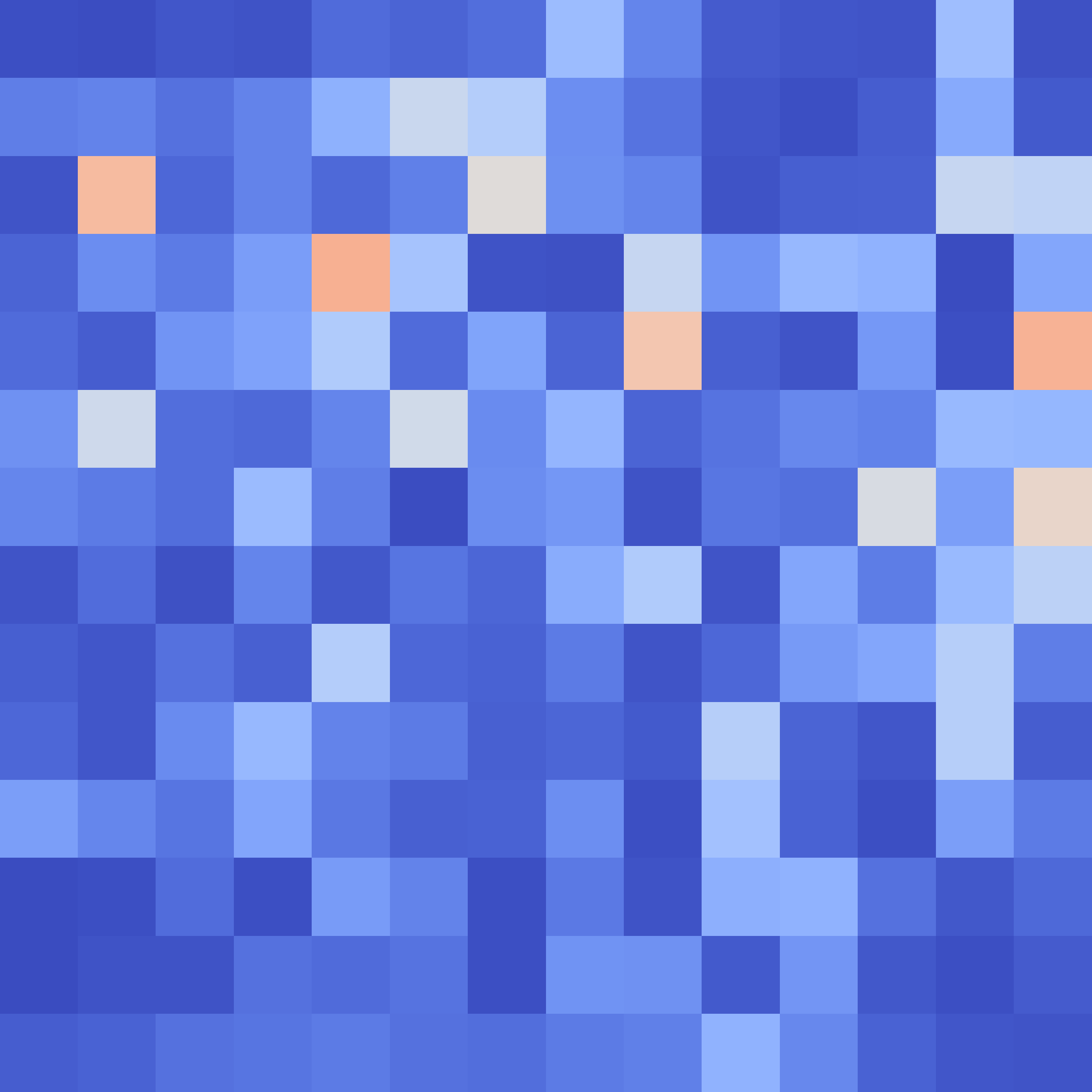} &
    \onebyfour{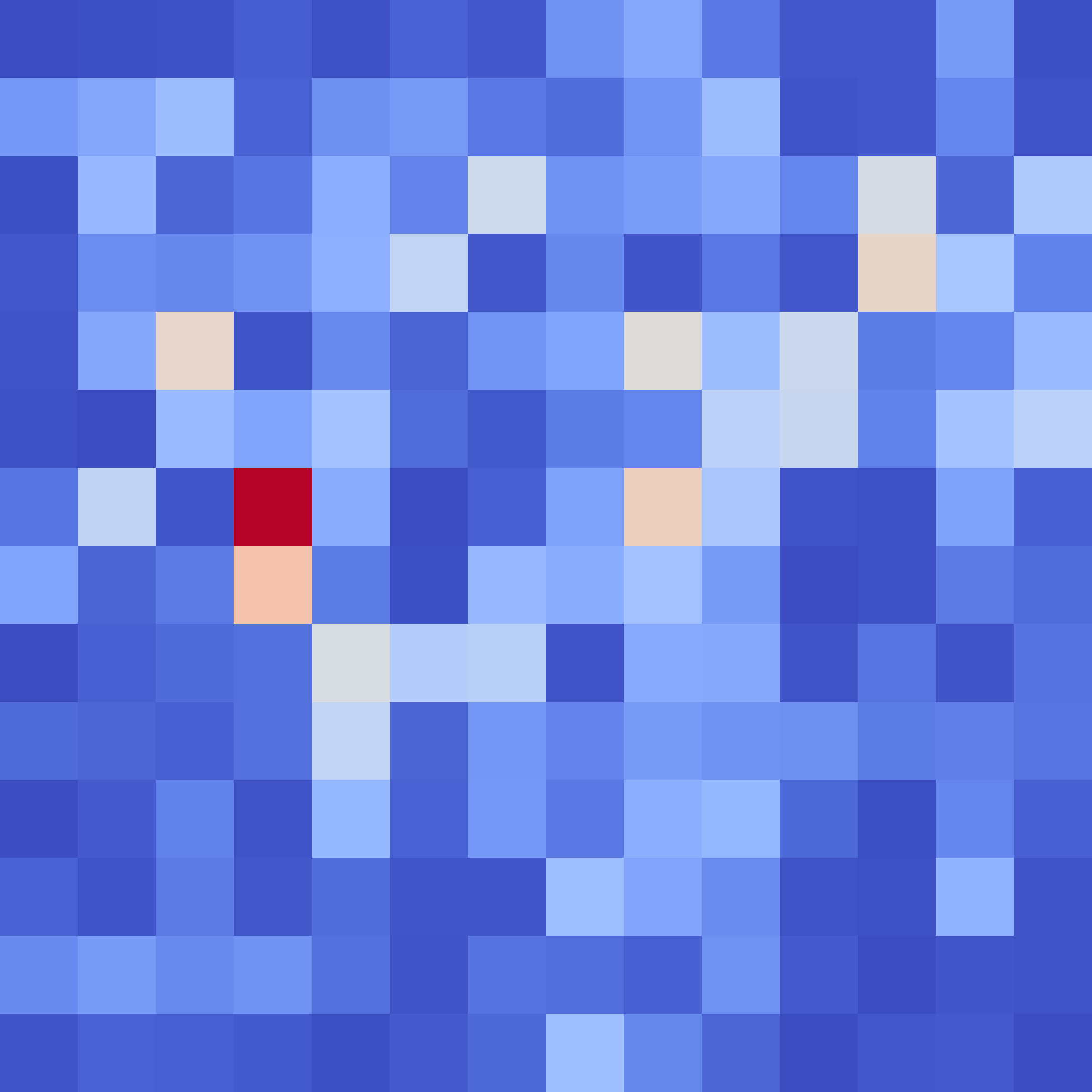}{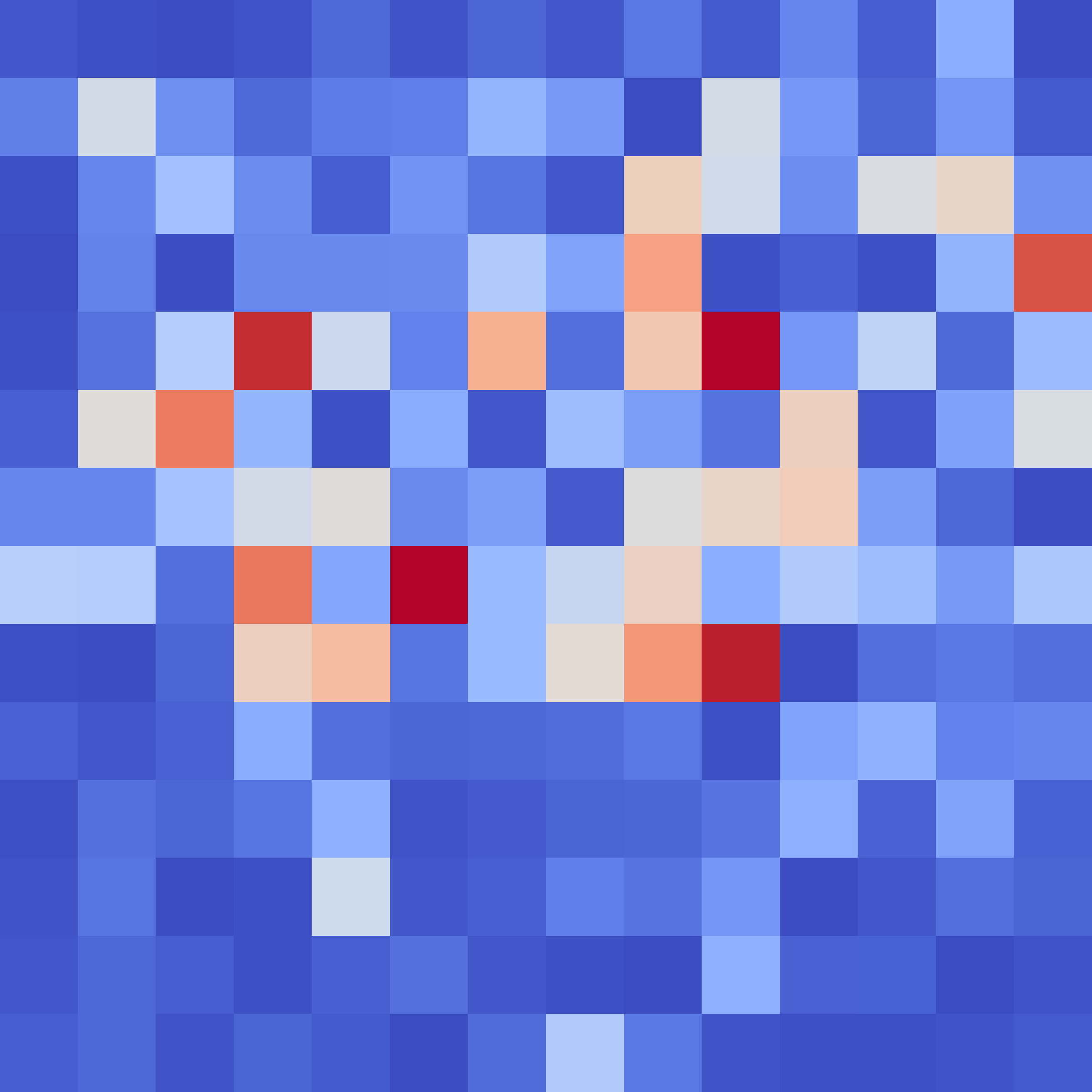}{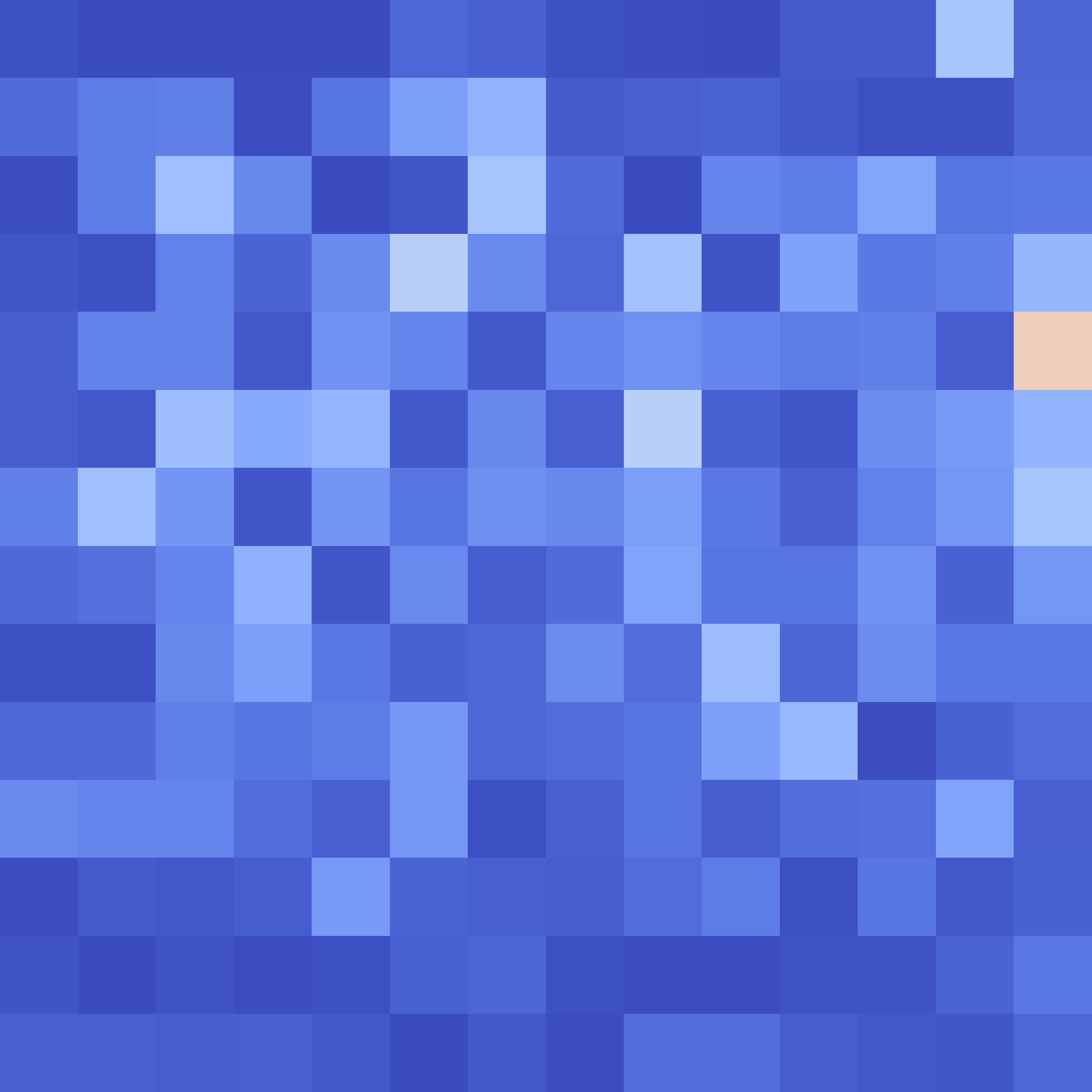}{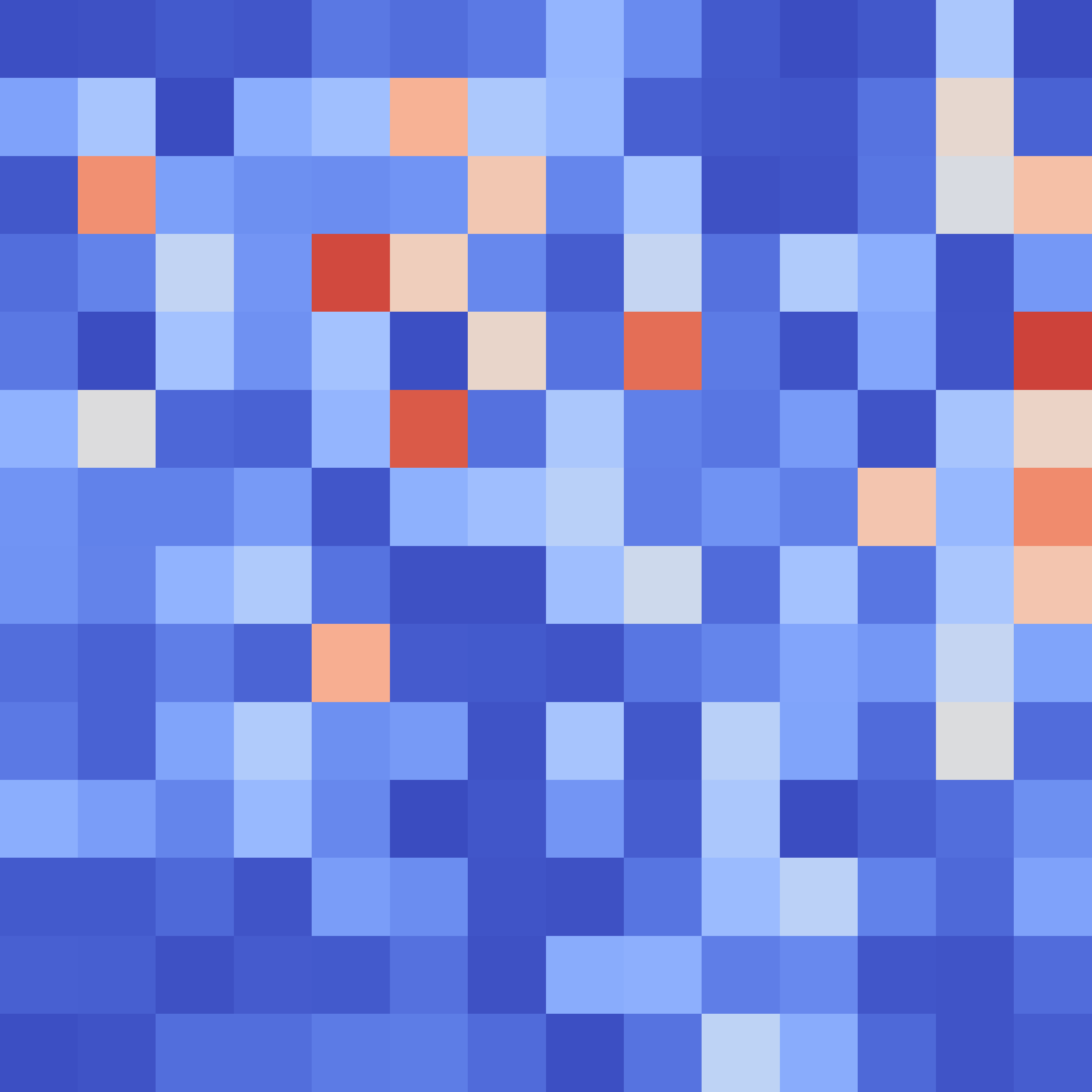} &
    \onebyfour{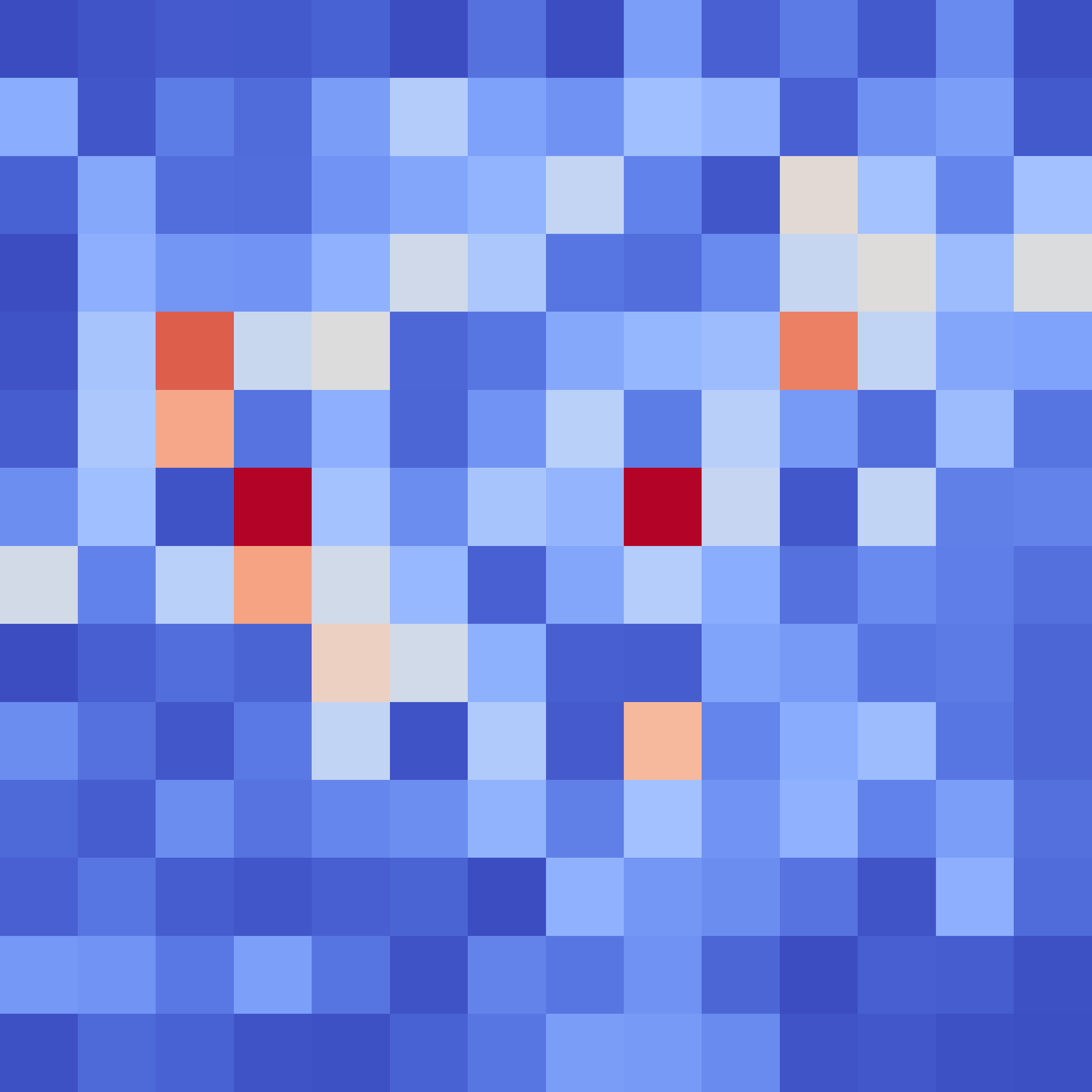}{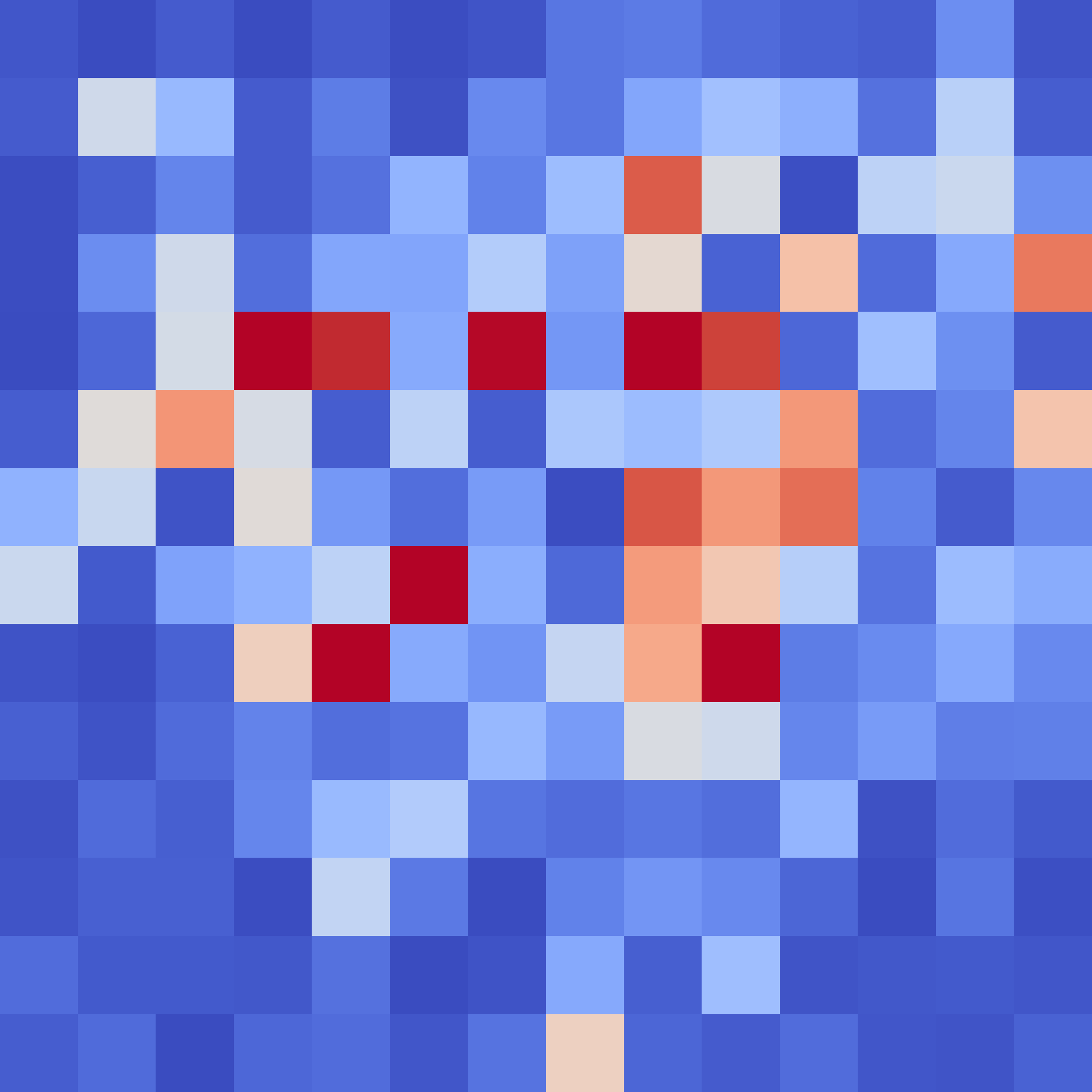}{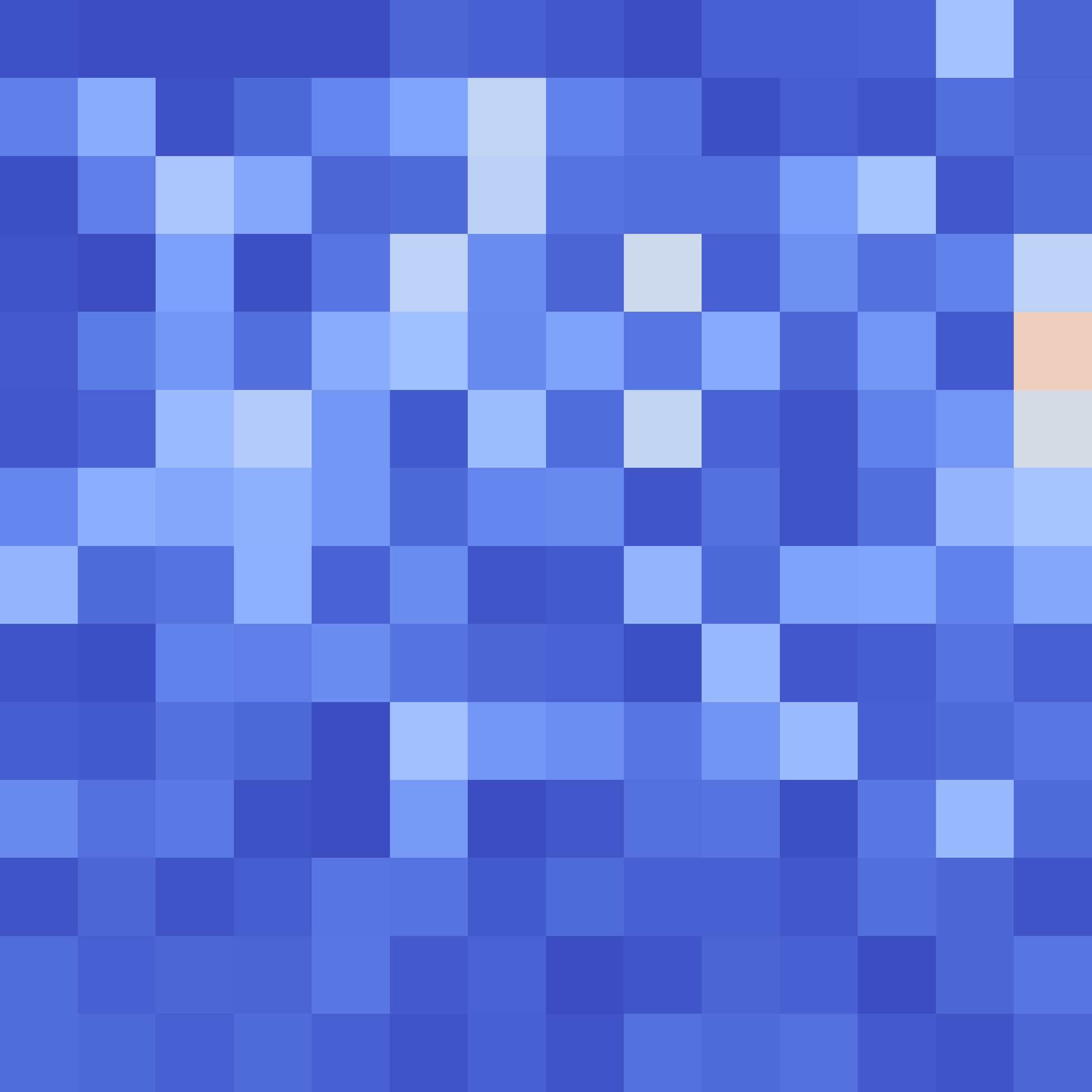}{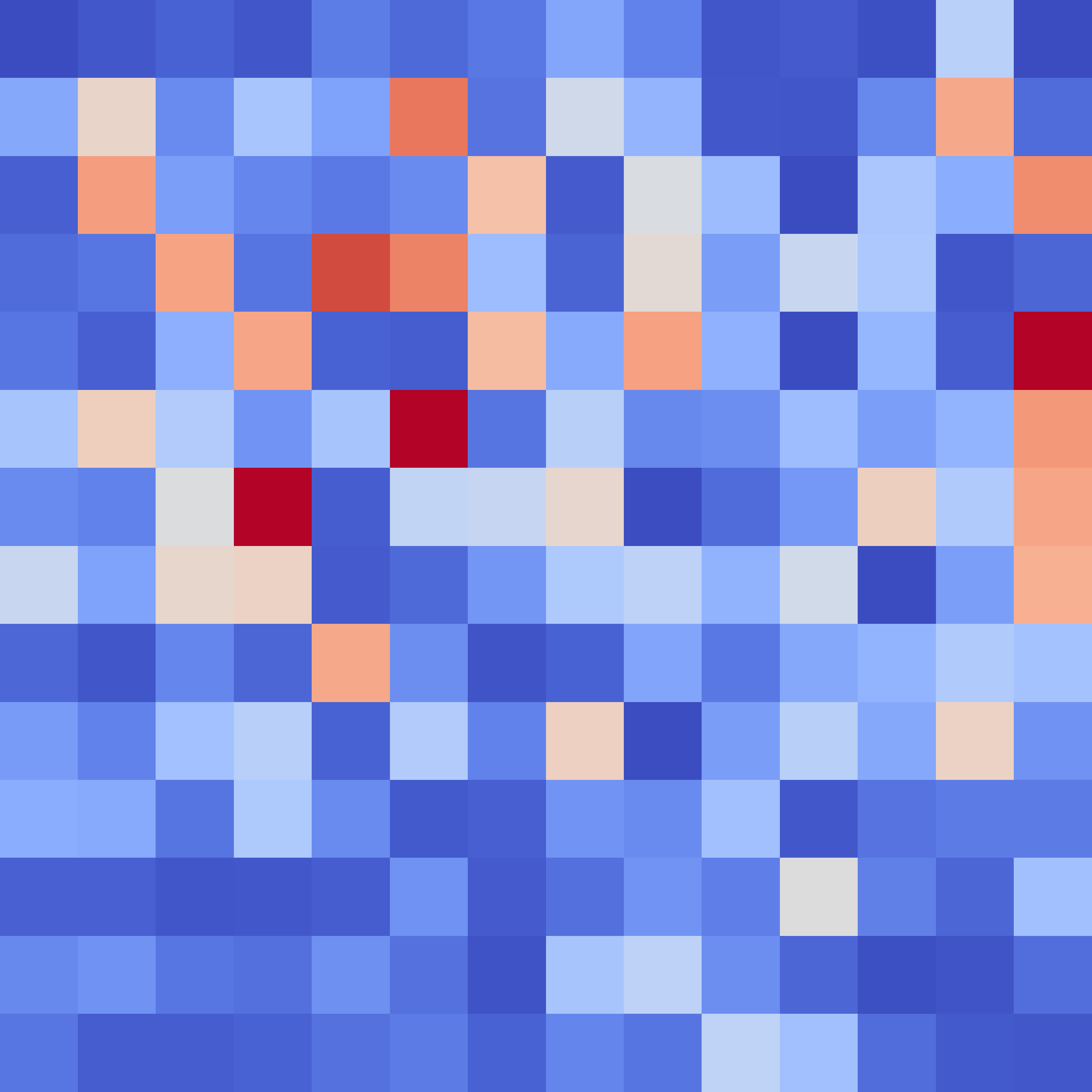}
\end{tabular}
}%
    \vspace{2pt}

\end{minipage}

\begin{minipage}{0.073\textwidth}
\hspace{1cm}
\end{minipage}
\begin{minipage}{0.927\textwidth}
    \vspace{2pt}
    \noindent\resizebox{\textwidth}{!}{%
    \begin{tikzpicture}[x=1\textwidth/13]
        \shade[left color={rgb,255:red,59;green,76;blue,192},
               right color=white]
               (0.61,-0.10) rectangle (7.00,-0.00);
        \shade[left color=white,
               right color={rgb,255:red,180;green,4;blue,38}]
               (7.20,-0.10) rectangle (13.4,-0.00);
        \node[anchor=south] at (1.12,-0.6) {\small Low error};
        \node[anchor=south] at (12.95,-0.645) {\small High error};
    \end{tikzpicture}
    }%
\end{minipage}
\vspace{-0.4cm}
    \captionof{figure}{{Visualization of the ViT~\cite{dosovitskiy2020image} weights with our soft equivariance layer (\wrt $90^\circ$ rotation) under different softness levels, along with the corresponding extracted features and the equivariance errors. Our tunable design allows the layers' weights to transition smoothly from perfectly equivariant to fully non-equivariant behavior in a controlled manner.
    }
    }
    \label{fig:teaser}
\begin{textblock*}{5cm}(1.8cm,9.1cm) %
  {\rotatebox{90}{\small Features}}
\end{textblock*}

%% file: src/abs.tex
Equivariance is a fundamental property in computer vision models, yet strict equivariance is rarely satisfied in real-world data, which can limit a model’s performance. Controlling the degree of equivariance is therefore desirable. We propose a general framework for constructing soft equivariant models by projecting the model weights into a designed subspace. The method applies to any pre-trained architecture and provides theoretical bounds on the induced equivariance error. Empirically, we demonstrate the effectiveness of our method across multiple pre-trained backbones, including ViT and ResNet, for image classification, semantic segmentation, and human-trajectory prediction. Notably, our approach improves the performance while simultaneously reducing equivariance error on the competitive ImageNet benchmark.

%% file: src/intro.tex
\section{Introduction}
\label{sec:intro}

A model is equivariant to a transformation if applying that transformation to the input results in a predictable transformation at the output. Consider image segmentation:  when an object in an image is shifted, the predicted mask is expected to shift by the same amount; this is known as shift equivariance. Although designing models with built-in equivariance has been well studied and shown to be effective in various applications~\cite{schneuing2024structure,cohen2016group,kipf2017semi,qi2017pointnet,worrall2019deep,van2020mdp,rahman2024truly,rahman2025local}, such architectures remain uncommon in mainstream vision systems. In practice, real-world data only approximately satisfies equivariance, and strictly enforcing it can reduce a model’s expressiveness.

This led to the development of soft equivariant models, \ie, models that are only approximately equivariant. Common approaches include augmentation~\cite{Simonyan15,szegedy2015going,benton2020learning} and regularization-based methods~\cite{simard1991tangent,kim2023regularizing,elhag2024relaxed}. However, these techniques do not offer guarantees on a model’s equivariance properties after training. Another direction~\cite{finzi2021residual,romero2021learning,van2022relaxing,van2023learning} achieves soft equivariance by adding non-equivariant components into equivariant models, providing a way to trade off between expressiveness and equivariance. Nonetheless, these methods still lack guarantees on the resulting equivariance and rely on specialized architectural designs that cannot be easily adapted from off-the-shelf models.

To address these challenges, we propose to construct soft-equivariant models through a generalized notion of ``blurring'' filters, which can be applied to any pre-trained model. This is inspired by the special case of shift-invariance in convolutional neural networks (CNNs) by~\citet{zhang2019making}, where anti-aliasing (blurring) filters are used to make CNNs more invariant. Our approach extends this idea beyond shift equivariance to other groups and further provides a bound on the equivariance error. This allows us to systematically tune the trade-off between equivariance and expressiveness in a principled manner; see illustration in~\figref{fig:teaser}.

In our experiments, we first demonstrate the tunability of the proposed soft equivariant models on small-scale image classification. We then incorporate the proposed layer into various pre-trained backbones, including ViT~\cite{dosovitskiy2020image}, DINOv2~\cite{oquab2023dinov2}, ResNet~\cite{he2016deep}, and Segformer~\cite{xie2021segformer} for image classification (CIFAR10/100~\cite{krizhevsky2009learning}, and ImageNet~\cite{deng2009imagenet}) and segmentation task (PASCAL VOC~\cite{everingham2010pascal}). We demonstrate that utilizing our soft equivariance layer further improves the model's performance and reduces equivariance errors. Finally, we go beyond image tasks and evaluate our layers on trajectory prediction~\cite{giuliari2021transformer} and a synthetic $O(5)$-invariant regression problem~\cite{finzi2021practical}.
{\bf\noindent Our main contributions:}
\begin{itemize}
    \item We introduce a novel framework for constructing soft equivariant layers by restricting the parameters via projections, applicable to any pre-trained model.
    \item We derive bounds on the equivariance error, %
    which guides the design of the tunable soft equivariant layers, allowing a controllable expressiveness-equivariance trade-off.
    \item Extensive experiments on three applications (classification, segmentation, and trajectory prediction) and four backbones demonstrate the practicality and effectiveness of the proposed approach.
\end{itemize}

%% file: src/rel.tex
\section{Related Work}

{\bf \noindent Group equivariant architectures.} Early work focused on group convolutions~\cite{cohen2016steerable, cohen2016group, bekkers2018roto}, while subsequent research extended equivariance to broader architectures, including transformers~\cite{thomas2018tensor, kundu2024steerable,bokman2025flopping} and graph neural networks~\cite{maron2018invariant, yeh2019diverse, liu2020pic, morris22a, liao2023equiformer, du2023new}. Later developments generalized equivariance beyond rotation and translation~\cite{zhang2019making,chaman2021truly,rojas2022learnable,rojas2024making} to include permutations~\cite{ravanbakhsh_sets, zaheer2017deep, hartford2018deep}, scaling~\cite{rahman2024pretraining, sosnovik2019scale}, and reflections~\cite{yeh2019chirality}. Recent works have explored ways to finetune pre-trained vision models to be equivariant to specific transformations~\cite{kaba2023equivariance,basu2023equi}.
Despite strong theoretical foundations, strict equivariance assumptions are often misaligned with real-world data and tasks, leading to suboptimal performance~\cite{wang2023general}.

\myparagraph{Soft equivariance.} Several works address misalignment by relaxing strict equivariance constraints. One line of research softens the rigid structure of group convolutions by introducing additional parameters or input-dependent modulation of the convolution operation~\cite{wang2022approximately, romero2021learning, veefkind2024probabilistic, van2022relaxing, samudre2024symmetry}. Another line employs loss-based regularization to encourage approximate equivariance~\cite{elhag2024relaxed, kim2023regularizing}. Residual mixing between equivariant and non-equivariant layers has also been explored, where the mixing weights are learned from data~\cite{finzi2021residual, van2023learning, van2022relaxing}. 

Recently, there have been works aiming to learn the underlying group representations directly from data~\cite{mcneela2023almost}, relating to broader efforts on discovering symmetries in data~\cite{yeh2022equivariance}. These techniques have been adopted to various applications~\cite{wu2024r2det, hofgard2024relaxed, jayanth2024eqnio}. However, these approaches are dependent on group equivariant architectures ~\cite{wang2022approximately, romero2021learning, veefkind2024probabilistic, van2022relaxing} or rely on kernels defined directly on the symmetry group \cite{samudre2024symmetry},  thus limiting their applicability to modern large-scale foundation models. In contrast, our framework is architecture-agnostic. While our approach shares conceptual similarities with \citet{finzi2021practical}, it generalizes the idea beyond exact equivariance, integrates seamlessly with modern pre-trained vision models, and provides explicit control over the level of equivariance.

{\noindent\bf Signal processing (SP).}
Traditional signal processing establishes a tight connection between low-pass filters, bandlimited subspaces, and shift-invariance: a bandlimited subspace remains bandlimited under shifts~\cite{vetterli2014foundations}. Equivalently, a low-pass (anti-aliasing) filter can be interpreted as a \emph{projection operator} onto a shift-invariant bandlimited space. This projection viewpoint has been generalized to graph signal processing via the graph Laplacian~\cite{chen2015discrete,chen2015sampling}, extended to arbitrary discrete groups~\cite{rahman2025group}, and recently adapted to image generative models~\cite{zhou2025alias}.

%% file: src/prelim.tex
\section{Preliminaries}
\label{sec:prelim}
For readers needing a refresher on the concept of groups, we provide a review in Appendix~\secref{supp:math}. Here, we will only discuss the most essential background information. 

{\bf \noindent A \emph{group}} $G$ is a set with a binary operation that is closed, associative, has an identity element $e$, and every element has an inverse. 
The \textit{group representation} $\rho : G \to \mathrm{GL}(V)$ maps group elements to linear transformations on a vector space $V$, which describes the \textit{action} of the group on $V$. 

\myparagraph{Lie algebra and Lie group.}
A \emph{Lie group} $G$ is a smooth manifold whose multiplication and inversion maps are smooth. 
The associated \emph{Lie algebra} $\mathfrak{g}$ %
is the tangent space at the identity, equipped with the Lie bracket $[\cdot,\cdot]: \mathfrak{g} \times \mathfrak{g} \to \mathfrak{g}$ (bilinear, antisymmetric, and satisfying the Jacobi identity).
Let $\{A_1, A_2, \ldots, A_N\}$ denote a basis of $\mathfrak{g}$, then any element $A \in \mathfrak{g}$ can be written as $A = \sum_{i=1}^N a_i A_i$ with real coefficients $a_i$. 

The \emph{exponential map} $\exp: \mathfrak{g} \to G$ connects the algebra to the group. For a matrix Lie group and $A \in \mathfrak{g}$, this is given by the matrix exponential
\begin{equation}
\exp(tA) = \sum_{n=0}^\infty \frac{(tA)^n}{n!},
\end{equation}
where the real parameter $t \in \R$ controls the magnitude.

A \emph{group representation} $\rho$ induces an 
\emph{Lie algebra representation} 
$d\rho : \mathfrak{g} \to \mathrm{End}(V)$ by differentiation at the identity:
\begin{equation}
d\rho(A) = \left. \frac{d}{dt} \rho(\exp(tA)) \right|_{t=0}, 
\qquad A \in \mathfrak{g}.
\end{equation}
Here, $d\rho(A)$ is the \emph{infinitesimal generator} describing the first-order derivative action %
near the identity on $V$. 

\myparagraph{Generalized Taylor expansion.}
The Taylor approximation of a smooth function $h: \mathbb{R} \to \mathbb{R}$ around $x_0=0$ is given as $h(x) = h(0) + h'(0)x + O(x^2)$. This can be extended to functions on a compact, connected Lie group $f: G \to \mathbb{R}$. Any $g \in G$ can be expressed as $g = \exp(A)$ for $A \in \mathfrak{g}$. The first-order Taylor approximation of $f$ around the identity element $e$, $f(\exp(\sum_{i=1}^k t_i A_i))$ is:
\bea
\label{eq:lie_taylor}
f(e) + \sum_{i=1}^k t_i \left.\frac{\partial f(\exp(\sum_i t_iA_i))}{\partial t_i}\right|_{t=0} 
	+ O(\|A\|_{\mathfrak{g}}^2),
\eea
where $\|\cdot\|_{\mathfrak{g}}$ denotes the norm on the Lie algebra.

\myparagraph{Equivariance.}
A function $F: \gX \rightarrow \gY$ is equivariant with respect to a group $G$ if 
\begin{equation}\label{eq:equivariance}
F(\rho_\gX(g)\vx) = \rho_\gY(g)F(\vx) \quad \forall g \in G,\; \vx \in \gX,
\end{equation}
where $\rho_\gY$ and $\rho_\gX$ are representations (group actions) of $G$ on $\gX$ and $\gY$, respectively. When $\rho_\gY(g)= \rho_\gY(e)= \rmI \;\;\forall g \in G$, then this is a special case called invariance.

%% file: src/method.tex
\section{Towards Soft Equivariant Networks}
\label{sec:method}
Soft equivariance can be viewed as a relaxation of the equality constraints (\equref{eq:equivariance}) into inequality constraints, \eg, the amount of violation measured in norm difference (\textit{a.k.a.} equivariance error):
\bea\| F( \rho_\gX(g) \vx) - \rho_\gY(g)  F(\vx) \| \leq \delta.
\eea
This formulation has been considered in prior work~\cite{mcneela2023almost,manolache2025learning}. However, such a definition is sensitive to the scale of the output $F(\vx)$, making it difficult to interpret the significance of the equivariance error $\delta$. Hence, we propose a relative notion of soft equivariance as follows:
\vspace{-0.2cm}
\begin{definition}[$\eta$-Soft Equivariant]
A function $F$ is $\eta$-soft equivariant with respect to a group $G$ if it satisfies:
\begin{equation}\label{eq:soft_eq}
    \frac{\| F( \rho_\gX(g)  \vx) - \rho_\gY(g)  F(\vx) \|}{\|\mJ_{F}(\vx)\|_{\tt F}\|\vx\|} \leq \eta, \quad \forall g \in G, \vx \in X.
\end{equation}
Here, $\mJ_{F}(\vx)$ is the Jacobian of $F$ at $\vx$, and $\eta \in \R^+$ is the soft equivariance constant.
When $\rho_\gY$ is the identity, then we say $F$ is $\eta$-soft invariant.
\end{definition} 
\vspace{-0.25cm}
Intuitively, the Jacobian $\norm{\mJ_{F}}$ represents the scaling locally around $\vx$. Hence,~\equref{eq:soft_eq} is measuring equivariance relative to $F$'s own local output variation, making $\eta$ easier to interpret. To avoid degeneracy, we assume $\|\mJ_{F}(\vx)\|_{\tt F} > 0$ and $\|\vx\| > 0$ for all $\vx$ in the domain (see \secref{sup:stability} for details). In the following, we will develop soft equivariant linear layers for both continuous and discrete groups. This is followed by practical considerations and guidance on incorporating them into pre-trained models.

\subsection{Soft equivariance for continuous groups}
\label{sec:continuous}
We describe our proposed soft invariant/equivariance linear layers for continuous groups. For readability, we present the layer for a single generator and a single output channel.

\myparagraph{Soft invariant fully connected layer.}  
A fully connected layer is defined as $\vy = F_{\tt FC}(\vx; \vw) \triangleq \vw^\top\vx$; here, we consider $\vx, \vw \in \R^d$.
To achieve $\eta$-soft invariance, we impose a structure on $\vw$ via a ``blurring''/ projection operator $\mB_{\tt inv} \in \R^{d\times d}$, \ie, 
\bea
\vy \triangleq (\mB_{\tt inv}\theta)^\top\vx, \text{ with } \vw \triangleq \mB_{\tt inv}\theta
\eea
where $\theta \in \R^d$ is the learnable parameter of the layer.
This projection operation is derived from the Lie algebra representation for a given group $G$. Let's denote the singular value decomposition (SVD) of the Lie algebra representation as 
\bea 
\label{eqn:inv_svd}
\mA \triangleq 
d\rho_{\gX}(A) = \mU \Sigma \mV^\top,
\eea  
where singular values are sorted, \ie, $0 \le \sigma_1 \le \sigma_2 \hdots$ with corresponding left and right singular vectors $\vu_i$ and $\vv_i$, respectively. We design the projection operator
\bea\label{eq:inv_blur}
\mB_{\tt inv} \triangleq \sum_{i: \sigma_i < b} \vu_i \vu_i^\top,
\eea
where it filters out the components with a cut-off value $b \in \R^+$. Intuitively, this projection operation removes the directions that are highly affected by the group action. 

We now formalize the soft invariance property of the proposed layer in the following claim.
\begin{myclaime}
\begin{restatable}[]{myclaim}{claimone}
\label{clm:soft_inv}
For any compact and connected Lie group $G$ with injective radius $r_G$ and $n_G$ number of generators, the function $F_{\tt FC}(\vx, \mB_{\tt inv} \theta)$ is $\eta_b$-soft invariant, \ie,
\bea 
\frac{\|(\mB_{\tt inv} \theta)^\top \vx - (\mB_{\tt inv} \theta)^\top \rho_\gX(g)\vx\|}{\|\mJ_{F_{\tt FC}(\cdot;\vw)}(\vx)\|_{\tt F} \|\vx \|} \le \eta_b,  \forall g \in G %
\eea 
where $\eta_b = b \sqrt{n_G} r_G + \varepsilon_G$, $b \in \R^+$ is the cut-off value for the projection operator 
, and $\varepsilon_G$ is the residual from the first-order Taylor approximation.
\end{restatable}
\end{myclaime}
\vspace{-0.15cm}
\begin{proof}  
 We use the Taylor expansion in~\equref{eq:lie_taylor} to express the invariance error in terms of the Lie algebra representation $d\rho$. Next, we relate the contribution of each singular vector of $d\rho(A)$ to the overall invariance error. Finally, we show that $\mB_{\tt inv}$ bounds this error by constraining $\vw$ to lie within a subspace spanned by a selected subset of singular vectors. See proof in Appendix~\ref{supp:soft_inv}.
  \vspace{-0.15cm}
\end{proof}
\clmref{clm:soft_inv} states that the invariance error is dependent on the cut-off value $b$ and other group properties $r_G$ and $n_G$ characterizing the group's size and complexity, \eg, for continuous $2D$ rotation, $r_G=\pi$ and $n_G=1$.

\myparagraph{\color{cvprblue}\textit{ Remarks.}} In the case of shift invariance, our construction of $\mB_{\tt inv}$ yields a blurring filter whose cut-off frequency is determined by the degree of invariance. While~\citet{zhang2019making} empirically showed that anti-aliasing (blurring filters) improves shift invariance in CNNs, our general framework provides a mathematical justification linking bandlimited signals to shift invariance.

\myparagraph{Multiple generators.} For groups with multiple generators $\{A_i\}_{i=1}^k$, we construct the combined projection operator $\mB_{\tt inv}$ by calculating the left singular vectors of the concatenated generators 
\bea
\mA \triangleq \left[d\rho(A_1)~|~d\rho(A_2)~|~\ldots ~|~d\rho(A_k)\right].
\eea 
The rest of the design remains the same following~\equref{eqn:inv_svd} and~\equref{eq:inv_blur}. See Appendix~\ref{supp:multi_generator} for proof.

\myparagraph{Soft equivariant fully connected layer.} A vector-valued fully connected layer is defined as $\vy = F_{\tt FC}(\vx; \mW) \triangleq \mW\vx$. Here, we consider $\vx \in \R^d$ and $\mW \in \R^{d'\times d}$. Similar to the invariant layer, to achieve $\eta$-soft equivariance, we impose a structure on $\mW$ via a projection operator $\mB_{\tt eq} \in \R^{d \cdot d' \times d \cdot d'}$ as follows:
\bea\label{eq:soft_eq_W}
\text{\tt vec}(\mW) \triangleq \mB_{\tt eq} \theta,
\eea
where $\theta \in \R^{d\cdot d'}$ is the learnable parameter of the layer and $\text{\tt vec}$ is the vectorization operator that stacks the columns of a matrix into a vector. 

Next, the equivariance constraint in~\equref{eq:equivariance} involving the input and output representations can be consolidated using the Kronecker product~\cite{finzi2021practical, horn2012matrix}. Similarly, we design a matrix $\mL$ using the Kronecker product of Lie algebra representations, which quantifies deviation from exact equivariance: %
\bea
\label{eq:equiv_condition_mat}
\mL \triangleq (d\rho_{\gX}(A)^\top \otimes \rmI_{d'} - \rmI_d \otimes d\rho_{\gY}(A)) \in \mathbb{R}^{d\cdot d' \times d\cdot d'}.
\eea
Let the SVD of $\mL$ be denoted as
\bea
\label{eq:equiv_svd}
\mL = \mU^{\tt \mL} \Sigma^{\tt \mL} {\mV^{\tt \mL}}^\top,
\eea
where the singular values are sorted, \ie, $0 \le \sigma_1 \le \sigma_2 \hdots$ with corresponding left and right singular vectors $\vu_i^{\tt L}$ and $\vv_i^{\tt L}$, respectively. We propose the projection operator for equivariance as
\bea
\label{eq:soft_equiv_filter}
\mB_{\tt eq} \triangleq \sum_{i: \sigma_i < b} \vv_i^{\tt L} {\vv_i^{\tt L}}^\top,
\eea
where it filters out the components with a cut-off value $b \in \R^+$.
We now formally state the soft equivariance property of this layer.
\begin{myclaime}
\begin{restatable}[]{myclaim}{claimtwo}
\label{clm:soft_eqv}
For any compact and connected Lie group $G$ with injective radius $r_G$ 
and $n_G$ generators, let $\mW$ be defined as in~\eqref{eq:soft_eq_W}. 
Then $F_{\tt FC}(\vx, \mW)$ is $\eta_b$-soft equivariant, i.e.,
\begin{align}
\frac{\|\mW \rho_{\gX}(g)\vx - \rho_{\gY}(g) \mW \vx\|}
     {\|\mJ_{F_{\tt FC}(\cdot;\mW)}(\vx)\|_{\tt F}\, \|\vx\|} 
  \le \eta_b, \quad \forall\, g \in G,
\end{align}
where $\eta_b = b\sqrt{n_G\, d'}\, r_G + \varepsilon_G$, \, 
$b \in \mathbb{R}^+$ is the cut-off value of the projection operator 
$\mB_{\tt eq}$, $d'$ is the output dimension, $\varepsilon_G$ is the residual from the first-order Taylor approximation.
\end{restatable}
\end{myclaime}
\begin{proof} We use the Taylor expansion in~\equref{eq:lie_taylor} to express the equivariance error in terms of the Lie algebra representation $d\rho$. Using properties of the Kronecker product and singular value decomposition, we separate the contribution of each of the singular vectors in the equivariance error. 
Complete proof in Appendix~\ref{supp:soft_eqv}.
\end{proof}

{\bf\noindent Efficient design of soft equivariant layer.} 
The computational complexity of performing the SVD on matrix $\mL$ in~\equref{eq:equiv_svd} is $O((d \cdot d')^3)$ for a parameter $\theta \in \R^{d \cdot d'}$. While the SVD is precomputed only once per group $G$ before training, it remains computationally expensive for large values of $d \cdot d'$.
To address this, we further propose an alternative design using the Schur decomposition~\cite{horn2012matrix} with time complexity of $O(\max(d,d')^3)$ 
for a group whose Lie algebra representations are normal matrices (commute with conjugate transpose), \eg, $2D$ or $3D$ rotations.

Denote the real Schur decomposition of $d\rho_\gX$ and $d\rho_\gY$ as
\bea
d\rho_\gX = \mU_\gX \mSigma_\gX \mU_\gX^\top \text{ and }
d\rho_\gY = \mU_\gY \mSigma_\gY \mU_\gY^\top.
\eea
As $d\rho_\gX$ and $d\rho_\gY$ are normal matrices, the $\mSigma_\gX$ and $\mSigma_\gY$ are block diagonal, \ie, %
$\mSigma_\gX=\operatorname{diag}(\{\mS_k\}_{k=1}^p)$ and $\mSigma_\gY=\operatorname{diag}(\{\mT_l\}_{l=1}^q)$, where $\{\mS_k\}_{k=1}^p$ and $\{\mT_l\}_{l =1}^q$ are sets of $1\times1$ or $2\times2$ Schur form blocks. %
The $\mU_\gX$ and $\mU_\gY$ are orthogonal matrices.

Let $\theta$ be the learnable parameter and $\Theta \in \R^{d' \times d}:\gX\to\gY $ with ${\tt vec} (\Theta) \triangleq \theta $.  We transform $\Theta$ into the Schur basis:
\bea 
\Theta' = \mU_\gY^\top \Theta \mU_\gX
\eea

\begin{mylemmae}
\begin{restatable}[]{mylemma}{lemmaone}
\label{clm:schur_equiv}
The weight matrix $\Theta$ is equivariant if it satisfies the condition
\bea 
\mSigma_\gY \Theta' - \Theta' \mSigma_\gX = \vzero 
\Longleftrightarrow 
\mT_l \Theta'_{lk} = \Theta'_{lk}\mS_k \quad \forall l,k,
\eea 
where $\Theta'_{lk}$ are blocks of $\Theta'$ corresponding to the blocks of $\mSigma_\gY$ and $\mSigma_\gX$ of dimensions $\dim(\mT_l)\times\dim(\mS_k)$.
\end{restatable}
\end{mylemmae}
\begin{proof}
We apply the change of basis using $\mU_\gX$ and $\mU_\gY$ to the equivariance condition, which results in block diagonalization of the Lie algebra representations. 
Complete proof in Appendix~\ref{supp:schur_equiv}.
\end{proof}
From Lemma~\ref{clm:schur_equiv}, we are motivated to leverage the block-wise diagonal structure in our design. Let the magnitude of the maximum eigenvalue of $\mS_k$ and $\mT_l$ be $\lambda_{\mS_k}$ and $\lambda_{\mT_l}$ respectively. We defined the Schur equivariance projection $\mB_{\tt Schur}$ as $\mW' = \mB_{\tt Schur} \left[\Theta'\right]$ with block-wise operation as
\bea
\label{eq:schur_conditions}
\mW'_{lk} = 
\begin{cases}
\vzero & \text{if~} \mT_l \not \simeq \mS_k,  \lambda_{\mS_k}+\lambda_{\mT_l} > b, \\
{\tt Sym}(\Theta'_{lk})  & \text{if }  \mT_l \simeq  \mS_k, \lambda_{\mS_k}+\lambda_{\mT_l} > b, \\
\Theta'_{lk}  & \text{otherwise}
\end{cases}
\eea 
where $b$ is the cut-off value, $\mT_l \simeq  \mS_k$ implies that $\mT_l$ and $\mS_k$ have common eigenvalues ($\mT_l \not \simeq \mS_k$ otherwise). For any $2 \times 2$ matrix $\Theta'_{lk} = \begin{pmatrix} a & b \\ c & d \end{pmatrix}$, we define %
\bea
{\tt Sym}(\Theta'_{lk}) = \begin{pmatrix} \frac{a+d}{2} & \frac{b-c}{2} \\ -\frac{b-c}{2} &
\frac{a+d}{2}\end{pmatrix}.
\eea 
This form is based on the observation from the condition in Lemma~\ref{clm:schur_equiv} that each block $\Theta'_{lk}$ solves a Sylvester equation. When $\mT_l \not \simeq \mS_k$ only zero matrix satisfies the condition~\cite{horn2012matrix}. When $\mT_l \simeq \mS_k$, by Schur's Lemma \cite{fulton2013representation}, for $1\times1$ blocks,  any unconstrained scalar satisfies the condition. For $2\times2$ blocks, the solutions have the form 
\bea 
\label{eq:2x2_commute}
\begin{pmatrix}\alpha & \beta\\ -\beta & \alpha\end{pmatrix},\quad \alpha,\beta\in\sR,
\eea
\ie, it has the weight symmetric form imposed by ${\tt Sym}(\cdot)$. Our proposed Schur projection eliminates the weight components that break the equivariance condition, depending on the cut-off value $b$. We now state the soft equivariance bound.

\begin{myclaime}
\begin{restatable}[]{myclaim}{claimfour}
\label{clm:schur_filter}
For any Lie group $G$ with normal Lie algebra representations with injective radius $r_G$ 
and $n_G$ generators. Let $\mW = \mU_\gY \mB_{\tt Schur}\left[\Theta'\right] \mU_\gX^\top$, then the function $F_{\tt FC}(\vx, \mW)$ is $\eta_b$-soft equivariant
\begin{align}
\frac{\| \mW \rho_\gX(g) \vx - \rho_\gY(g) \mW \vx \|}{\|\mJ_{F_{\tt FC}(\cdot; \mW)}(\vx)\|_{\tt F}\|\vx\|} \leq \eta_b, \forall g \in G %
\end{align}
with $\eta_b = b \sqrt{n_G} r_G+ \varepsilon_G$, $b$ is the cut-off value of the Schur filter $\mB_{\tt Schur}$ and $\varepsilon_G$ is the residual from the first-order Taylor approximation.
\end{restatable}
\end{myclaime}
\vspace{-0.2cm}
\begin{proof}
Complete proof in Appendix~\ref{supp:schur_filter}. The proof follows the same overall structure as the previous claims.
\end{proof}

\subsection{Soft equivariance for discrete groups} 
The projection design in~\secref{sec:continuous} builds on the Lie group Taylor expansion in~\equref{eq:lie_taylor} and its Lie algebra representation, which do not apply to discrete groups. 
We extend this formulation by introducing a \emph{group forward-difference} operator, the discrete analogue of the Lie-algebra representation, and use it to derive the first-order Taylor expansion for discrete groups.

\myparagraph{Taylor approximation for discrete groups.}
For a finite discrete group $G$ with generating set $\sS$ (reviewed in Appendix~\ref{supp:math}), we defined the forward difference operator $\Delta_s$ as action of function $f: G \to \R$ along generator $s \in \sS$ as
\bea
\Delta_s f(g) = f(sg) - f(g), \forall g \in G.
\eea

Using the forward difference operator, we can state the first-order Taylor approximation for discrete groups as follows:
\begin{mylemmae}
\begin{restatable}[]{mylemma}{lemmatwo}
\label{lemma:forward_diff} 
Let $G$ be a finite discrete group with generating set $\sS = \{s_1, \ldots, s_k\}$, and $f: G \to \R$ be $h$-Lipschitz with respect to the word metric $d_{\sS}$, then the first-order Taylor approximation of $f$ at the identity element $e$ defined as $\hat{f}(g) \triangleq f(e) + \sum_{i=1}^k n_{s_i} \,\Delta_{s_i} f(e)$  
satisfies the following point-wise error bound:
\begin{equation}
    \left| f(g) - \hat{f}(g) \right| \leq 2h \cdot d_{\sS}(e, g),
\end{equation}
where $n_{s_i}$ number of occurrence of $s_i$ in the canonical word representation of $g$.
\end{restatable}
\end{mylemmae}
\begin{proof}
The error $|f(g) - \hat{f}(g)|$ decomposes into two terms via the triangle inequality: the global displacement $|f(g) - f(e)|$ and contribution of each generators $\sum_{i} n_i |\Delta_{s_i} f(e)|$. Each term is then bounded using the $h$-Lipschitz condition. The complete proof is in Appendix~\ref{supp:forward_diff}.
\end{proof}
With Taylor approximations for discrete groups, the previously introduced designs can be extended to a discrete group by replacing the Lie algebra representation $d\rho$ with the forward difference operator $\Delta_s$.

\subsection{Soft equivariant layers in practice}\label{sec:using}
The proposed soft equivariant/invariant layers can be seamlessly integrated into existing architectures, whether pre-trained or trained from scratch, without introducing additional learnable parameters.

{\bf \noindent Integration into existing architectures.} For vision models, we incorporate soft equivariant and invariant layers to improve consistency under spatial transformations. These layers are applied to operations defined on structured grids, \eg, convolution, patch embedding, positional encoding, and fully connected layers over flattened grids. The point-wise non-linearities (\eg, ReLU) are equivariant by design. For other modalities, such as 2D/3D point clouds or geometric inputs (\eg, velocity or flow direction), we apply the soft equivariant and invariant layers to the fully connected layers operating on point features.

\myparagraph{Softness control.} The degree of softness is controlled by the cut-off value $b$ in the projection operator $\mB_{\tt inv}$ and $\mB_{\tt eq}$. A smaller $b$ results in a stronger bias towards equivariance, while a larger $b$ allows for more flexibility at the cost of increased equivariance error. In practice, we treat $b$ as a hyperparameter that can be tuned based on validation performance or specific requirements of the task at hand.

We defer the multi-generator equivariant layer design, smooth cut-off with soft threshold, and additional implementation details to Appendix~\ref{sec:sup_method_details}.

%% file: src/exp.tex
\section{Experiments}
We empirically validate the effectiveness of our soft equivariance method. %
In~\secref{sec:exp_tunability}, we demonstrate the ability to control a model's equivariance error. Next, we show that our method can adapt pre-trained vision models to improve their equivariance consistency and performance in image classification (\secref{sec:exp_classification}) and semantic segmentation (\secref{sec:exp_segmentation}). We also showcase an application on human trajectory prediction (\secref{sec:exp_human_traj}). 
For experiments in Sec.~\ref{sec:exp_classification},~\ref{sec:exp_segmentation} and~\ref{sec:exp_human_traj}, we focus on $2D$ rotation equivariance following prior works~\cite{finzi2021residual,kaba2023equivariance, mondal2023equivariant}.

\input{figs/tunability}

\subsection{Validating tunable softness level}
\label{sec:exp_tunability}

\myparagraph{Experiment setup.} We aim to demonstrate the tunability of our soft equivariance approach. We evaluate on the MNIST classification dataset~\cite{lecun2002gradient}. 
All the models here are trained from scratch. Our model and RPP are trained with varying rotation equivariance \emph{`softness'} $ \in [0,1)$, where `softness' of $1$ corresponds to a non-equivariant model and $0$ to a fully equivariant model. The `softness' value controls the fraction of the total number of basis vectors in the projection operator by adjusting the cutoff value $b$. 
See Appendix~\ref{sup:exp_details} for details.

\myparagraph{Evaluation metric.} We report the following metrics:
\begin{itemize}
    \item Test accuracy (Acc): The standard test set's accuracy.
    \item Augmented test accuracy (aAcc): Accuracy evaluated on augmented test samples, where augmentations are uniformly drawn from an augmented set $\tilde{G}$, a subset of the group $G$ to which the model is designed to be equivariant.
    \item Combined accuracy (cAcc): To have a single metric, we report the geometric mean between Acc and aAcc.
    \bea 
    \label{eq:cacc}
    \text{cAcc} \triangleq (\text{Acc} \cdot  \text{aAcc})^{1/2}.
    \eea 
    \item Invariance error (iErr): As the classifiers output probabilities, we use the KL divergence to measure the invariance error, \ie,
    \bea 
    \label{eq:con}
    \text{iErr} \triangleq \E_{\substack{\vx \sim \gD,  g \sim \tilde{G}}} \mathrm{KL}\big[F(\vx) \,\|\ F(\rho(g) \vx)\big].
    \eea
\end{itemize}
In this experiment, we consider two groups: rotations and roto-reflections to design the models. For evaluation, the augmented set for rotations is $\tilde{G}_1$, which consists of rotations within $\pm60^\circ$, while for the roto-reflection group it is $\tilde{G}_2$, which includes the same range of rotations plus reflections.

\myparagraph{Baseline.} We compare with the residual-based soft equivariance method (RPP)~\cite{finzi2021residual}. This approach starts from an equivariant model and introduces a non-equivariant branch through residual connections, roughly doubling the model size. For tunability, we incorporate a scalar parameter on their residual path to balance between equivariance and non-equivariance.

\myparagraph{Results.} In~\figref{fig:mnist_exp}, we observe that for both the rotation and the roto-reflection group, on different degrees of softness of equivariance, our method has lower invariance error (iErr) while maintaining higher combined accuracy (cAcc). We also observe that RPP has a worse trade-off between accuracy and invariance error compared to ours.

\input{tables/cifar.tex}

\input{tables/imagenet.tex}

\input{src/exp_image_class}

\input{src/exp_segmentation}
\input{src/exp_traj}

\subsection{Ablation studies}
\label{sec:ablation}
\myparagraph{Schur decomposition vs. SVD.}
The runtime comparison between Schur decomposition and SVD is reported in~\tabref{tab:schur_vs_svd}. We observe that the cost of constructing the projection operator via SVD grows rapidly, reaching nearly 15 minutes for an input size of $14 \times 14$. In contrast, the Schur-based operator requires less than one second. %

\myparagraph{Hard threshold vs. smooth cut-off in projection.}
In~\tabref{tab:hard_vs_soft}, we provide the ablation results comparing the hard vs. smooth thresholding method. The experiment follows the same setup as in the segmentation experiment using the ViT backbone. We observe that using the smooth cut-off in projection leads to improved performance (mIoU, aIoU, cIoU) and lower equivariance error (eErr).  
We defer additional results and ablations to the Appendix~\ref{sec:additional_res}.
\input{tables/hard-vs-soft.tex}

%% file: figs/tunability.tex
\begin{figure}[t]
    \centering
    \setlength{\tabcolsep}{1pt}
    \begin{tabular}{cc}
    \multicolumn{2}{ c }{Results on Rotations $\tilde G_1$}\\
    \includegraphics[width=0.49\columnwidth, trim=0.0cm 0.0cm 0.0cm 0.0cm, clip]{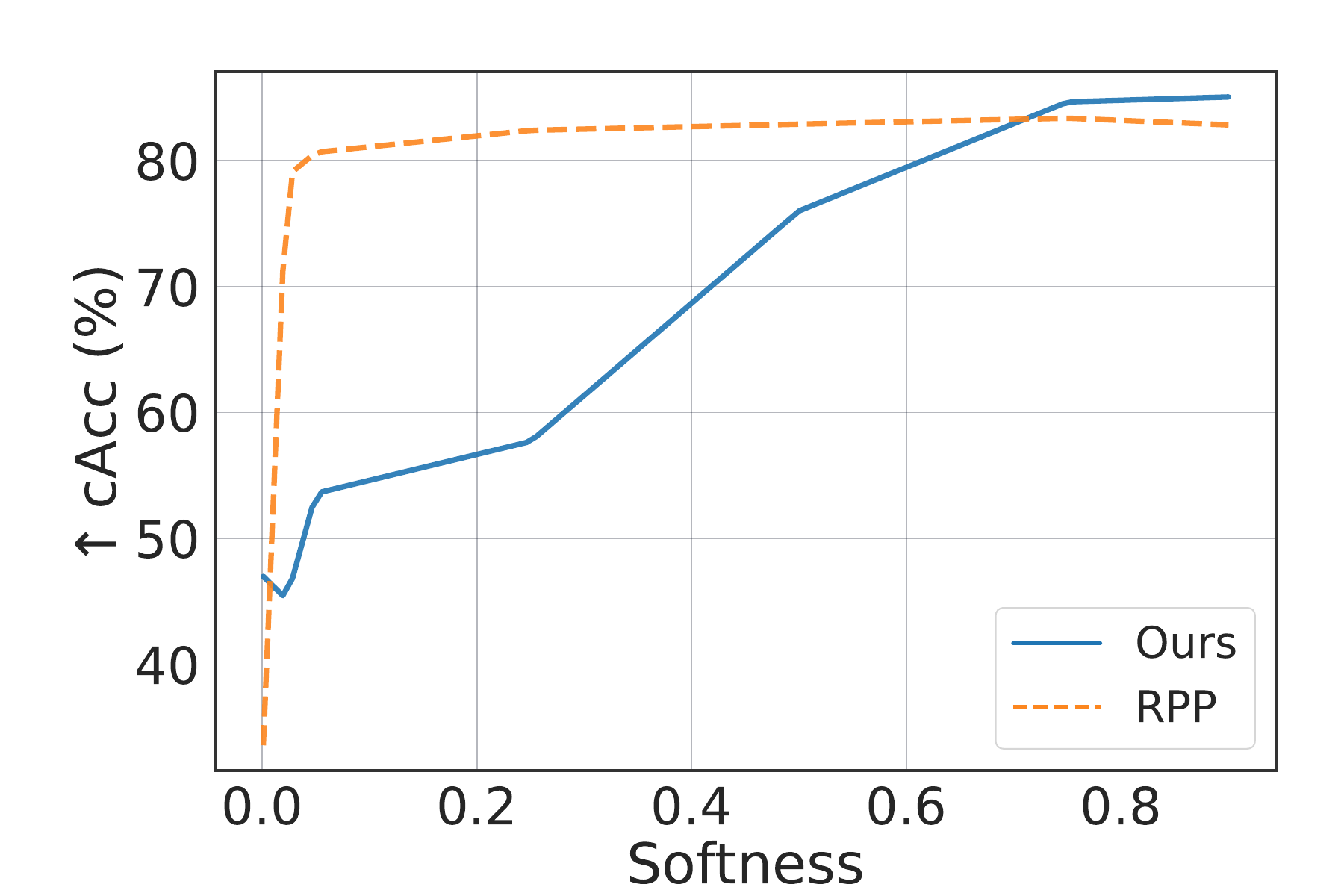} &
    \includegraphics[width=0.49\columnwidth, trim=0.0cm 0.0cm 0.0cm 0.0cm, clip]{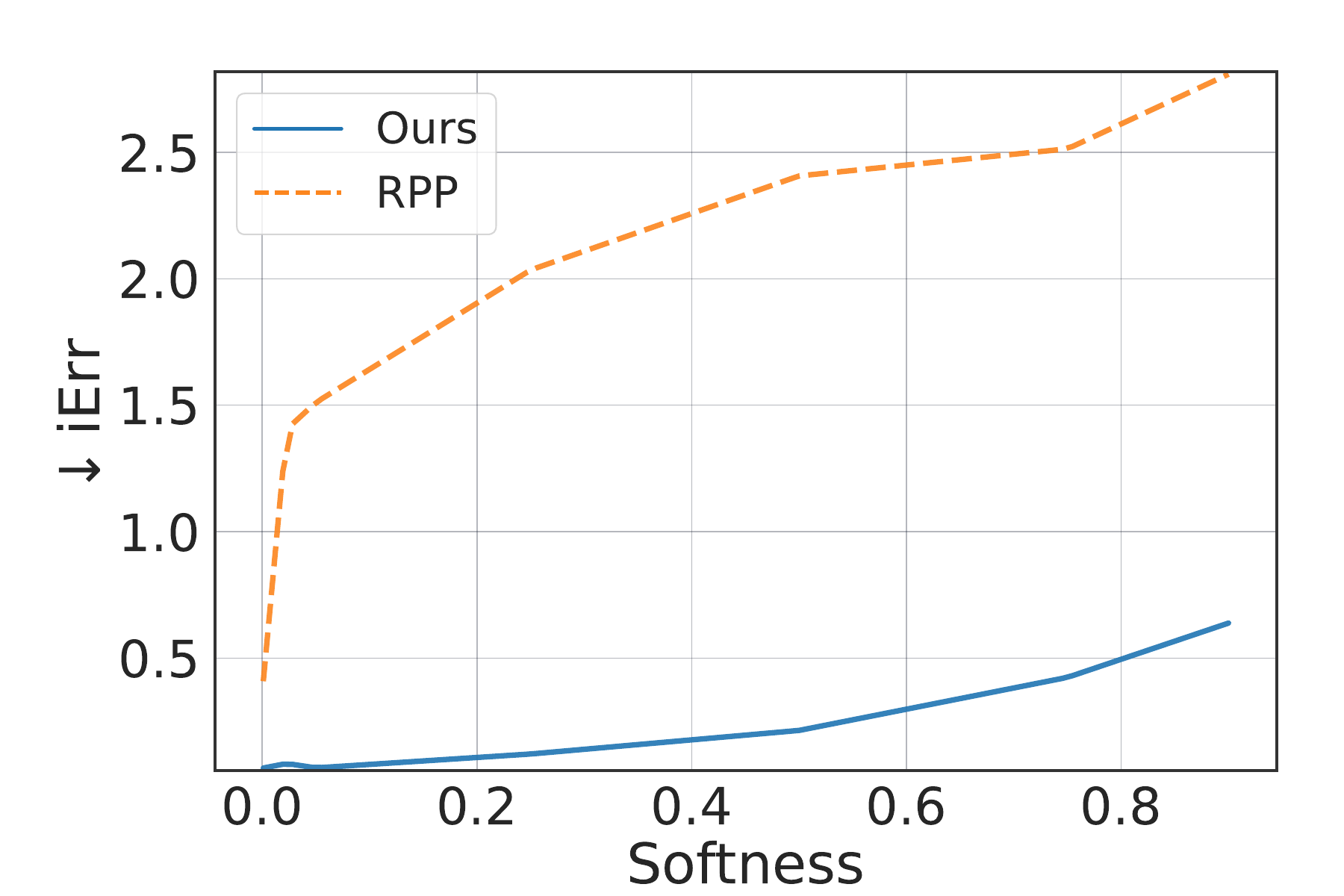} \\
    \multicolumn{2}{ c }{Results on Roto-Reflection $\tilde G_2$}\\
    \includegraphics[width=0.49\columnwidth, trim=0.0cm 0.0cm 0.0cm 0.0cm, clip]{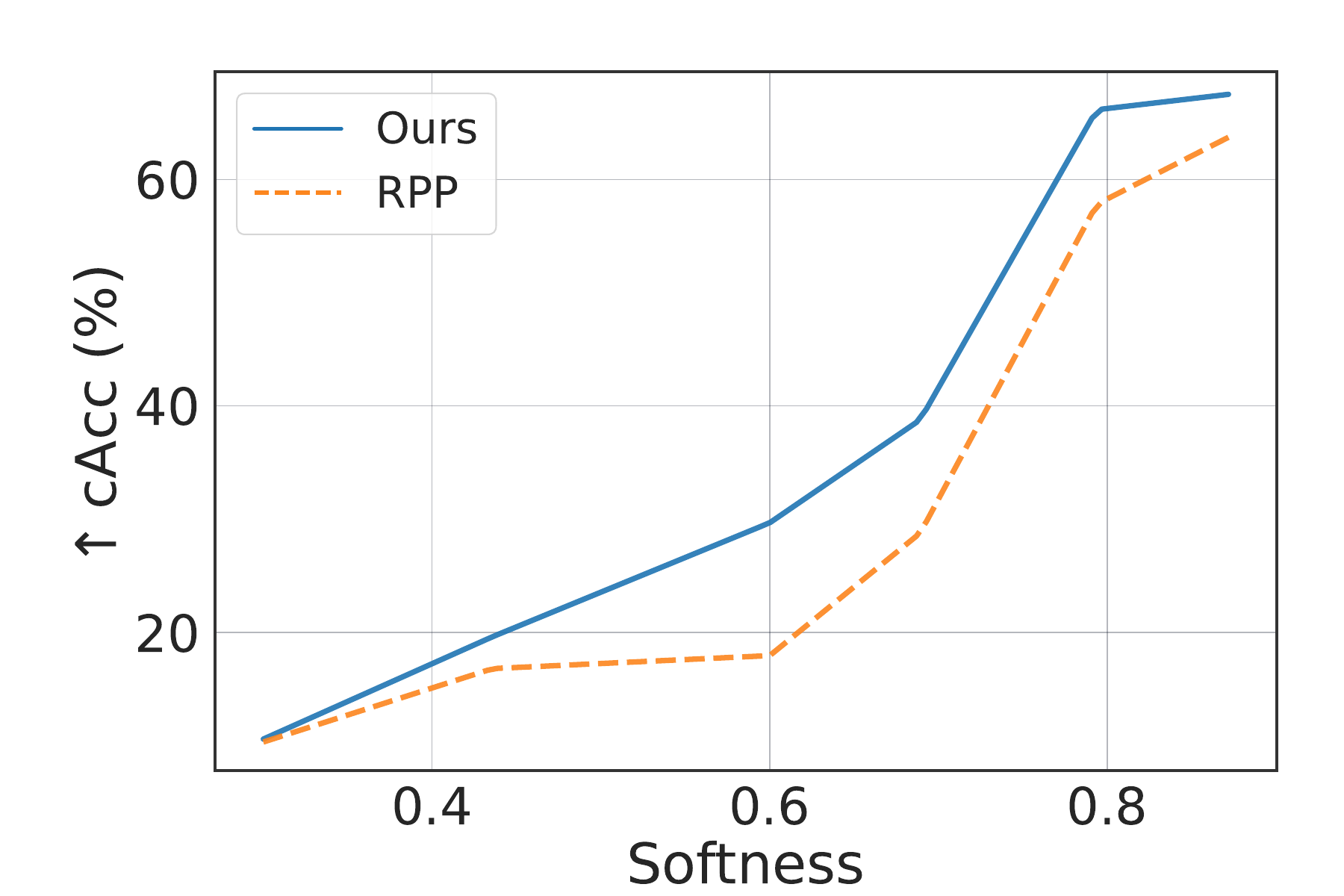} &
    \includegraphics[width=0.49\columnwidth, trim=0.0cm 0.0cm 0.0cm 0.0cm, clip]{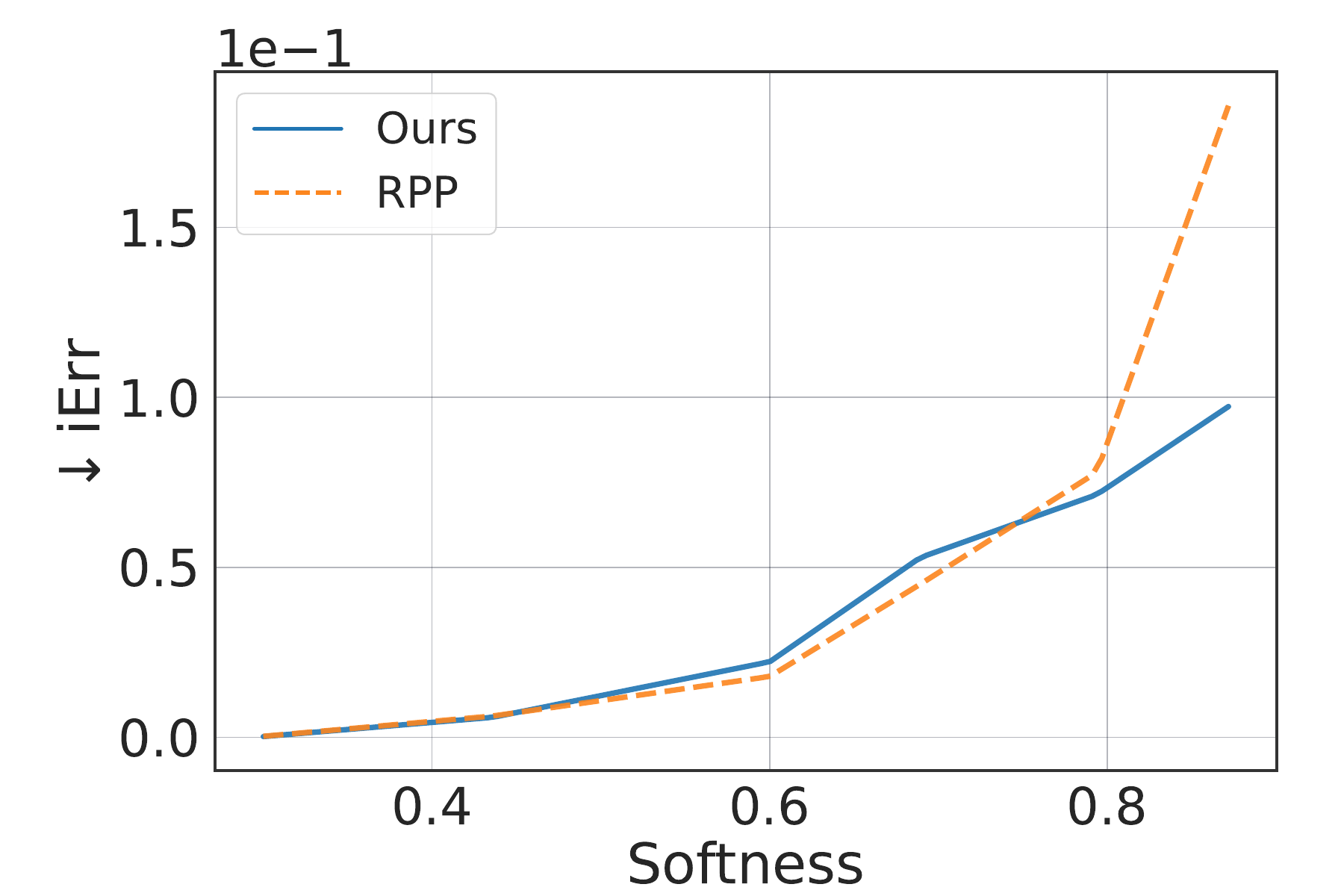}\\
    \end{tabular}
    \vspace{-0.25cm}
    \caption{{Tunable softness results (cAcc \& iErr).} Compared to RPP, ours achieves comparable performance with better iErr across two groups.
    }
    \label{fig:mnist_exp}
    \vspace{-0.25cm}
\end{figure}

%% file: tables/cifar.tex
\begin{table*}[t]
\centering
\small
\setlength{\tabcolsep}{2.5pt}
\caption{\textbf{Performance on CIFAR10 and CIFAR100 across various backbones.} Acc($\Delta$) is top-1 accuracy with $\Delta$ vs. Base; aAcc is accuracy under augmentation; cAcc is the combined accuracy; iErr is invariance error ($\times 10^{-2}$, lower is better). We observe that Ours offers a strong performance while maintaining low invariance error.
}
\vspace{-0.15cm}
\label{tab:cifar_results}
\begin{tabular}{@{}ll >{\hskip 4pt} cccc S cccc S cccc @{}}
\specialrule{.15em}{.05em}{.05em}
& \textbf{Arch.}
& \multicolumn{4}{c}{\textbf{ViT}~\cite{dosovitskiy2020image}} & &
  \multicolumn{4}{c}{\textbf{DINOv2}~\cite{oquab2023dinov2}} & &
  \multicolumn{4}{c}{\textbf{ResNet-50}~\cite{he2016deep}} \\
& &
Acc($\Delta$){\small$\uparrow$} & aAcc{\small$\uparrow$} & cAcc{\small$\uparrow$} & iErr{\small$\downarrow$} & &
Acc($\Delta$){\small$\uparrow$} & aAcc{\small$\uparrow$} & cAcc{\small$\uparrow$} & iErr{\small$\downarrow$} & &
Acc($\Delta$){\small$\uparrow$} & aAcc{\small$\uparrow$} & cAcc{\small$\uparrow$} & iErr{\small$\downarrow$} \\
\hline
\multirow{3}{*}{\rotatebox{90}{\scriptsize{CIFAR10\;}}}
& Base
& \bf 97.79{\scriptsize(00.00)} & 96.62 & 97.20 & 07.21 & &
  98.81{\scriptsize(00.00)} & 97.27 & 98.04 & 08.16 & &
  96.32{\scriptsize(00.00)} & 91.15 & 93.70 & 27.29 \\
& Canon.
& 93.23{\scriptsize(-04.56)} & 92.62 & 92.92 & 10.97 & &
  97.16{\scriptsize(-01.65)} & 96.88 & 97.02 & \bf 05.58 & &
  91.11{\scriptsize(-05.21)} & 90.66 & 90.89 & \bf 10.32 \\
 & \cellcolor{myblue!15!White} Ours 
& \cellcolor{myblue!15!White} \bf 97.79{\scriptsize(00.00)} 
& \cellcolor{myblue!15!White} \bf 96.69 
& \cellcolor{myblue!15!White} \bf 97.24 
& \cellcolor{myblue!15!White} \bf 07.02 
& \cellcolor{myblue!15!White} 
& \cellcolor{myblue!15!White}
  \bf 98.82{\scriptsize(00.01)} 
& \cellcolor{myblue!15!White}\bf 97.43 
& \cellcolor{myblue!15!White} \bf 98.12 
& \cellcolor{myblue!15!White} 07.03 
& \cellcolor{myblue!15!White} 
& \cellcolor{myblue!15!White}
  \bf 96.61{\scriptsize(00.29)} 
& \cellcolor{myblue!15!White}
\bf 91.28 
& \cellcolor{myblue!15!White} \bf 93.91 
& \cellcolor{myblue!15!White} 24.20 \\
\hline
\multirow{3}{*}{\rotatebox{90}{\scriptsize{CIFAR100\;}}}
& Base
& 86.36{\scriptsize(00.00)} & 84.09 & 85.22 & 23.81 & &
  91.01{\scriptsize(00.00)} & 85.52 & 88.22 & 43.02 & &
  83.10{\scriptsize(00.00)} & 73.73 & 78.28 & 59.36 \\
& Canon.
& 78.42{\scriptsize(-07.94)} & 76.83 & 77.62 & 24.55 & &
  84.39{\scriptsize(-06.62)} & 83.66 & 84.02 & \bf 20.86 & &
  72.90{\scriptsize(-10.20)} & 72.11 & 72.50 & \bf 19.07 \\
& \cellcolor{myblue!15!White} Ours
& \cellcolor{myblue!15!White} \bf 86.60{\scriptsize(00.24)} 
& \cellcolor{myblue!15!White} \bf 84.13 
& \cellcolor{myblue!15!White} \bf 85.36 
& \cellcolor{myblue!15!White} \bf 23.80 
& \cellcolor{myblue!15!White} 
& \cellcolor{myblue!15!White} \bf 91.03{\scriptsize(00.02)} 
& \cellcolor{myblue!15!White} \bf 86.81 
& \cellcolor{myblue!15!White} \bf 88.90 
& \cellcolor{myblue!15!White} 35.17 
& \cellcolor{myblue!15!White} 
& \cellcolor{myblue!15!White} \bf 83.30{\scriptsize(00.20)} 
& \cellcolor{myblue!15!White} \bf 74.27 
& \cellcolor{myblue!15!White} \bf 78.66 
& \cellcolor{myblue!15!White} 58.50 \\
\specialrule{.15em}{.05em}{.05em}
\end{tabular}
\vspace{-0.1cm}
\end{table*}

%% file: tables/imagenet.tex
\begin{table*}[t]
\centering
\small
\setlength{\tabcolsep}{2.5pt}
\caption{\textbf{ImageNet results with various backbones.} Acc($\Delta$) is top-1 accuracy with $\Delta$ vs. Base; aAcc is accuracy under augmentation; cAcc is the combined accuracy; iErr is invariance error ($\times 10^{-2}$, lower is better).
}
\vspace{-0.15cm}
\label{tab:imagenet_results}
\begin{tabular}{@{}l>{\hskip 4pt} cccc S cccc S cccc @{}}
\specialrule{.15em}{.05em}{.05em}
\textbf{Arch.} &
\multicolumn{4}{c}{\textbf{ViT}~\cite{dosovitskiy2020image}} & &
\multicolumn{4}{c}{\textbf{DINOv2}~\cite{oquab2023dinov2}} & &
\multicolumn{4}{c}{\textbf{ResNet-50}~\cite{he2016deep}} \\
& Acc($\Delta$){\small$\uparrow$} & aAcc{\small$\uparrow$} & cAcc{\small$\uparrow$} & iErr{\small$\downarrow$} & &
Acc($\Delta$){\small$\uparrow$} & aAcc{\small$\uparrow$} & cAcc{\small$\uparrow$} & iErr{\small$\downarrow$} & &
Acc($\Delta$){\small$\uparrow$} & aAcc{\small$\uparrow$} & cAcc{\small$\uparrow$} & iErr{\small$\downarrow$} \\
\hline
Base
& 81.67{\scriptsize(00.00)} & 77.29 & 79.40 & 00.36 & &
  84.27{\scriptsize(00.00)} & 82.82 & 83.52 & 00.13 & &
  77.91{\scriptsize(00.00)} & 75.12 & 76.48 & 00.24 \\
Canon.
& 76.51{\scriptsize(-05.16)} & 75.81 & 76.15 & \textbf{00.15} & &
  83.22{\scriptsize(-01.06)} & 82.54 & 82.87 & 00.07 & &
  72.07{\scriptsize(-05.84)} & 70.89 & 71.46 & 00.24 \\
\rowcolor{myblue!15!White}
Ours
& \bf 82.28{\scriptsize(00.61)} & \bf 80.56 & \bf 81.40 & \textbf{00.15} & &
  \bf 85.31{\scriptsize(01.04)} & \bf 84.44 & \bf 84.87 & \bf 00.05 & &
  \bf 77.96{\scriptsize(00.06)} & \bf 75.52 & \bf 76.72 & \bf 00.11 \\
\specialrule{.15em}{.05em}{.05em}
\end{tabular}
\vspace{-0.25cm}
\end{table*}

%% file: src/exp_image_class.tex
\subsection{Image classification}
\label{sec:exp_classification}
We demonstrate that our proposed method can be effectively incorporated into pre-trained models through fine-tuning.

\myparagraph{Setup.} 
We evaluate on CIFAR10/100~\cite{krizhevsky2009learning} and ImageNet-1K~\cite{deng2009imagenet}. For pre-trained backbones, we use models from PyTorch Image Models~\cite{wightman2019pytorch}, including ViT-B/16~\cite{dosovitskiy2020image}, DINOv2-Base~\cite{oquab2023dinov2,facebook_dinov2}, and ResNet-50~\cite{he2016deep}. For CIFAR-10/100, the classification head is replaced to match the number of classes, and the image is interpolated to match the expected input size of the pre-trained model. While for ImageNet-1K, we fine-tune the released models without modification.

\input{tables/imnet_angel_ablation.tex}
\myparagraph{Evaluation metric.} We follow the same metrics of Acc, aAcc, cAcc, and iErr as described in~\secref{sec:exp_tunability}. The main results are reported with the augmentation set $\tilde{G}$ that consists of rotations between $\pm 30^\circ$. We also report on $\pm 15^\circ, 45^\circ,$ and $60^\circ$.
\input{tables/segmentation.tex}

\myparagraph{Baselines.} Alongside the original model (Base) for which we applied our method, we also compare against the canonicalizer (Canon.) baseline~\cite{mondal2023equivariant}. An equivariant network is used to predict the group element that standardizes the image before passing it to the base model. Building on prior work, we employ discrete rotations to construct the models \cite{kaba2023equivariance, mondal2023equivariant, cohen2016steerable}.  Note that ResEq~\cite{finzi2021residual}'s method builds on top of an equivariant model, so we are unable to adapt a non-equivariant backbone.

\myparagraph{CIFAR10/100 results.} In~\tabref{tab:cifar_results}, we present the results on CIFAR10 and CIFAR100. We observe that Ours achieves the best overall performance across CIFAR10/100 and all backbones (ViT, DINOv2, ResNet-50). In every setting, Ours consistently improves both aAcc and cAcc, and reduces iErr relative to the Base model. While Canon. achieved a smaller iErr for DINOv2 or ResNet-50, it does so with a substantial drop, roughly 5 to 9\%, in performance. We note that Canon. does not achieve zero invariance error due to boundary effects under rotation. When rotating an image, corner pixels move outside the field of view, and these missing pixels can change the canonicalizer prediction. 

\myparagraph{ImageNet results.} 
\tabref{tab:imagenet_results} shows results on ImageNet.
As in the CIFAR experiments, we observe that Ours is best across all backbones on ImageNet, both in terms of accuracy and invariance error. Ours achieved the best aAcc and cAcc in every case. At the same time, it matches or improves the invariance error (iErr), tying the best with ViT and outperforming DINOv2 and ResNet. In this case, there is no trade-off between performance and invariance, \ie, Ours improves both simultaneously. Finally, in~\tabref{tab:imagenet_angle} we report additional experiments on different ranges of rotation angles. As expected, the performance generally drops when a larger degree of rotation is used. In general, the same trend in~\tabref{tab:imagenet_results} holds for Ours.

%% file: tables/imnet_angel_ablation.tex
\begin{table}[t]
\centering
\setlength{\tabcolsep}{2pt}
\caption{\textbf{Aug. accuracy (aAcc) with various angles on ImageNet.}
}
\label{tab:imagenet_angle}
\vspace{-0.2cm}
\resizebox{\columnwidth}{!}{
\begin{tabular}{@{}l>{\hskip 3pt}ccc>{\hskip 8pt}ccc>{\hskip 8pt}ccc@{}}
    \toprule
    \textbf{Arch.} & \multicolumn{3}{c}{\textbf{ViT}~\cite{dosovitskiy2020image}} & \multicolumn{3}{c}{\textbf{DINOv2}~\cite{oquab2023dinov2}} & \multicolumn{3}{c}{\textbf{ResNet-50}~\cite{he2016deep}} \\
    & $15^\circ$ & $45^\circ$ & $60^\circ$ & $15^\circ$ & $45^\circ$ & $60^\circ$&$15^\circ$ & $45^\circ$ & $60^\circ$ \\
\midrule
Base                                      & 79.04 & 74.83 & 72.66 & 83.63 & 81.78 & 80.37 & 75.77 & 73.42 & 70.70 \\
Canon. & 75.91 & 76.01 & 75.92 & 82.57 & 82.57 & 82.42 & 70.93 & 71.01 & 70.78 \\
\rowcolor{myblue!15!White} Ours                                      & \bf 81.23 & \bf 79.36 & \bf 77.55 & \bf 84.80 & \bf 84.08 & \bf 82.74 & \bf 76.47 & \bf 73.70 & \bf 71.27 \\
\bottomrule
\end{tabular}
}
\vspace{-0.35cm}
\end{table}

%% file: tables/segmentation.tex
\begin{table*}[t]
\centering
\small
\setlength{\tabcolsep}{2pt}
\caption{\textbf{Segmentation Performance on PASCAL VOC~\cite{everingham2010pascal}.}
mIoU ($\Delta$) is the mean intersection over union with $\Delta$ vs. Base; aIoU is the augmented mIoU; cIoU is the combined mIoU; eErr is equivariance error ($\times 10^{-2}$, lower is better.)
}
\vspace{-0.15cm}
\label{tab:voc_seg_results}
\begin{tabular}{@{}l >{\hskip 4pt} cccc S cccc S cccc @{}}
\specialrule{.15em}{.05em}{.05em}
\textbf{Arch.} &
\multicolumn{4}{c}{\textbf{ViT}~\cite{dosovitskiy2020image}} & &
\multicolumn{4}{c}{\textbf{DINOv2}~\cite{oquab2023dinov2}} & &
\multicolumn{4}{c}{\textbf{SegFormer}~\cite{xie2021segformer}} \\
& mIoU($\Delta$){\small$\uparrow$} & aIoU{\small$\uparrow$} & cIoU{\small$\uparrow$} & eErr{\small$\downarrow$} & &
mIoU($\Delta$){\small$\uparrow$} & aIoU{\small$\uparrow$} & cIoU{\small$\uparrow$} & eErr{\small$\downarrow$} & &
mIoU($\Delta$){\small$\uparrow$} & aIoU{\small$\uparrow$} & cIoU{\small$\uparrow$} & eErr{\small$\downarrow$} \\
\hline
Base
& 73.40{\scriptsize(00.00)} & 70.09 & 71.73 & 12.31 & &
  \bf 89.57{\scriptsize(00.00)} & 88.10 & 88.83 & 04.06 & &
  65.34{\scriptsize(00.00)} & 61.17 & 63.22 & 11.03 \\
Canon.
& 65.36{\scriptsize(-08.03)} & 61.93 & 63.62 & 20.39 & &
  83.48{\scriptsize(-06.09)} & 82.73 & 83.11 & 44.93 & &
  57.50{\scriptsize(-07.84)} & 55.23 & 56.36 & 24.82 \\
\rowcolor{myblue!15!White}
Ours
& \bf 74.78{\scriptsize(01.38)} & \bf 71.61 & \bf 73.18 & \bf 11.12 & &
  89.48{\scriptsize(-00.09)} & \bf 88.70 & \bf 89.09 & \bf 03.70 & &
  \bf 66.34{\scriptsize(00.99)} & \bf 62.52 & \bf 64.40 & \bf 10.64 \\
\specialrule{.15em}{.05em}{.05em}
\end{tabular}
\vspace{-0.1cm}
\end{table*}

%% file: src/exp_segmentation.tex
\subsection{Semantic segmentation}
\label{sec:exp_segmentation}

\myparagraph{Setup.} We report on the PASCAL VOC (2012) dataset~\cite{everingham2010pascal}, which comprises 21 object classes for semantic segmentation.
For the pretrained backbones, we choose ViT~\cite{dosovitskiy2020image}, DINOv2~\cite{oquab2023dinov2} and SegFormer~\cite{xie2021segformer} (trained on the ADE20K dataset~\cite{zhou2019semantic}). Following the linear probing experiment in DINOv2, we add a linear classification head on top of features extracted from ViT and DINOv2. For SegFormer, we replace the output head to match the number of classes. %

\myparagraph{Evaluation metrics.} We report the standard mean Intersection over Union (mIoU). Analogous to the image classification setting, we report mIoU on the augmented set (aIoU) and the combined mIoU (cIoU). As segmentation models $F$ are equivariant, we report the equivariance error (eErr):
\bea
\label{eq:con_seg}
\E_{(u,v), \vx, g \sim\tilde{G}}  \text{KL}\Big[F(\rho_{\gX}(g) \vx)[u,v] \,\|\ \rho_{\gY}(g) F(\vx)[u,v]\Big],
\eea
where $(u,v)$ indexes the set of valid pixels, excluding those that go out of bounds after rotation. A lower eErr. indicates the model is more consistent in its prediction under the augmentation. We study $\tilde{G}$ consisting of $\pm 30^\circ$ degree rotations. Note, larger rotation leads to significant boundary issues for segmentation, which are less meaningful to study.

\myparagraph{Baselines.} Same as in the classification experiment, we compare with the Base model, and with the Canonicalization-based approach~\cite{mondal2023equivariant}.

\myparagraph{Results.} 
\tabref{tab:voc_seg_results} shows segmentation results. Our method improves aIoU, cIoU, and reduces eErr across all settings. For ViT and SegFormer, there is an increase in mIoU, and a mild drop for DINOv2. In contrast, the Canon. baseline reduces mIoU significantly and worsens consistency. We found that the canonicalizer was unable to effectively predict the rotation. In summary, our method offers a better, or even no, trade-off between performance and equivariance error.

%% file: src/exp_traj.tex
\input{tables/human_traj.tex}

\subsection{Human trajectory prediction} 
\label{sec:exp_human_traj}

\myparagraph{Setup.} Beyond image-based tasks, we evaluate our method on human trajectory prediction using the ETH~\cite{pellegrini2009you} and UCY~\cite{lerner2007crowds} datasets, which together comprise five scenes (ETH, Hotel, Zara1, Zara2, and Univ) with varying crowd densities. The task predicts the next 10 time steps of pedestrian positions~$\vy$ given the past 10 time steps~$\vx$, where each position is represented by 2D coordinates.

{\bf \noindent Evaluation metrics.} Following prior work~\cite{giuliari2021transformer}, we report standard performance metrics of the Average Displacement Error (ADE) and Final Displacement Error (FDE) to quantify performance. As with the other tasks, we report the augmented metrics (aADE, aFDE). For the combined metrics (cADE, cFDE), as these are no longer probabilities, we directly combine the standard and augmented metrics with their average.

Next, for equivariance error, as the output $\vy$ is 2D coordinates, we use the $\ell_2$-loss to capture differences, \ie, eErr is defined as:
\bea
\E_{\vx \sim \gD, g \sim \tilde{G}}  \norm{F(\rho_\gX(g)\vx)-\rho_\gY(g) F(\vx)}_2,
\eea
where $\gD$ is the test set and $\tilde{G}$ consists of $\pm 30^\circ$ rotations.

\input{tables/schur_vs_SVD.tex}
\myparagraph{Baselines.}
We adopt the autoregressive transformer model by~\citet{giuliari2021transformer} as our base architecture (Base) and apply our proposed method (Ours) on top of it. We also compare against an equivariant baseline (EqAuto), an equivariant autoregressive transformer based on vector neurons~\cite{deng2021vector}. We adopt continuous rotation to design the models.  %

\myparagraph{Results.} In~\tabref{tab:traj_results}, we report the quantitative metrics. Our method outperforms both the standard and equivariant transformer baselines in terms of cADE and cFDE across most scenes. We observe that EqAuto with full equivariance does not necessarily lead to better performance. Our approach achieves the best cADE and cFDE on four of the scenes.

%% file: tables/human_traj.tex
\begin{table*}[t]
\centering
\setlength{\tabcolsep}{2pt}
\caption{\textbf{Prediction Error on Human Trajectory Datasets.}
cADE and cFDE are combined metrics. eErr is equivariance error in $\times 10^{-2}$.
}
\small
\label{tab:traj_results}
\vspace{-0.15cm}
\begin{tabular}{@{}l>{\hskip 3pt} ccc S ccc S ccc S ccc S ccc @{}}
\specialrule{.15em}{.05em}{.05em}
\textbf{Scenes}
& \multicolumn{3}{c}{\textbf{ETH}} & &
  \multicolumn{3}{c}{\textbf{UNIV}} & &
  \multicolumn{3}{c}{\textbf{ZARA1}} & &
  \multicolumn{3}{c}{\textbf{ZARA2}} & &
  \multicolumn{3}{c}{\textbf{HOTEL}} \\
& {\small cADE}{\small$\downarrow$} & {\small cFDE}{\small$\downarrow$} & {\small eErr}{\small$\downarrow$} & &
  {\small cADE}{\small$\downarrow$} & {\small cFDE}{\small$\downarrow$} & {\small eErr}{\small$\downarrow$} & &
  {\small cADE}{\small$\downarrow$} & {\small cFDE}{\small$\downarrow$} & {\small eErr}{\small$\downarrow$} & &
  {\small cADE}{\small$\downarrow$} & {\small cFDE}{\small$\downarrow$} & {\small eErr}{\small$\downarrow$} & &
  {\small cADE}{\small$\downarrow$} & {\small cFDE}{\small$\downarrow$} & {\small eErr}{\small$\downarrow$} \\
\hline
Base~\cite{giuliari2021transformer}
  & 4.73 & \bf 6.15 & 1.68 & &
    7.91 & 8.16 & 0.73 & &
    3.61 & 4.68 & 1.14 & &
    3.17 & 3.79 & 1.17 & &
    5.84 & 6.67 & 2.50 \\
EqAuto~\cite{deng2021vector}
  & 5.40 & 7.33 & \bf 0.00 & &
    8.16 & 8.33 & \bf 0.00 & &
    3.66 & 4.98 & \bf 0.00 & &
    2.94 & 3.63 & \bf 0.00 & &
    6.16 & 6.78 & \bf 0.00 \\
\rowcolor{myblue!15!White}
Ours
  & \bf 4.58 & 6.23 & 1.42 & &
    \bf 7.85 & \bf 8.07 & 0.69 & &
    \bf 3.40 & \bf 4.67 & 0.39 & &
    \bf 2.91 & \bf 3.60 & 0.24 & &
    \bf 5.69 & \bf 6.26 & 0.41 \\
\specialrule{.15em}{.05em}{.05em}
\end{tabular}
\vspace{-0.15cm}
\end{table*}

%% file: tables/schur_vs_SVD.tex
\begin{table}[t]
\centering
\setlength{\tabcolsep}{3pt}
\renewcommand{\arraystretch}{0.9}
\caption{\textbf{Runtime~(s) of Schur vs. SVD to compute the projection operator across input sizes.}}
\label{tab:schur_vs_svd}
\vspace{-0.2cm}
\resizebox{\columnwidth}{!}{
\begin{tabular}{@{}lccccccc@{}}
\specialrule{.15em}{.05em}{.05em}
\textbf{Size} & %
& {4×4} & {6×6} & {8×8} & {10×10} & {12×12} & {14×14} \\ 
\hline
SVD   & %
& $1.7\! \cdot\!10^{-1}$ & $5.0\!\cdot\!10^{-1}$ & $1.0\!\cdot\!10^{1}$ & $4.5\!\cdot\!10^{1}$ & $2.8\!\cdot\!10^{2}$ & $8.9\!\cdot\!10^{2}$ \\
Schur & %
& $\bf 4.0\!\cdot\!10^{-3}$ & $ \bf 1.0\!\cdot\!10^{-2}$ &  $ \bf 2\!\cdot\!10^{-2}$ &  $ \bf 5\!\cdot\!10^{-2}$ & $\bf 1.3\!\cdot\!10^{-1}$ & $ \bf 2.5\!\cdot\!10^{-1}$ \\
\specialrule{.15em}{.05em}{.05em}
\end{tabular}
}
\vspace{-0.1cm}
\end{table}

%% file: tables/hard-vs-soft.tex
\begin{table}[t]
\centering
\setlength{\tabcolsep}{1.5pt}
\caption{\textbf{Hard vs. Soft threshold performance for image segmentation on PASCAL VOC~\cite{everingham2010pascal}.}
}
\small
\vspace{-0.15cm}
\label{tab:hard_vs_soft}
\begin{tabular}{@{} >{\hskip 4pt} cccc S cccc @{}}
\specialrule{.15em}{.05em}{.05em}
\multicolumn{4}{c}{\textbf{Hard Threshold}} & &
\multicolumn{4}{c}{\textbf{Soft Threshold}} \\
mIoU{\small$\uparrow$} & aIoU{\small$\uparrow$} & cIoU{\small$\uparrow$} & eErr{\small$\downarrow$} & &
mIoU{\small$\uparrow$} & aIoU{\small$\uparrow$} & cIoU{\small$\uparrow$} & eErr{\small$\downarrow$} \\
\hline
73.92 & 69.70 & 71.78 & 11.74 & & \bf 74.78 & \bf 71.61 & \bf 73.18 & \bf 11.12 \\
\specialrule{.15em}{.05em}{.05em}
\end{tabular}
\vspace{-0.2cm}
\end{table}

%% file: src/conc.tex
\section{Conclusion}
\label{sec:conc}
We propose a principled framework for designing soft invariant and equivariant layers with tunable and theoretically bounded equivariance error. This enables the creation of a family of layers with controllable softness levels. Our layer can be used to adapt existing non-equivariant pre-trained models to soft-equivariant ones. Extensive experiments on image classification, segmentation, trajectory prediction, and synthetic $O(5)$ invariant tasks demonstrate that our approach improves task performance while reducing equivariance error. Notably, our method achieves gains on the competitive ImageNet benchmark. Overall, our framework provides a practical solution for incorporating soft equivariance into modern vision systems with guarantees.

%% file: supp/appendix.tex
\onecolumn
\appendix
\label{sec:appendix}

{\noindent \bf \Large Appendix}

\vspace{6pt}
{\noindent The appendix is organized as follows:}
\begin{itemize}
    \item In~\secref{supp:more_fig1}, we provide additional visualizations following~\figref{fig:teaser}'s format.
    \item In~\secref{supp:math}, we review additional mathematical background.
    \item In~\secref{sec:sup_method_details}, we provide additional details on the method.
    \item In~\secref{sec:additional_clarifications}, we provide clarifications on our design choices and provide examples.
    \item In~\secref{supp:proofs}, we provide the complete proofs of the claims and lemma in the main paper.
    \item In~\secref{sec:additional_res}, we provide additional results on the human trajectory experiment. 
    \item In~\secref{sup:exp_details}, we document the experimental details. Code is available at \url{https://github.com/ashiq24/soft-equivariance}
\end{itemize}

\section{Additional Visualization following~\figref{fig:teaser}}
\label{supp:more_fig1}

Following the format of~\figref{fig:teaser}, we provide more visualization on the model weights, and with respect to different rotations in~\figref{fig:teaser2} and~\figref{fig:teaser3}.

{\noindent\input{figs/feature_viz_90}}
{\noindent\input{figs/feature_viz_10}}

\input{supp/math_prelim}

\input{supp/method_details}

\input{supp/math_proofs}
\input{supp/additional_results}

\input{supp/training_detail.tex}

%% file: figs/feature_viz_90.tex
\setlength{\tabcolsep}{1.8pt}%
\renewcommand{\arraystretch}{0.8}
\begin{minipage}{\textwidth}
\newcommand{\onebyfour}[4]{%
\begin{minipage}{0.20\textwidth}\centering
    \includegraphics[width=0.30\linewidth]{#1}\hspace{0.0095\linewidth}%
    \includegraphics[width=0.30\linewidth]{#2}\hspace{0.0095\linewidth}%
    \includegraphics[width=0.30\linewidth]{#3}\hspace{0.0095\linewidth}%
\end{minipage}}

\begin{minipage}{0.073\textwidth}
\hspace{1cm}
\end{minipage}
\begin{minipage}{0.927\textwidth}
    \noindent\resizebox{\textwidth}{!}{%
    \begin{tikzpicture}[x=1\textwidth/10]
        \draw[thick, <->] (0.,0) -- (10.,0);
        \node[anchor=north] at (0.42,+0.6) {\small Equivariant};
        \node[anchor=north] at (5,+0.6) {\small Soft equivariant};
        \node[anchor=north] at (9.45,+0.6) {\small Non-equivariant};
    \end{tikzpicture}
    }%
\vspace{6pt}
\end{minipage}
\resizebox{\textwidth}{!}{%
\begin{tabular}{cccccc}
    Weights & \onebyfour{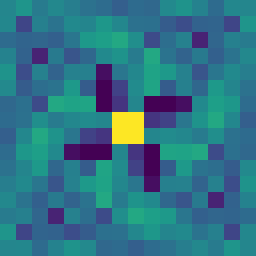}{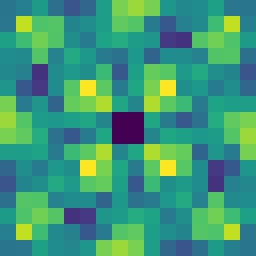}{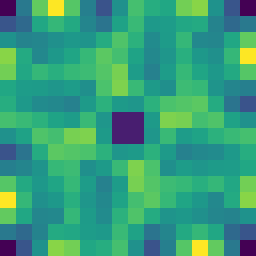}{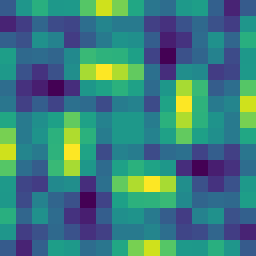} &
    \onebyfour{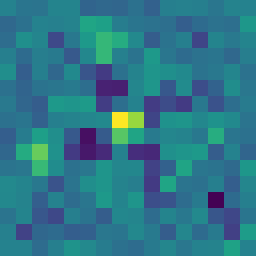}{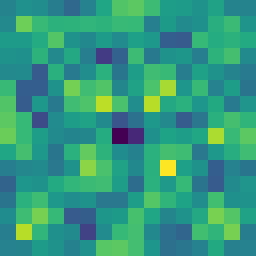}{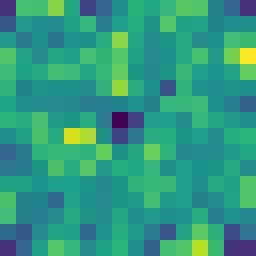}{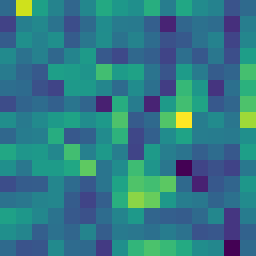} &
    \onebyfour{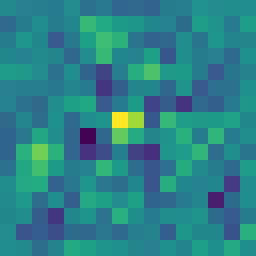}{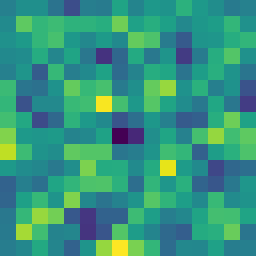}{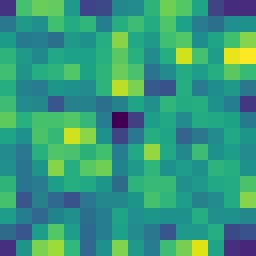}{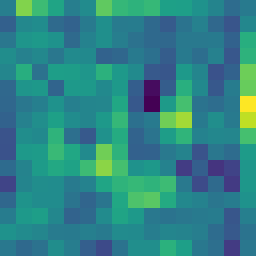} &
    \onebyfour{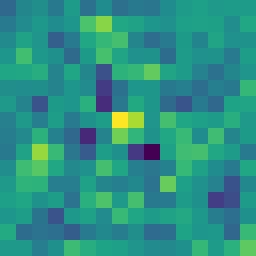}{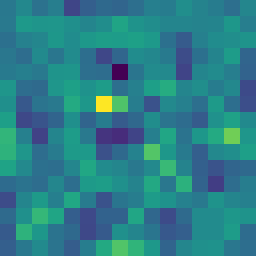}{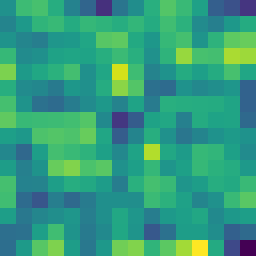}{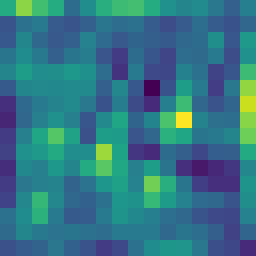} &
    \onebyfour{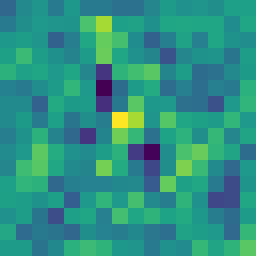}{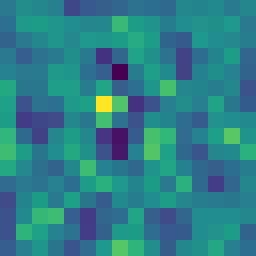}{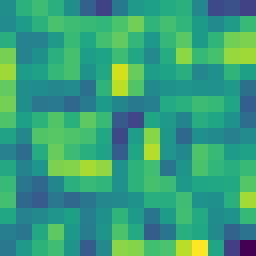}{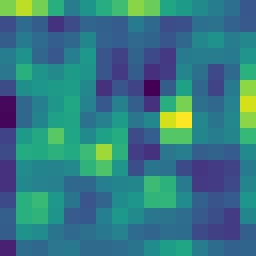} \\[0.42cm]
    \begin{minipage}{0.065\textwidth}\centering\includegraphics[width=0.91\linewidth,angle=90,origin=c]{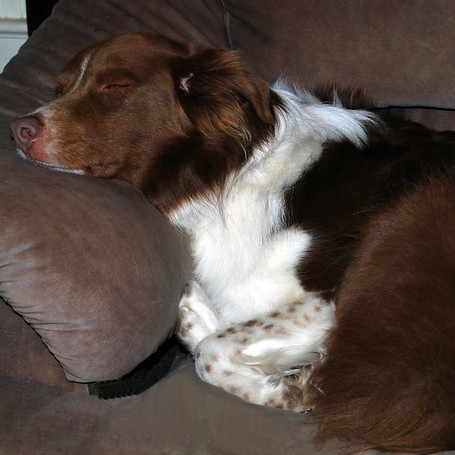}\end{minipage} & \onebyfour{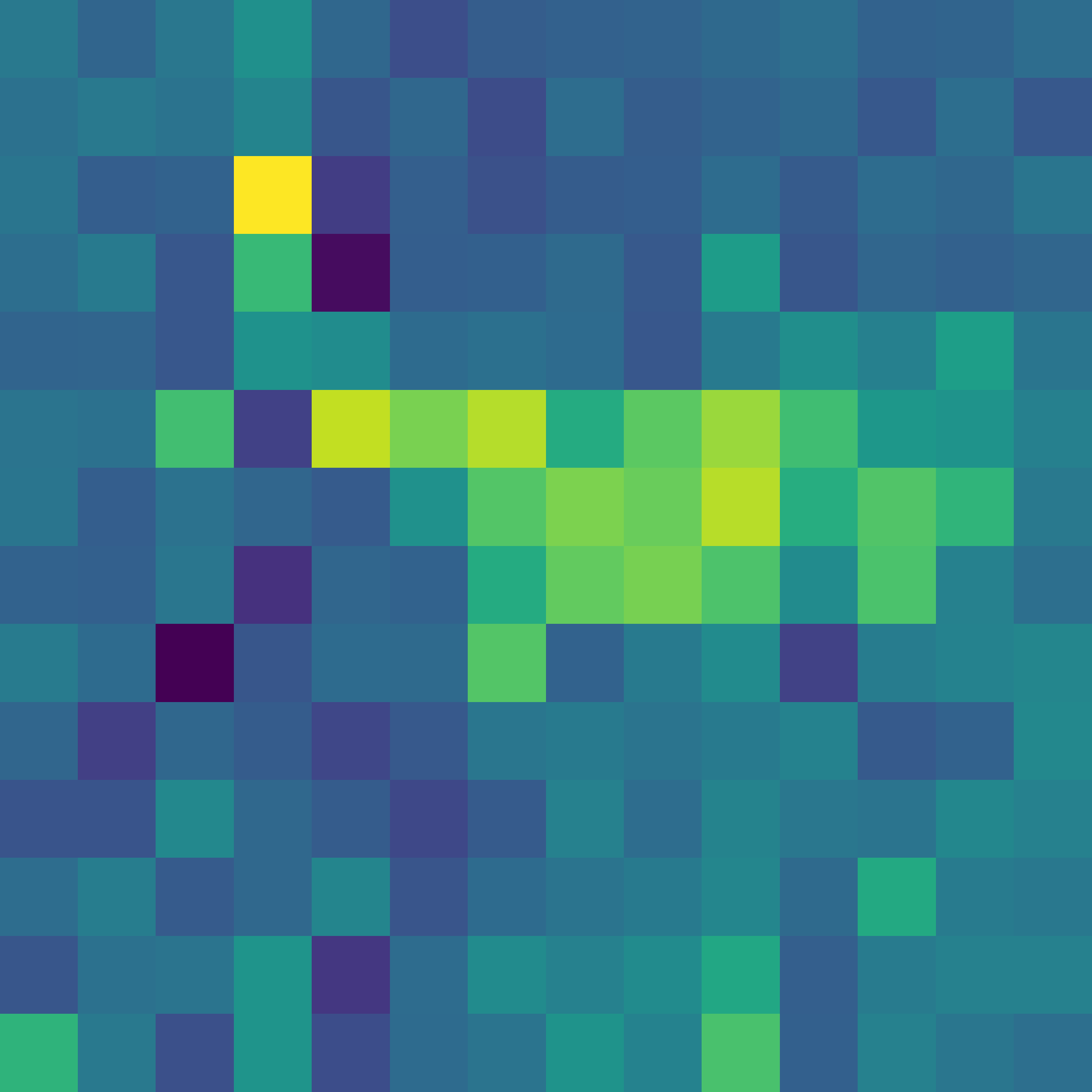}{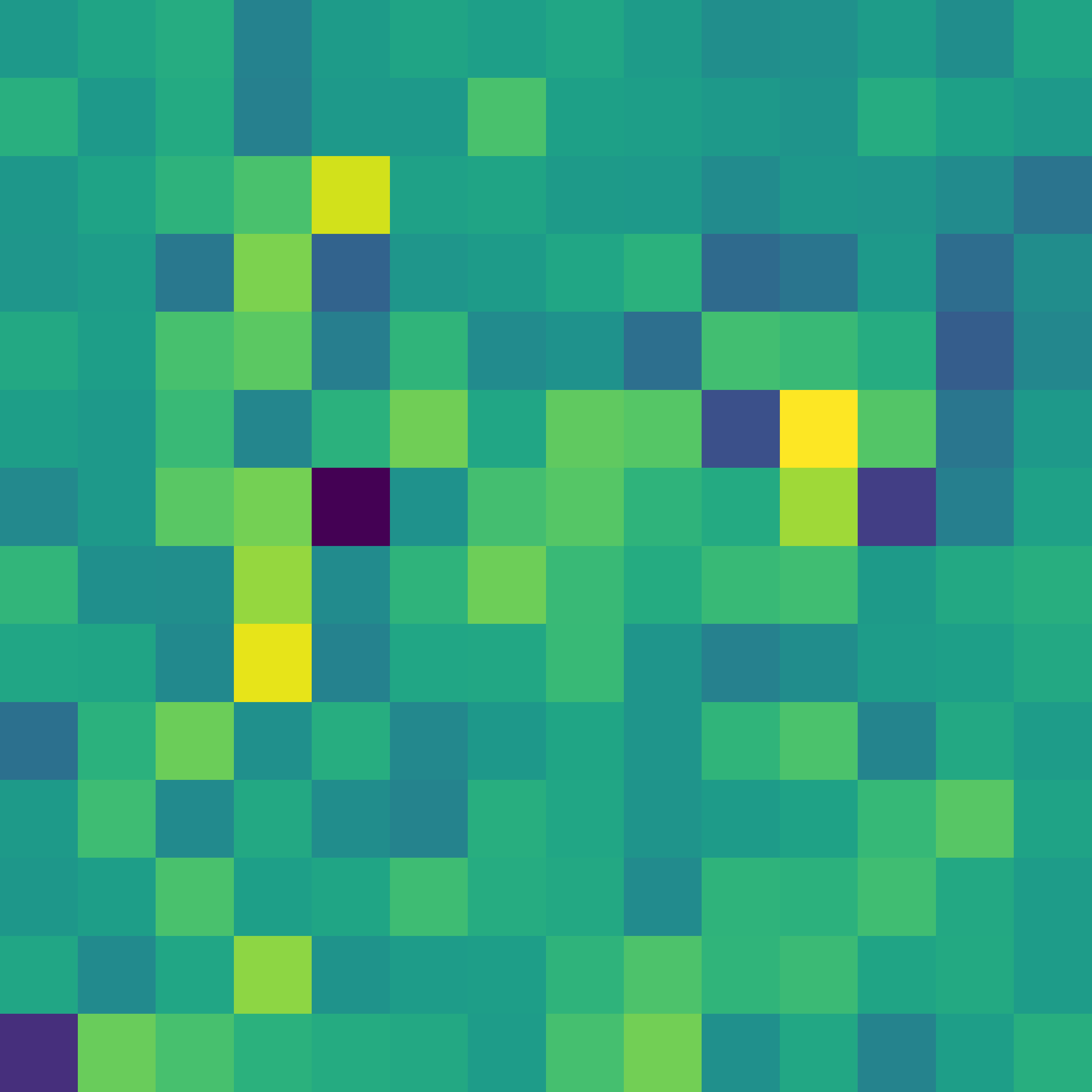}{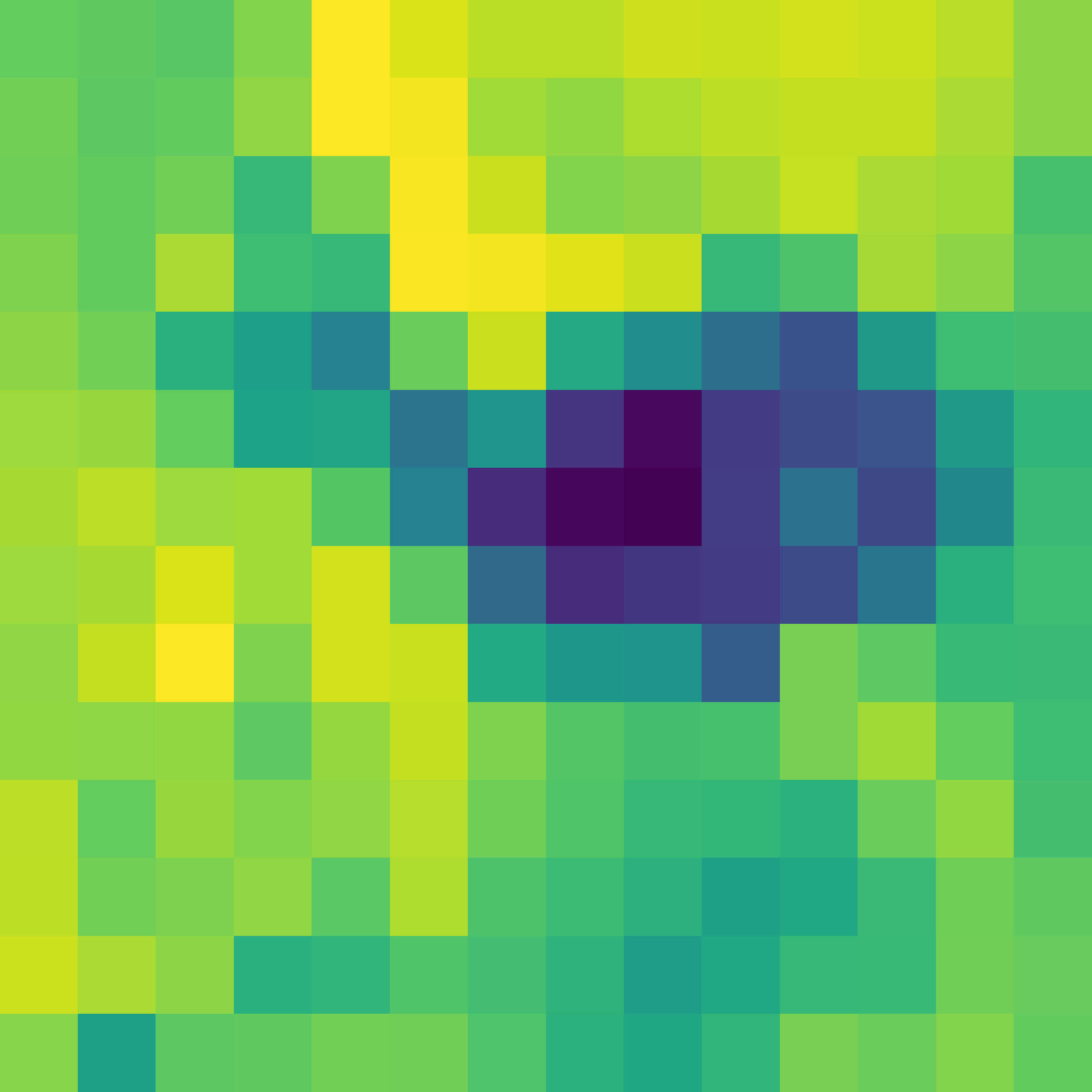}{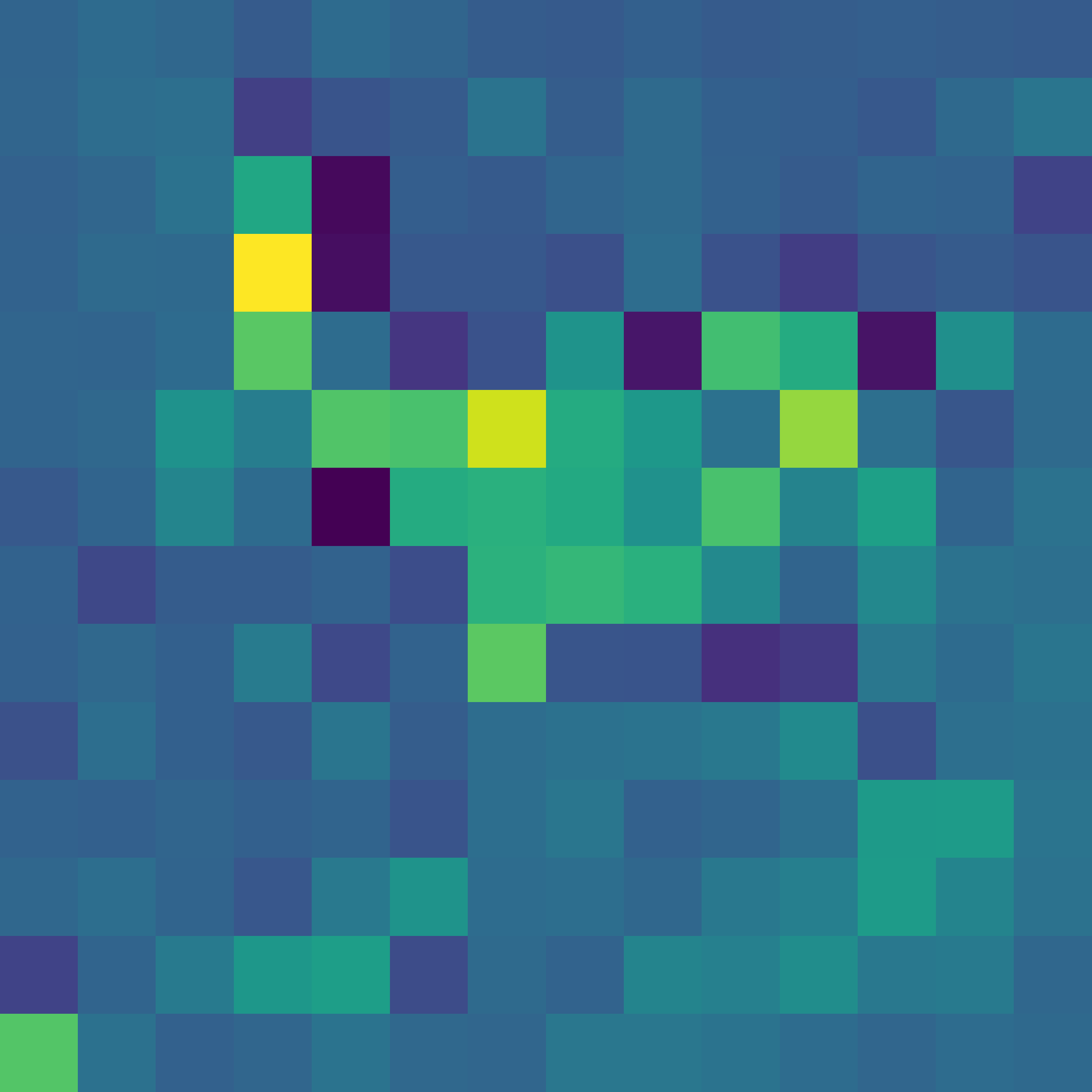} &
    \onebyfour{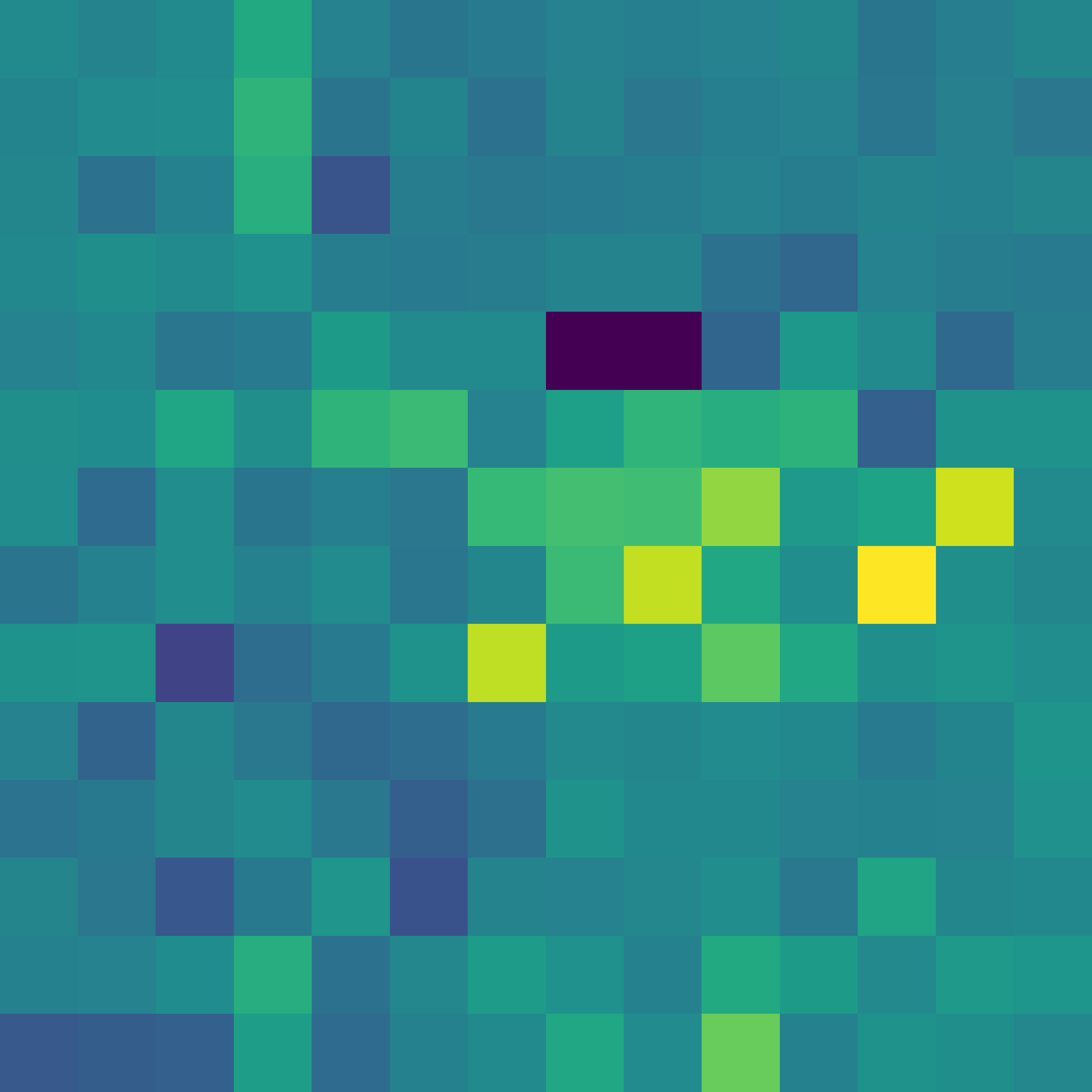}{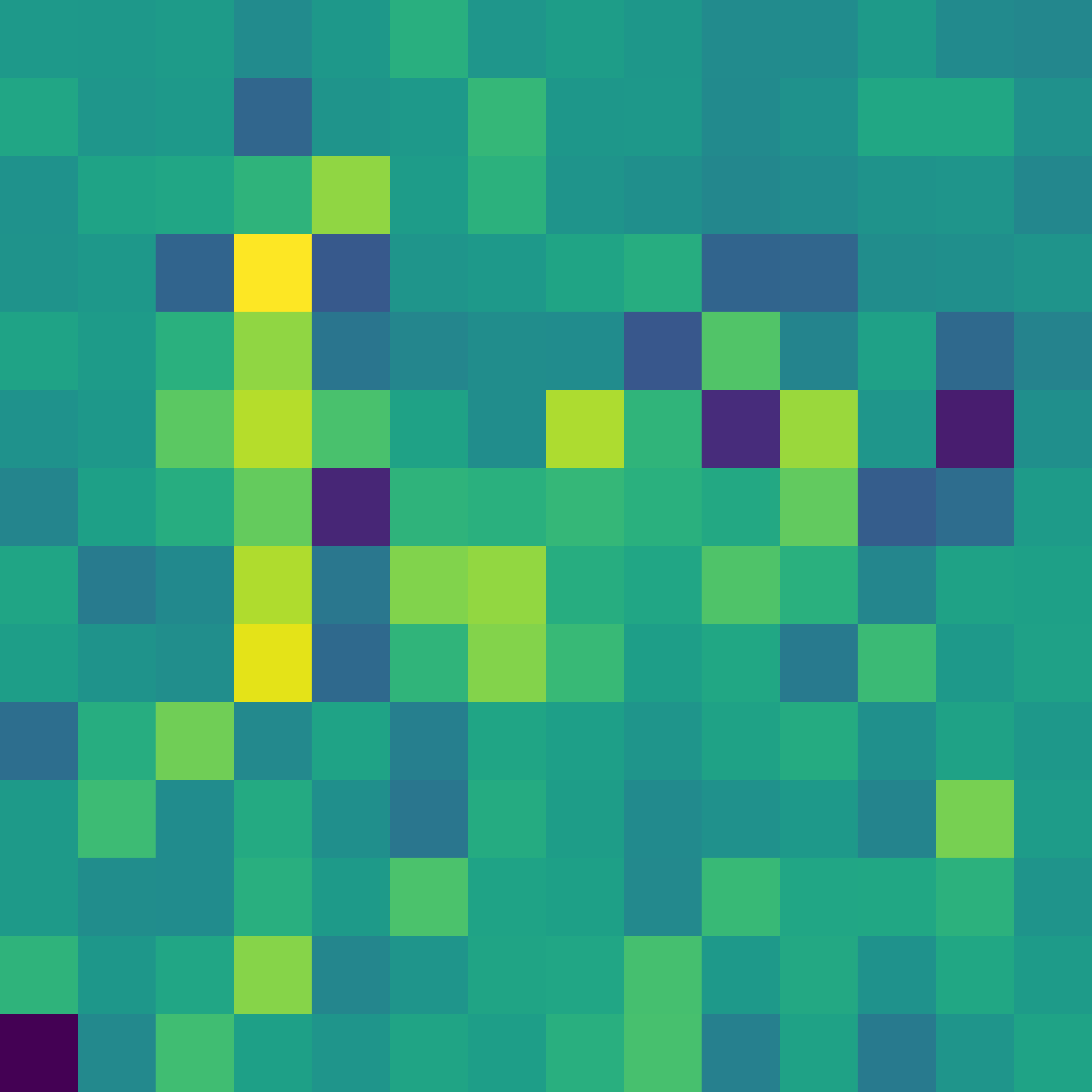}{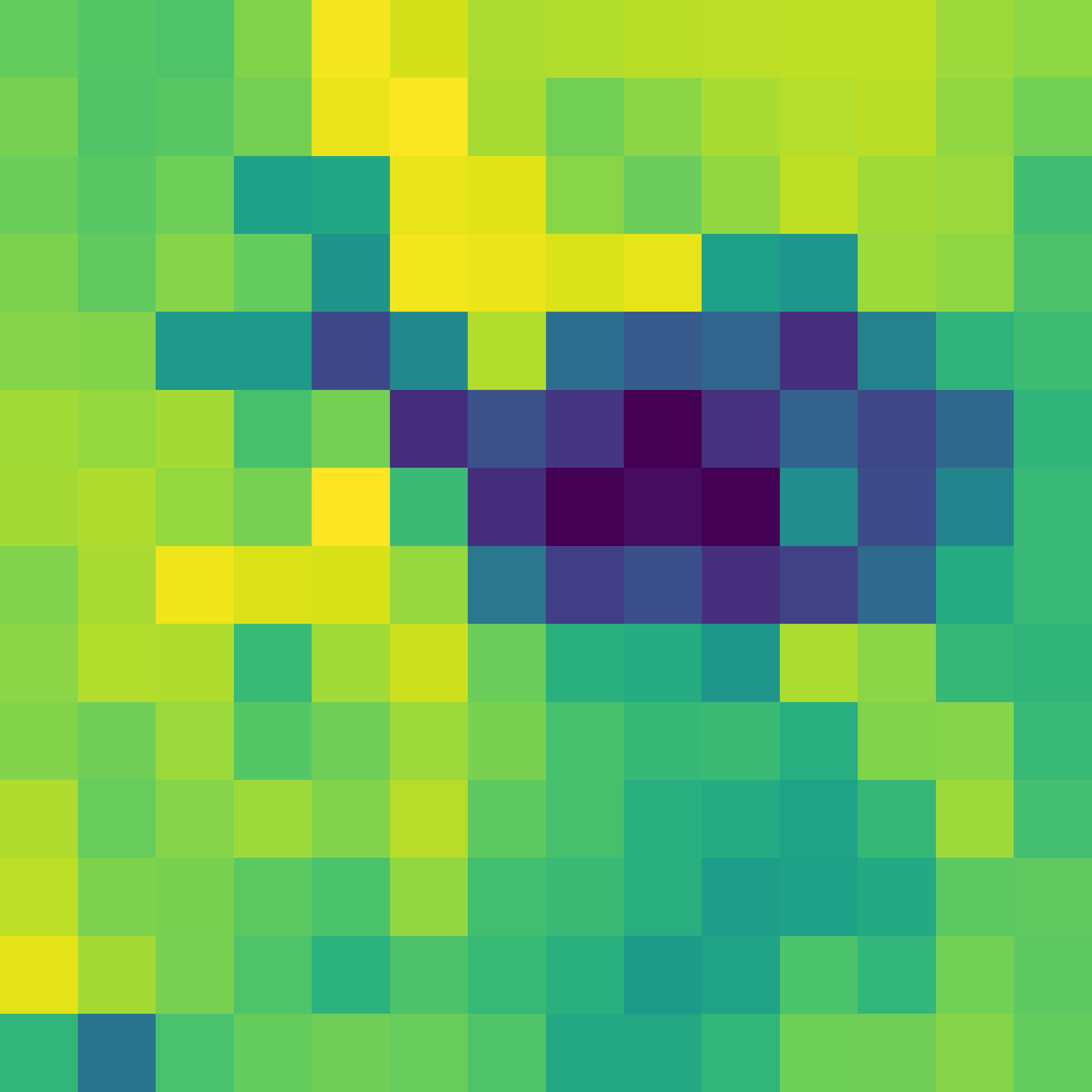}{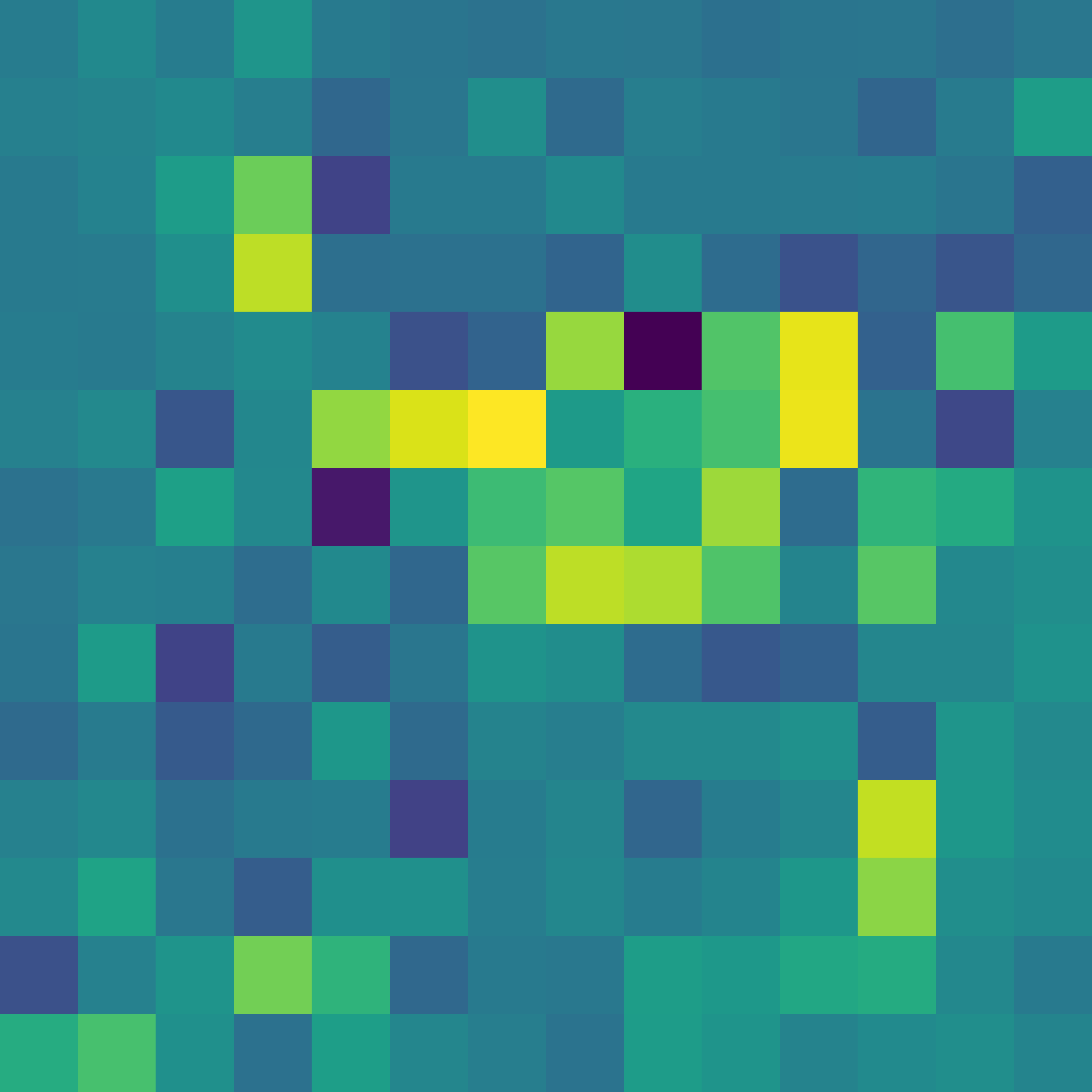} &
    \onebyfour{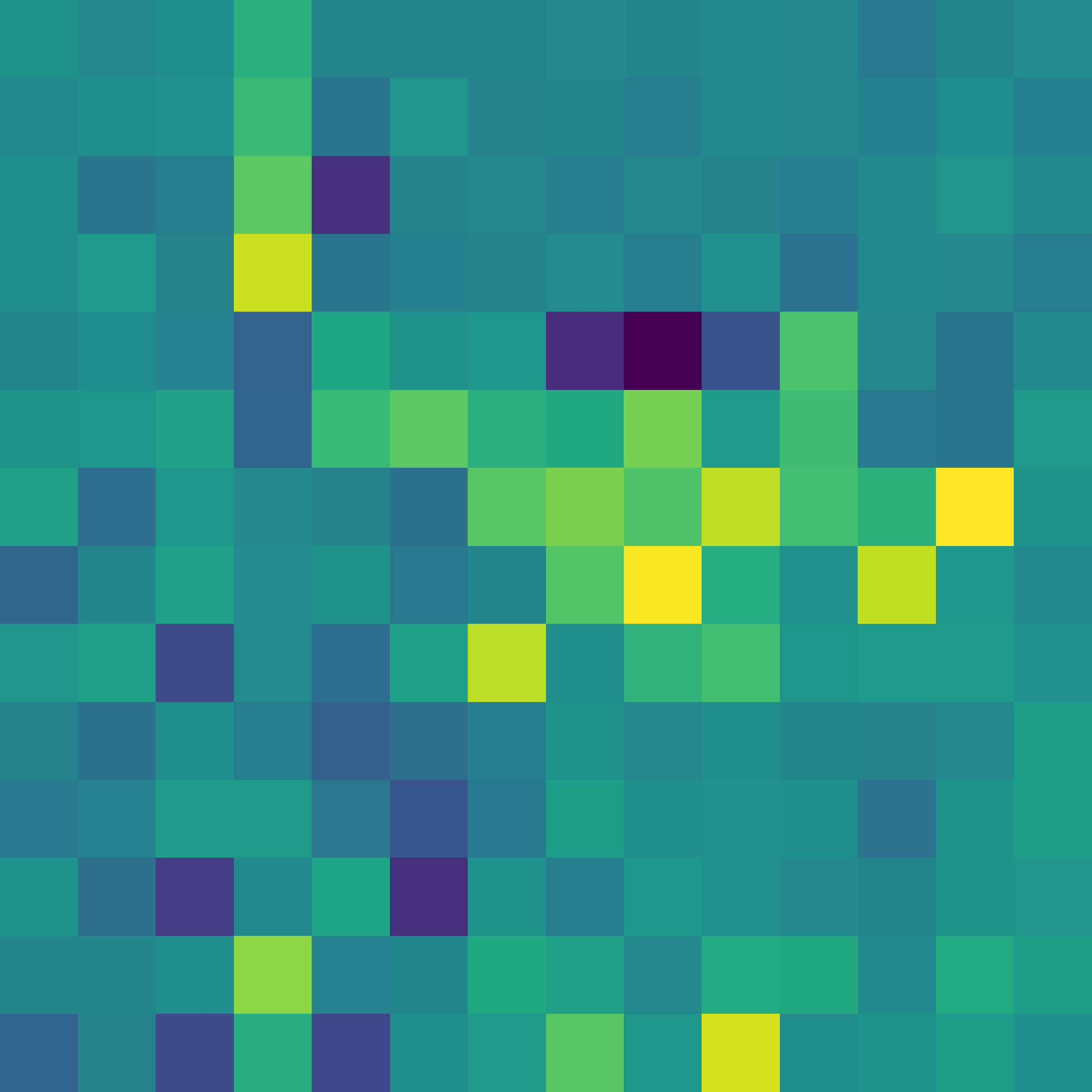}{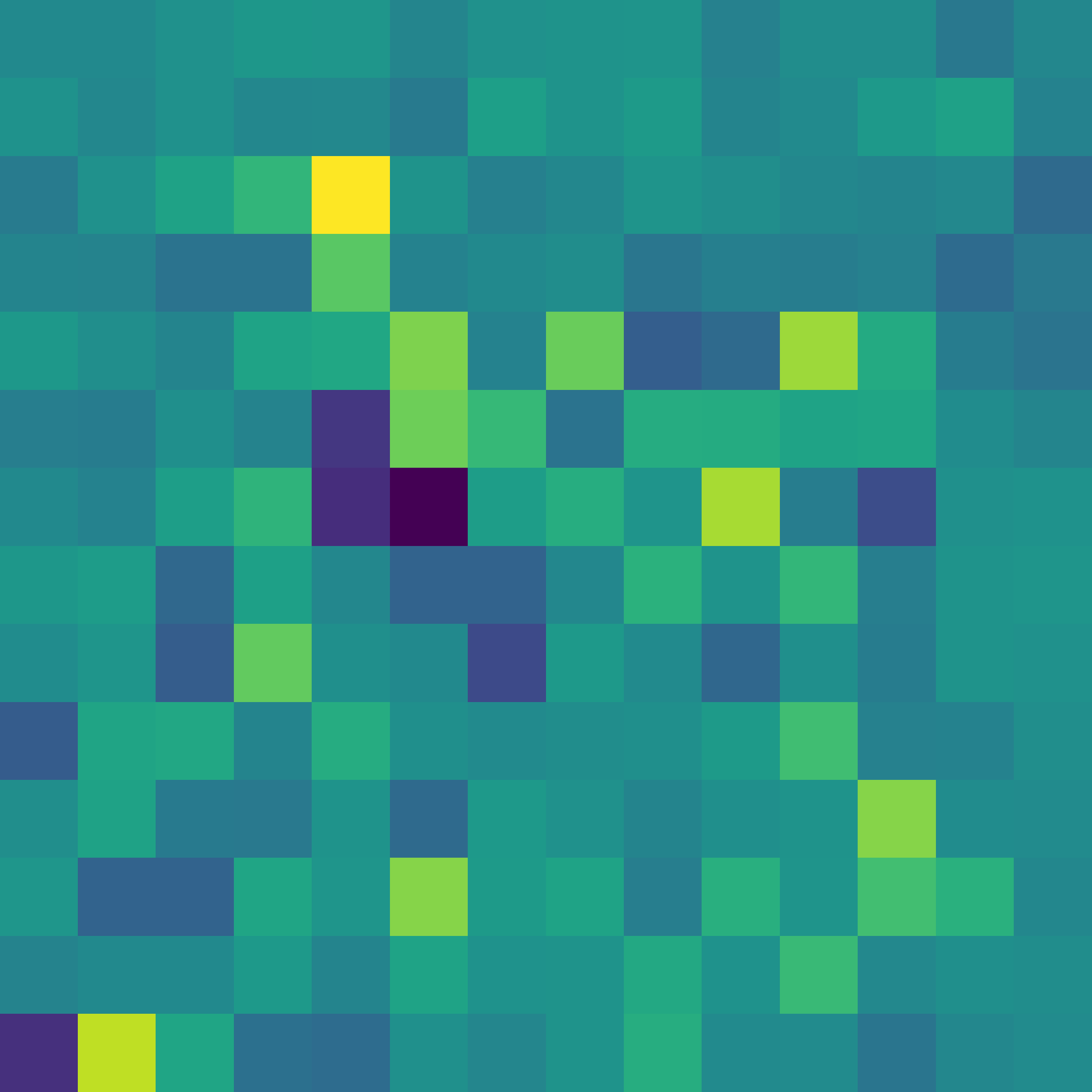}{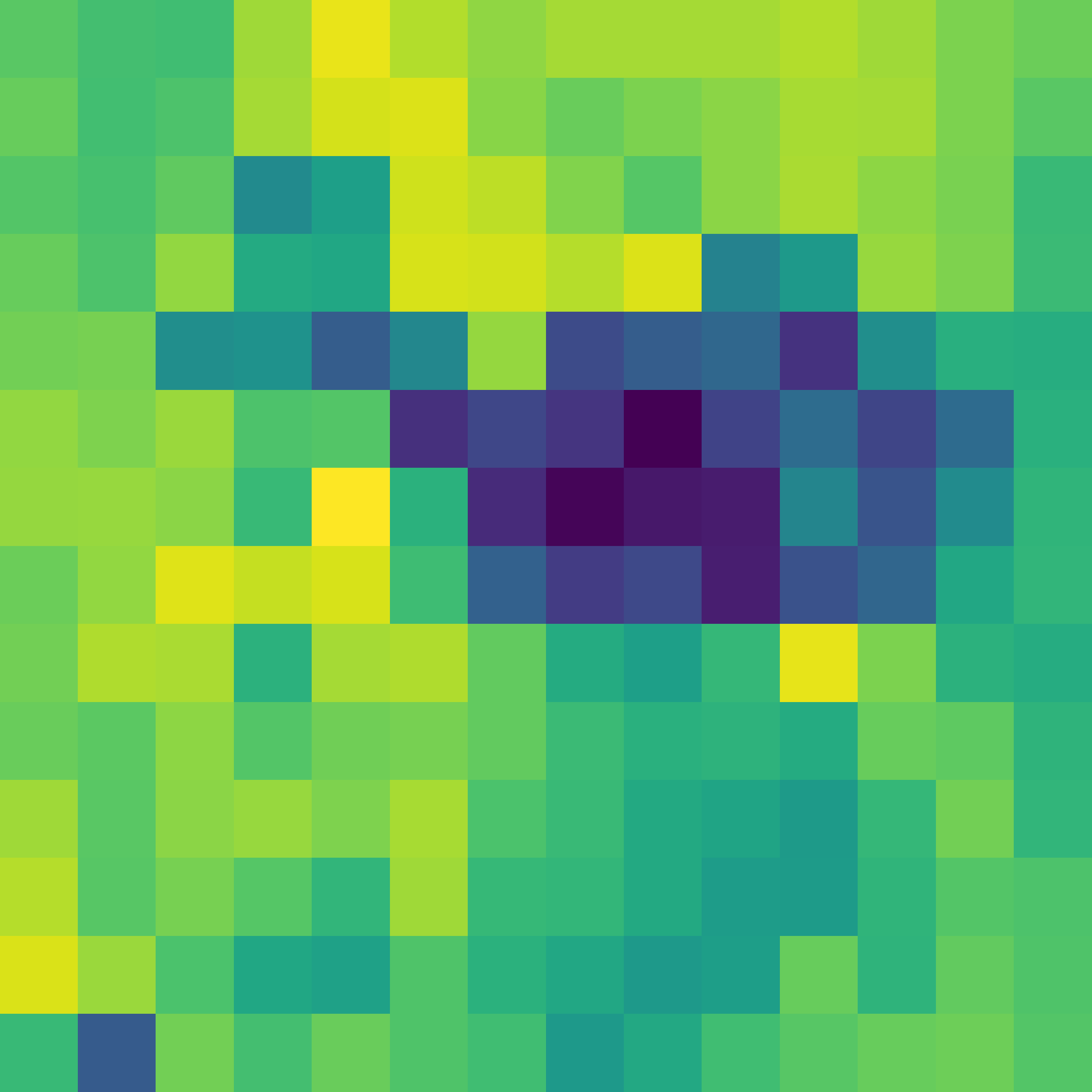}{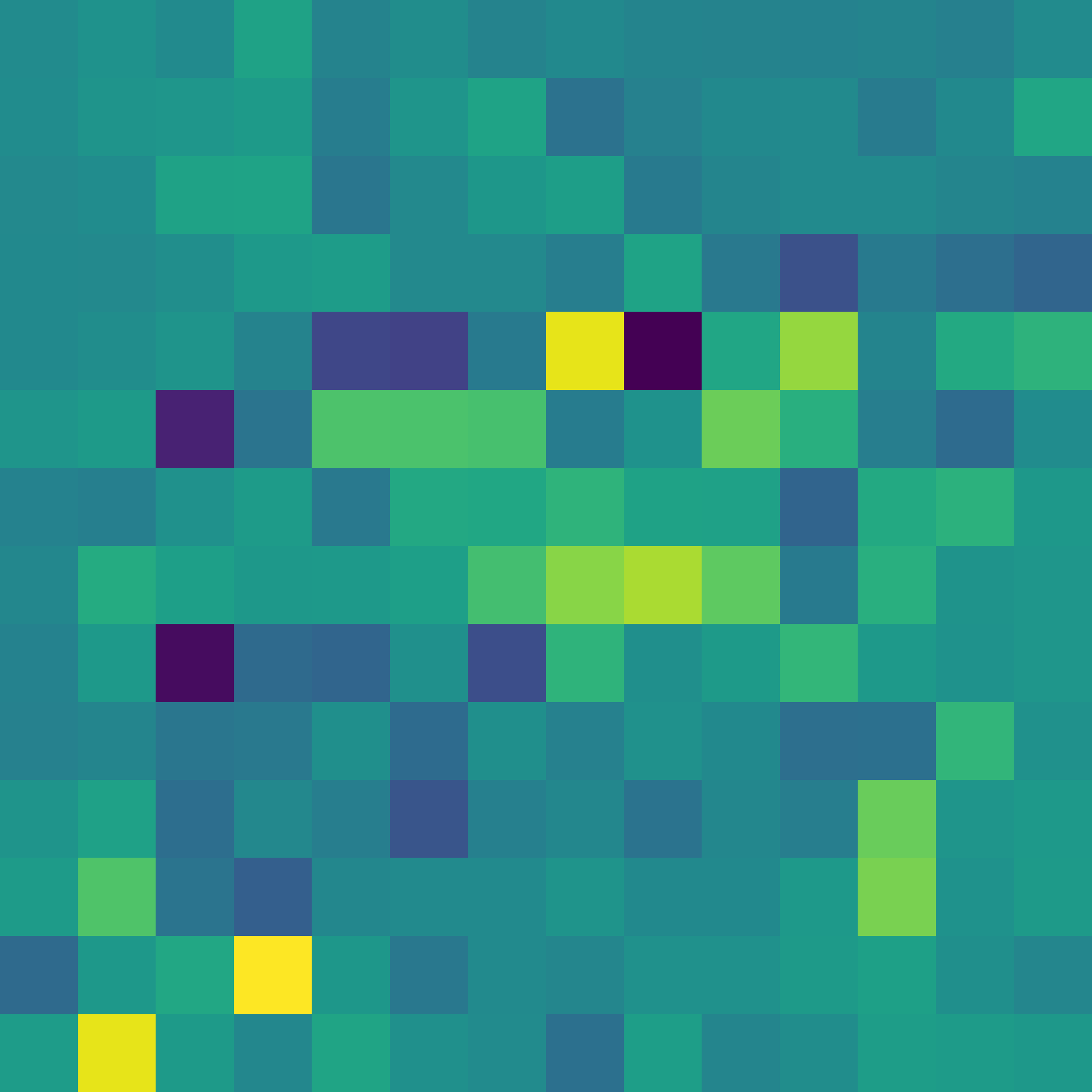} &
    \onebyfour{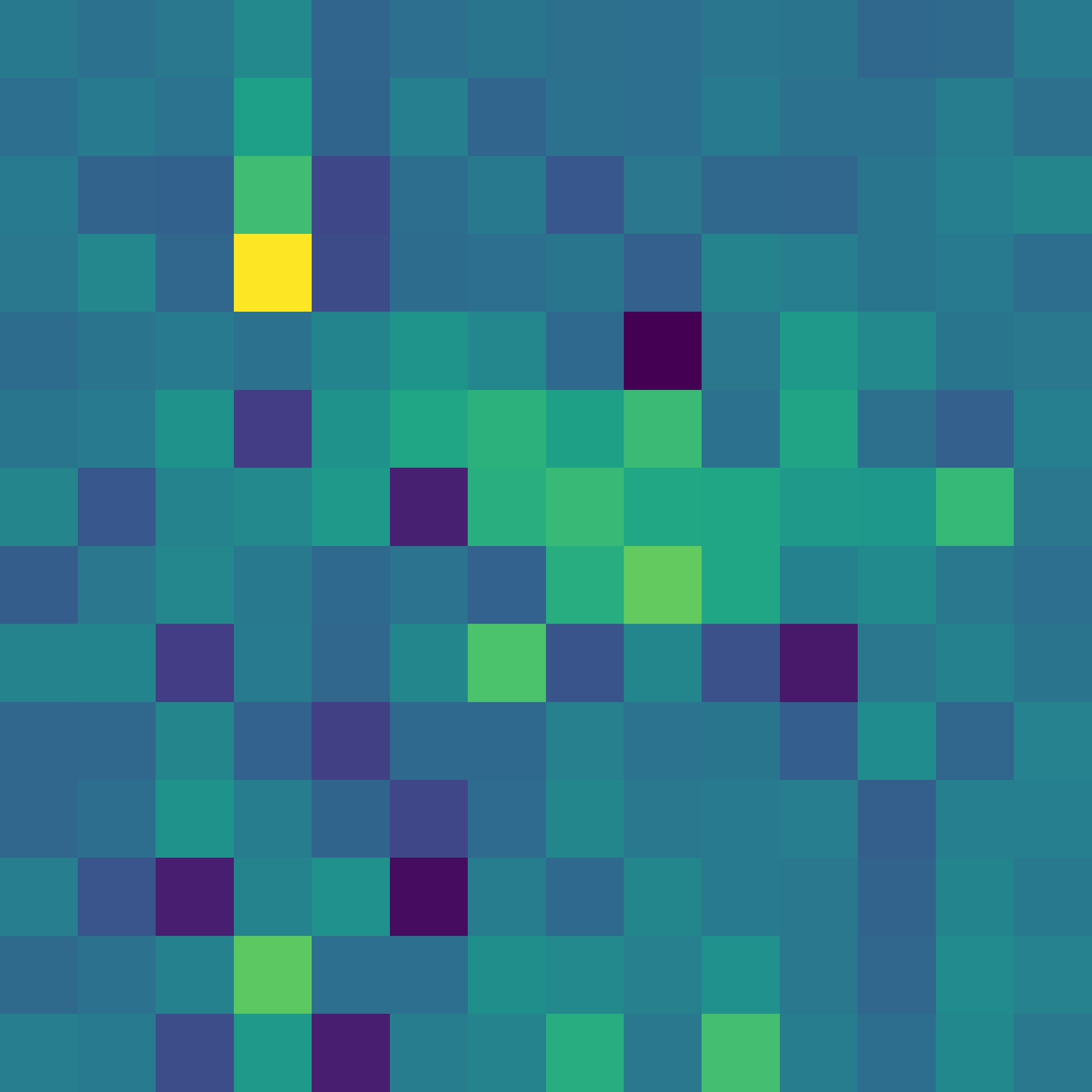}{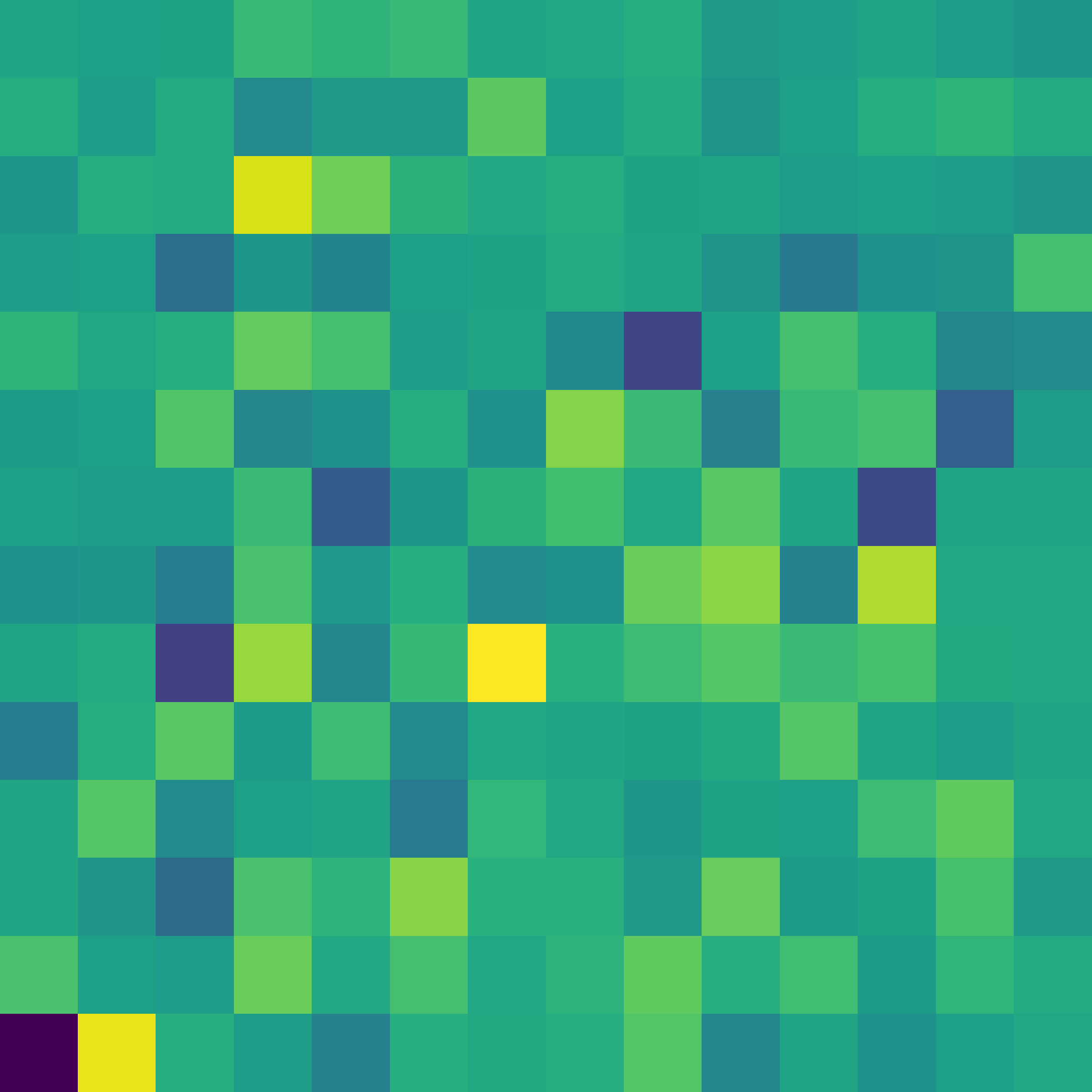}{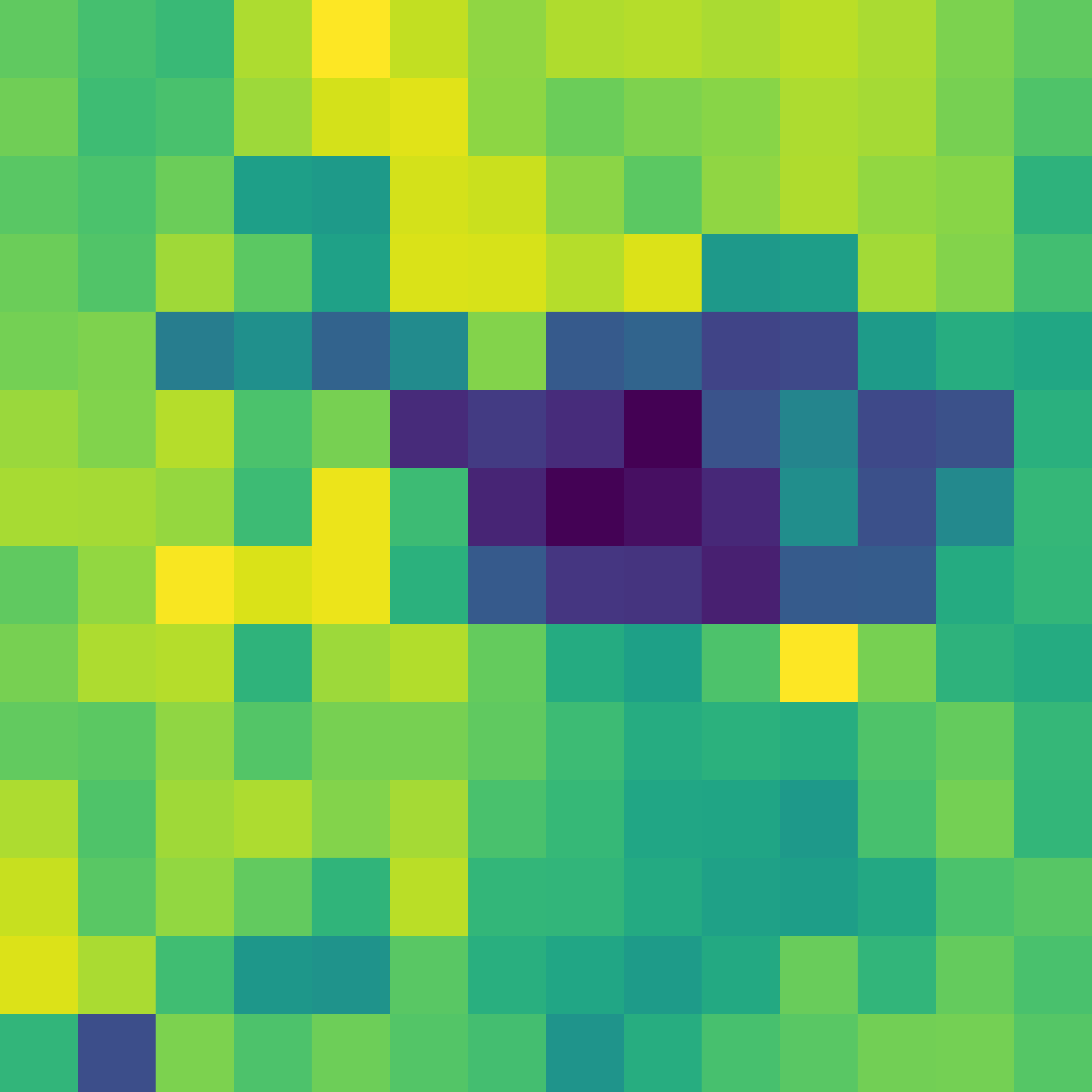}{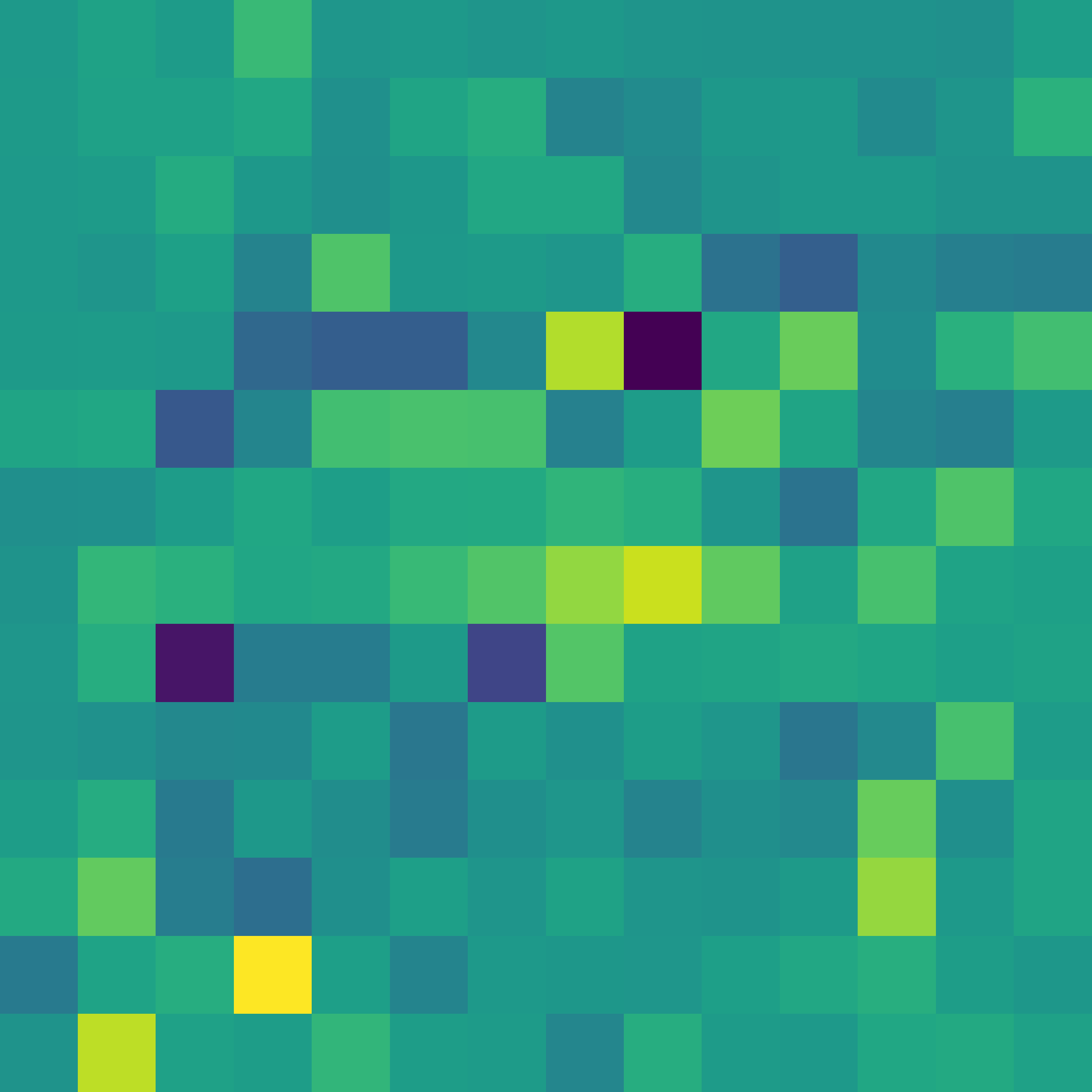} &
    \onebyfour{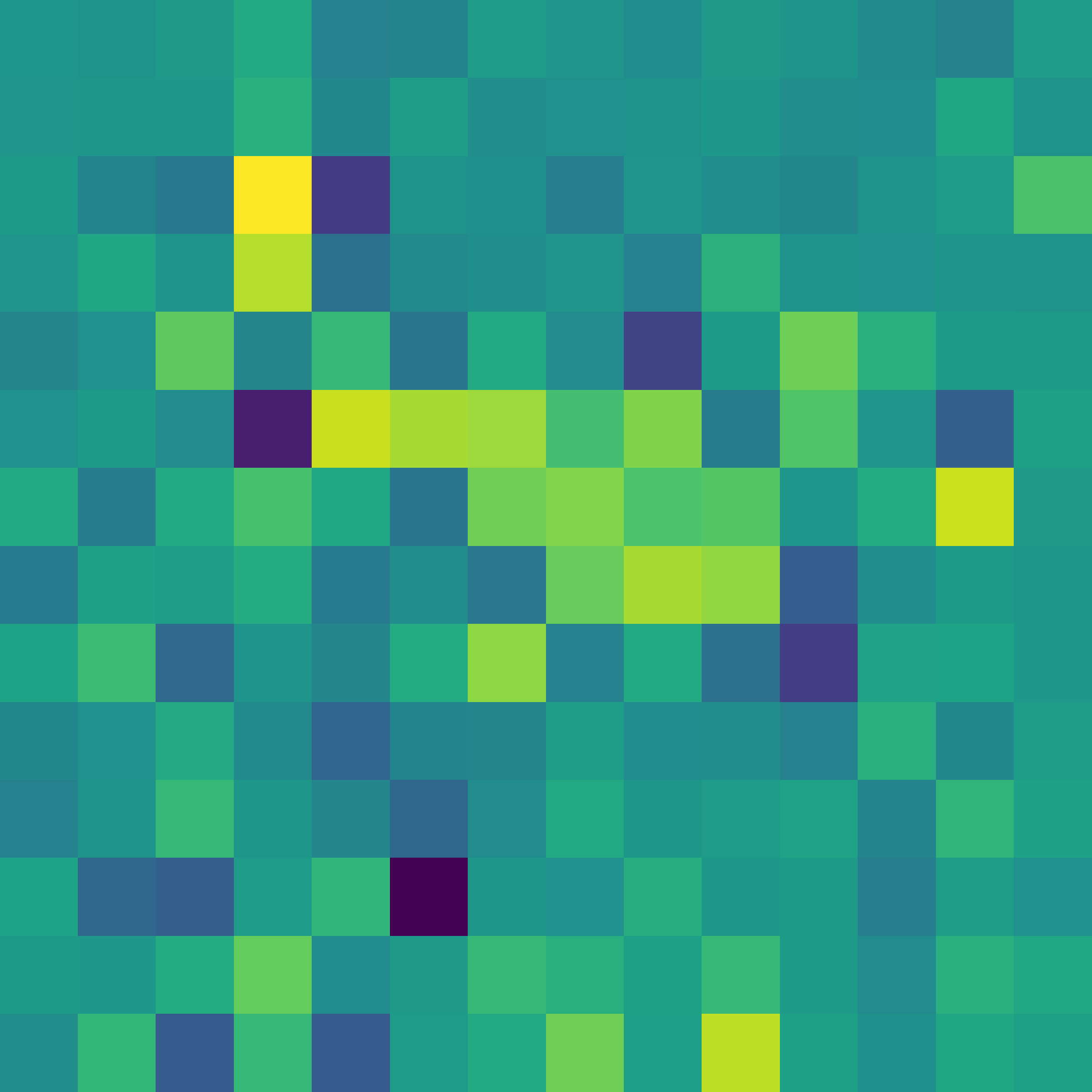}{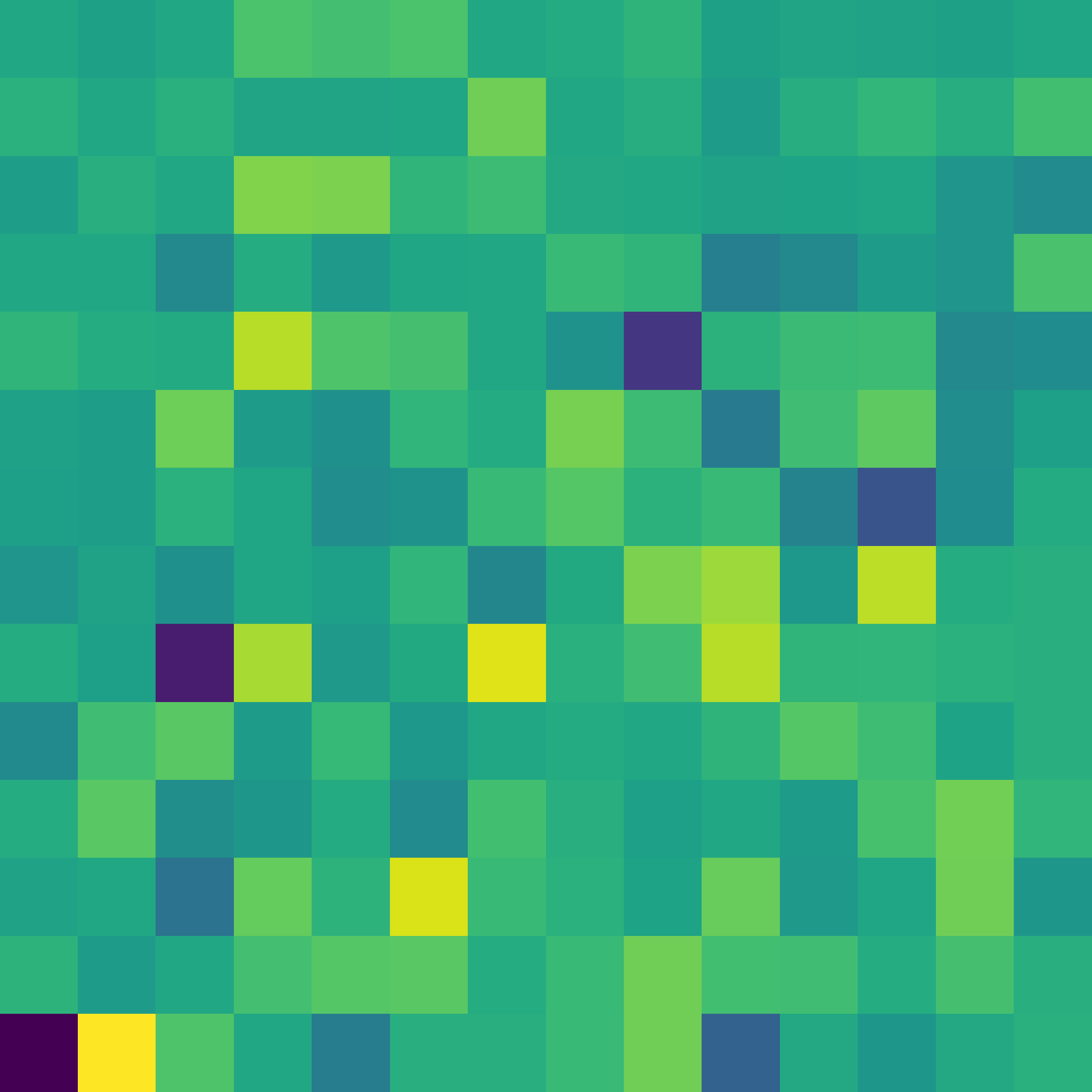}{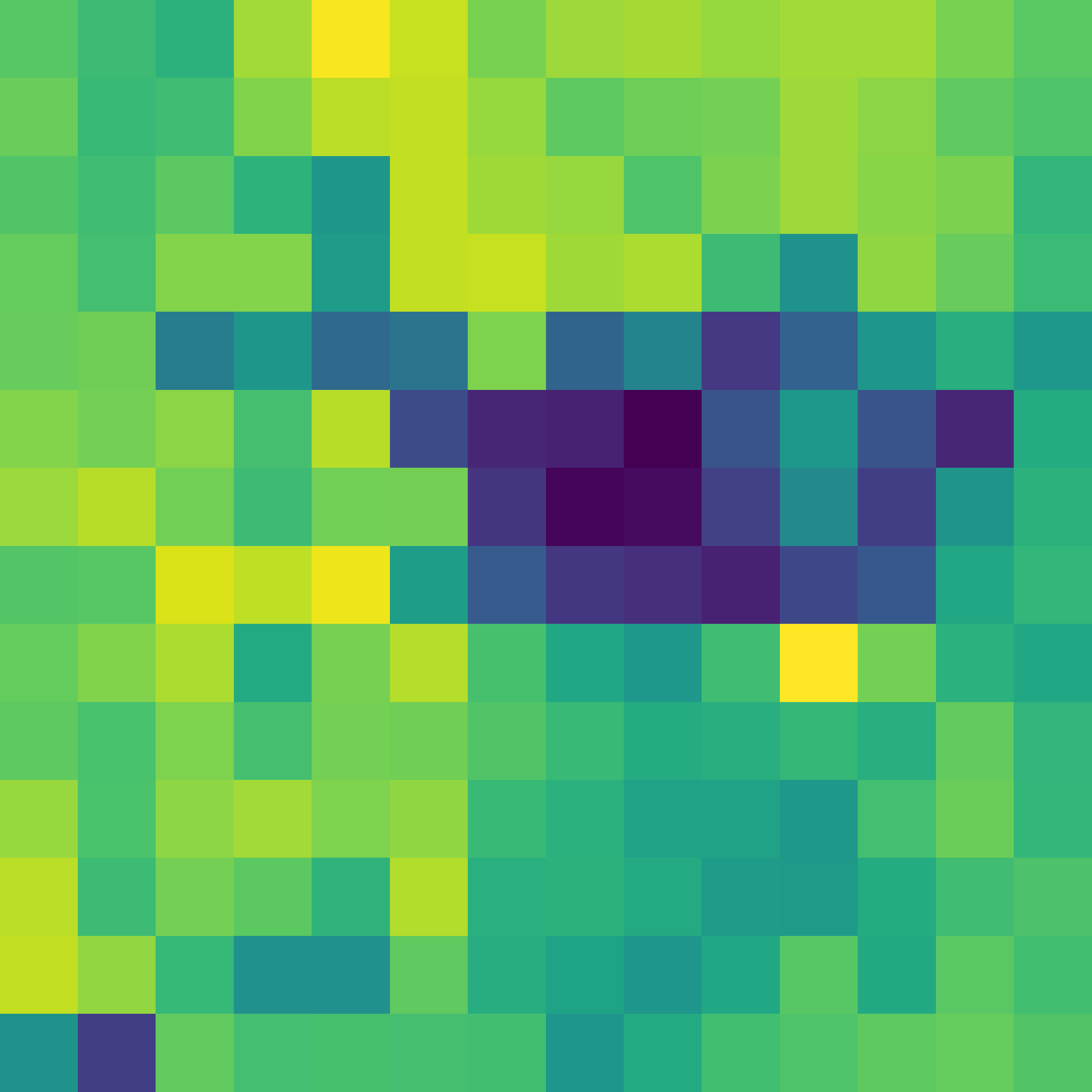}{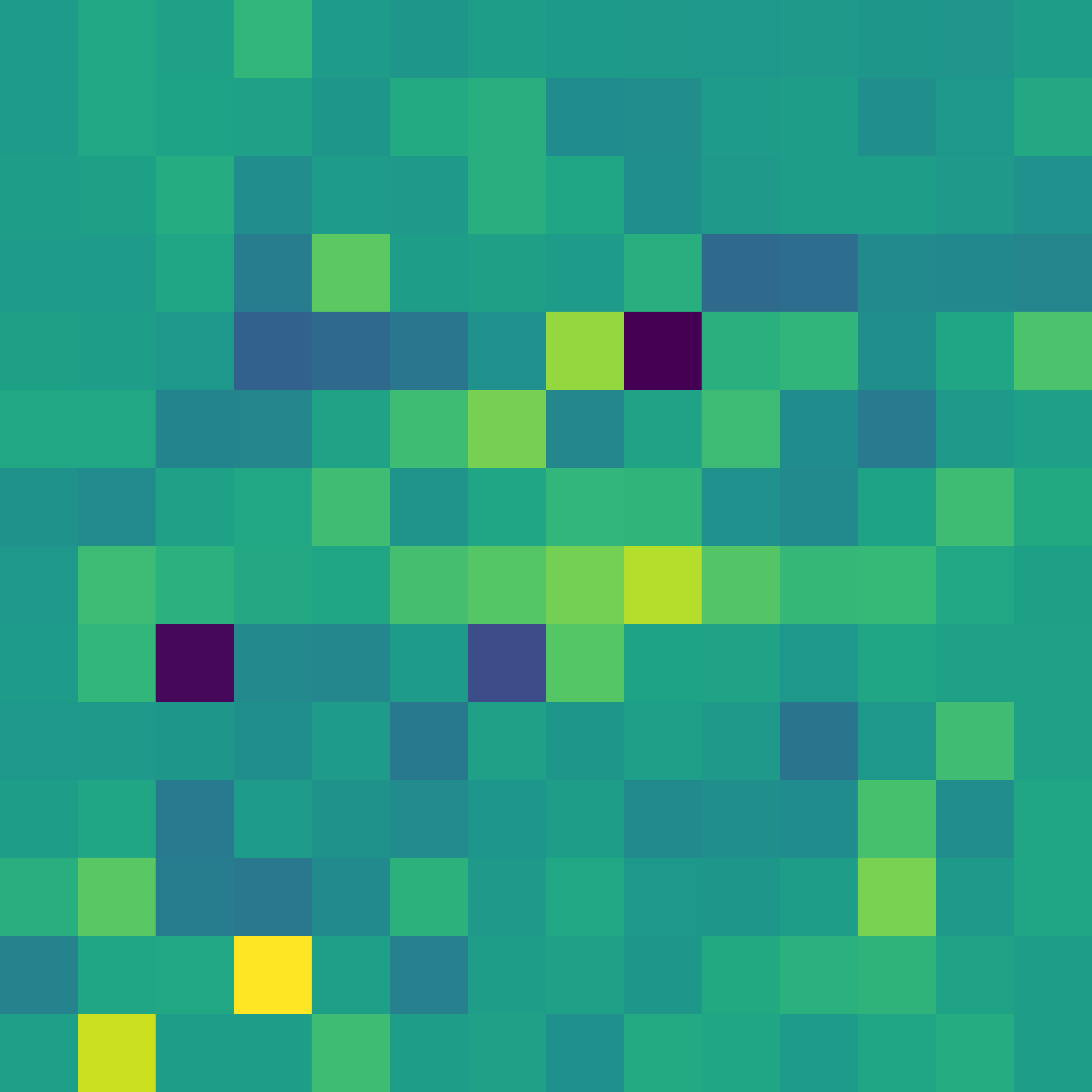} \\[0.4cm]
    \begin{minipage}{0.065\textwidth}\centering\includegraphics[width=0.91\linewidth]{figs/nrot4_v2/reference_images/3043284246_cb2b545a43_z.png}\end{minipage} & \onebyfour{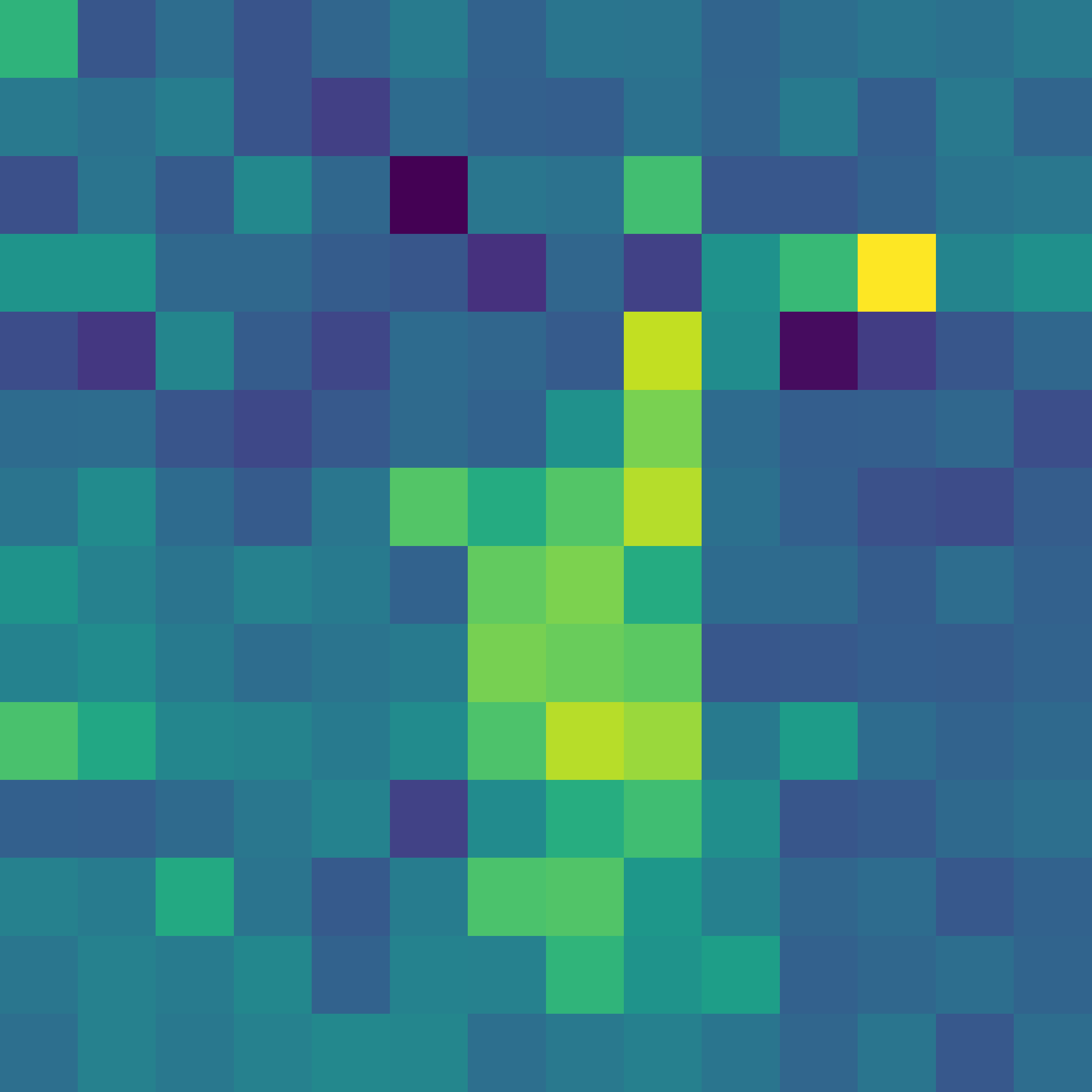}{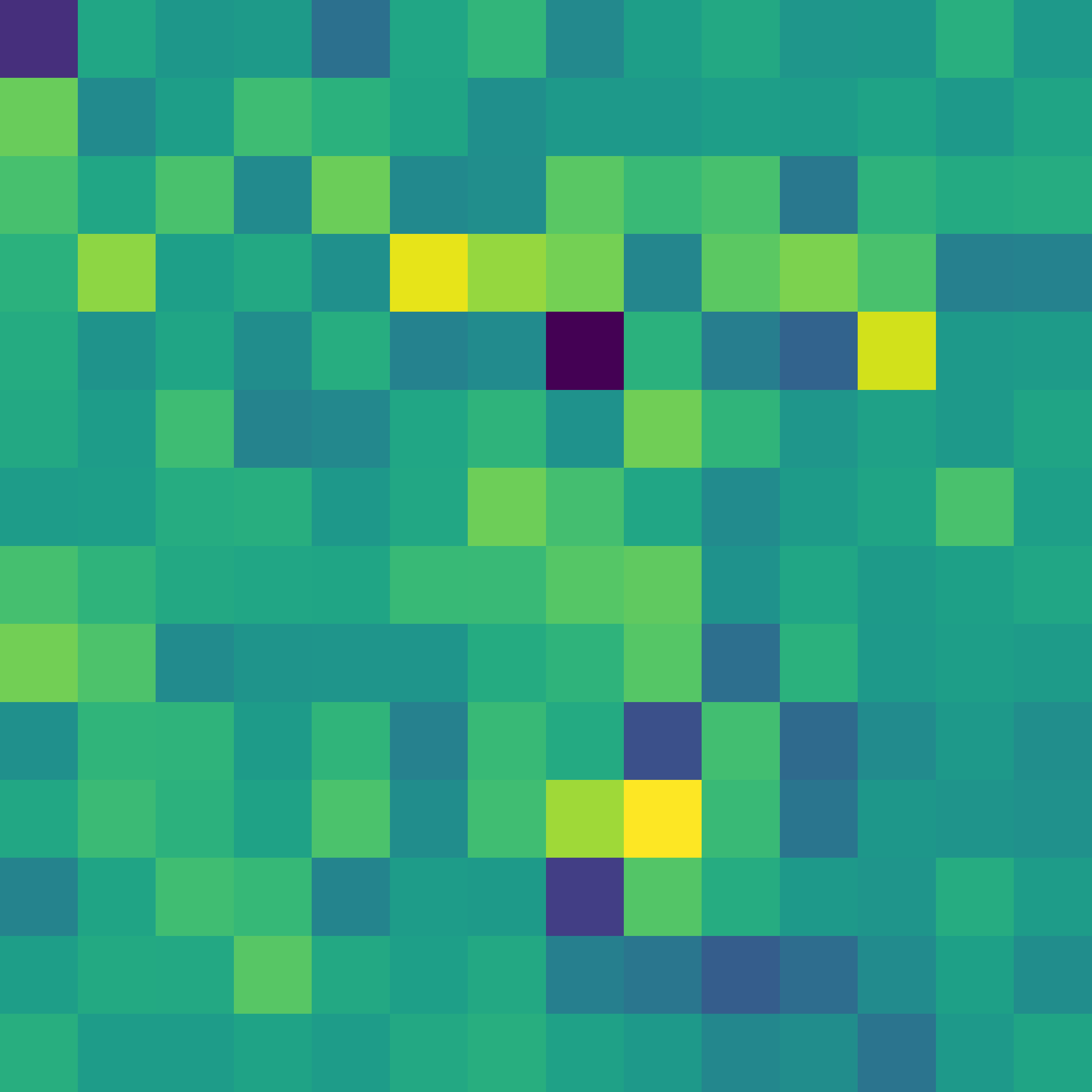}{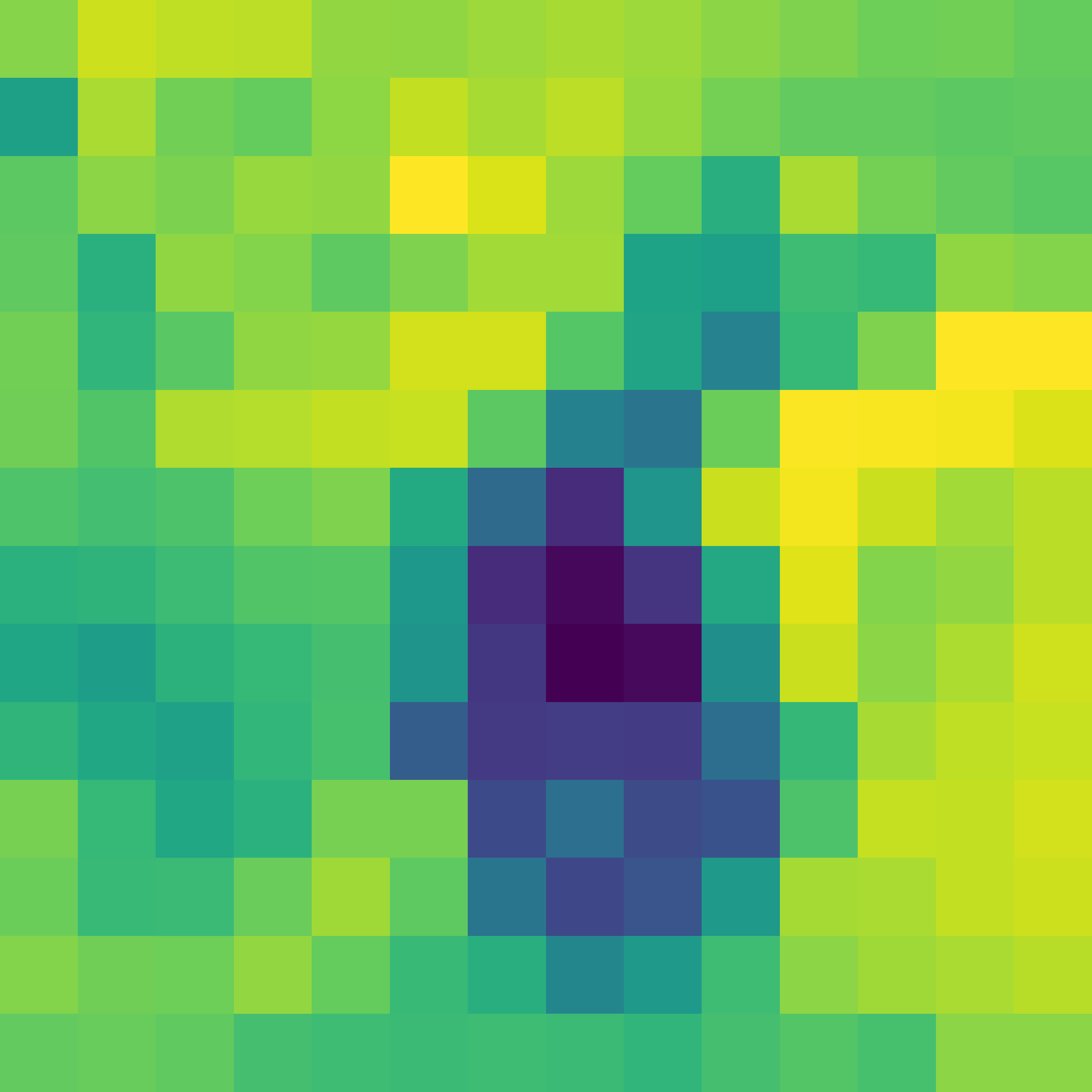}{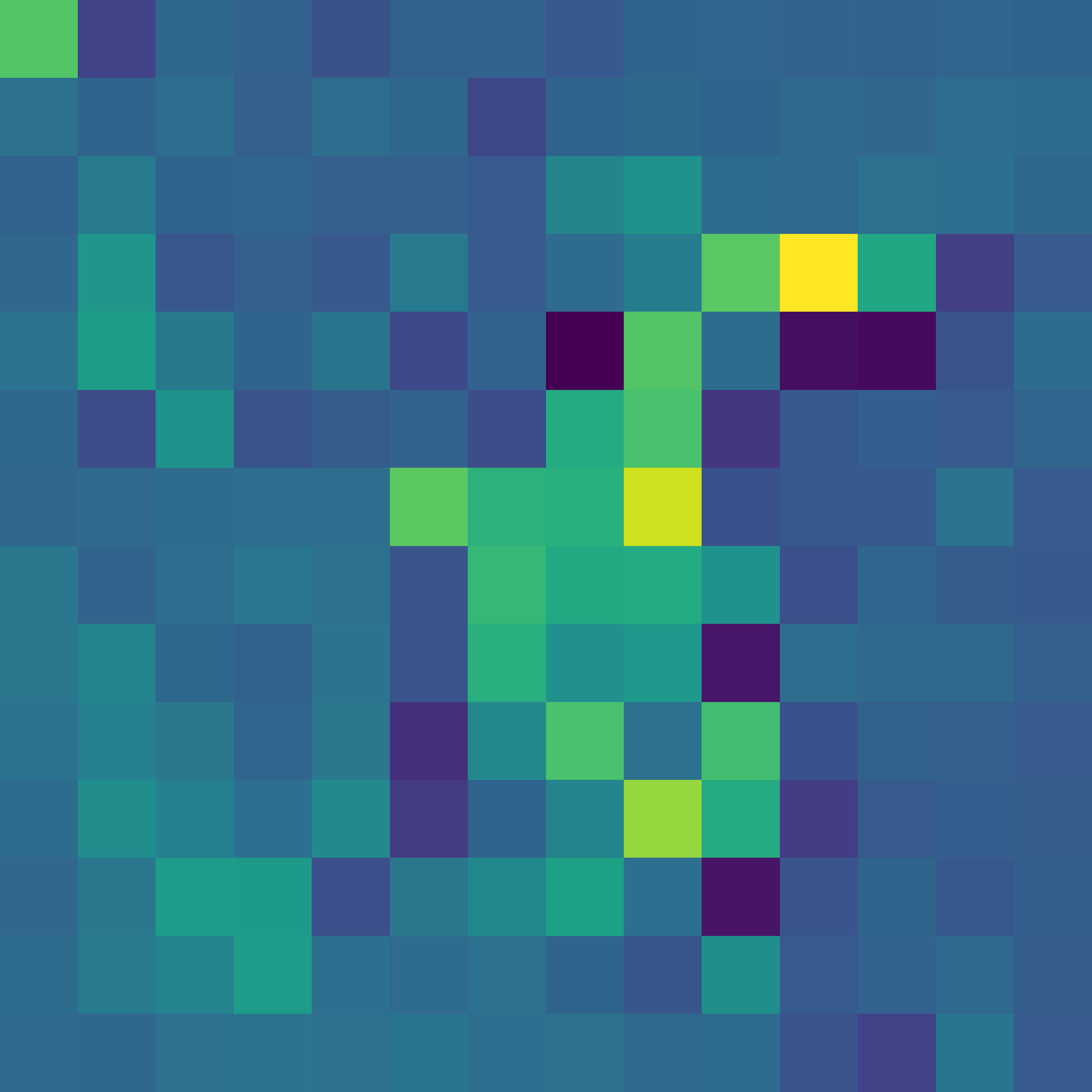} &
    \onebyfour{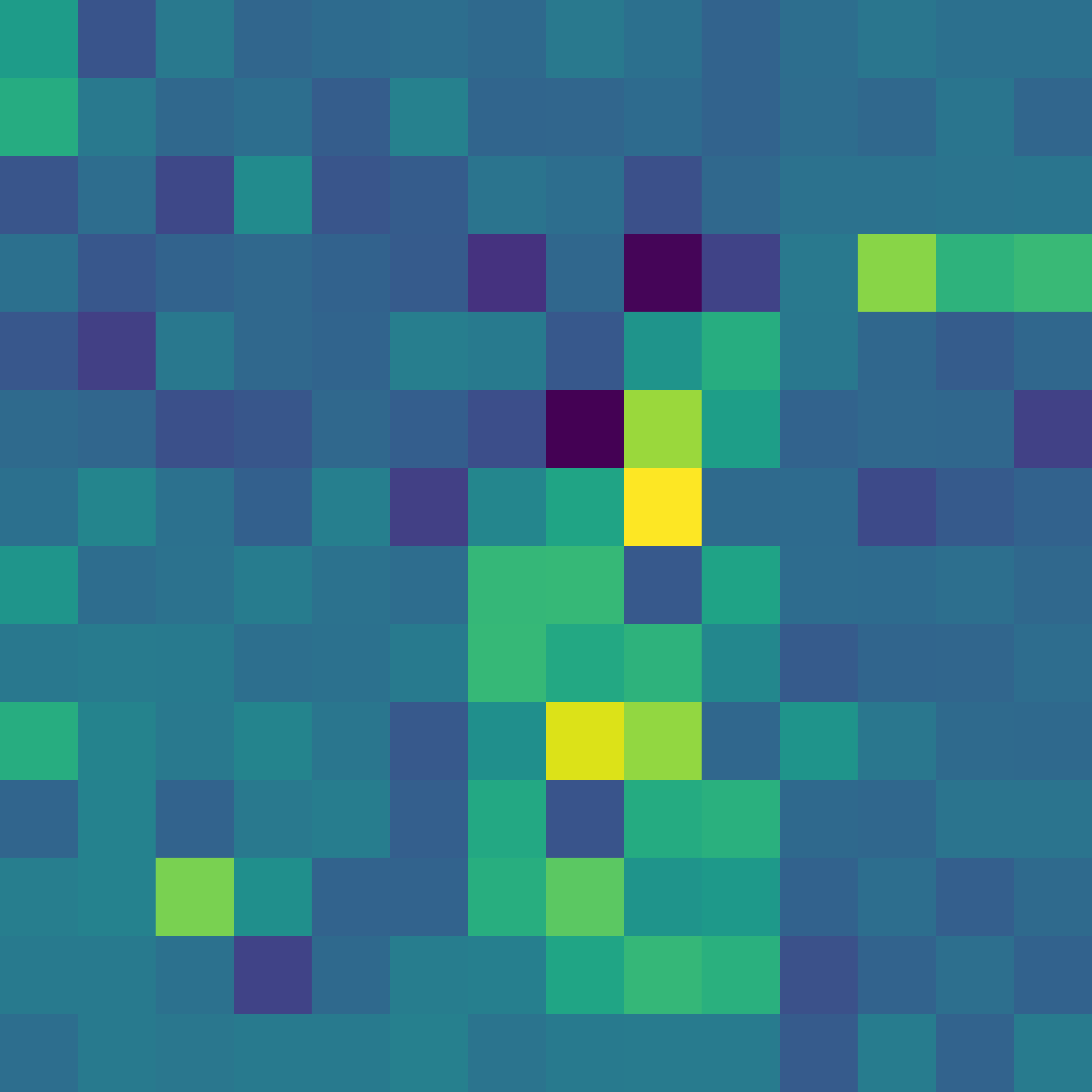}{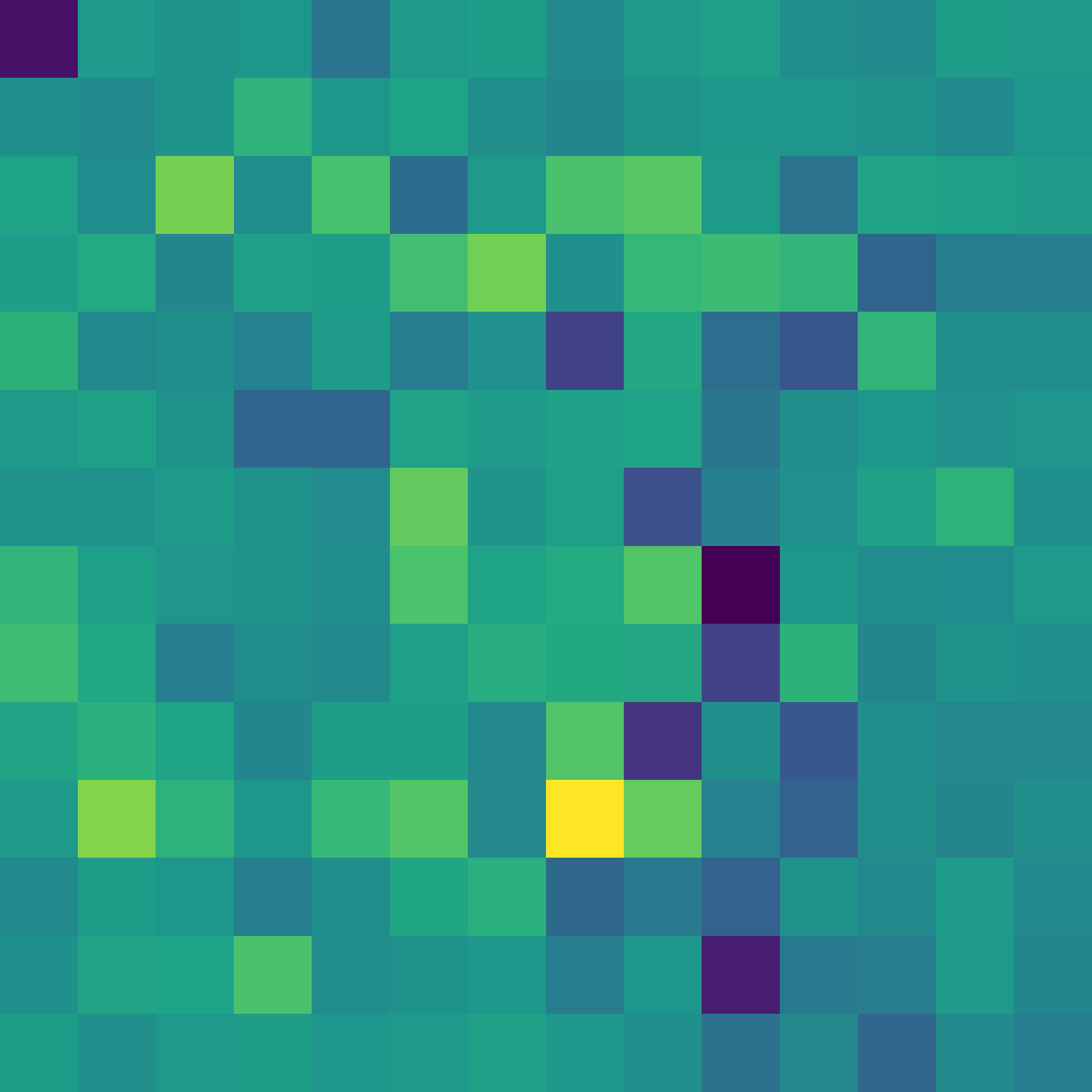}{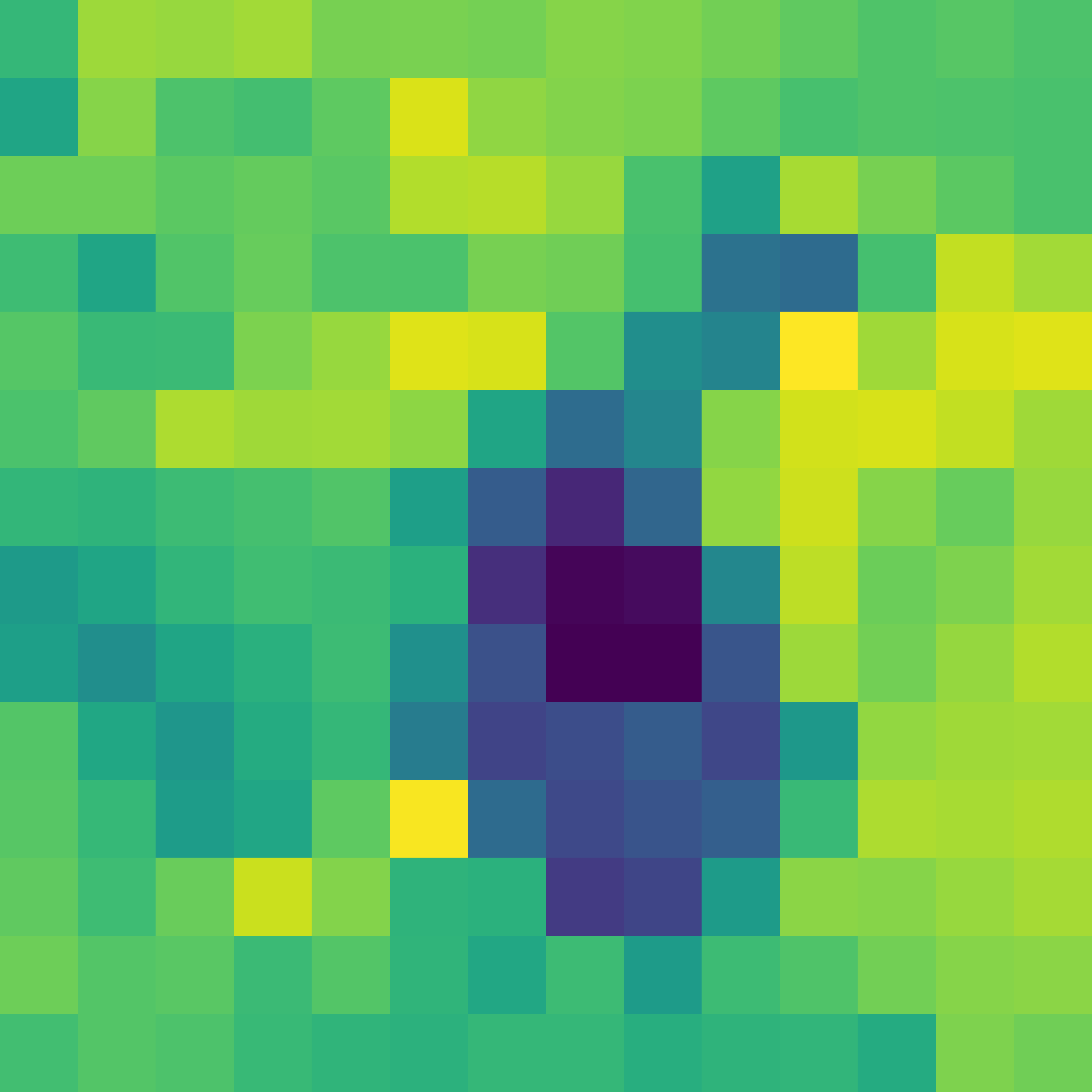}{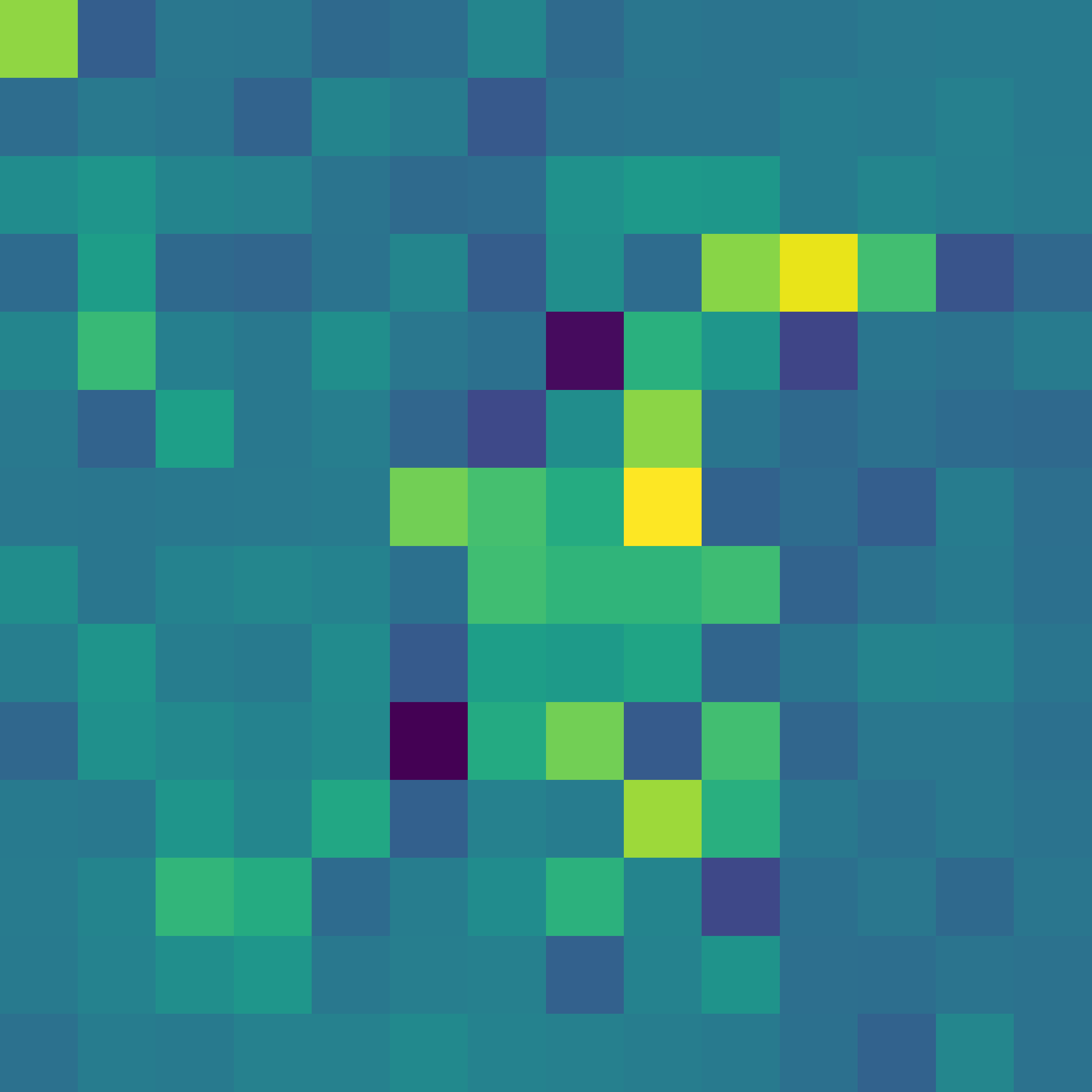} &
    \onebyfour{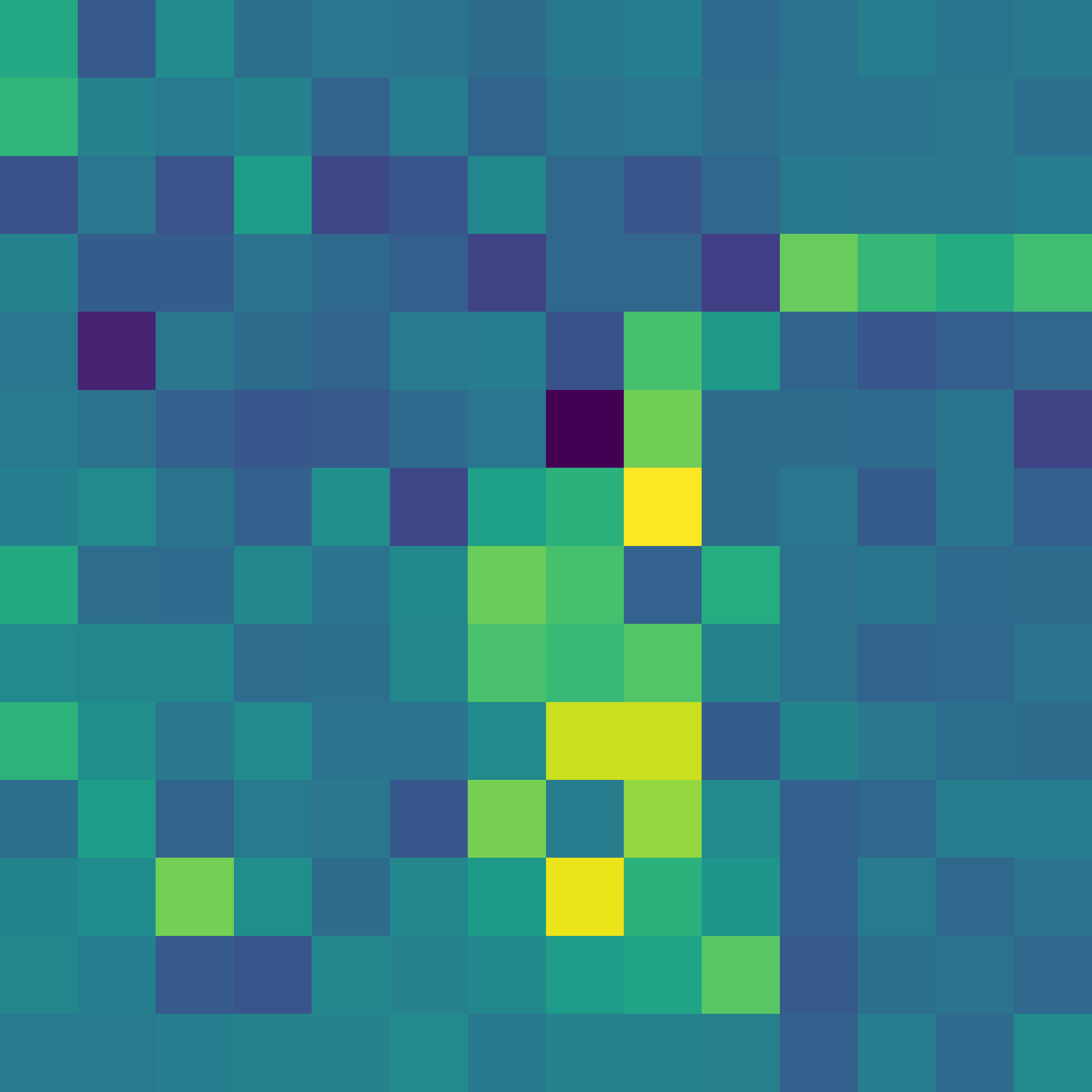}{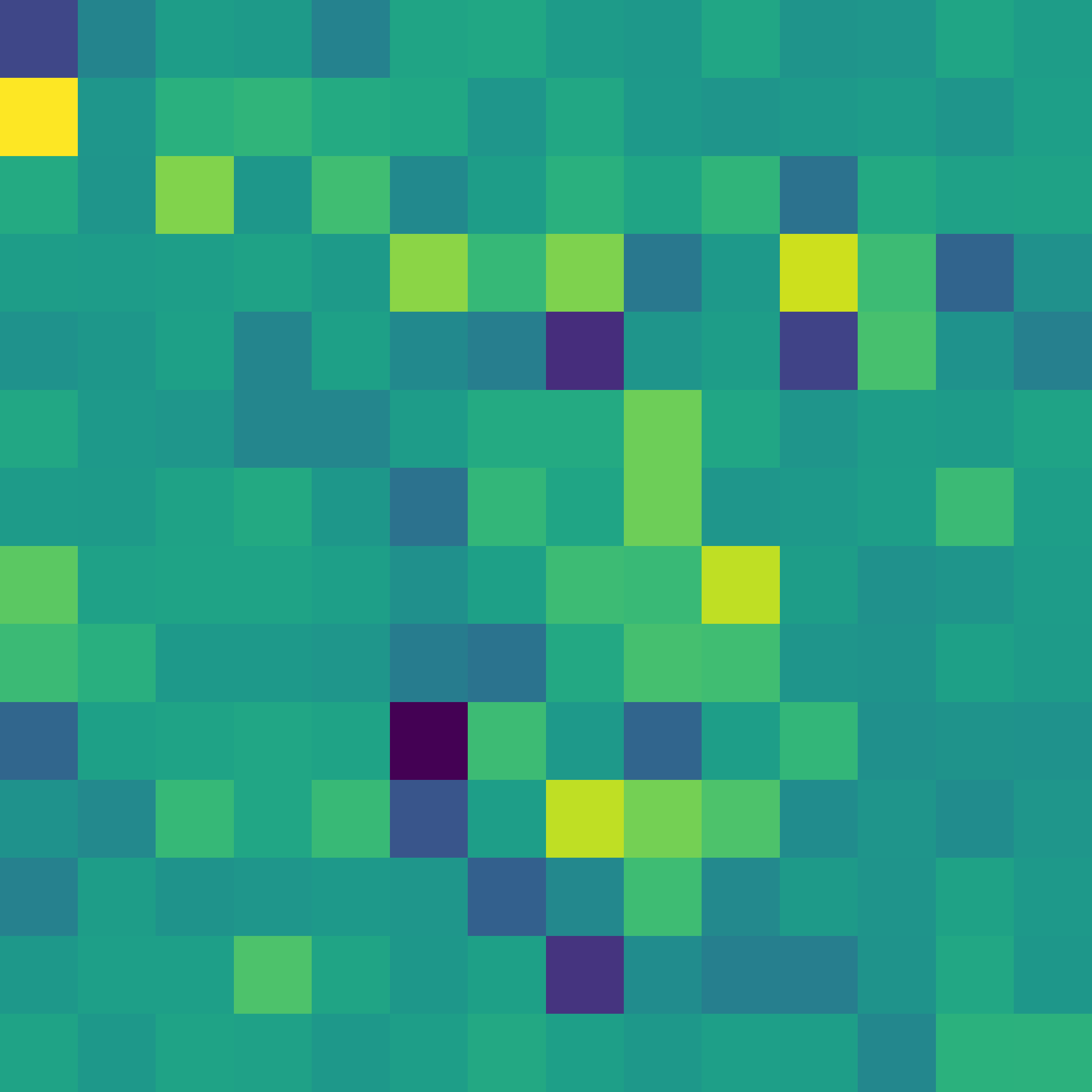}{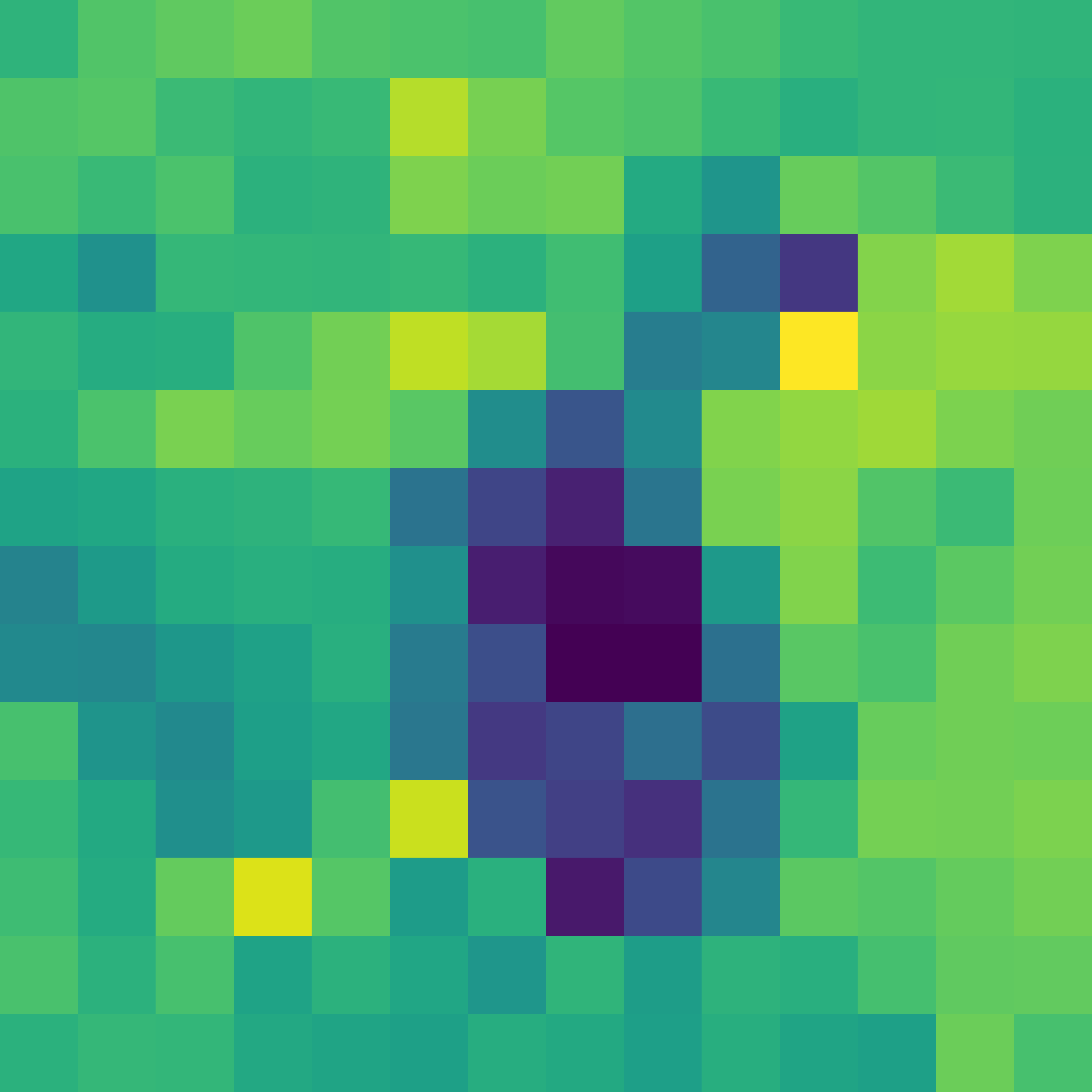}{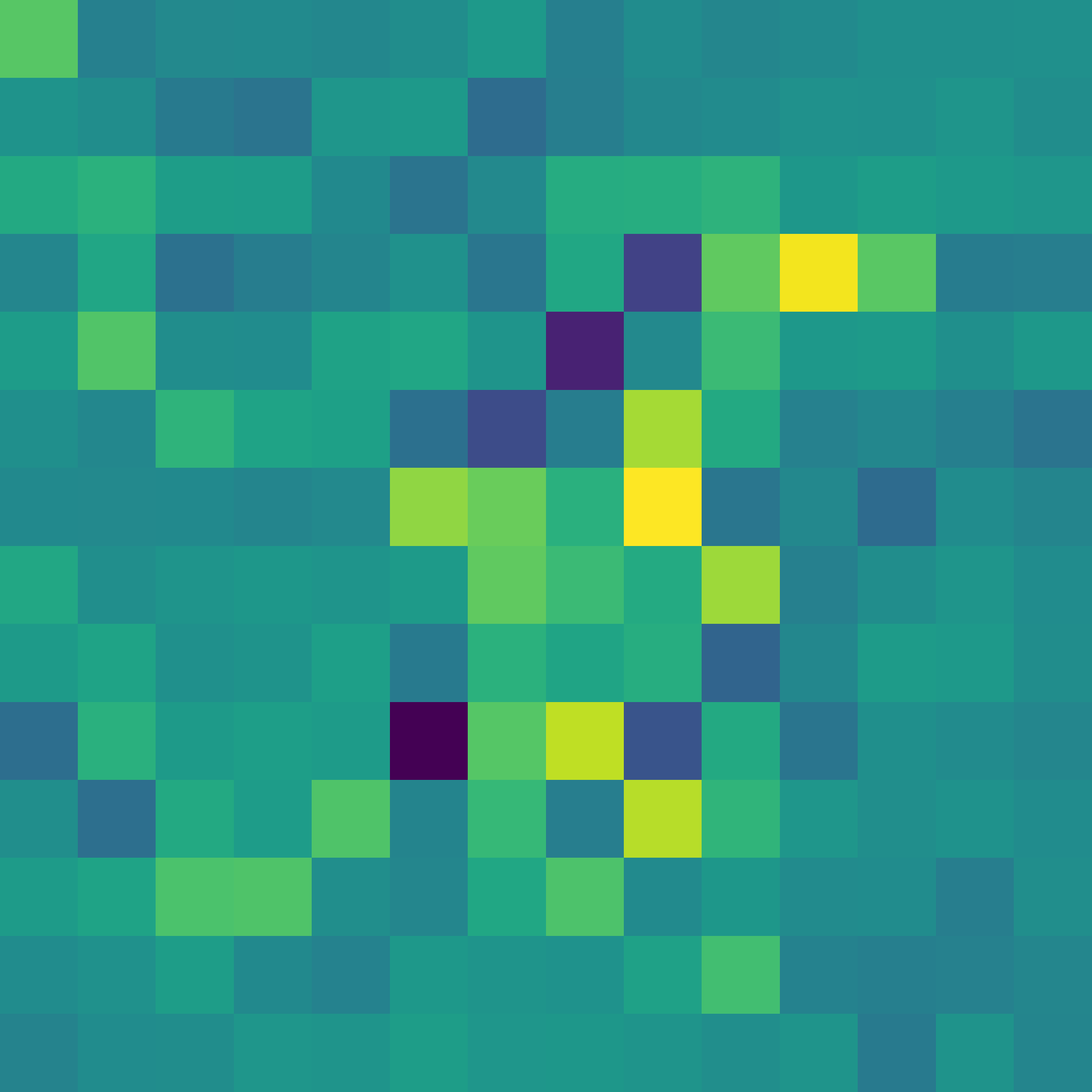} &
    \onebyfour{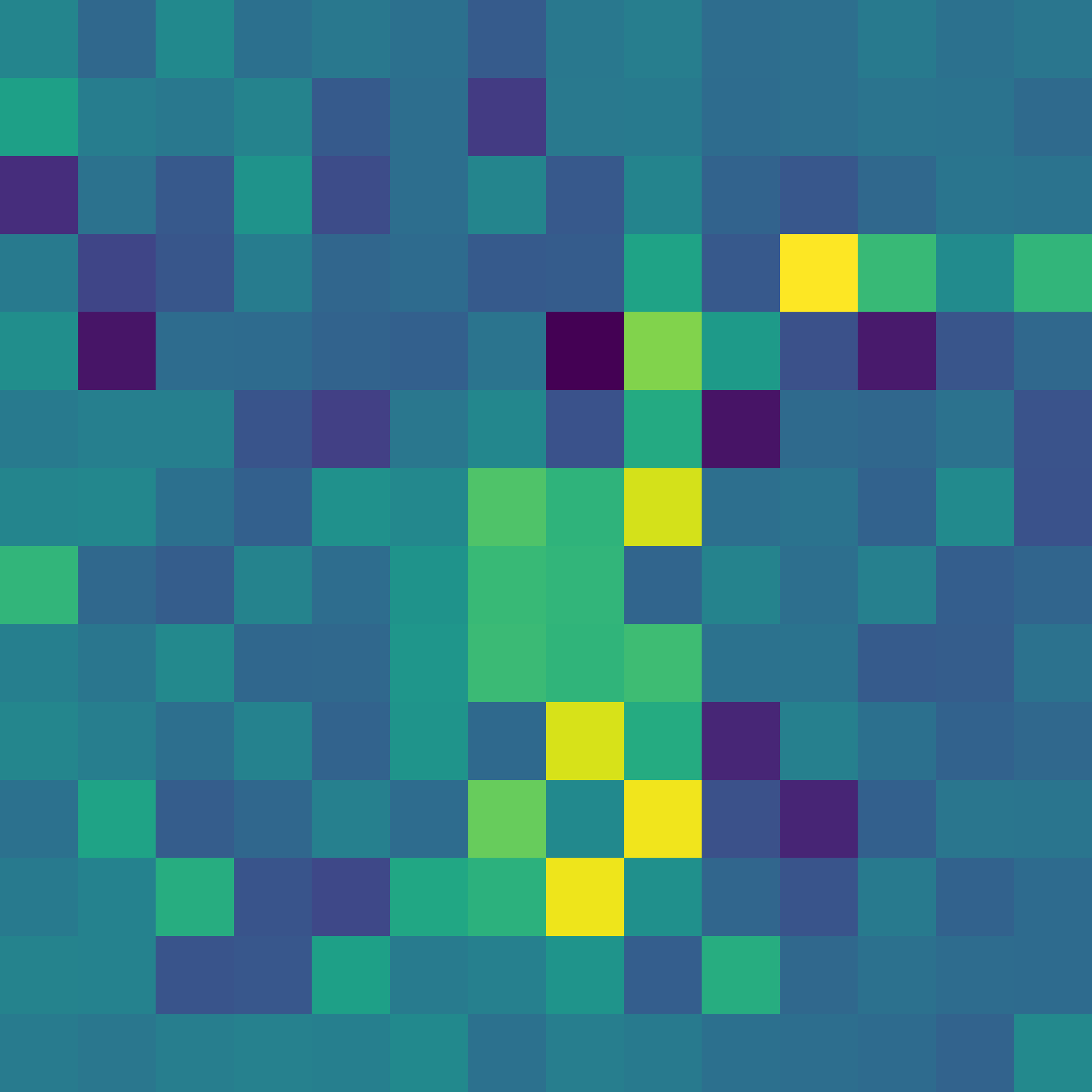}{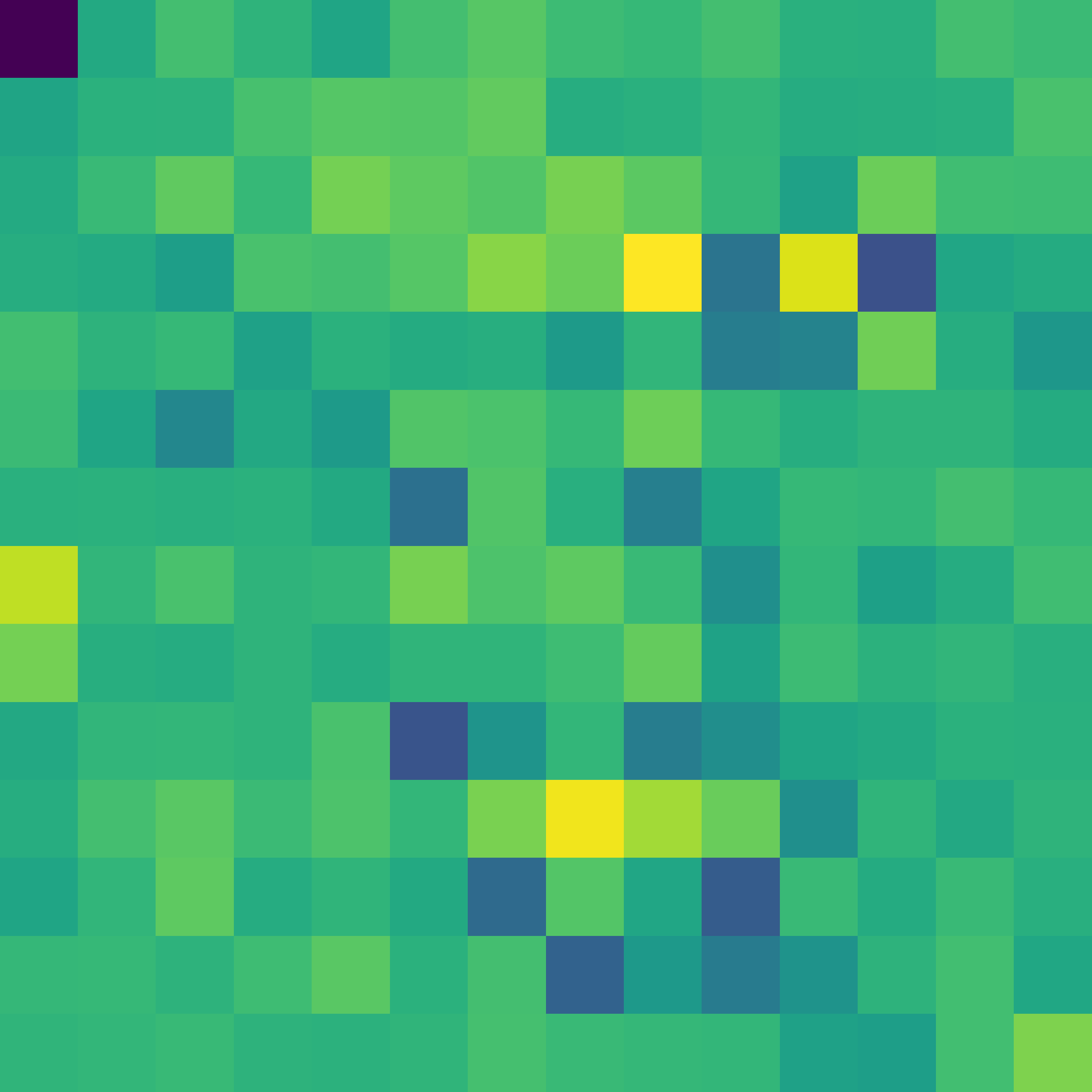}{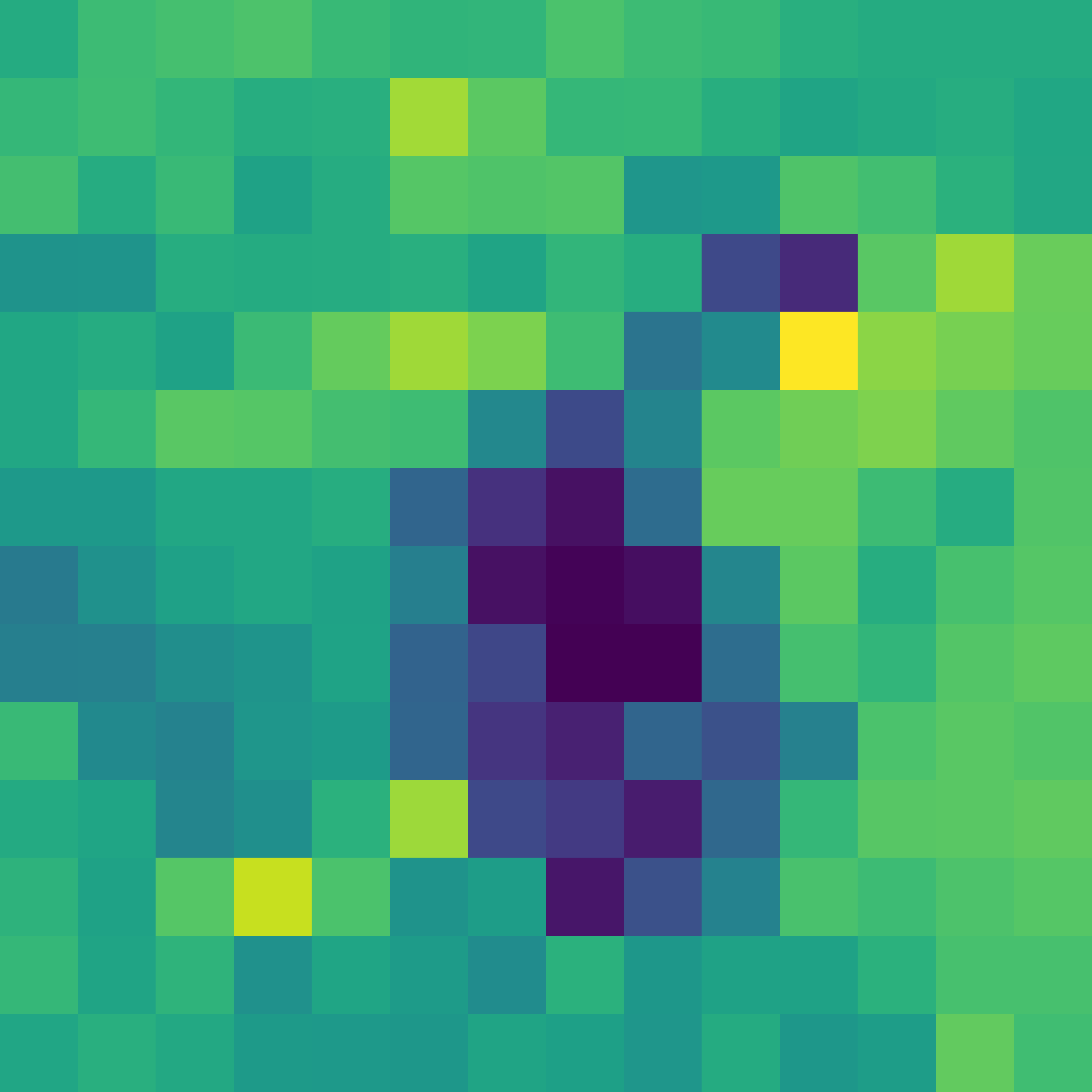}{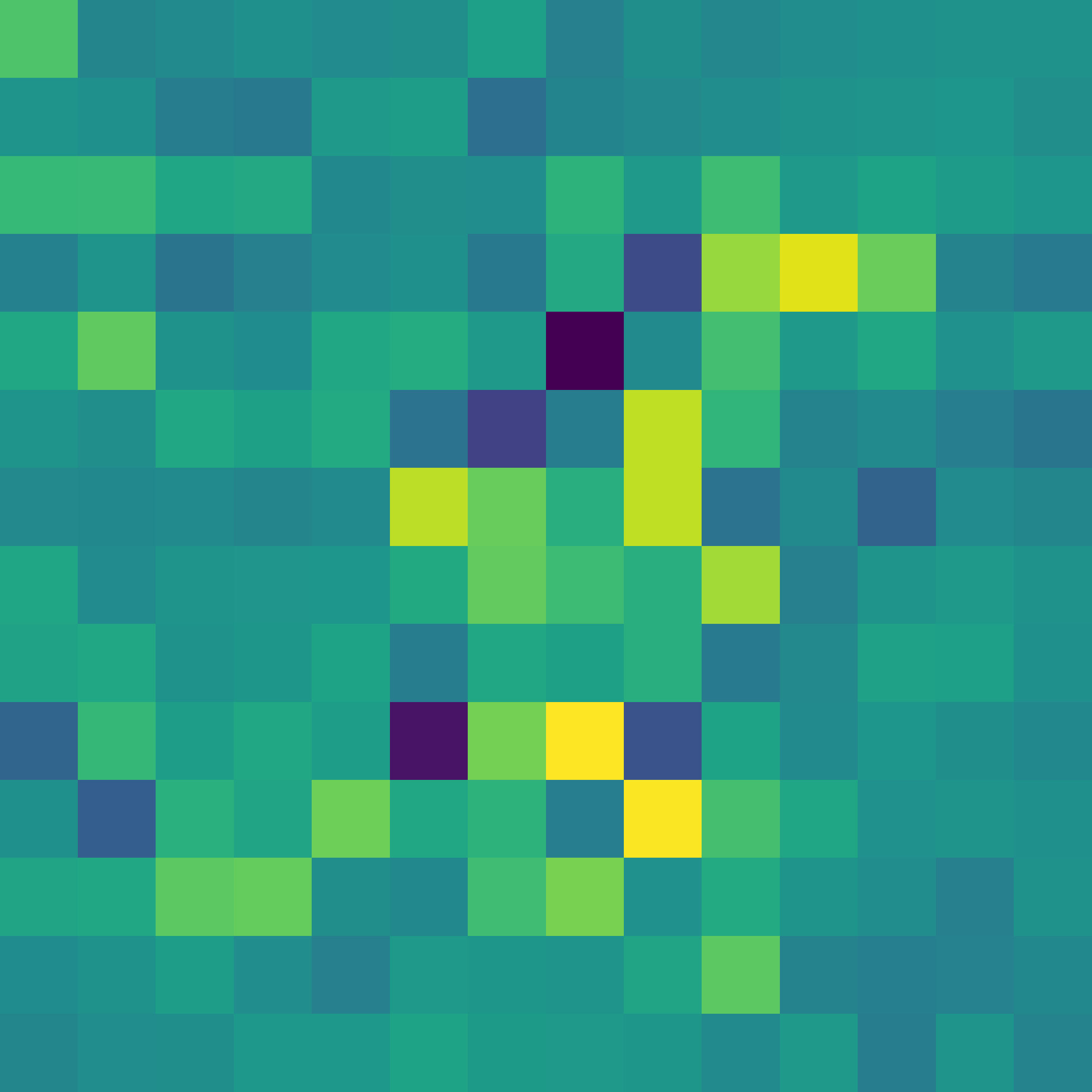} &
    \onebyfour{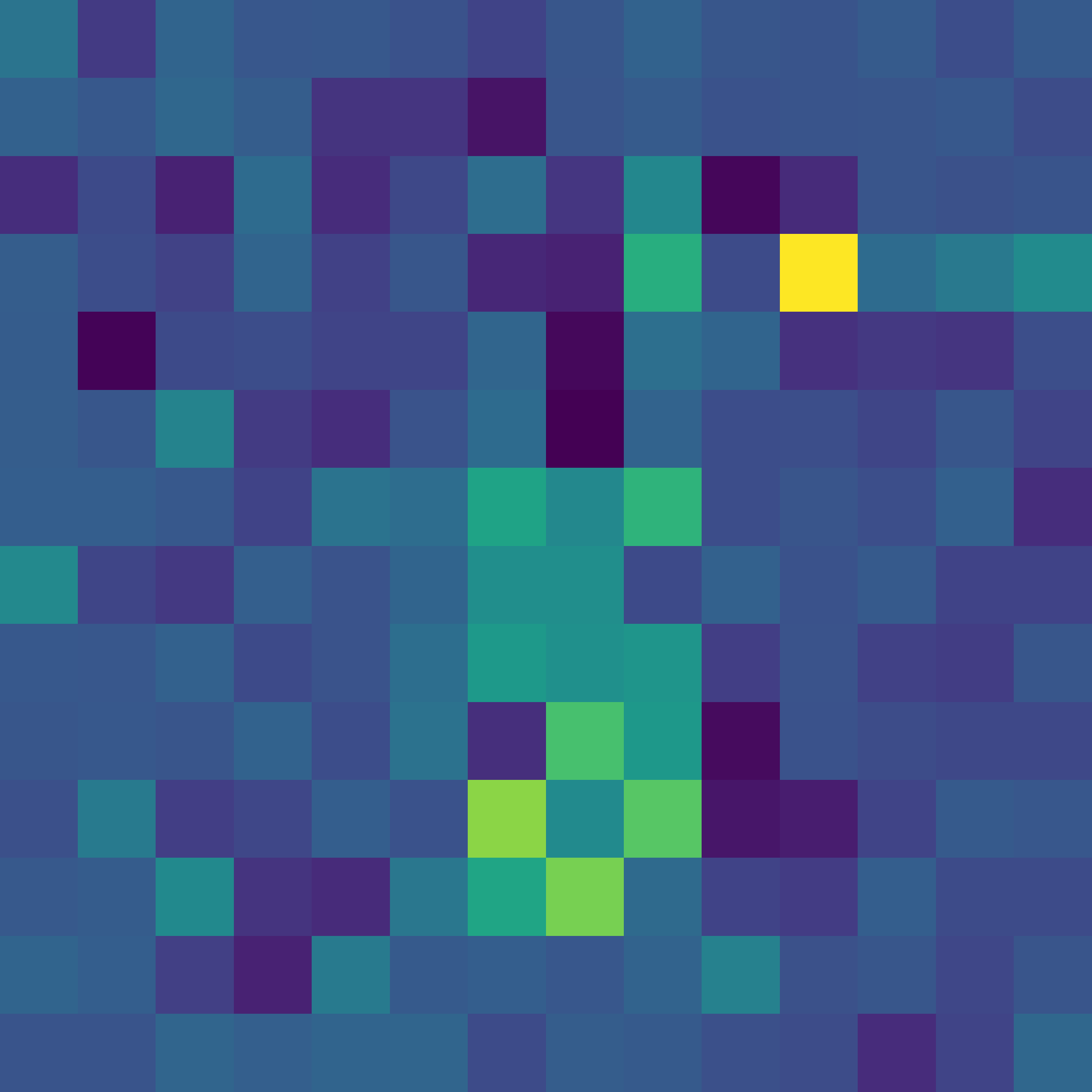}{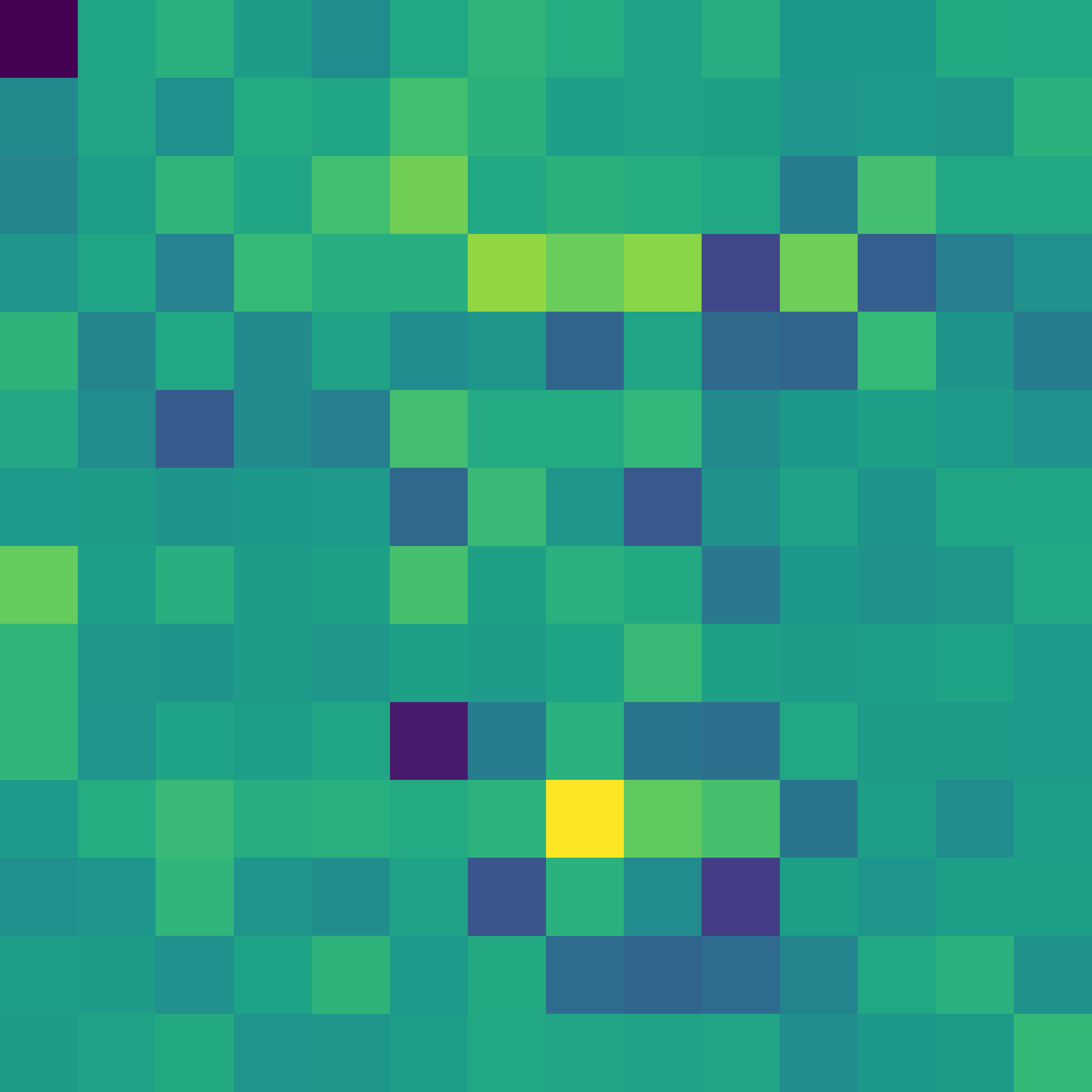}{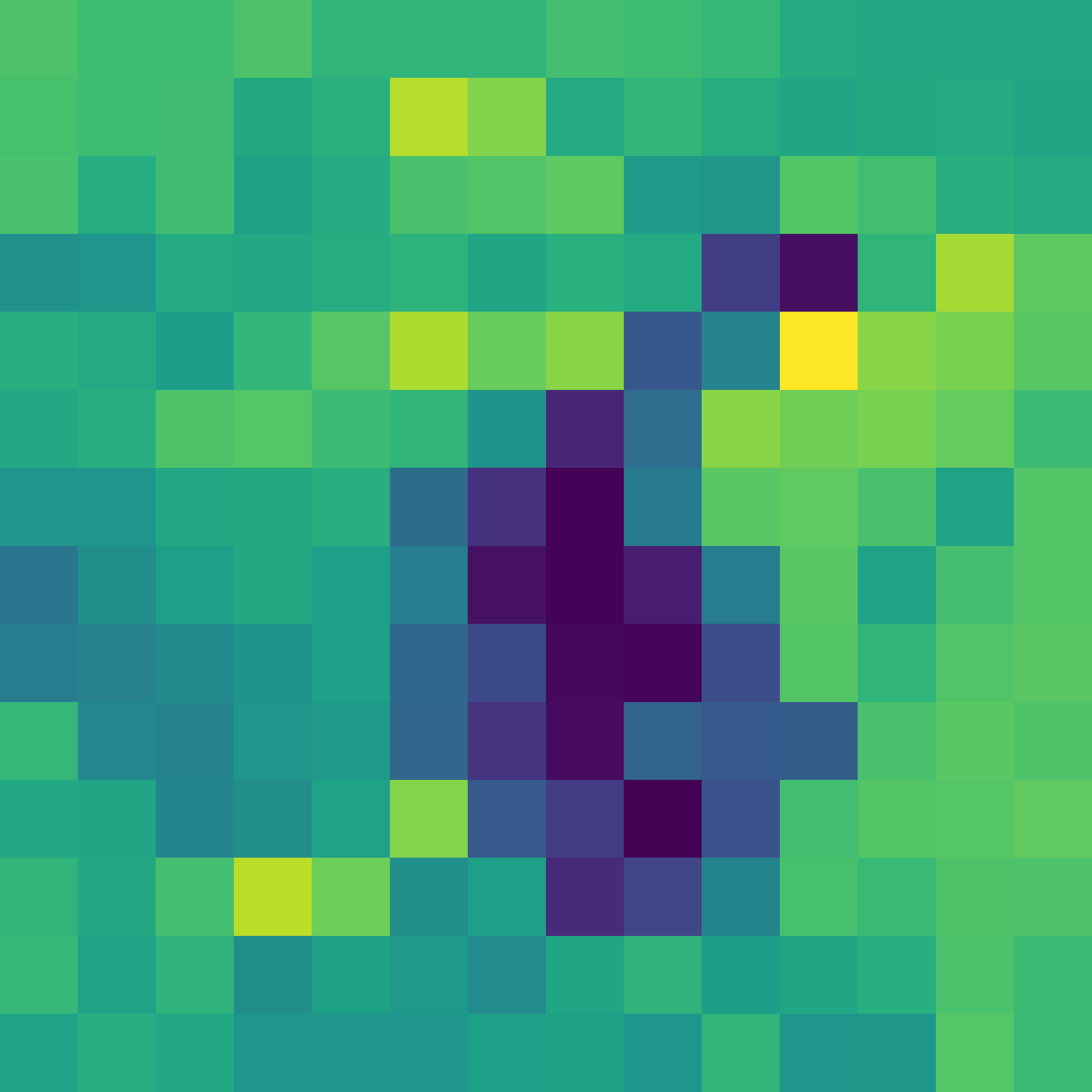}{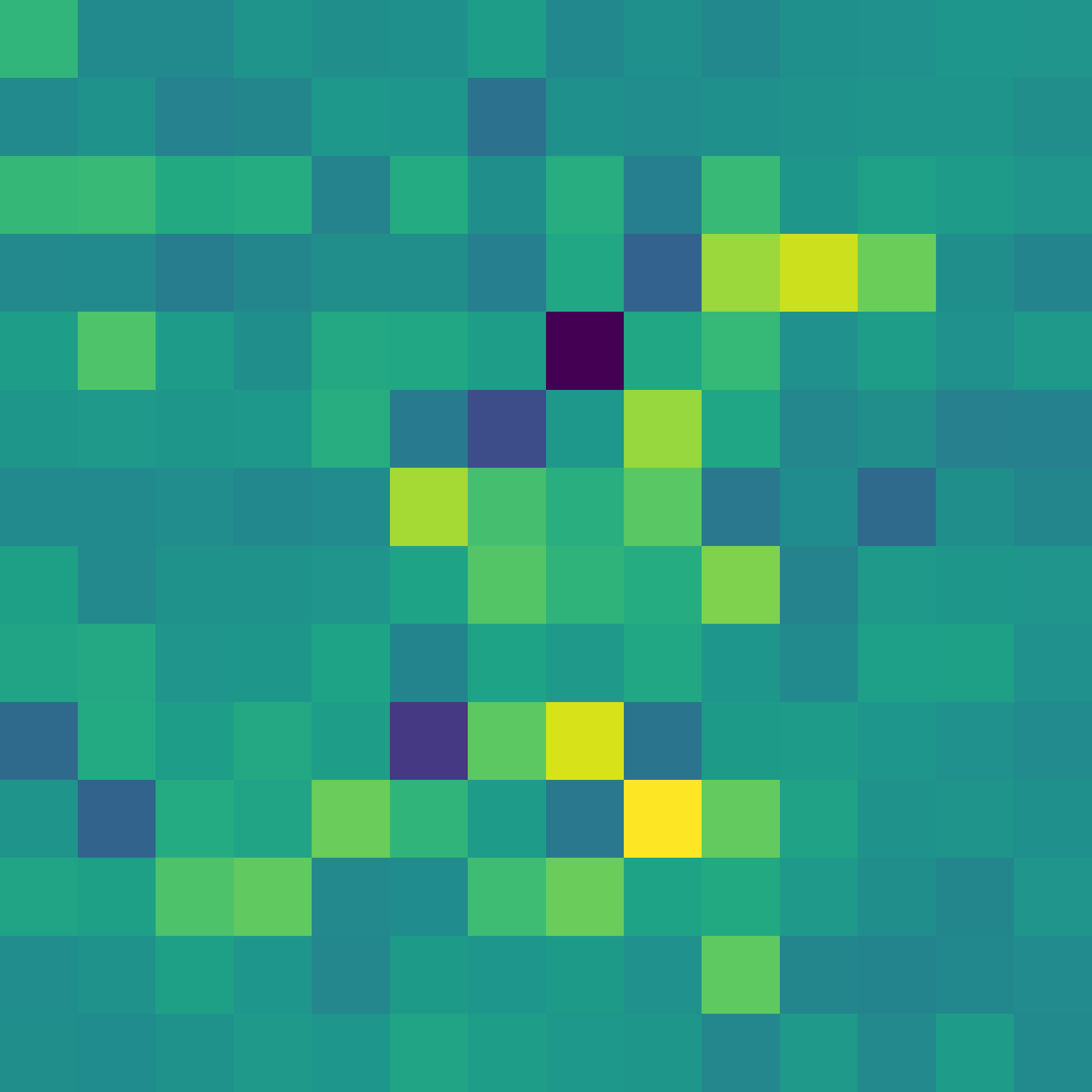} \\[0.45cm]
    Eq. Error & \onebyfour{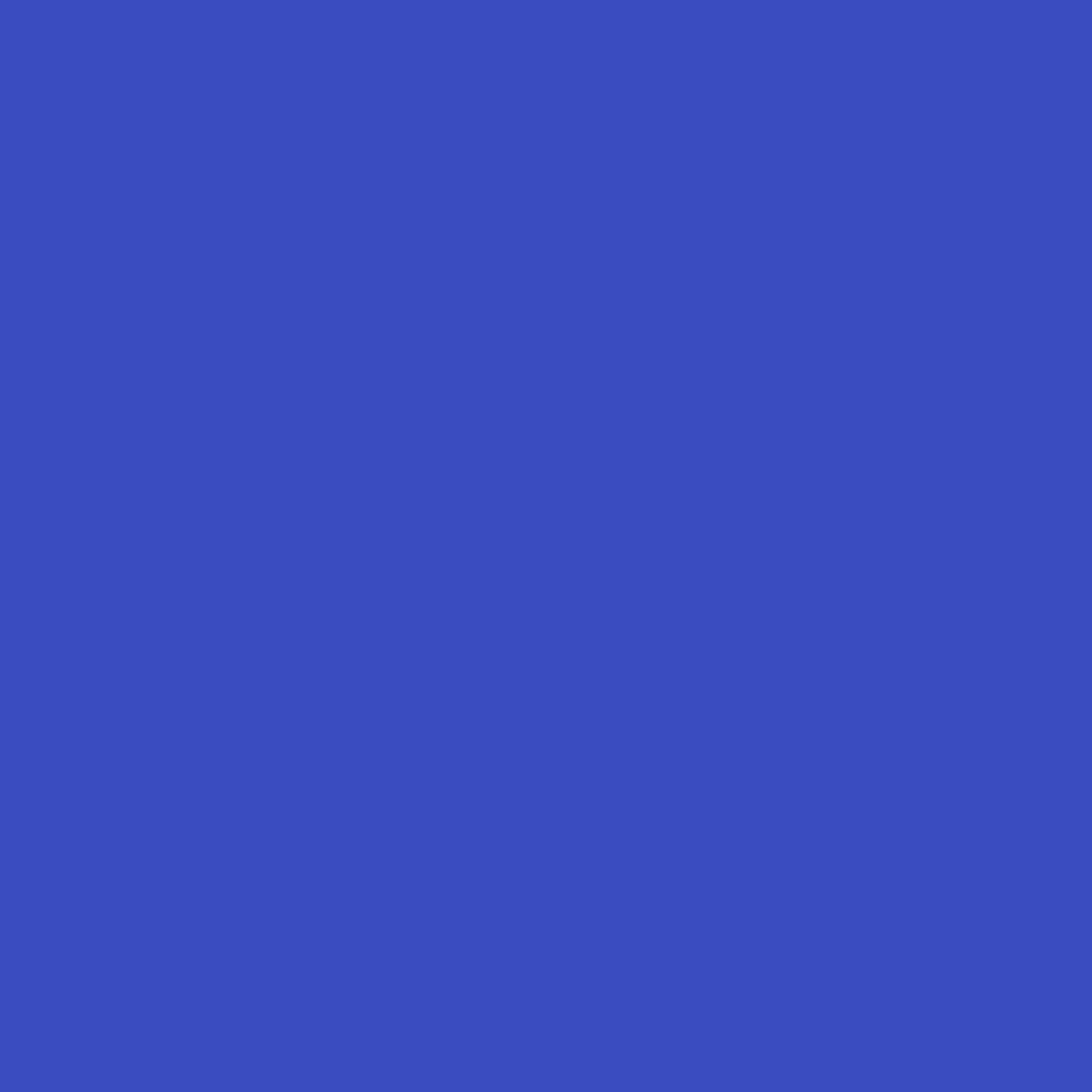}{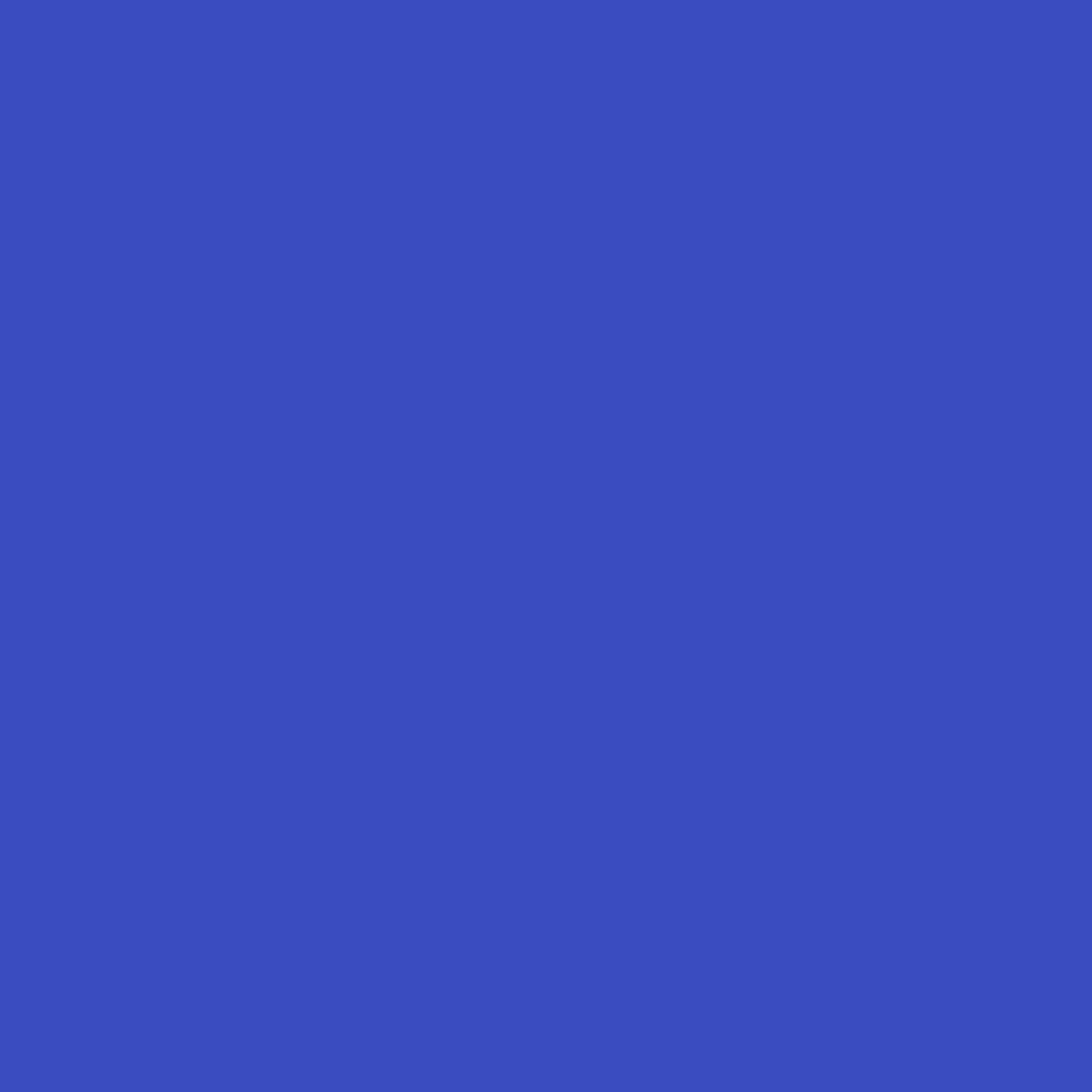}{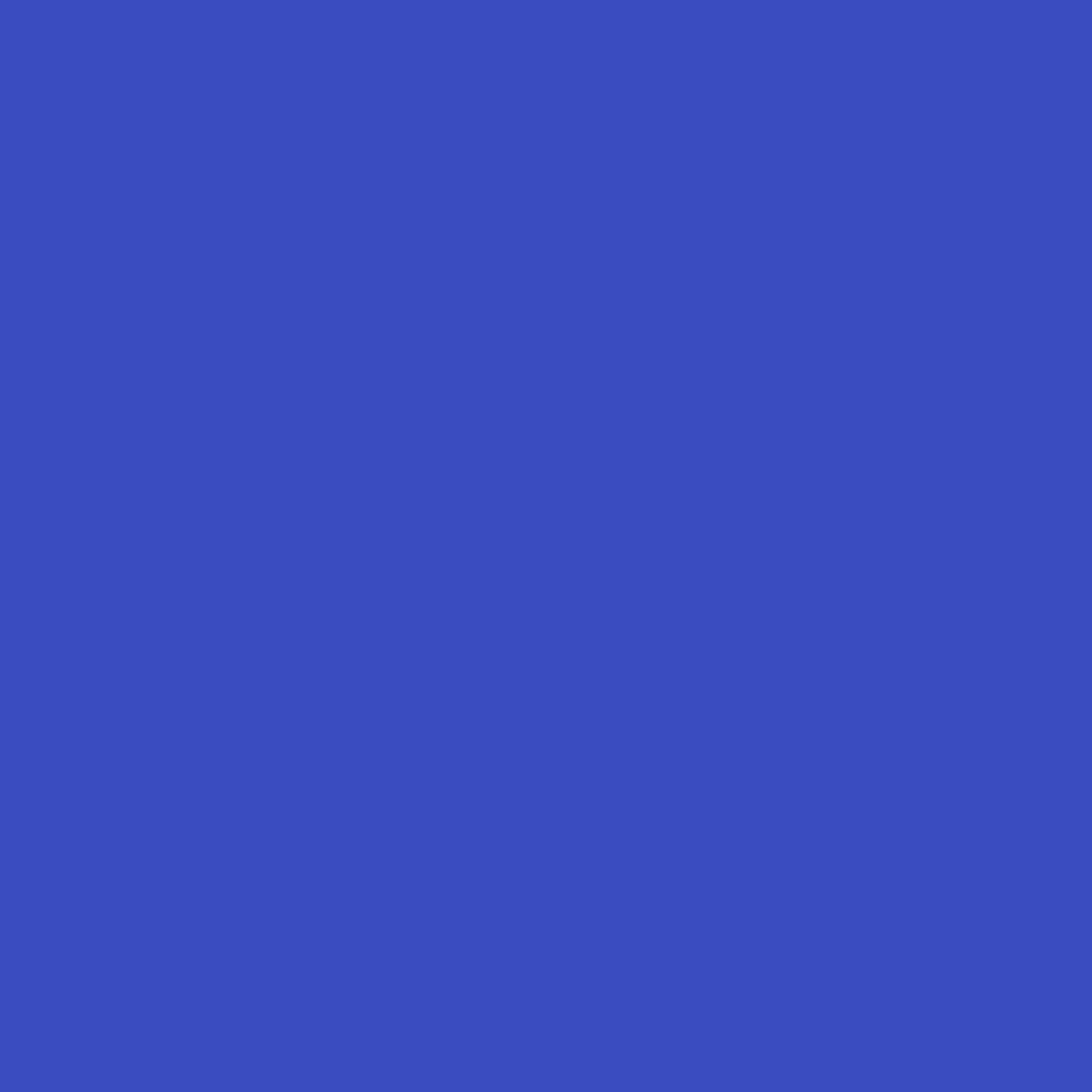}{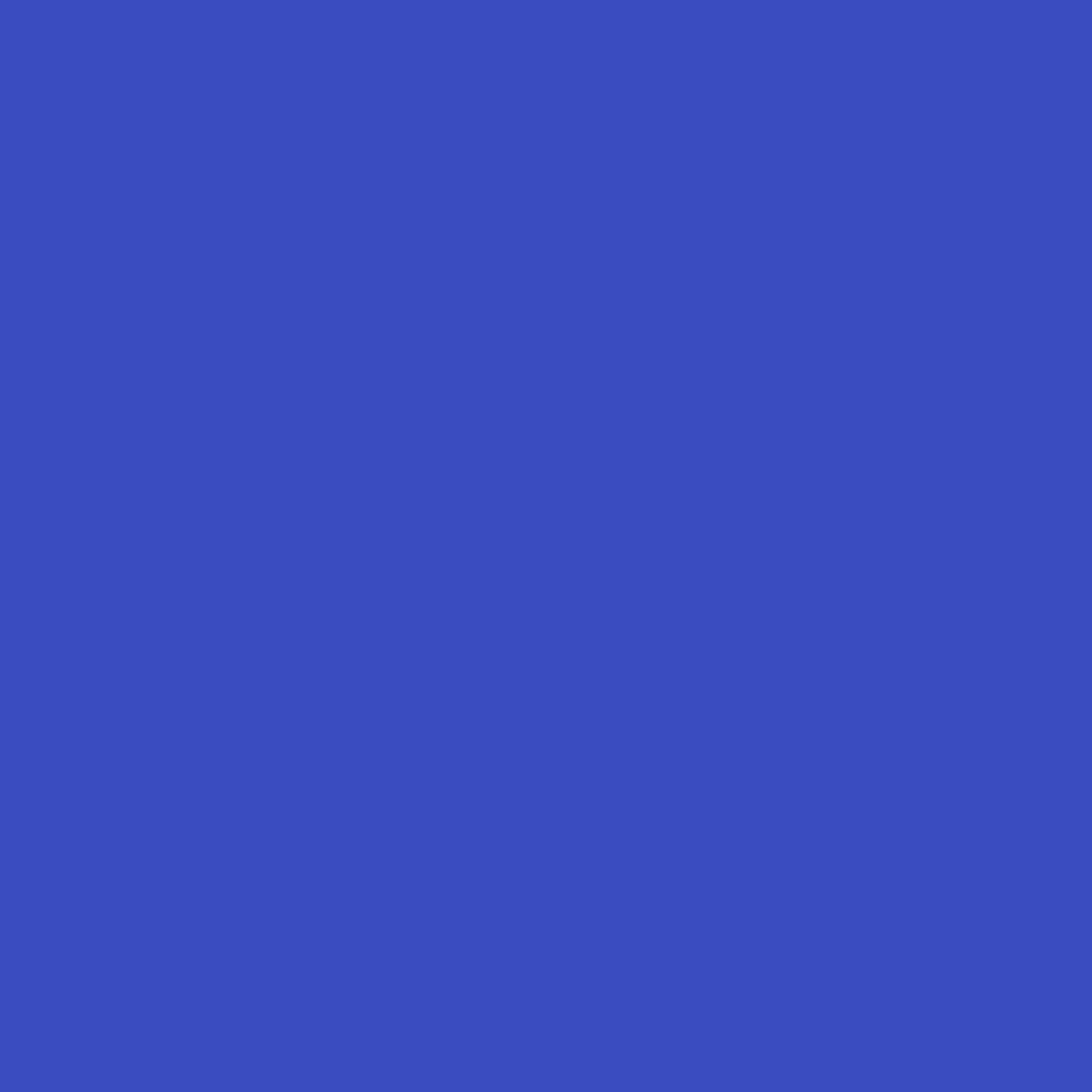} &
    \onebyfour{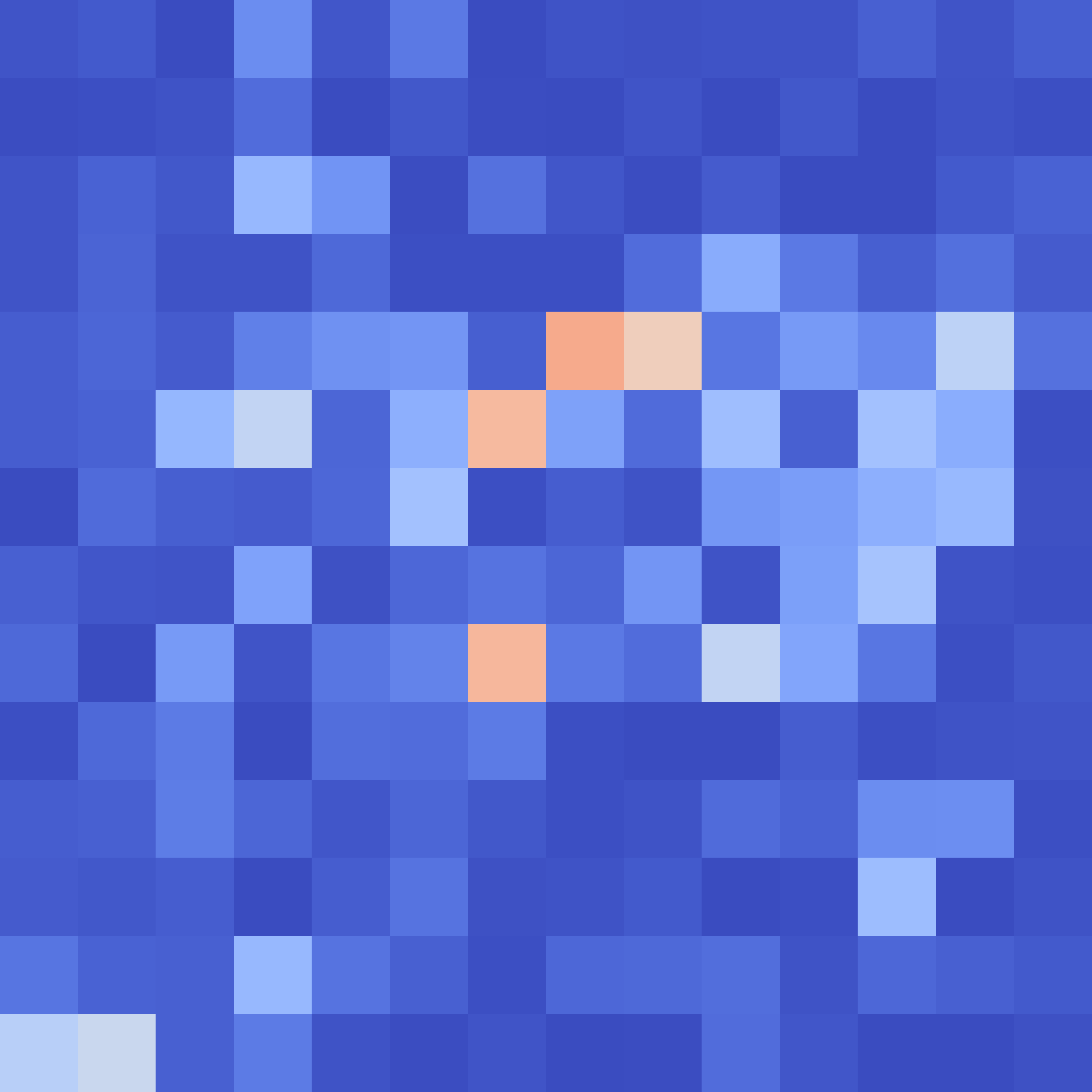}{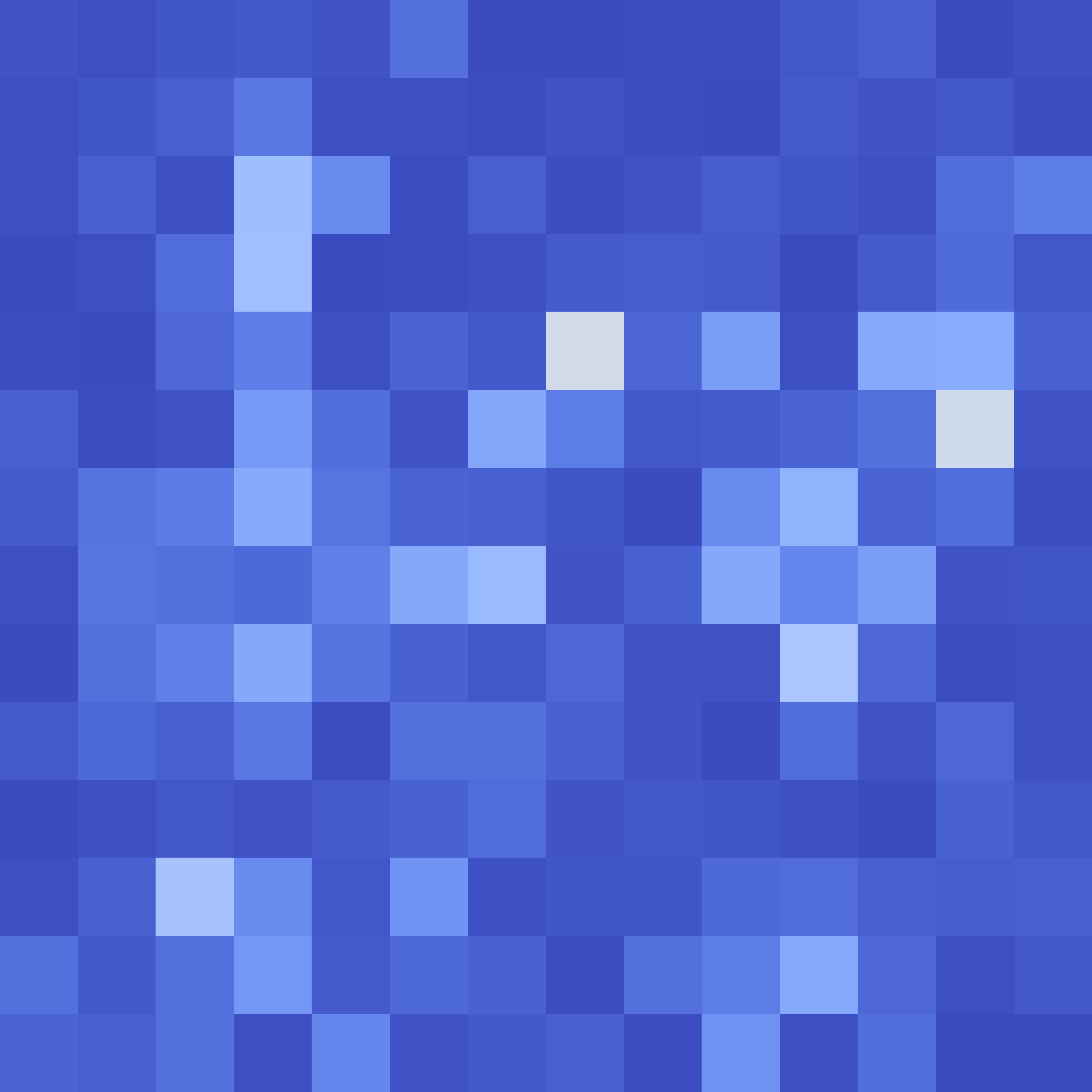}{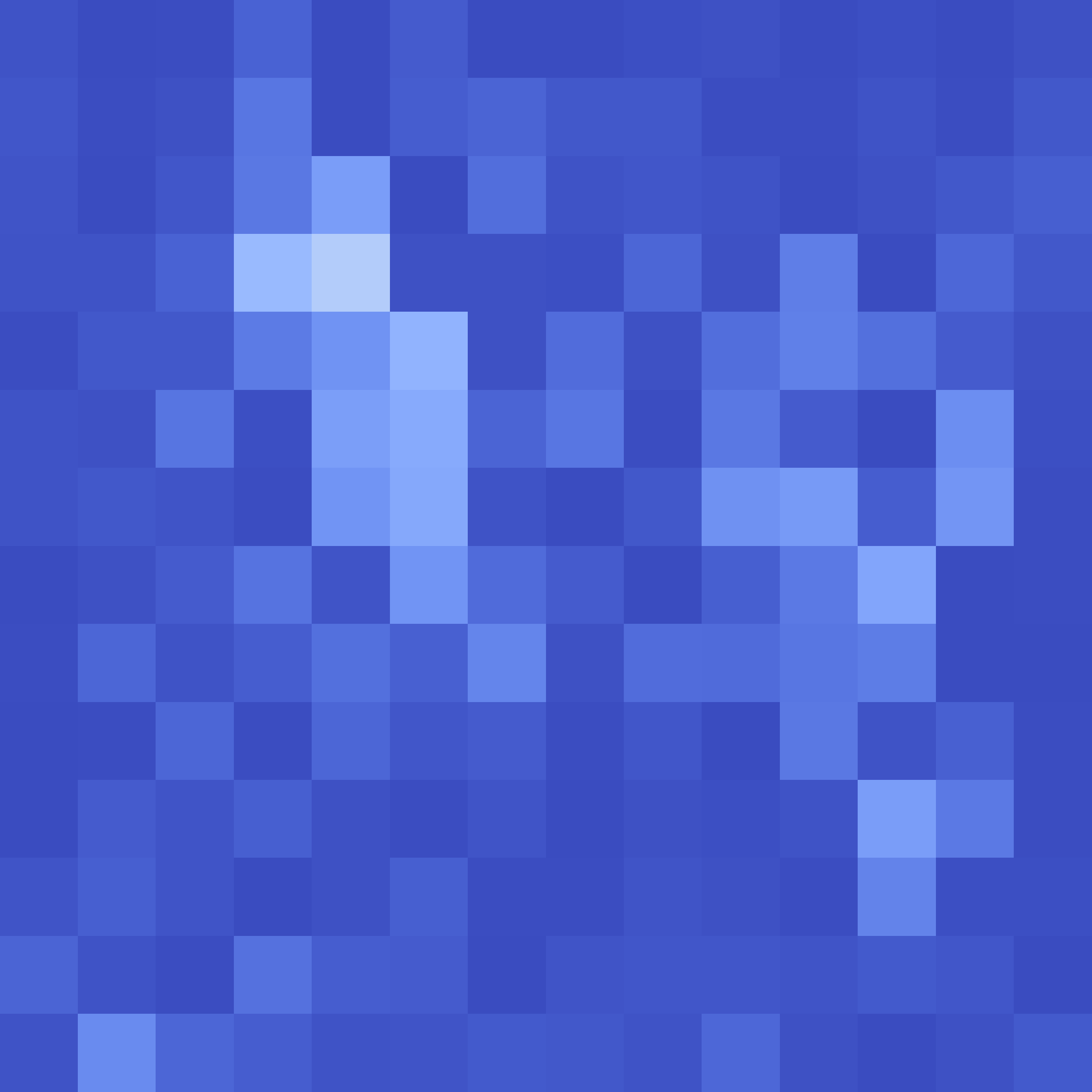}{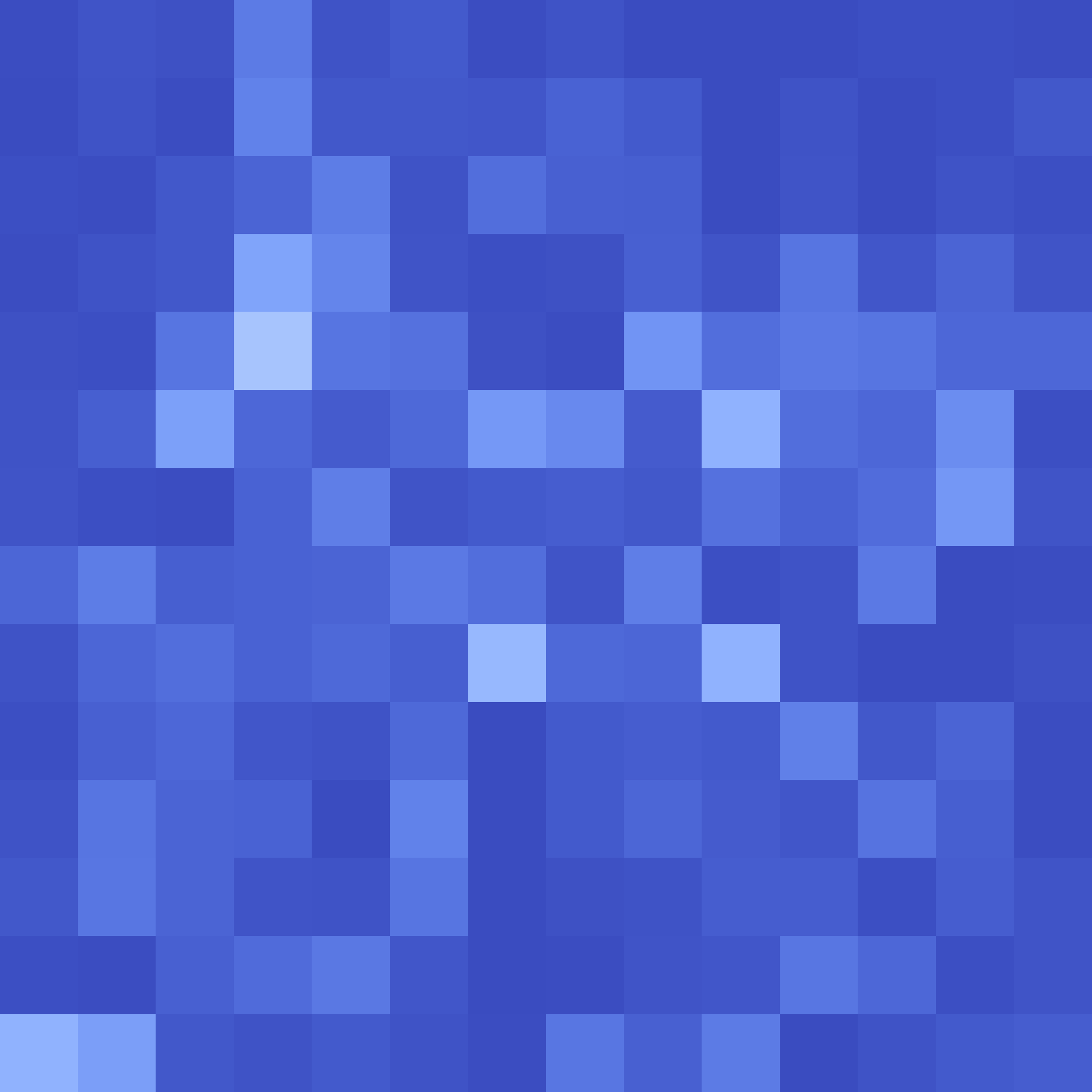} &
    \onebyfour{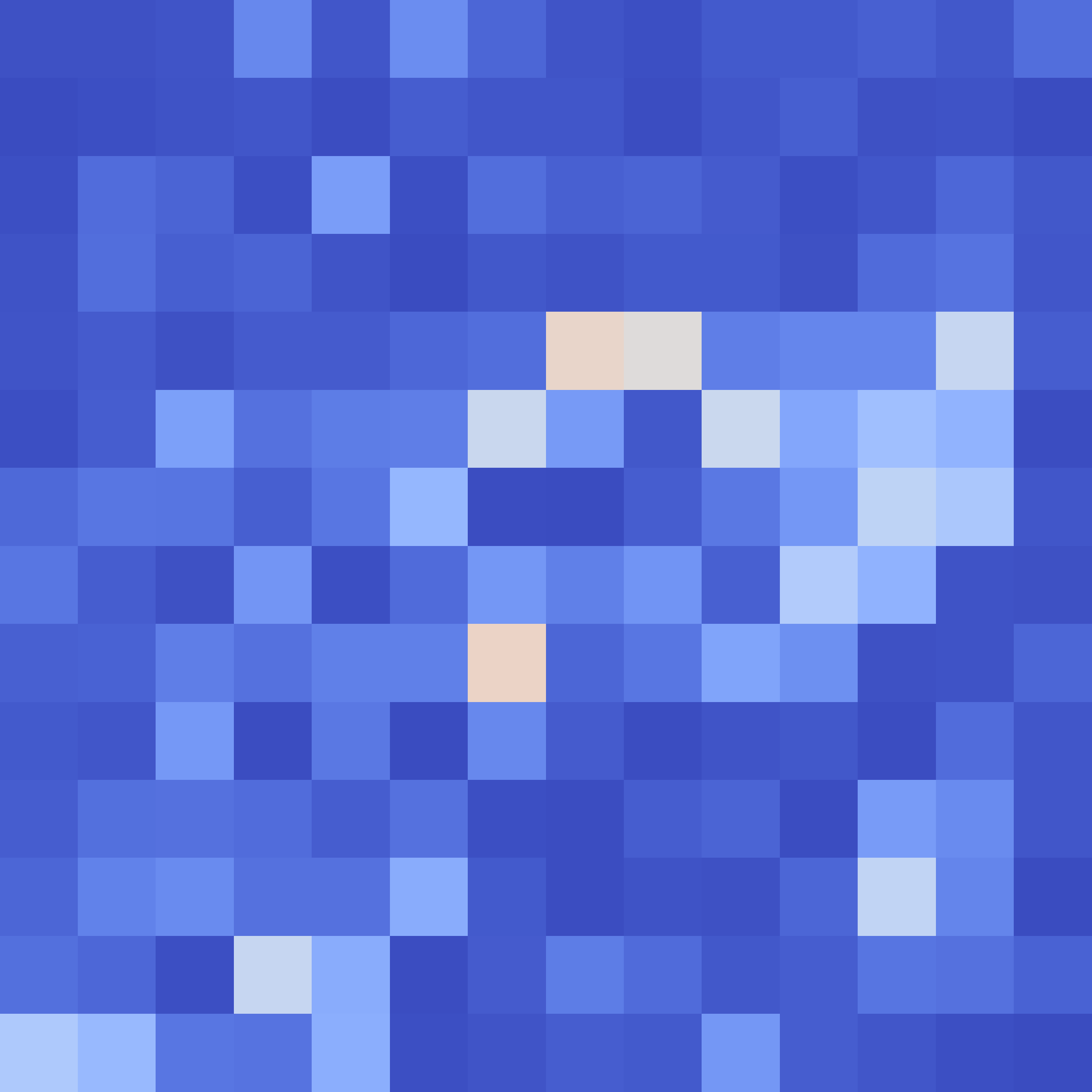}{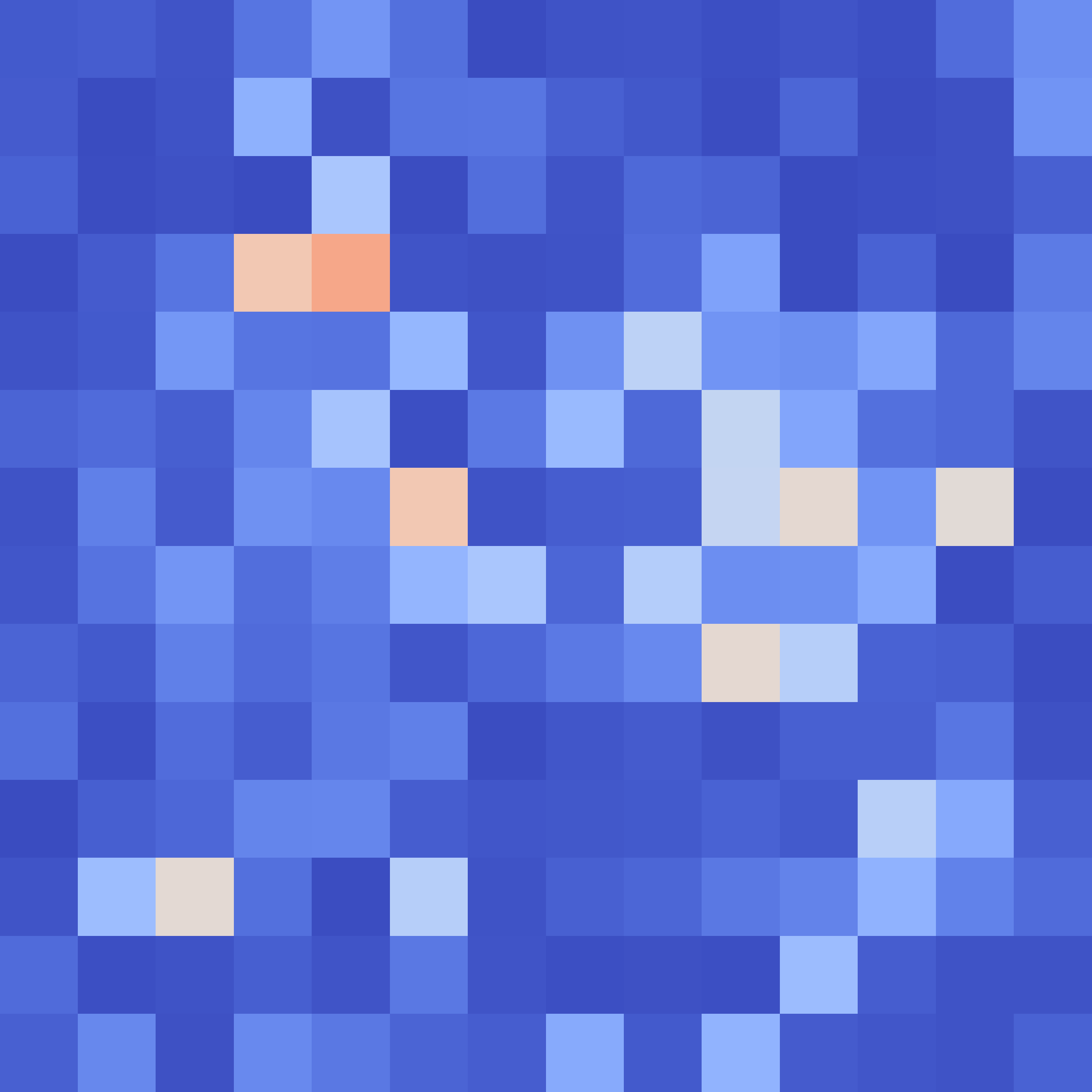}{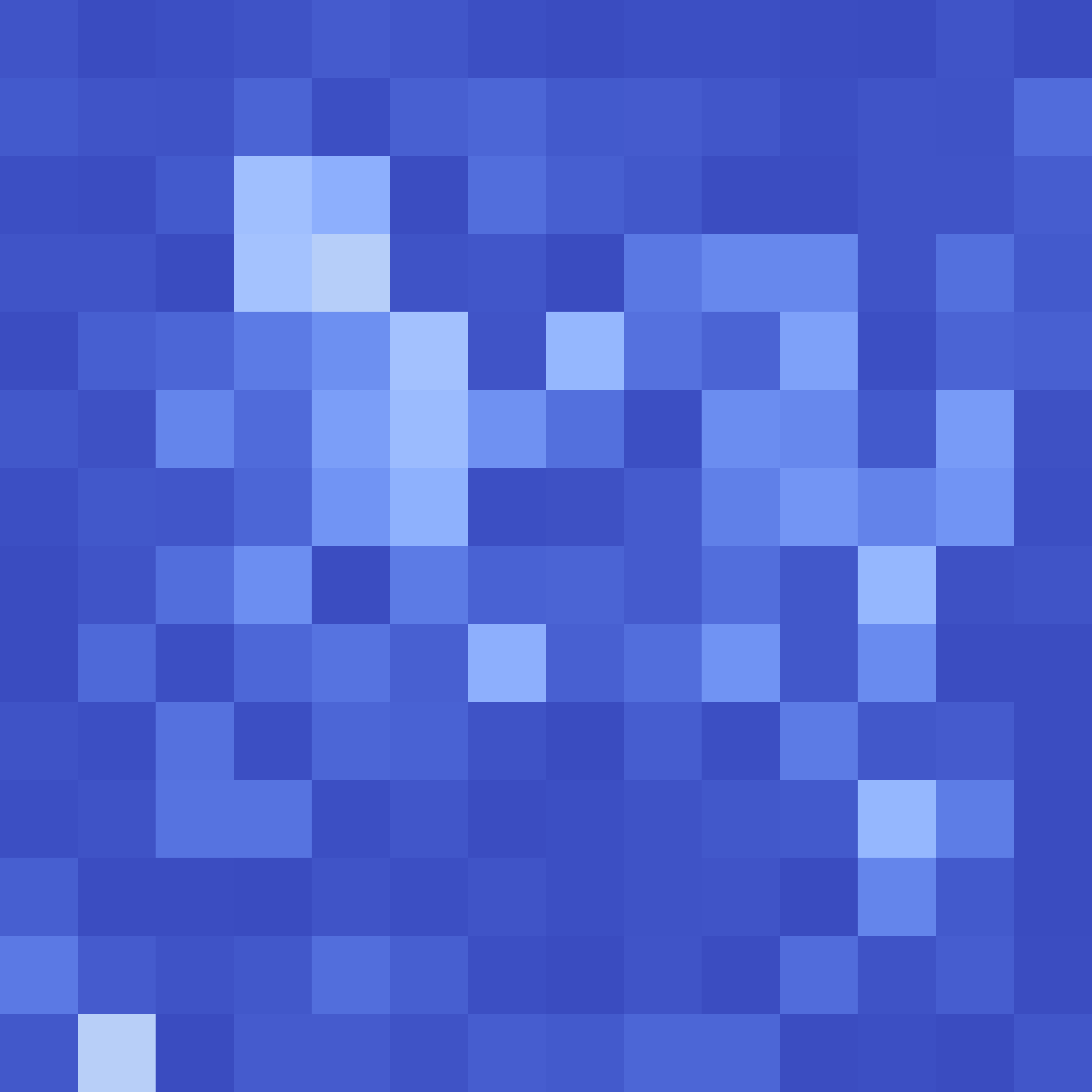}{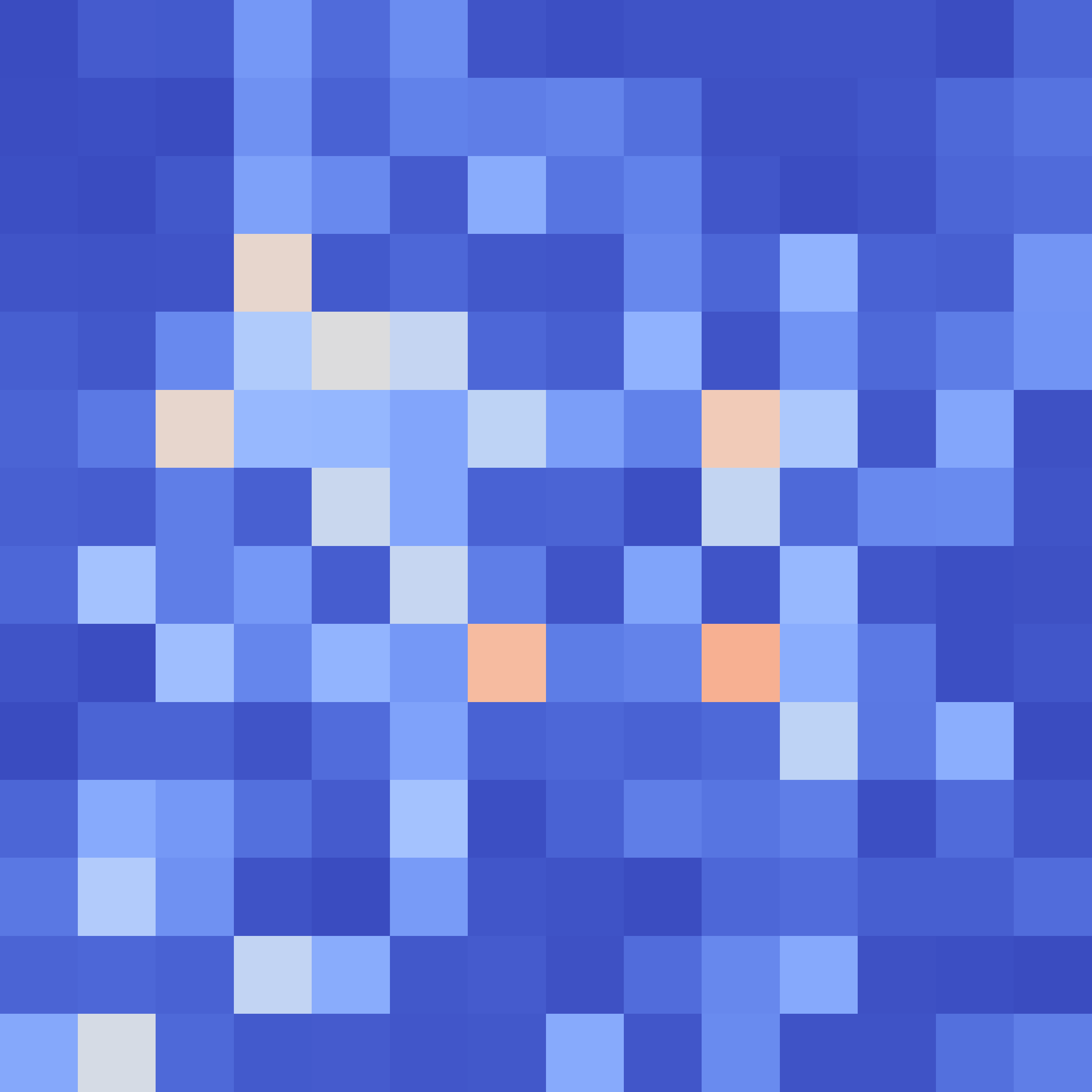} &
    \onebyfour{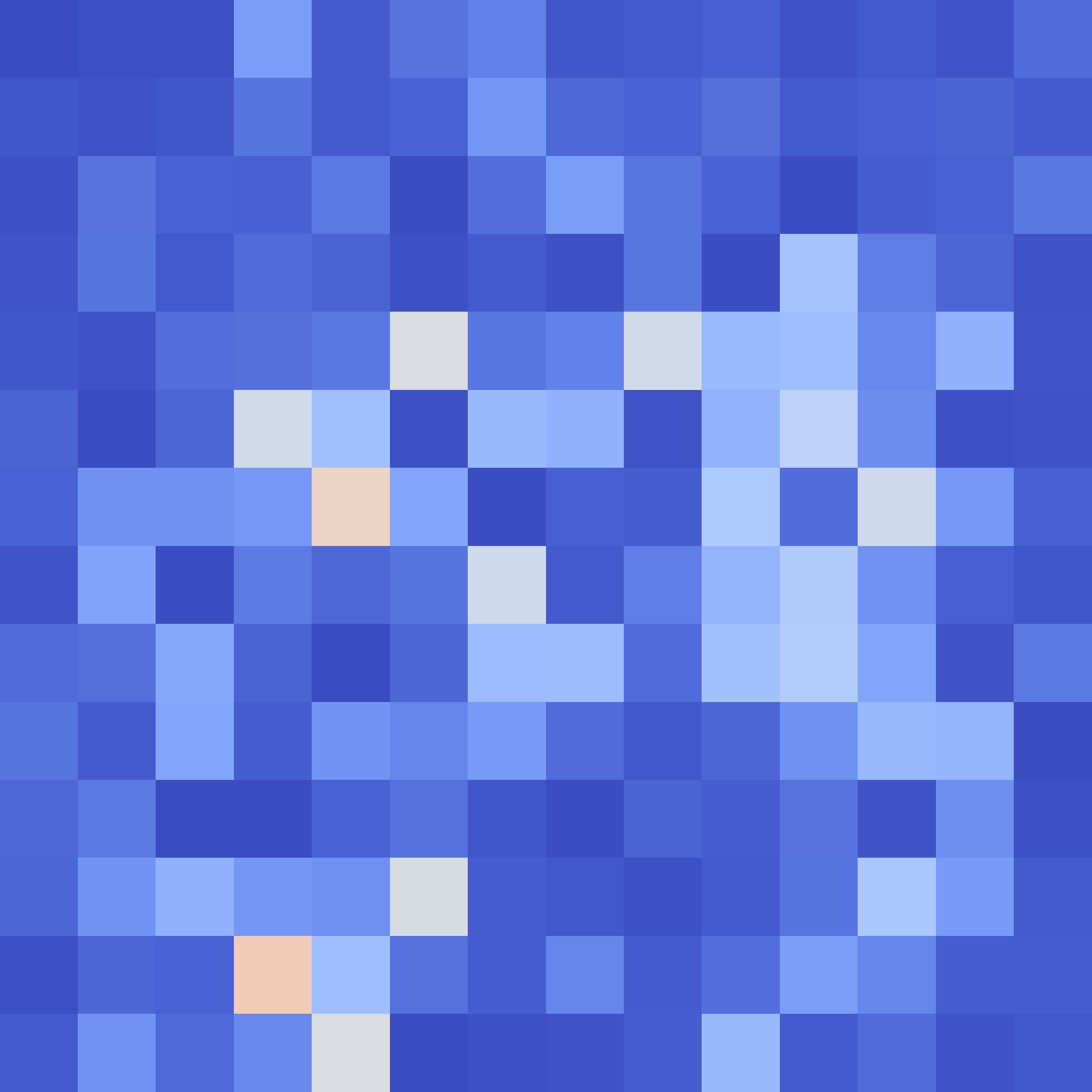}{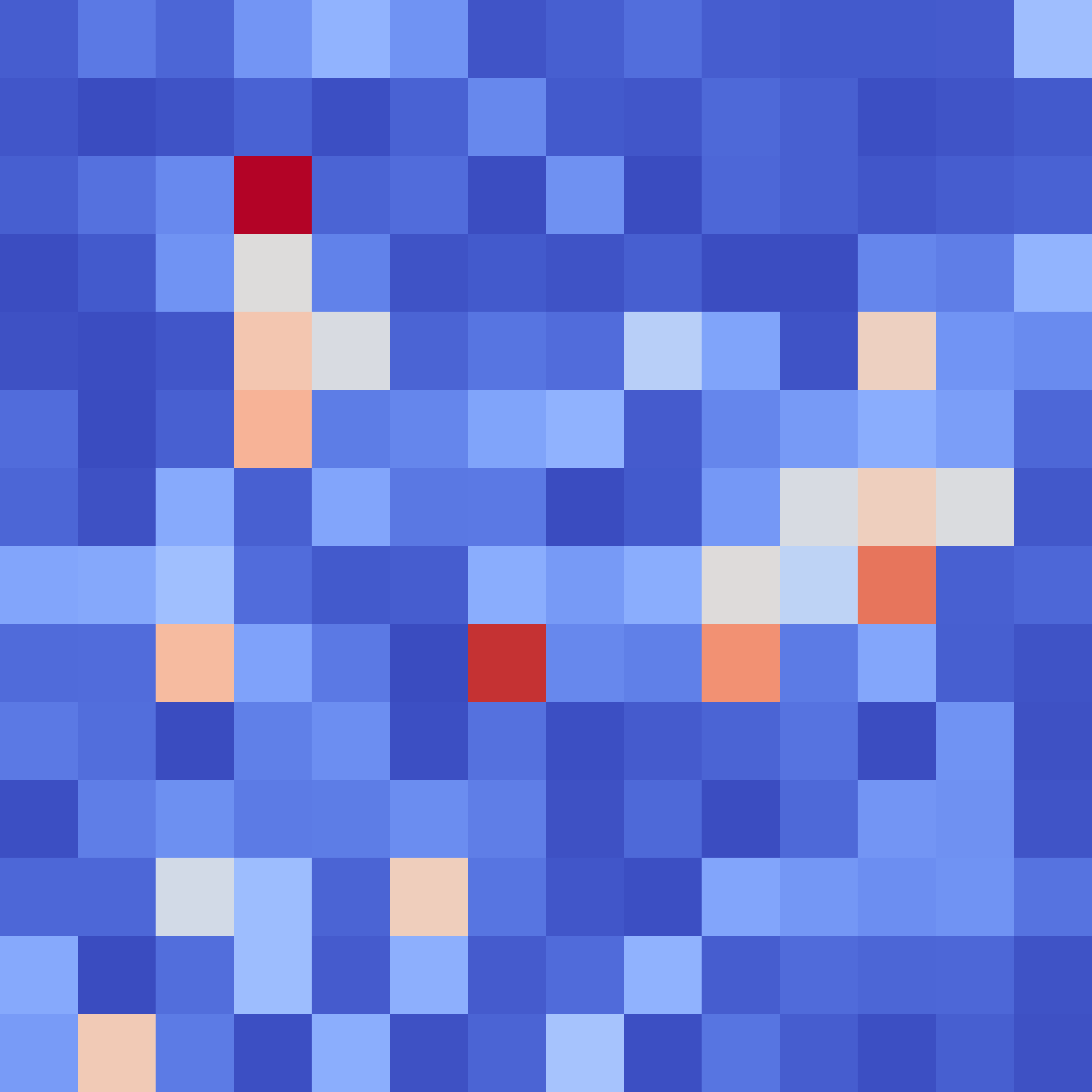}{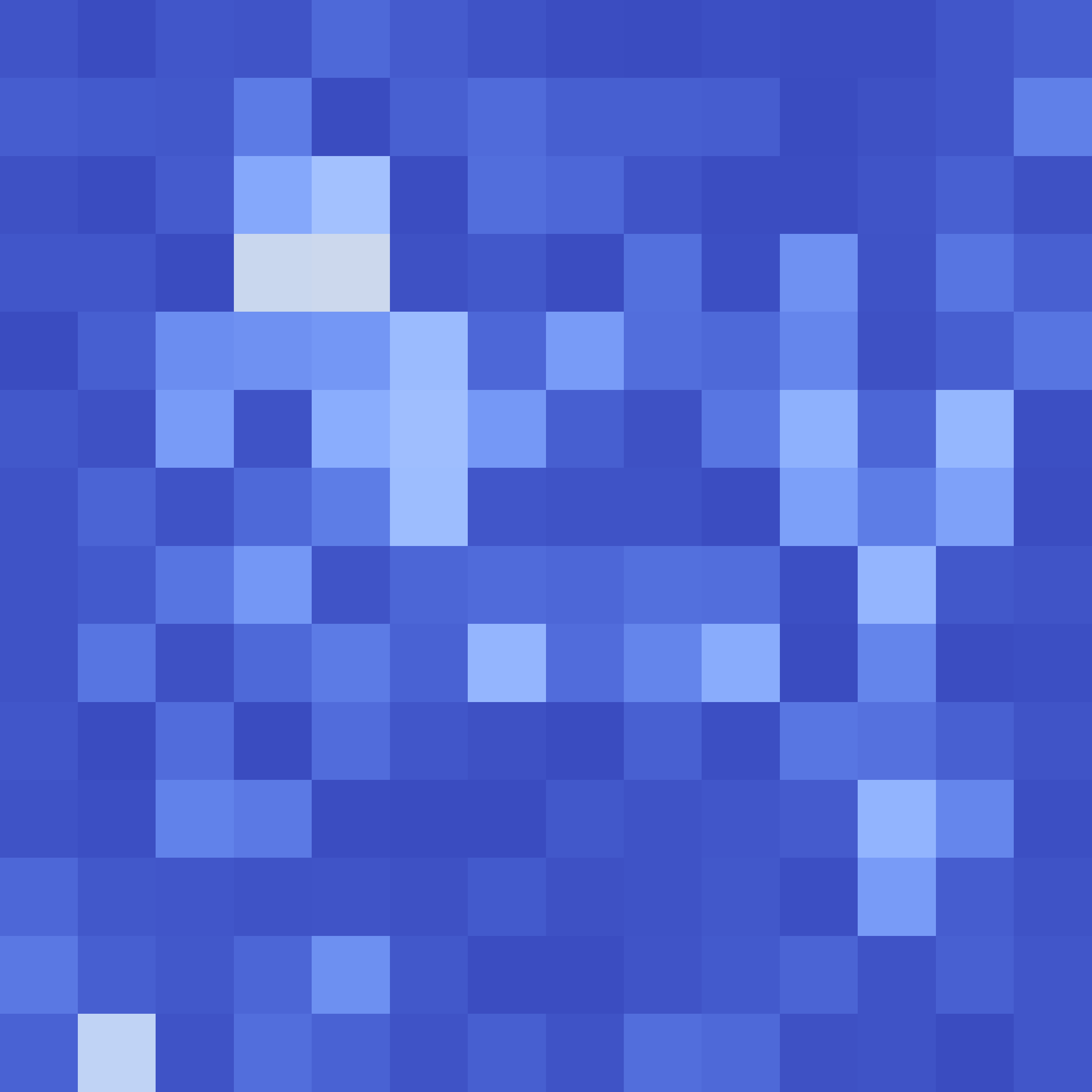}{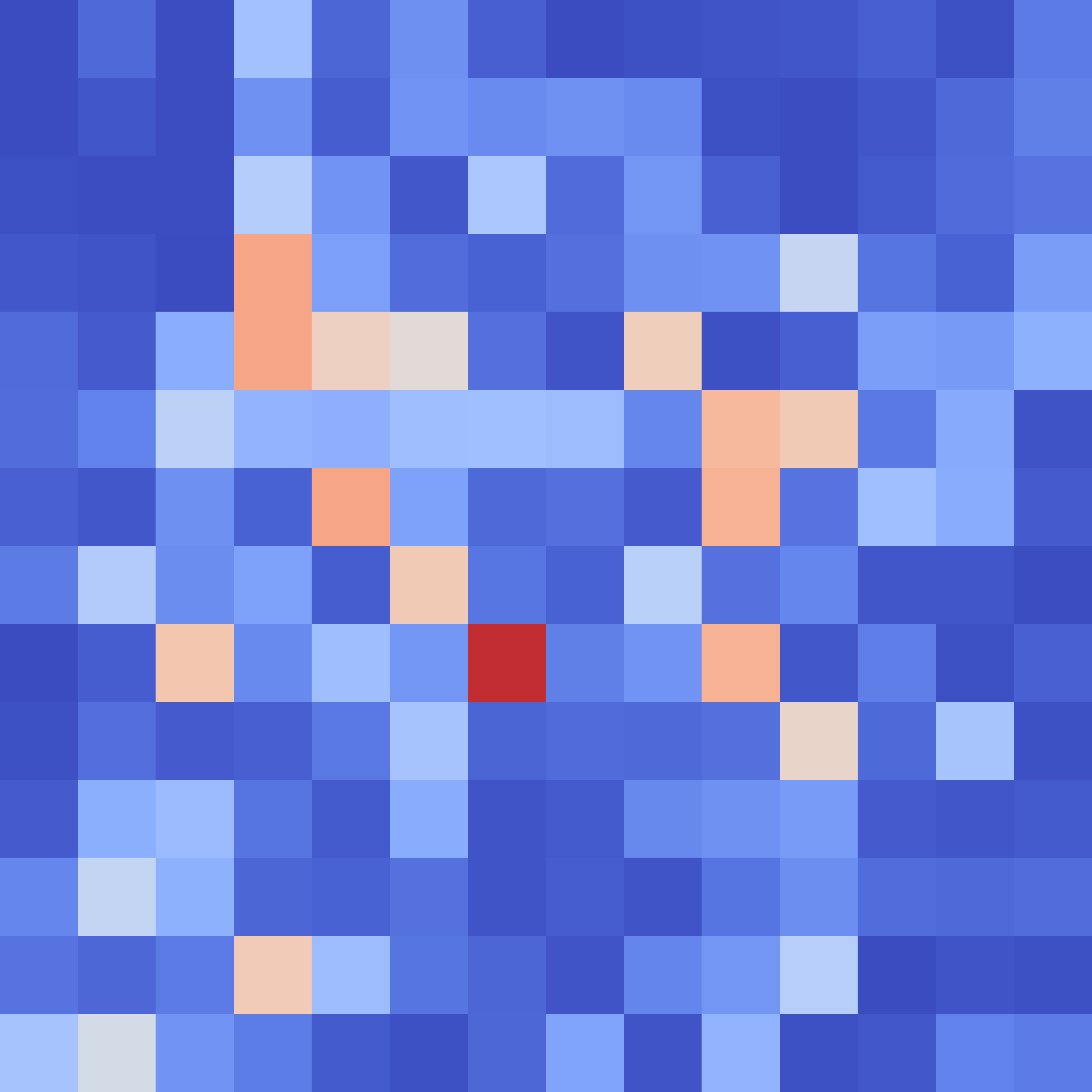} &
    \onebyfour{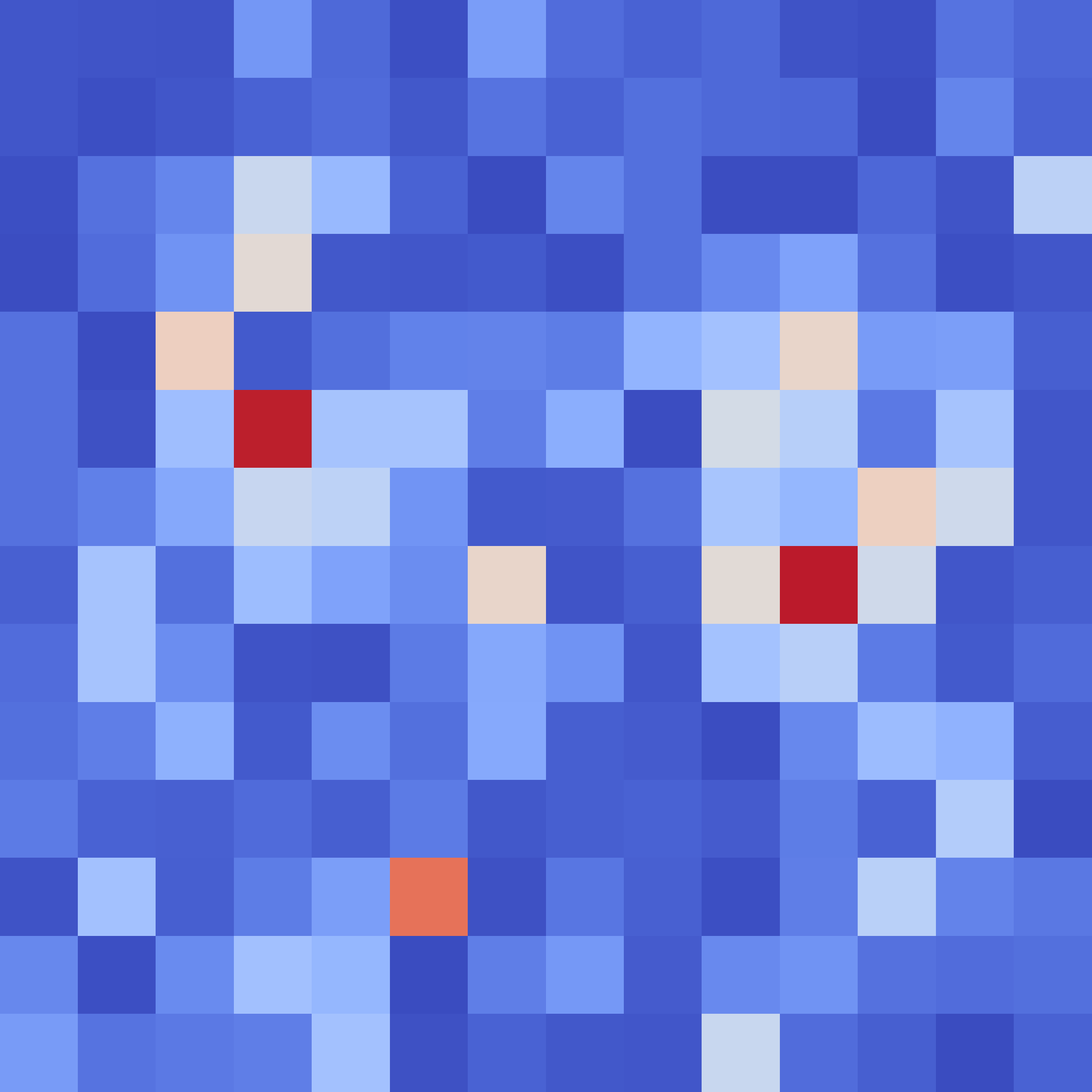}{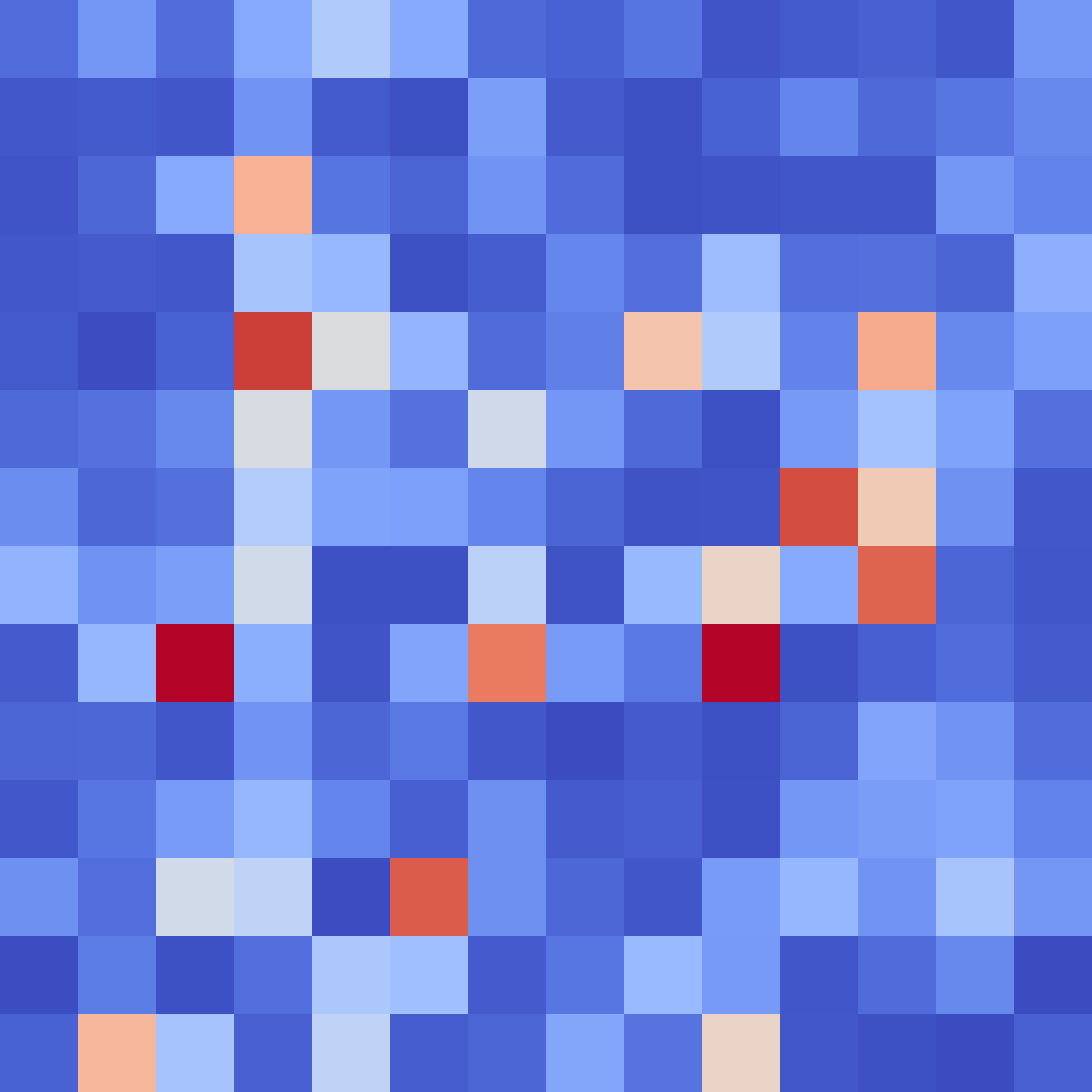}{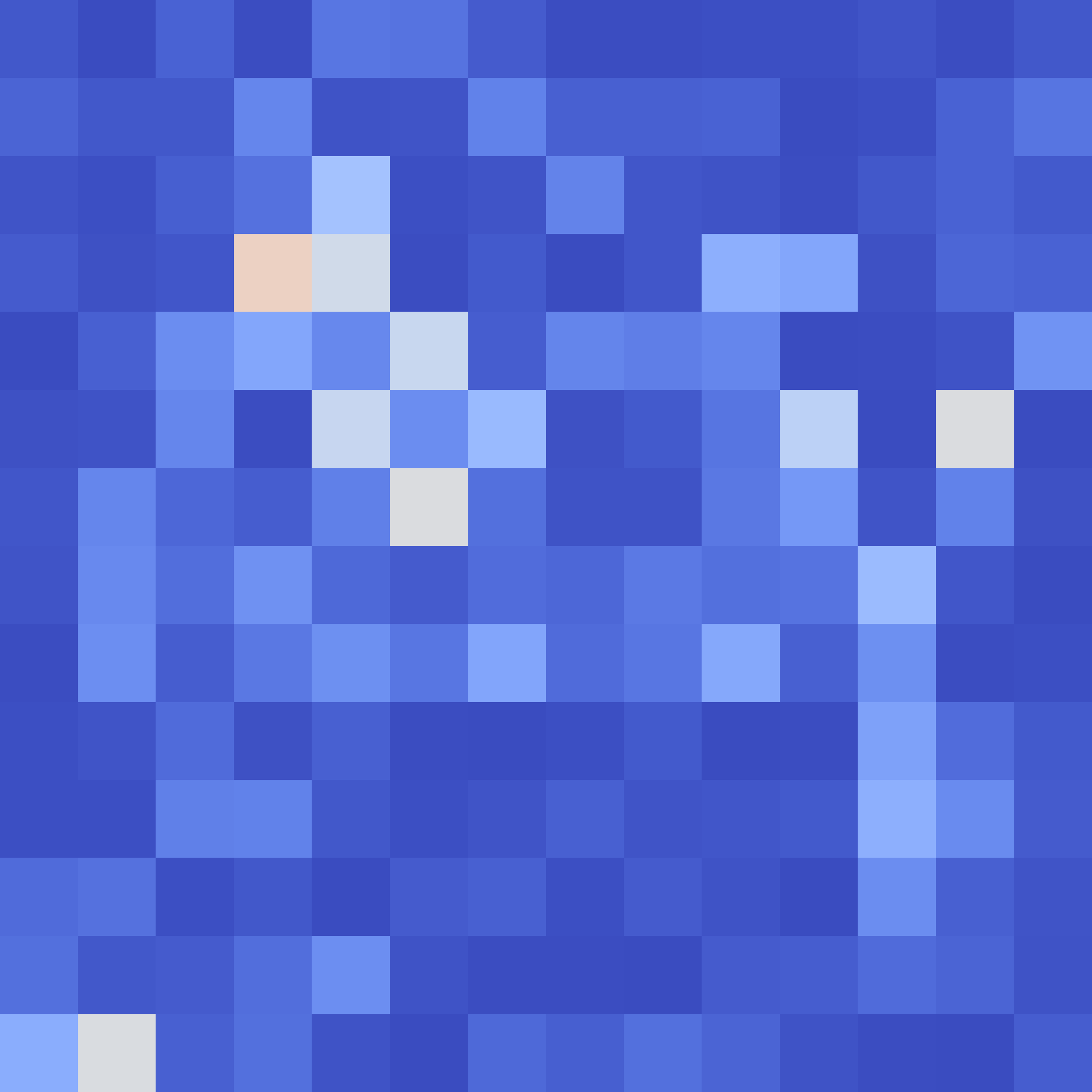}{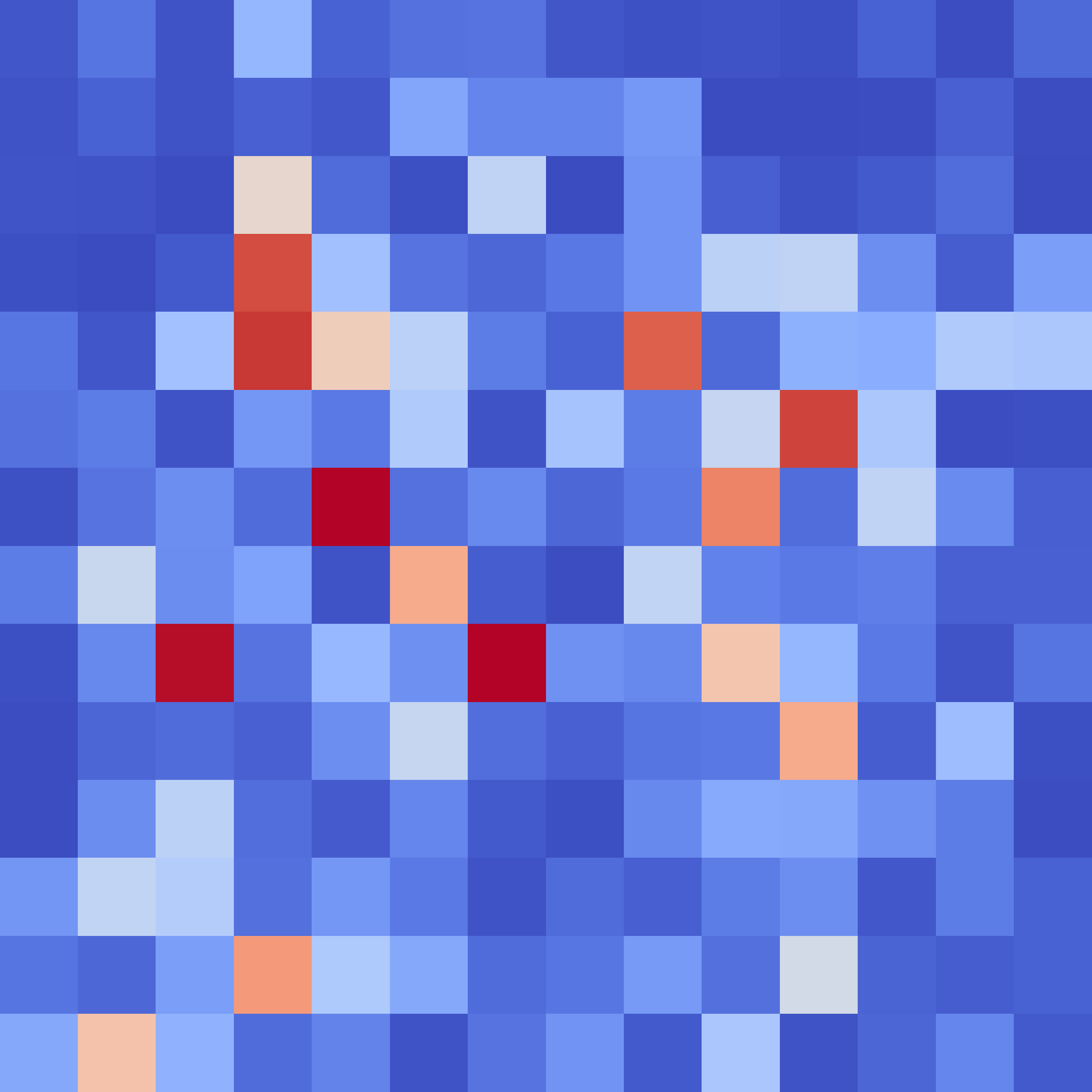}
\end{tabular}
}%
    \vspace{2pt}

\end{minipage}
\hfill
\begin{minipage}{0.079\textwidth}
\hspace{1.5cm}
\end{minipage}
\begin{minipage}{0.93\textwidth}
    \vspace{2pt}
    \noindent\resizebox{\textwidth}{!}{%
    \begin{tikzpicture}[x=1\textwidth/13]
        \shade[left color={rgb,255:red,59;green,76;blue,192},
               right color=white]
               (0.61,-0.10) rectangle (7.00,-0.00);
        \shade[left color=white,
               right color={rgb,255:red,180;green,4;blue,38}]
               (7.20,-0.10) rectangle (13.3,-0.00);
        \node[anchor=south] at (1.12,-0.6) {\small Low error};
        \node[anchor=south] at (12.95,-0.645) {\small High error};
    \end{tikzpicture}
    }%
\end{minipage}
\vspace{-0.38cm}
    \captionof{figure}{Visualization of the ViT~\cite{dosovitskiy2020image} weights with our soft equivariance layer (\wrt $90^\circ$ rotation) under different softness levels, along with the corresponding extracted features and the equivariance errors. Our tunable design allows the layers' weights to transition smoothly from perfectly equivariant to fully non-equivariant behavior in a controlled manner.
    }
    \label{fig:teaser2}
\begin{textblock*}{5cm}(1.8cm,10.1cm) %
  {\rotatebox{90}{\small Features}}
\end{textblock*}

%% file: figs/feature_viz_10.tex
\setlength{\tabcolsep}{1.8pt}%
\renewcommand{\arraystretch}{0.8}
\begin{minipage}{\textwidth}
\newcommand{\onebyfour}[4]{%
\begin{minipage}{0.20\textwidth}\centering
    \includegraphics[width=0.30\linewidth]{#1}\hspace{0.0095\linewidth}%
    \includegraphics[width=0.30\linewidth]{#2}\hspace{0.0095\linewidth}%
    \includegraphics[width=0.30\linewidth]{#3}\hspace{0.0095\linewidth}%
\end{minipage}}

\begin{minipage}{0.073\textwidth}
\hspace{1cm}
\end{minipage}
\begin{minipage}{0.927\textwidth}
    \noindent\resizebox{\textwidth}{!}{%
    \begin{tikzpicture}[x=1\textwidth/10]
        \draw[thick, <->] (0.,0) -- (10.,0);
        \node[anchor=north] at (0.42,+0.6) {\small Equivariant};
        \node[anchor=north] at (5,+0.6) {\small Soft equivariant};
        \node[anchor=north] at (9.45,+0.6) {\small Non-equivariant};
    \end{tikzpicture}
    }%
\vspace{6pt}
\end{minipage}

\resizebox{\textwidth}{!}{%
\begin{tabular}{cccccc}
    Weights & \onebyfour{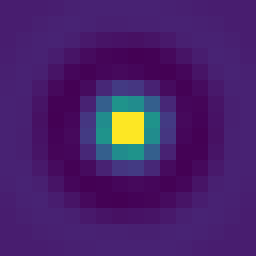}{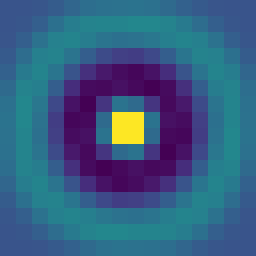}{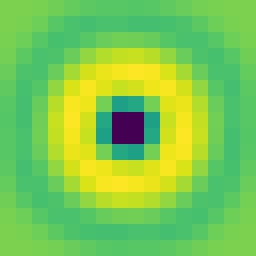}{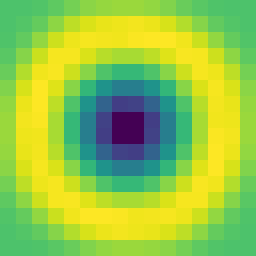} &
    \onebyfour{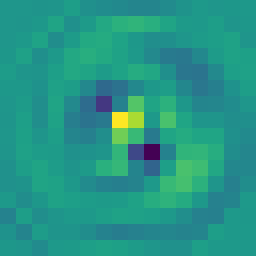}{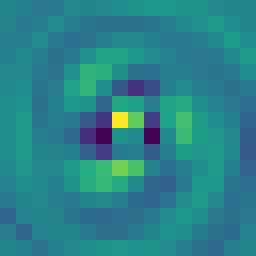}{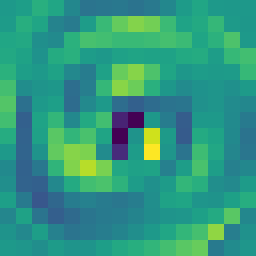}{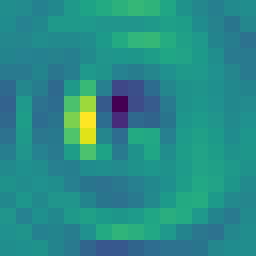} &
    \onebyfour{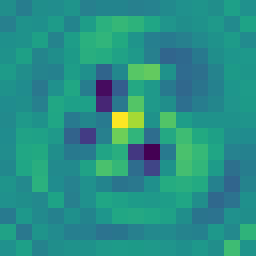}{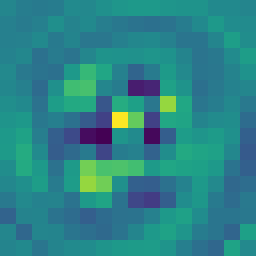}{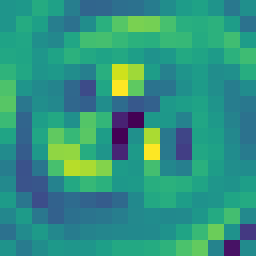}{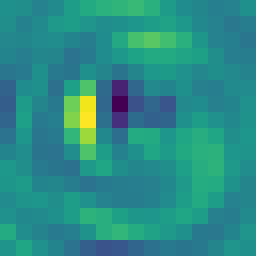} &
    \onebyfour{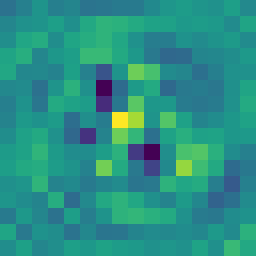}{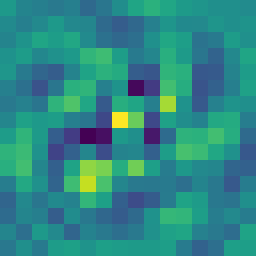}{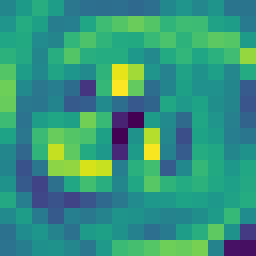}{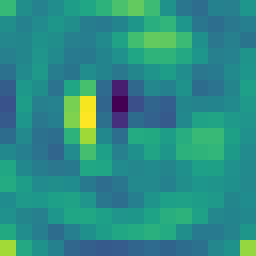} &
    \onebyfour{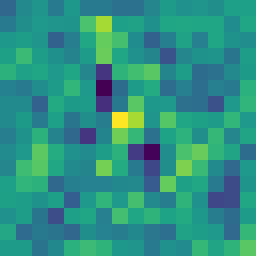}{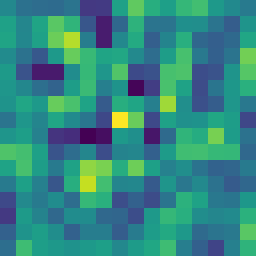}{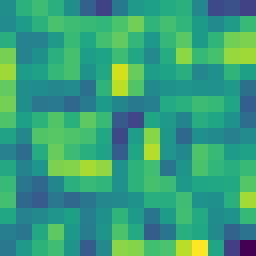}{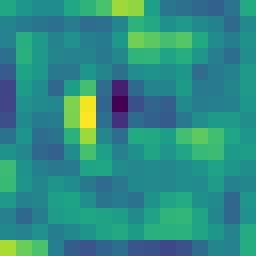} \\[0.42cm]
    \begin{minipage}{0.065\textwidth}\centering\includegraphics[width=0.91\linewidth]{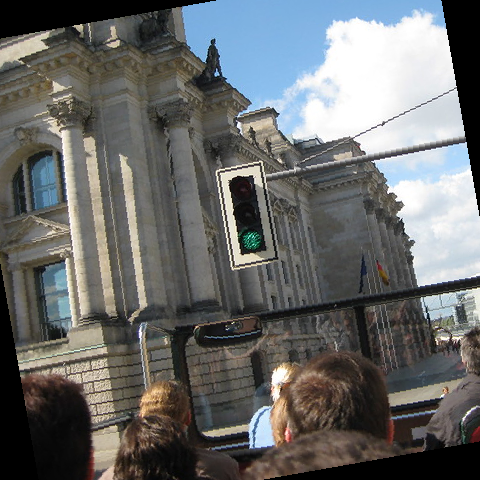}\end{minipage} & \onebyfour{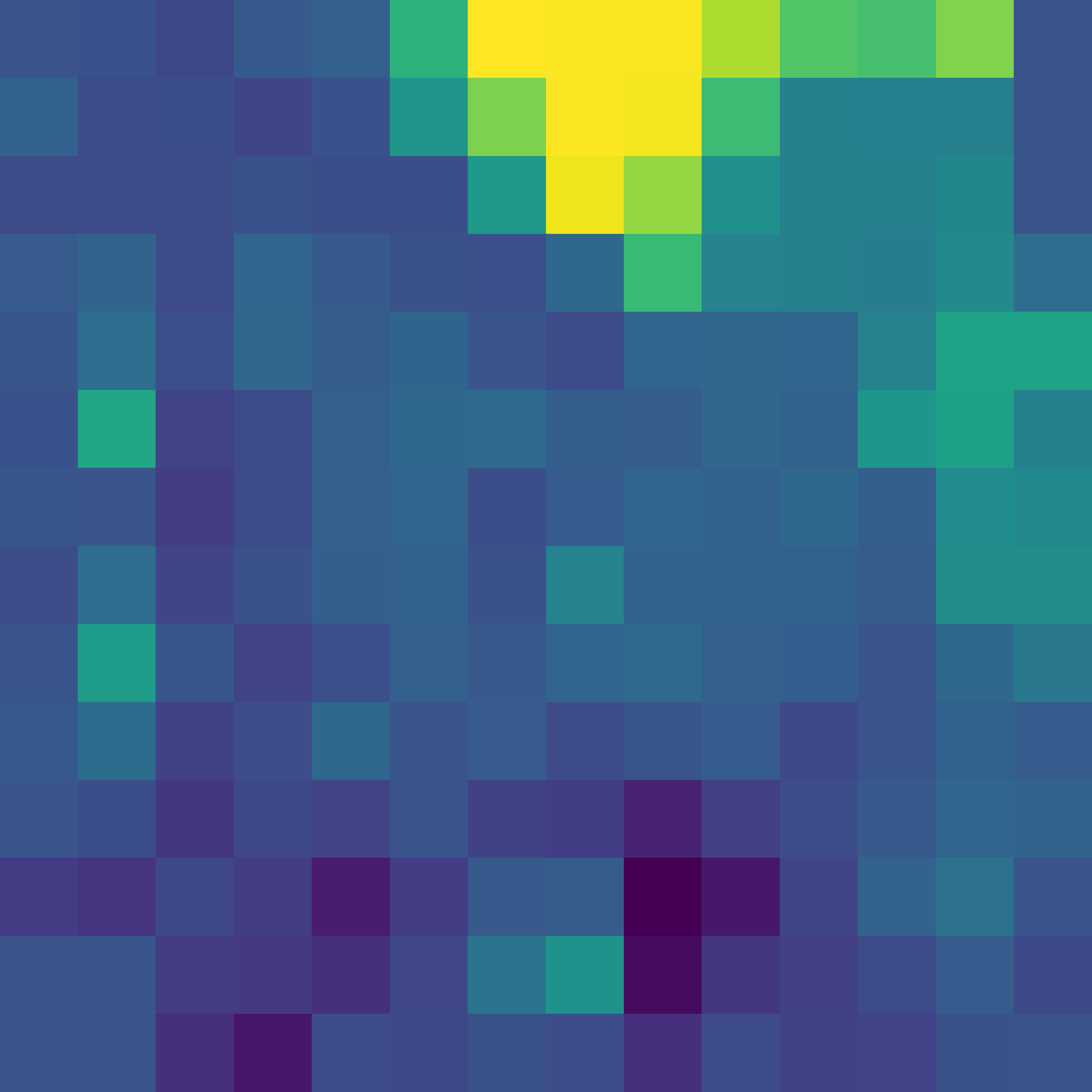}{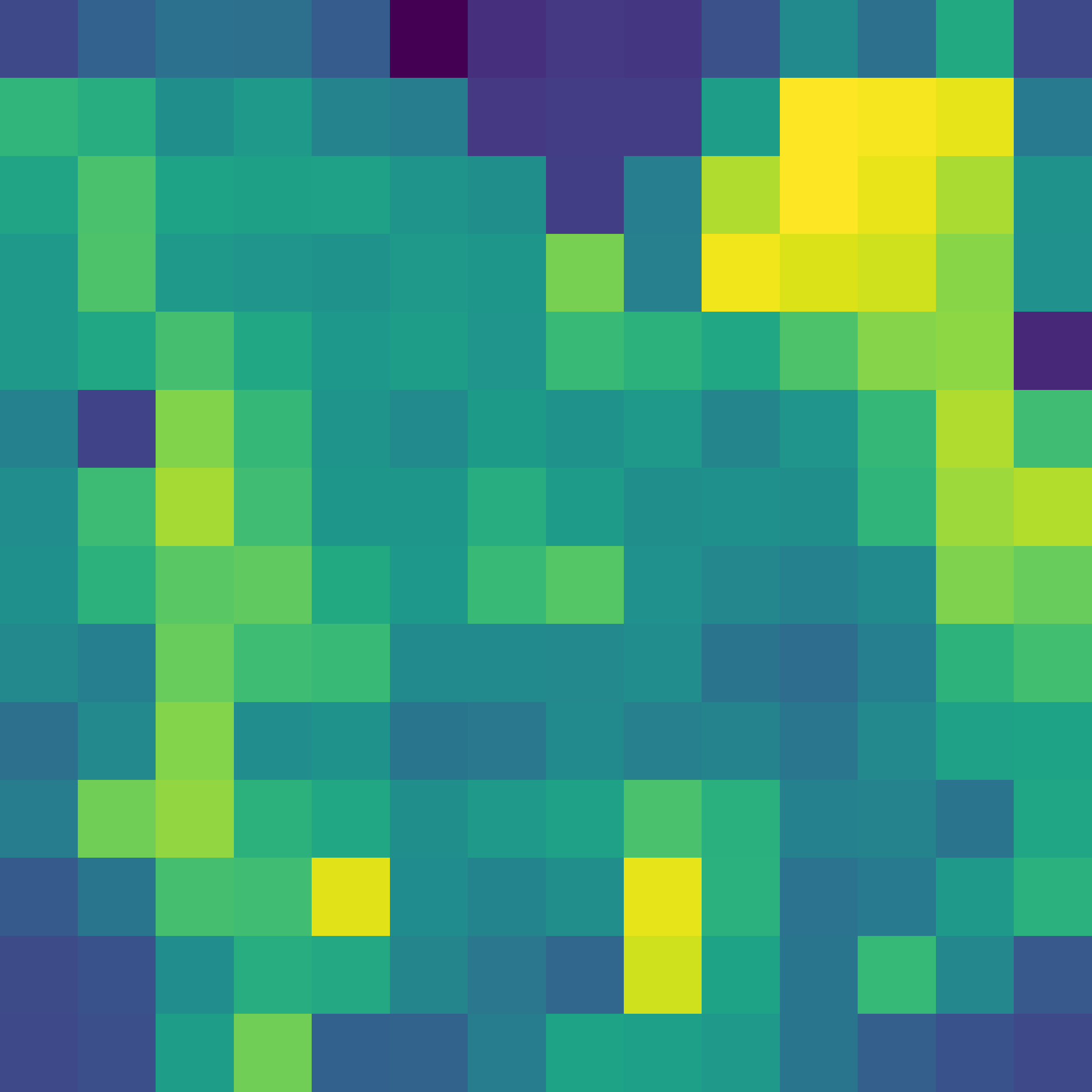}{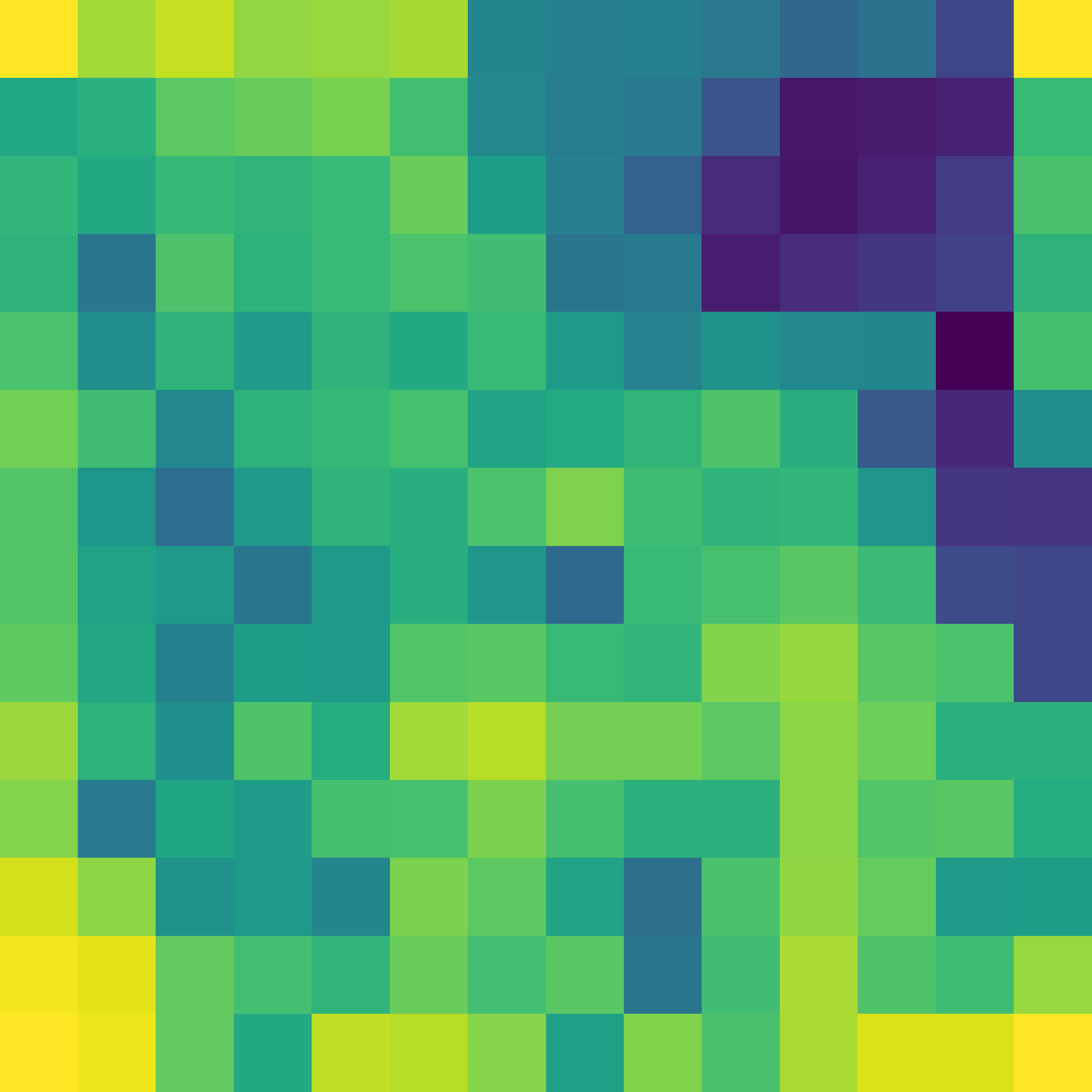}{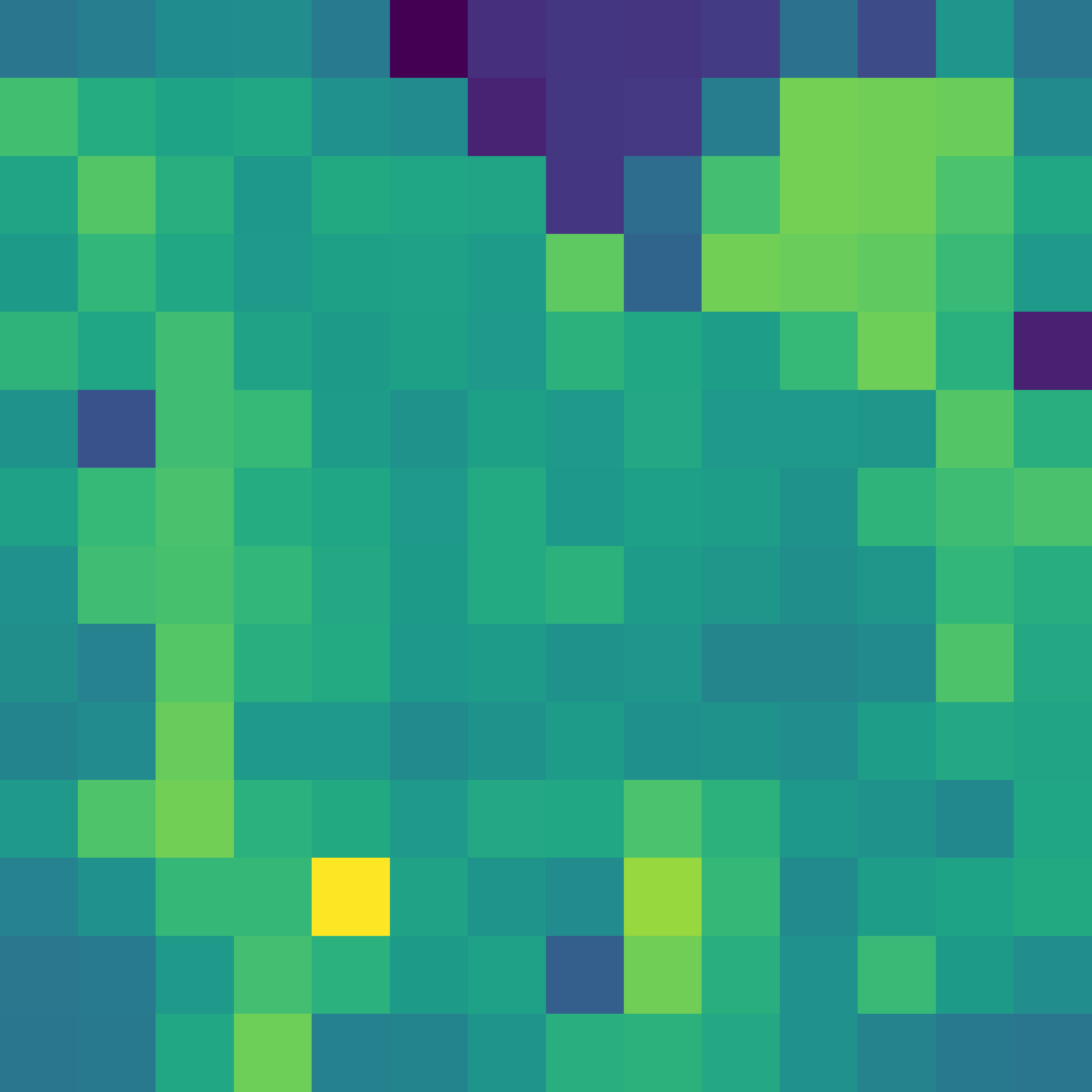} &
    \onebyfour{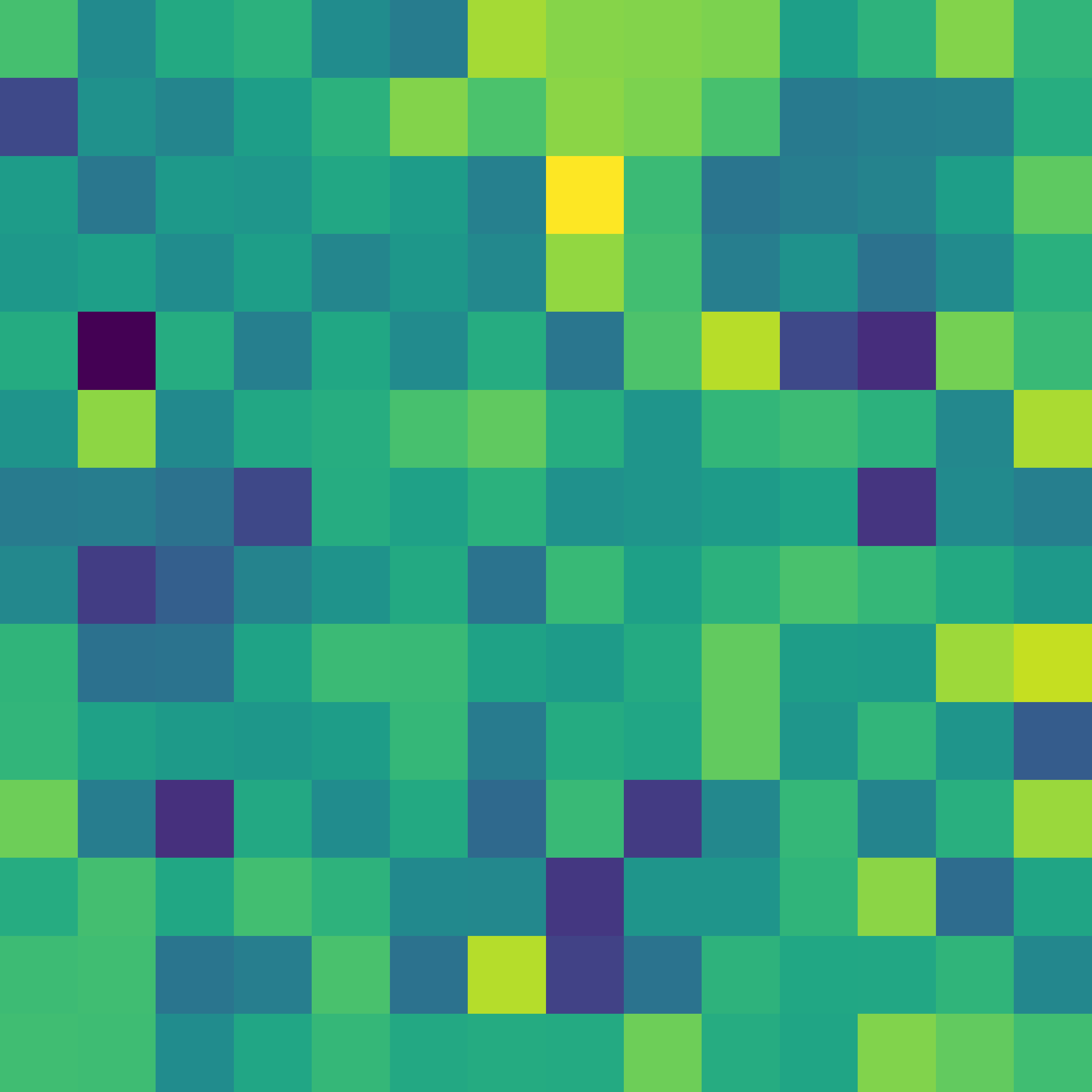}{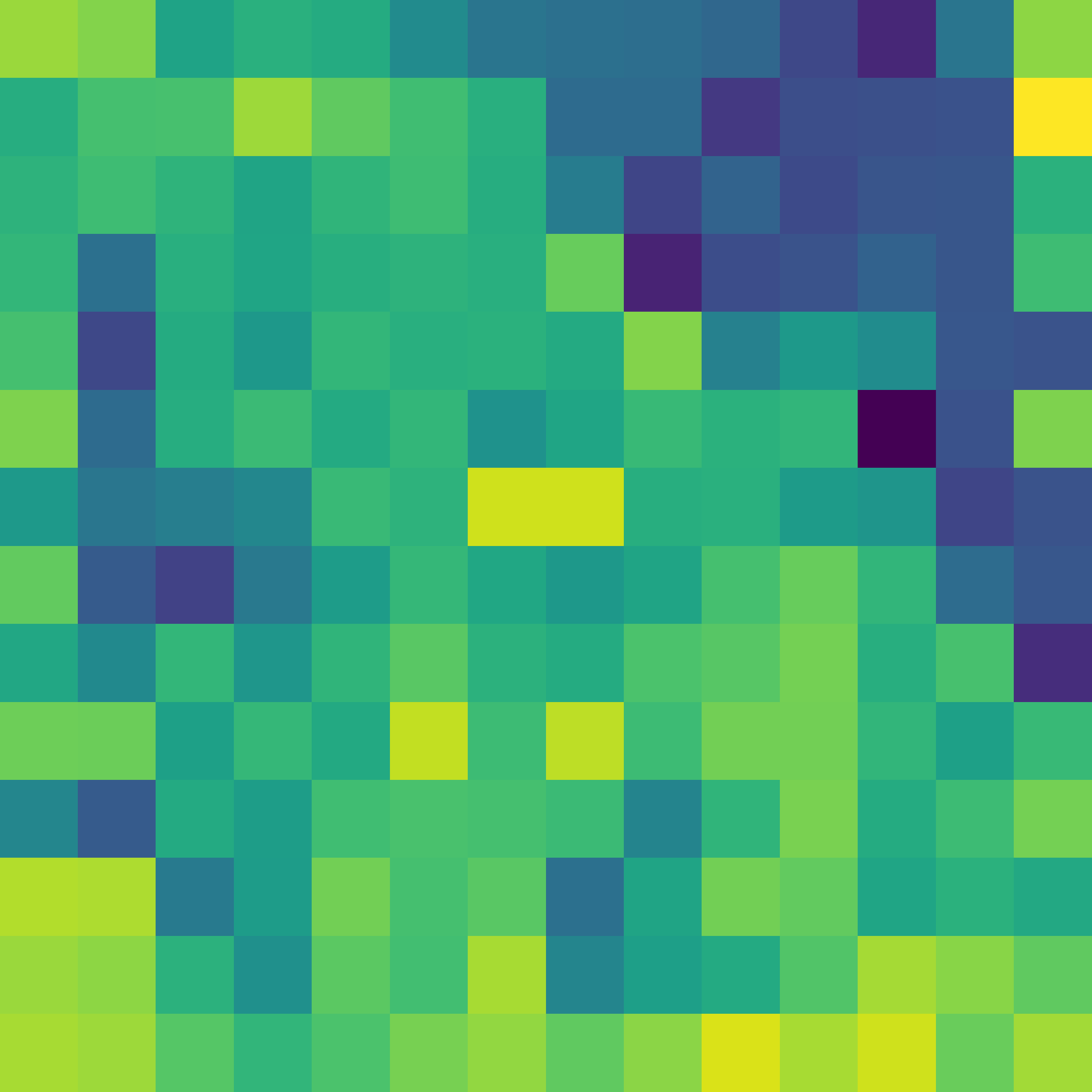}{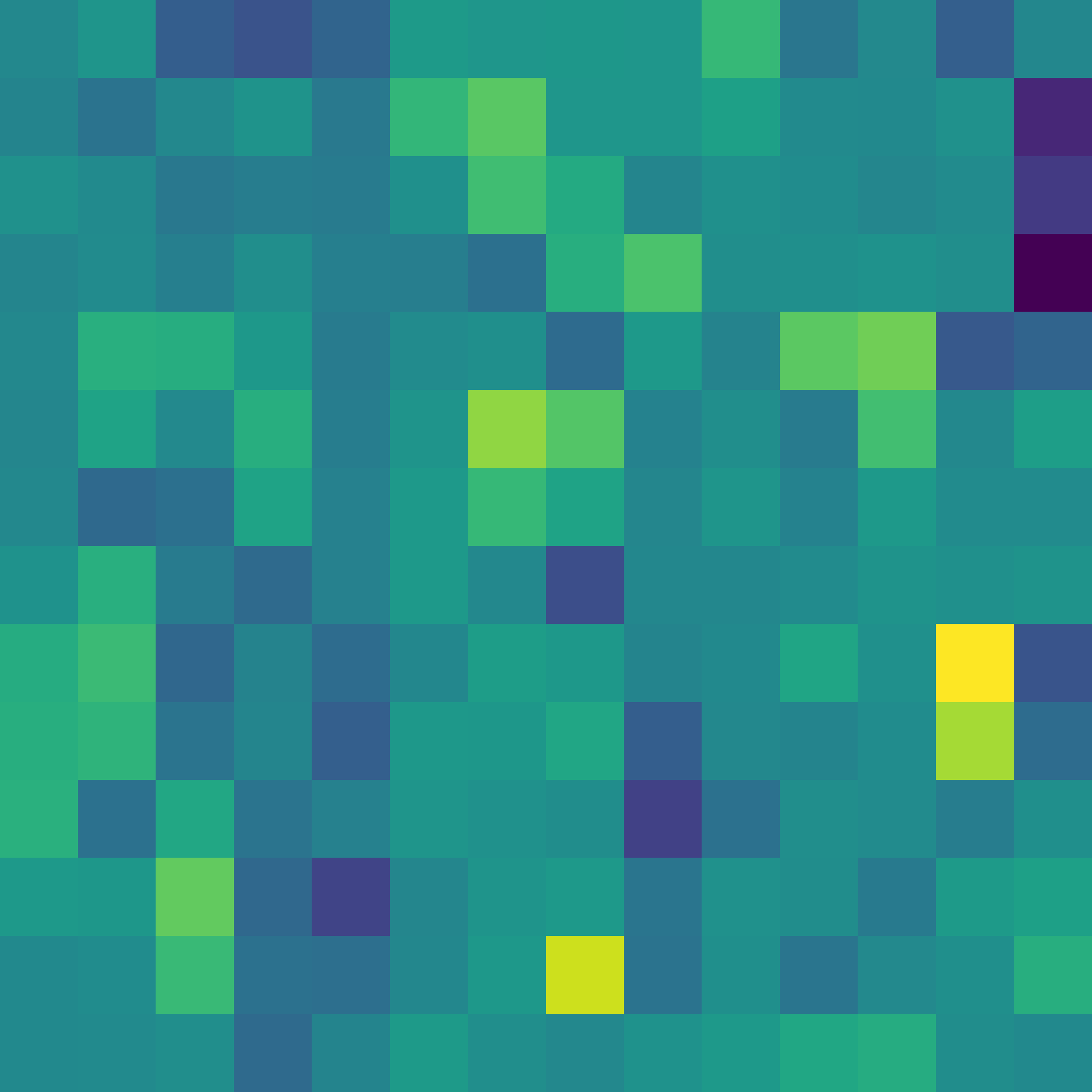}{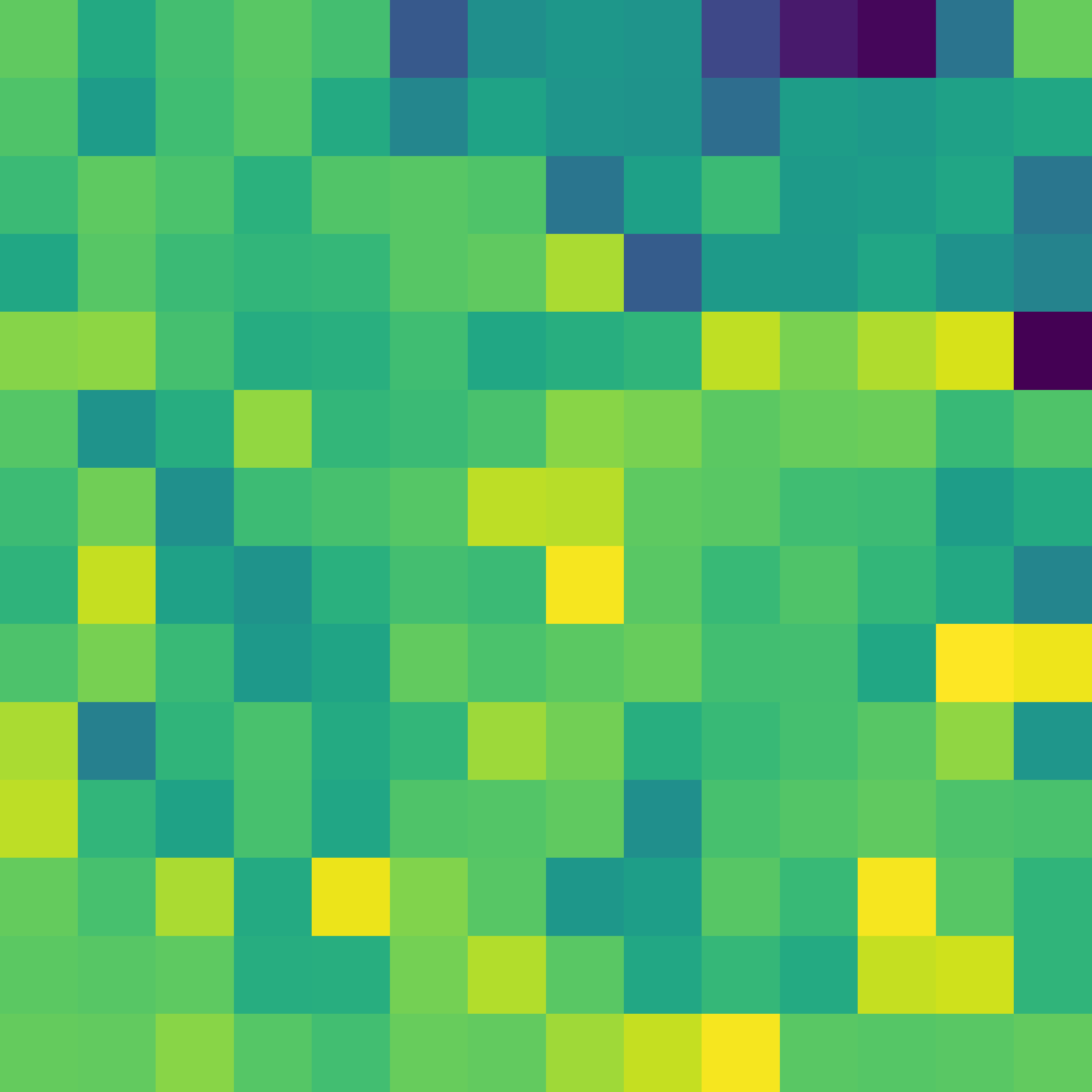} &
    \onebyfour{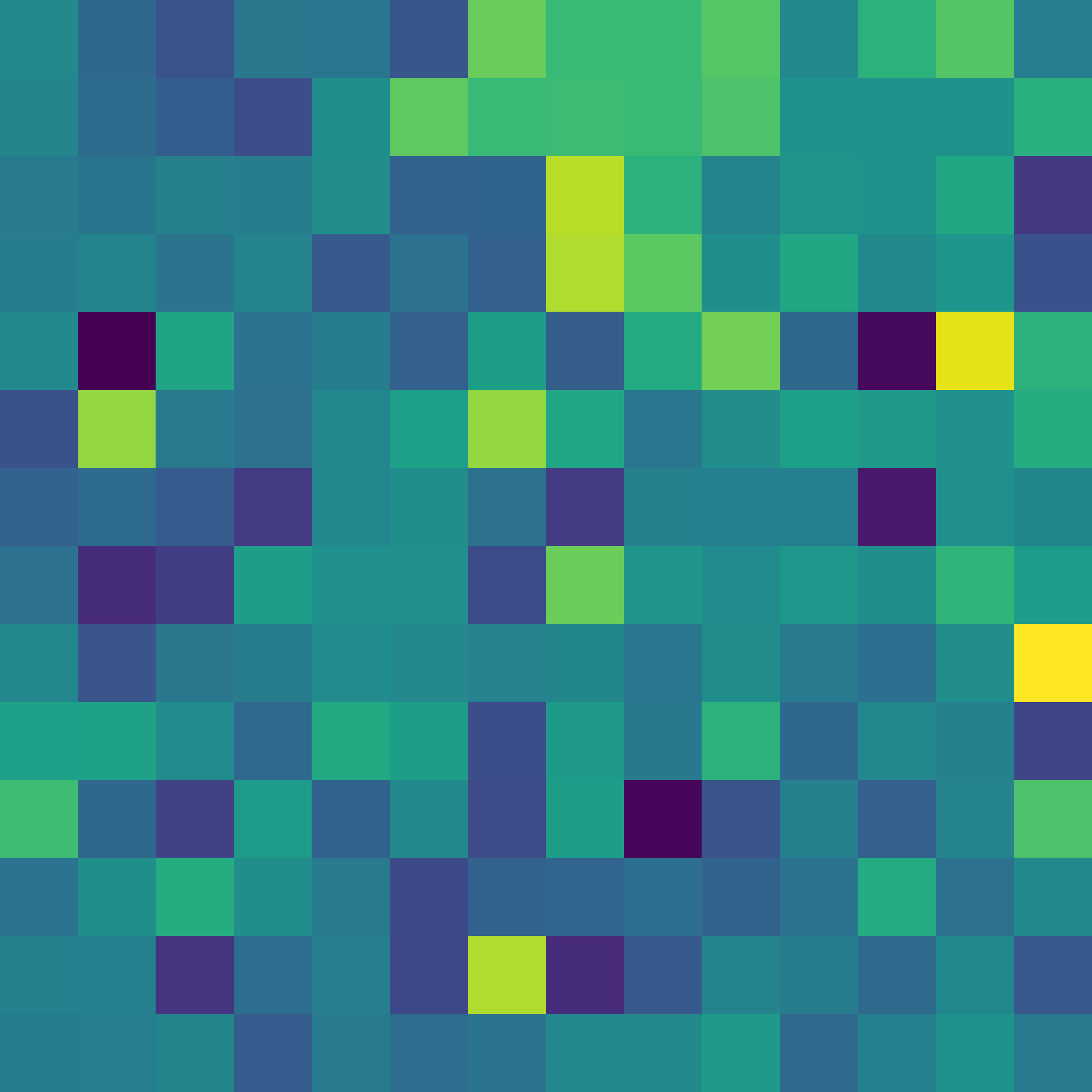}{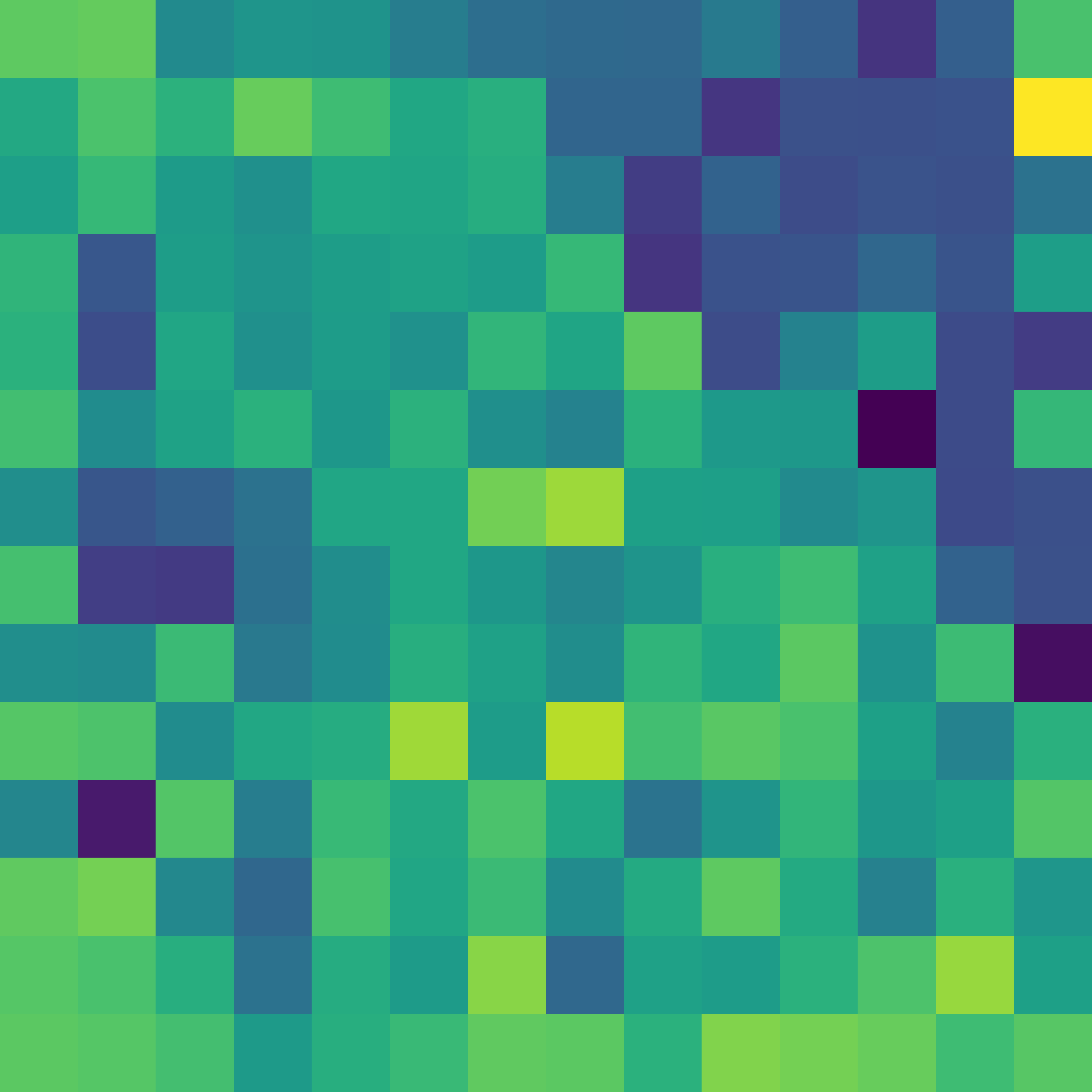}{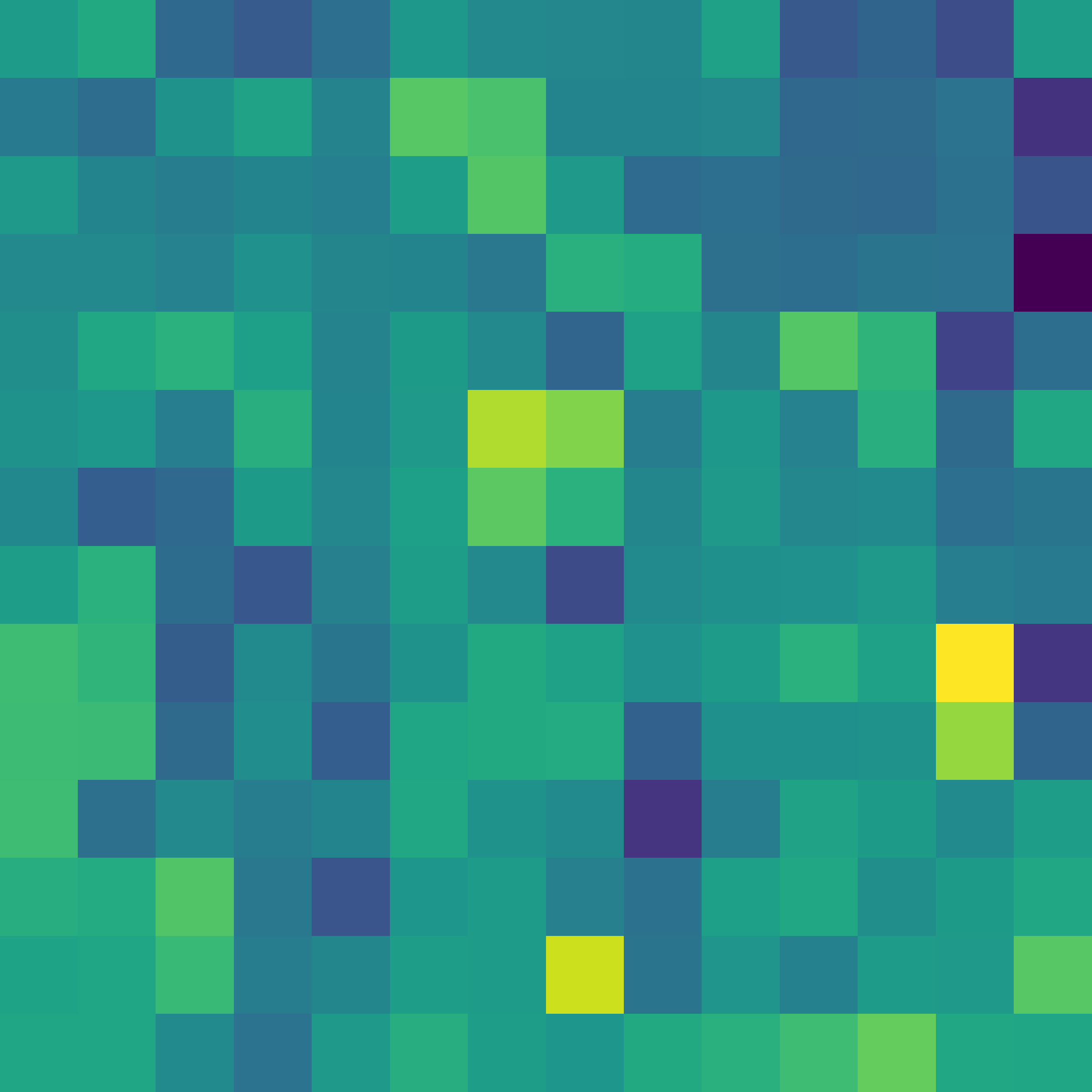}{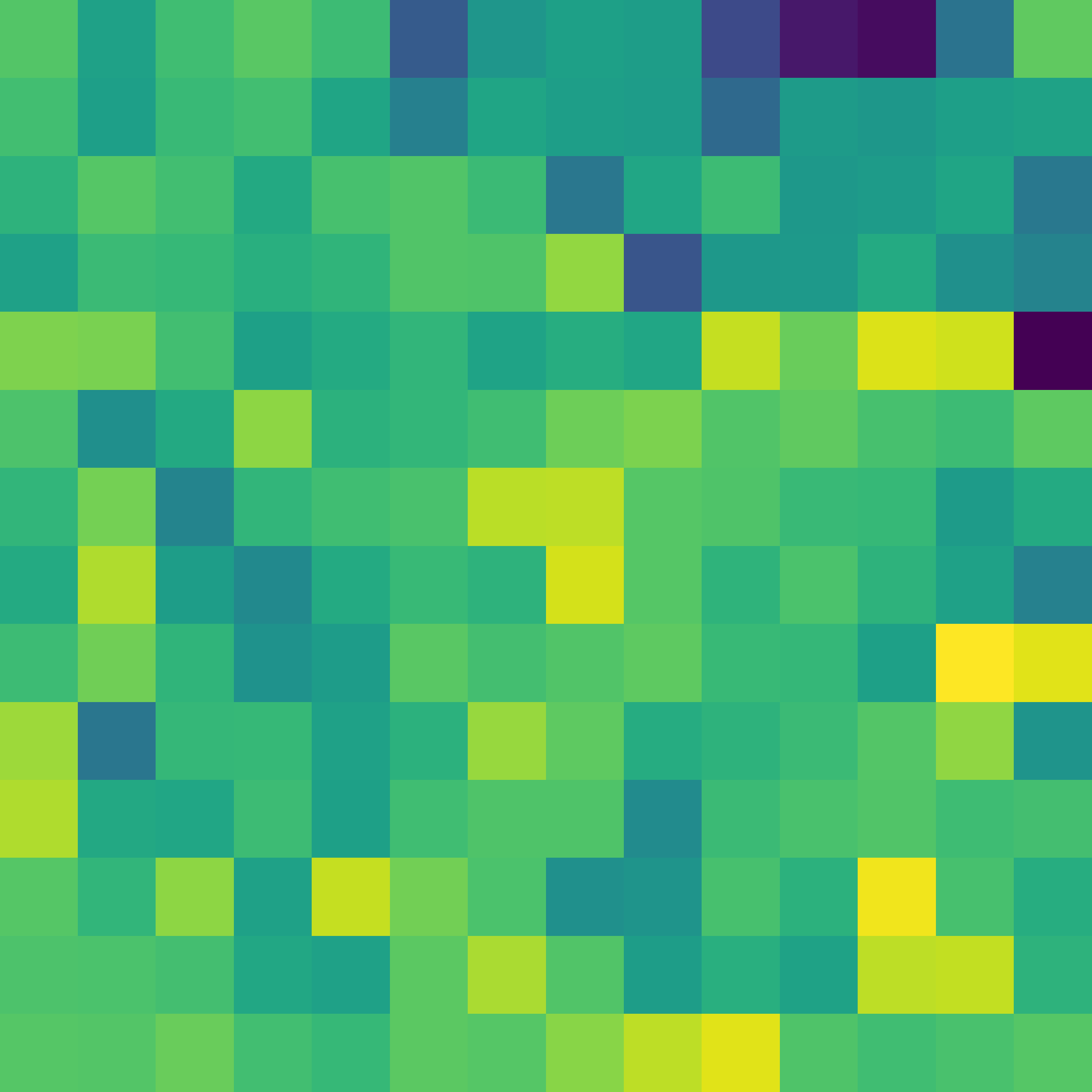} &
    \onebyfour{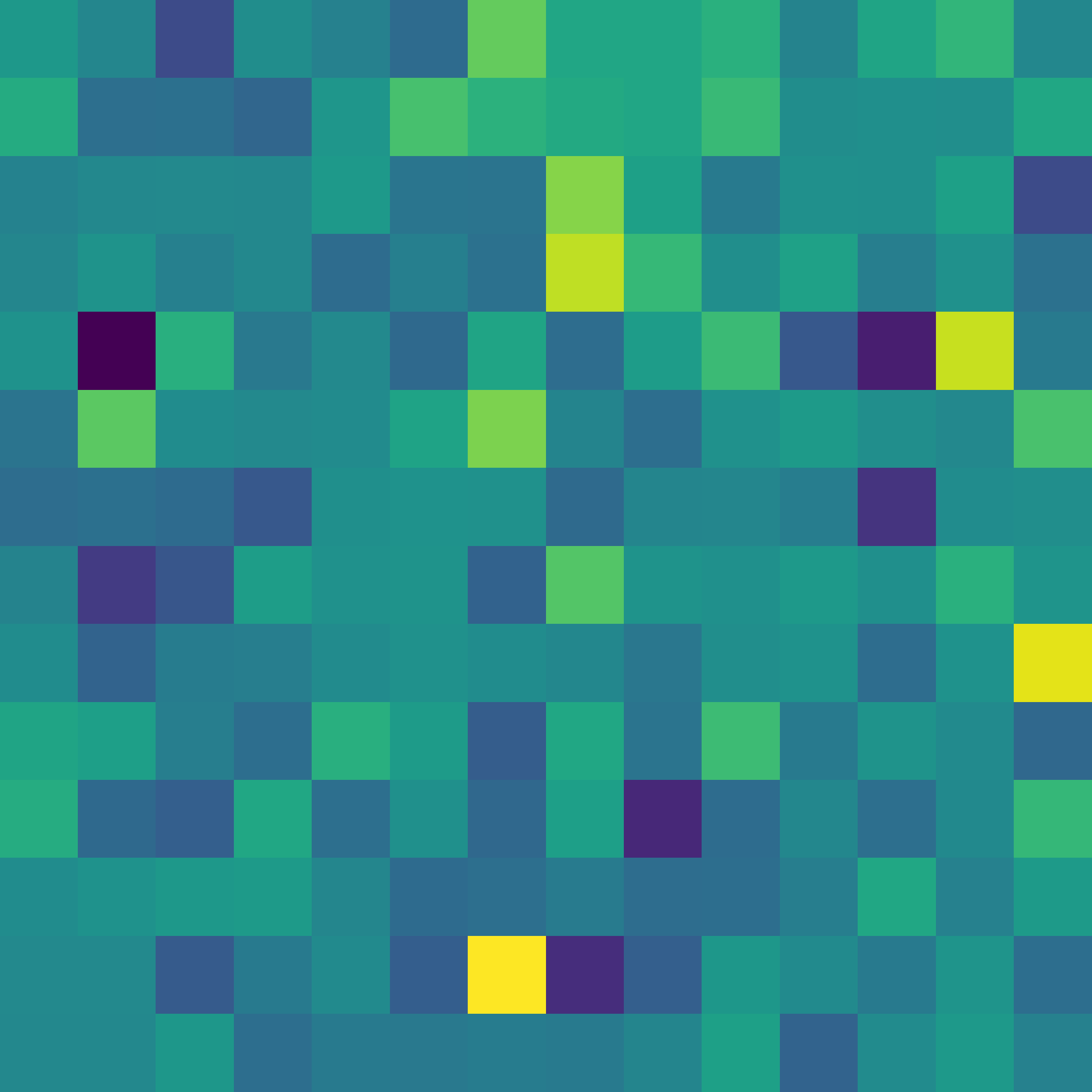}{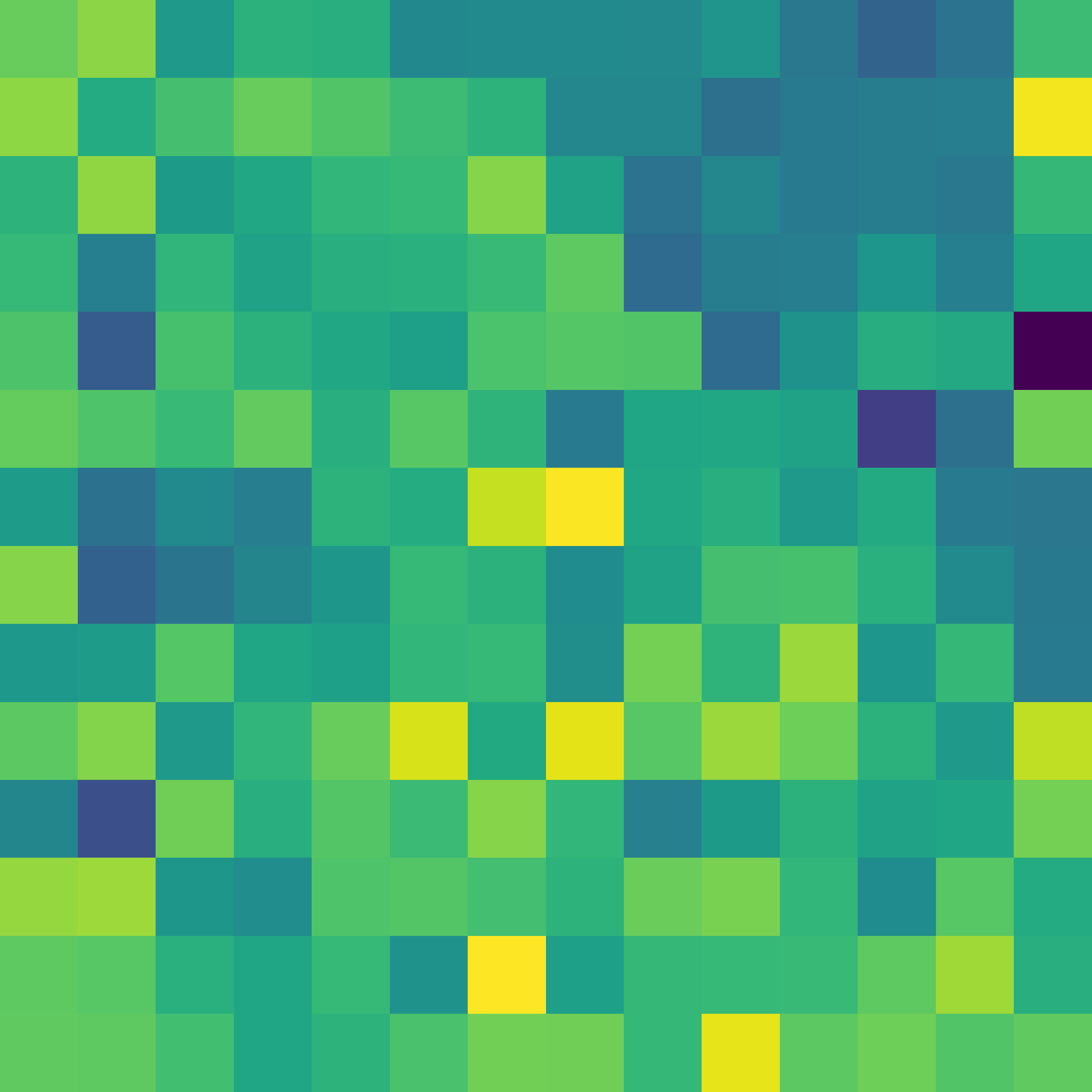}{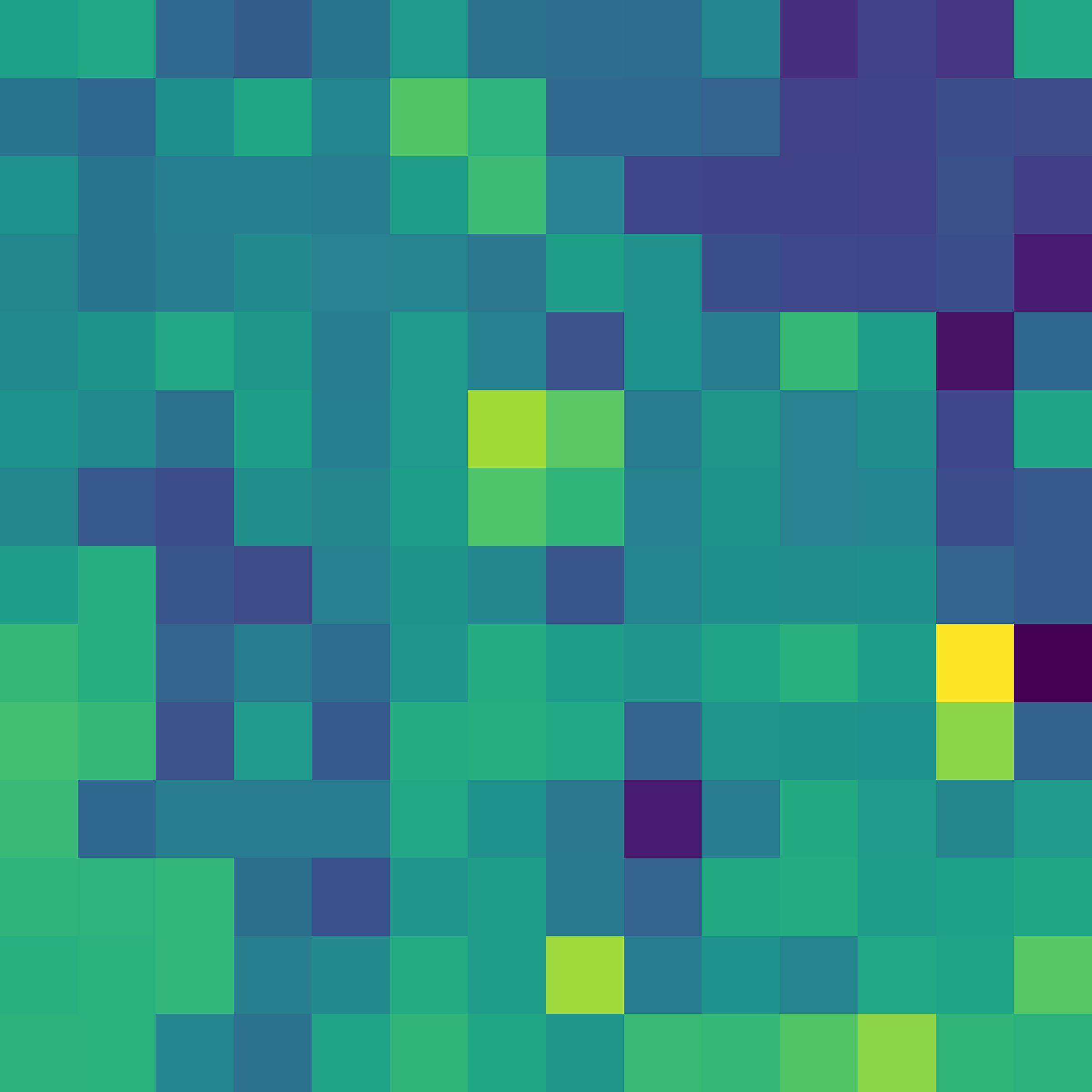}{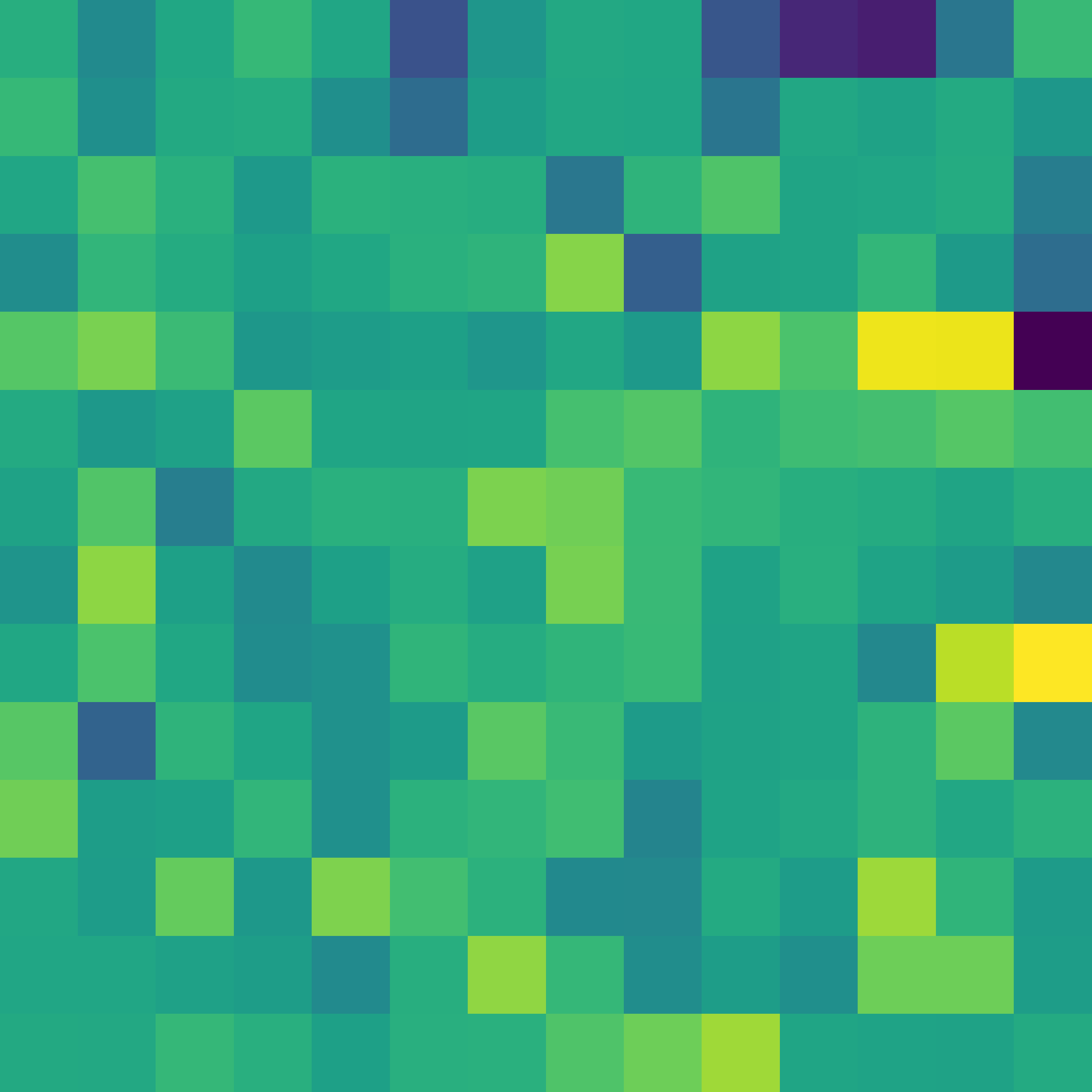} &
    \onebyfour{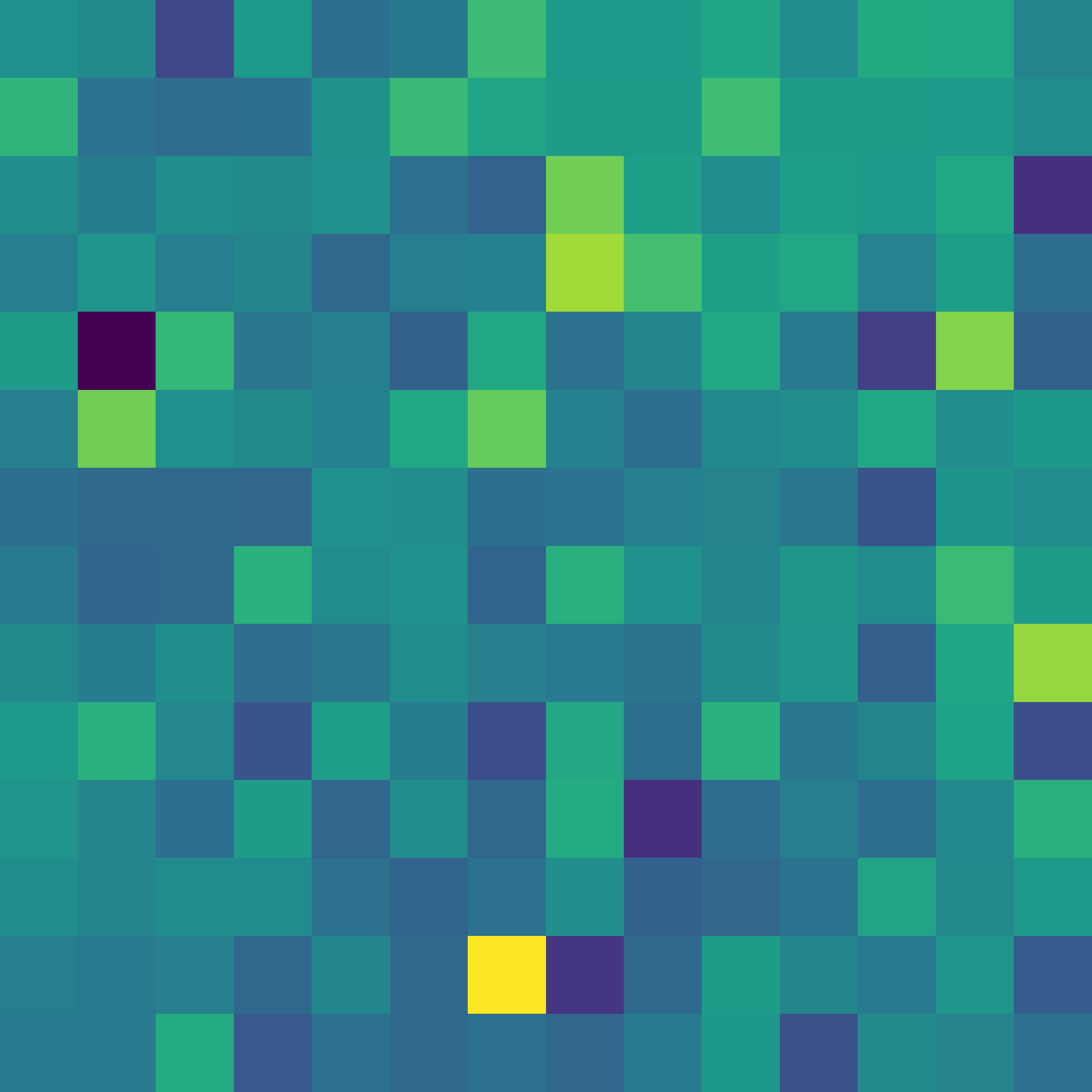}{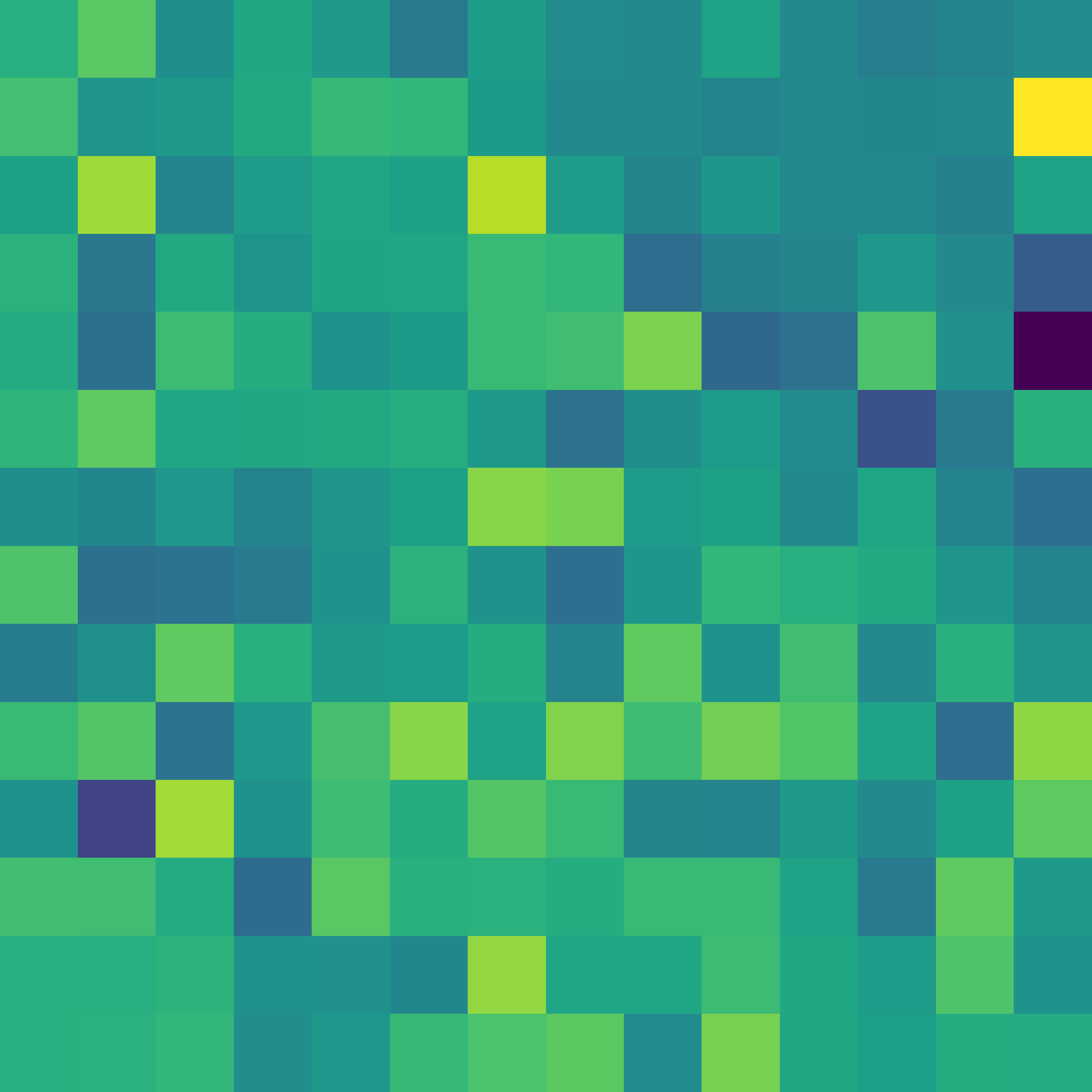}{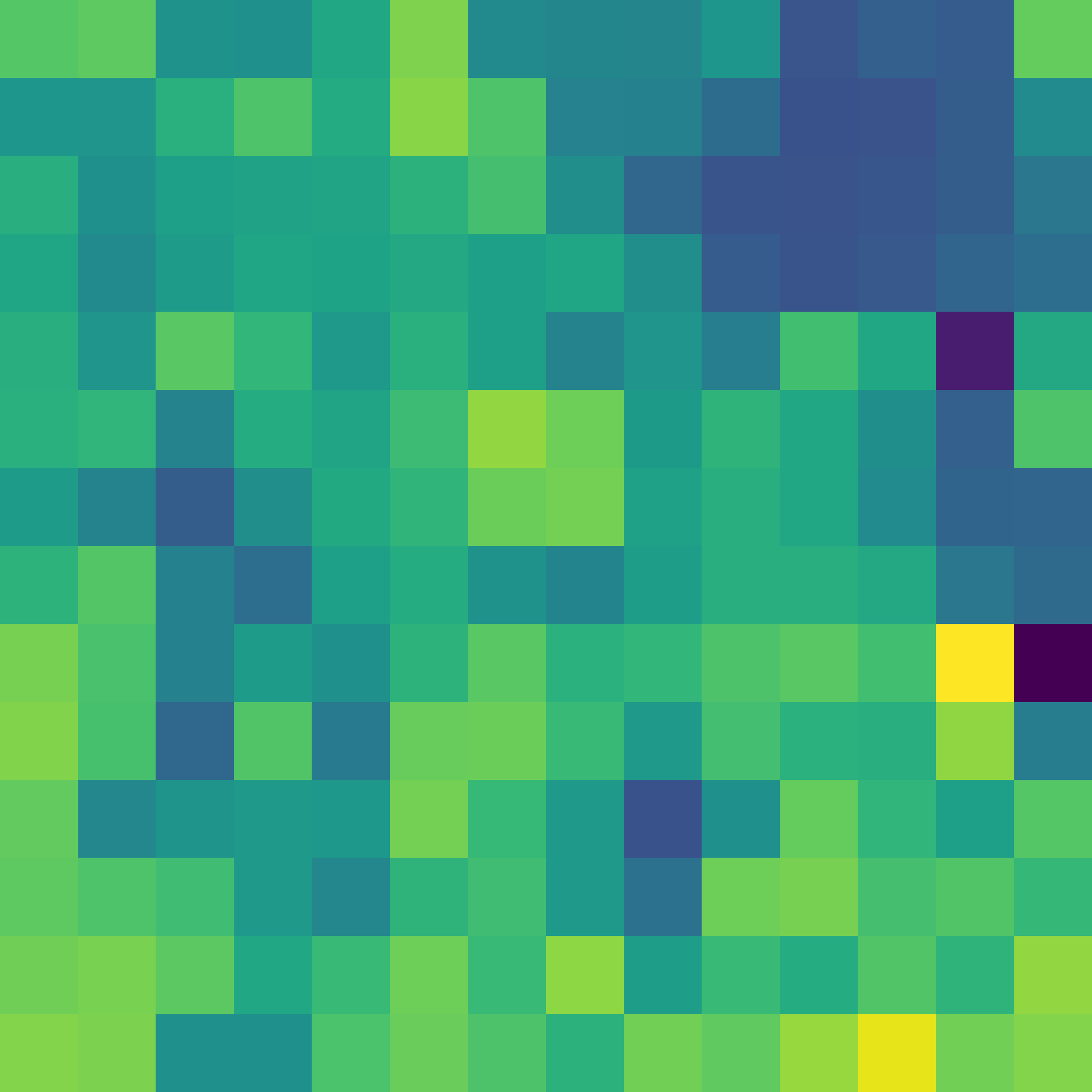}{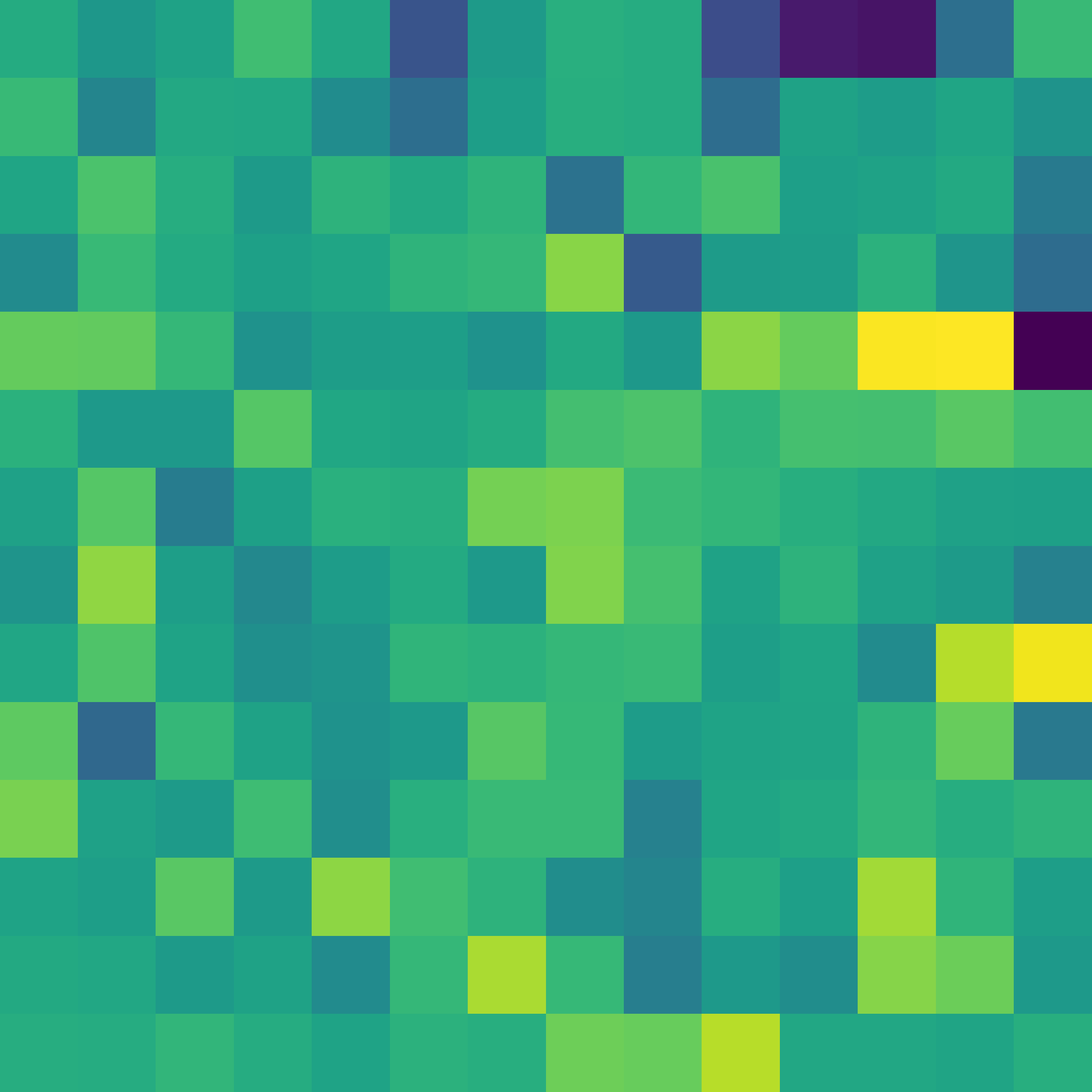} \\[0.4cm]
    \begin{minipage}{0.065\textwidth}\centering\includegraphics[width=0.91\linewidth]{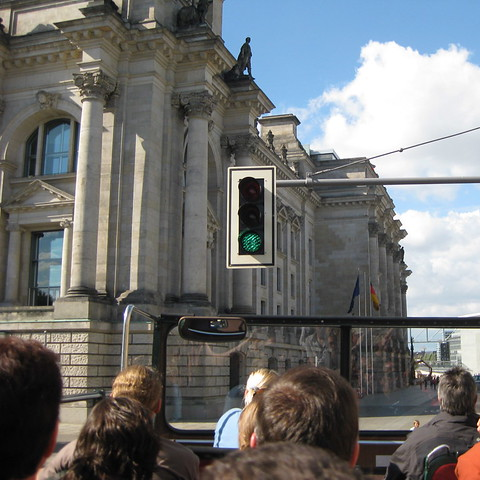}\end{minipage} & \onebyfour{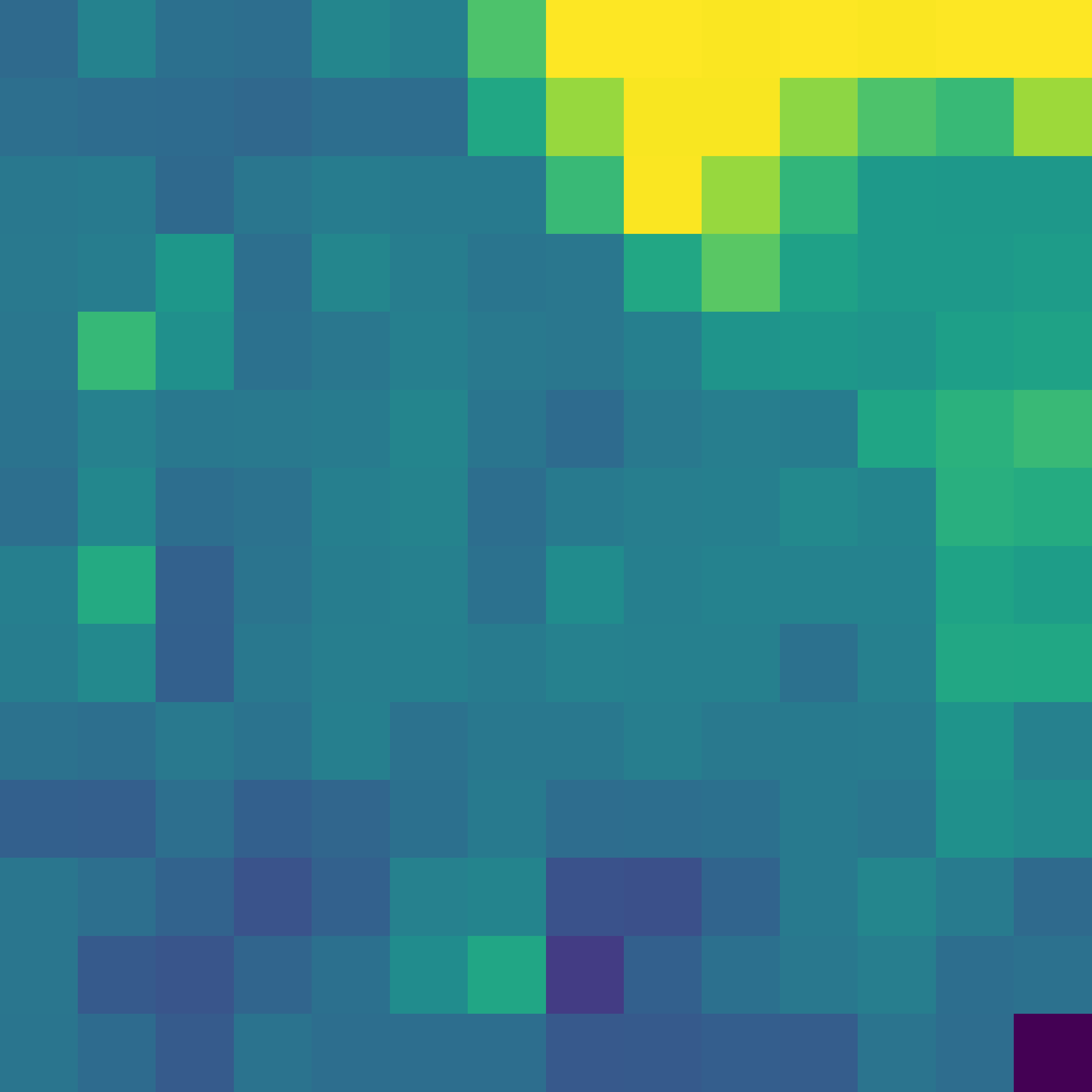}{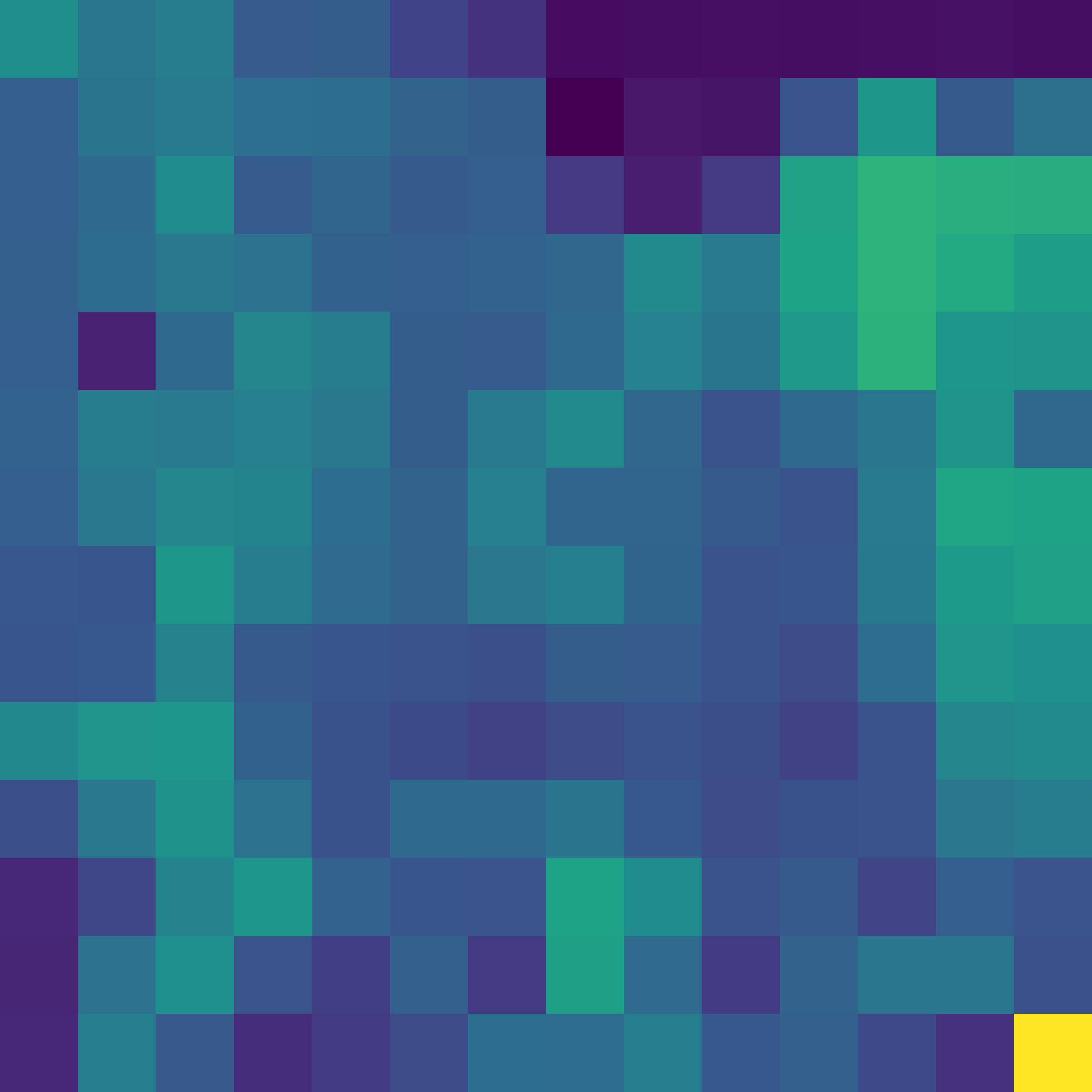}{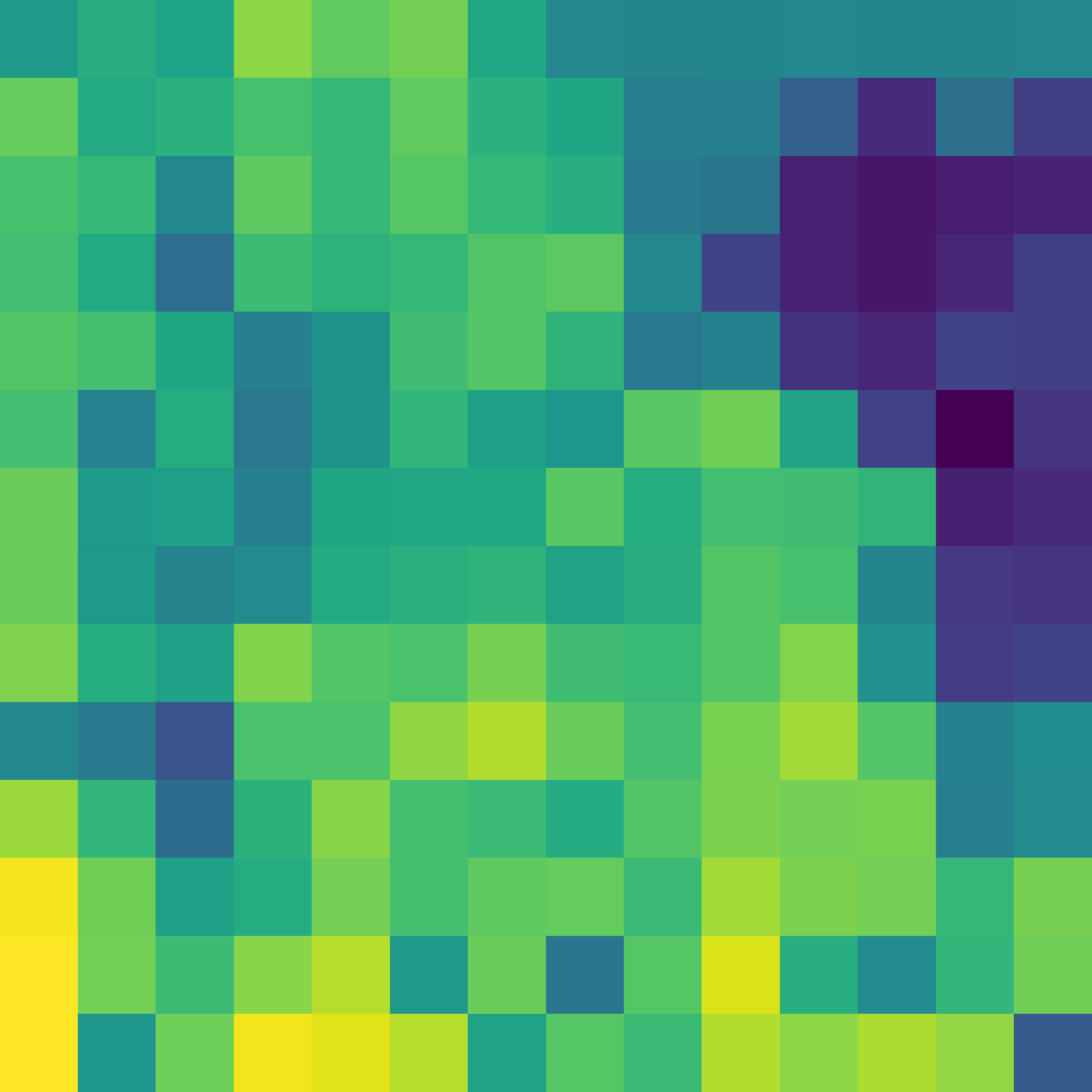}{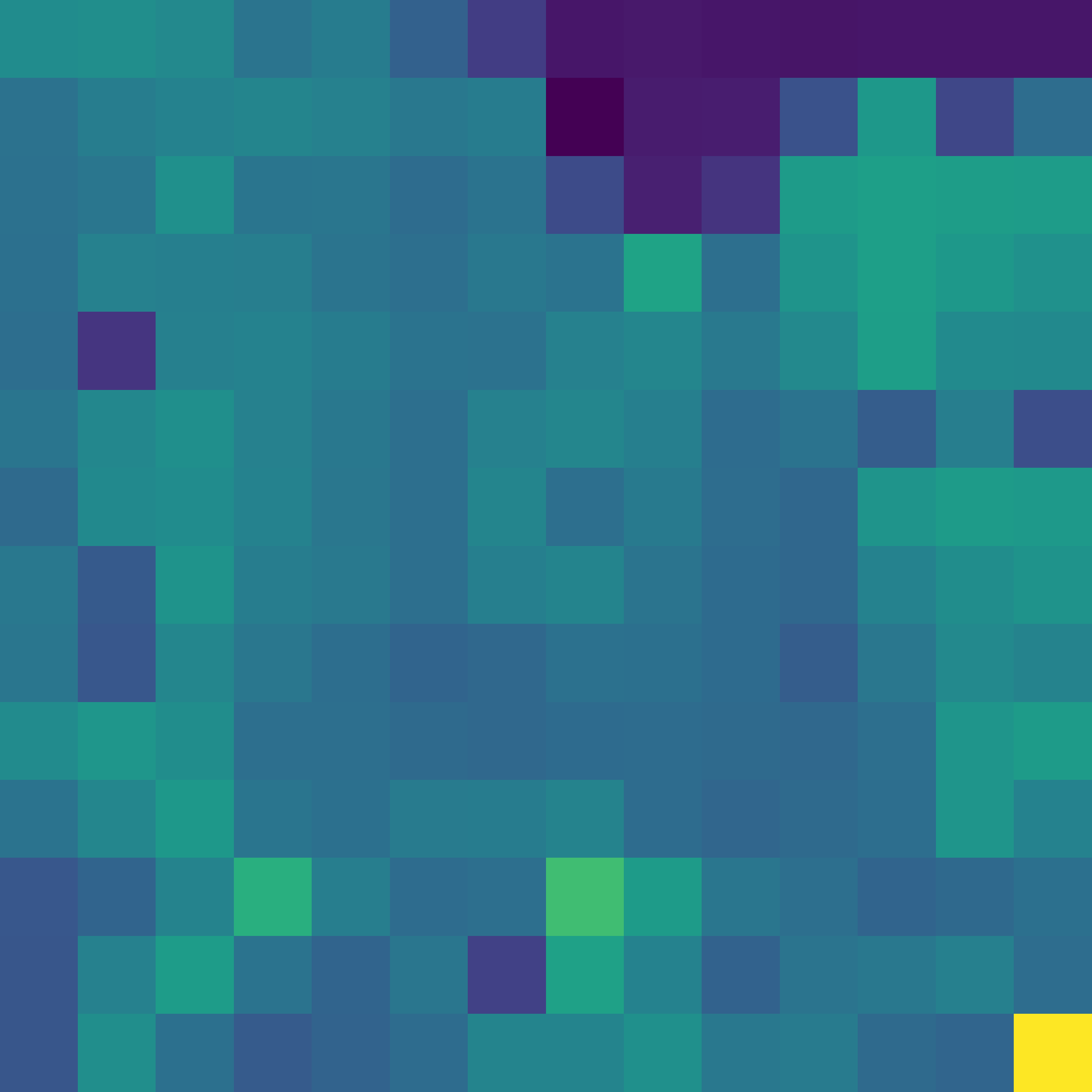} &
    \onebyfour{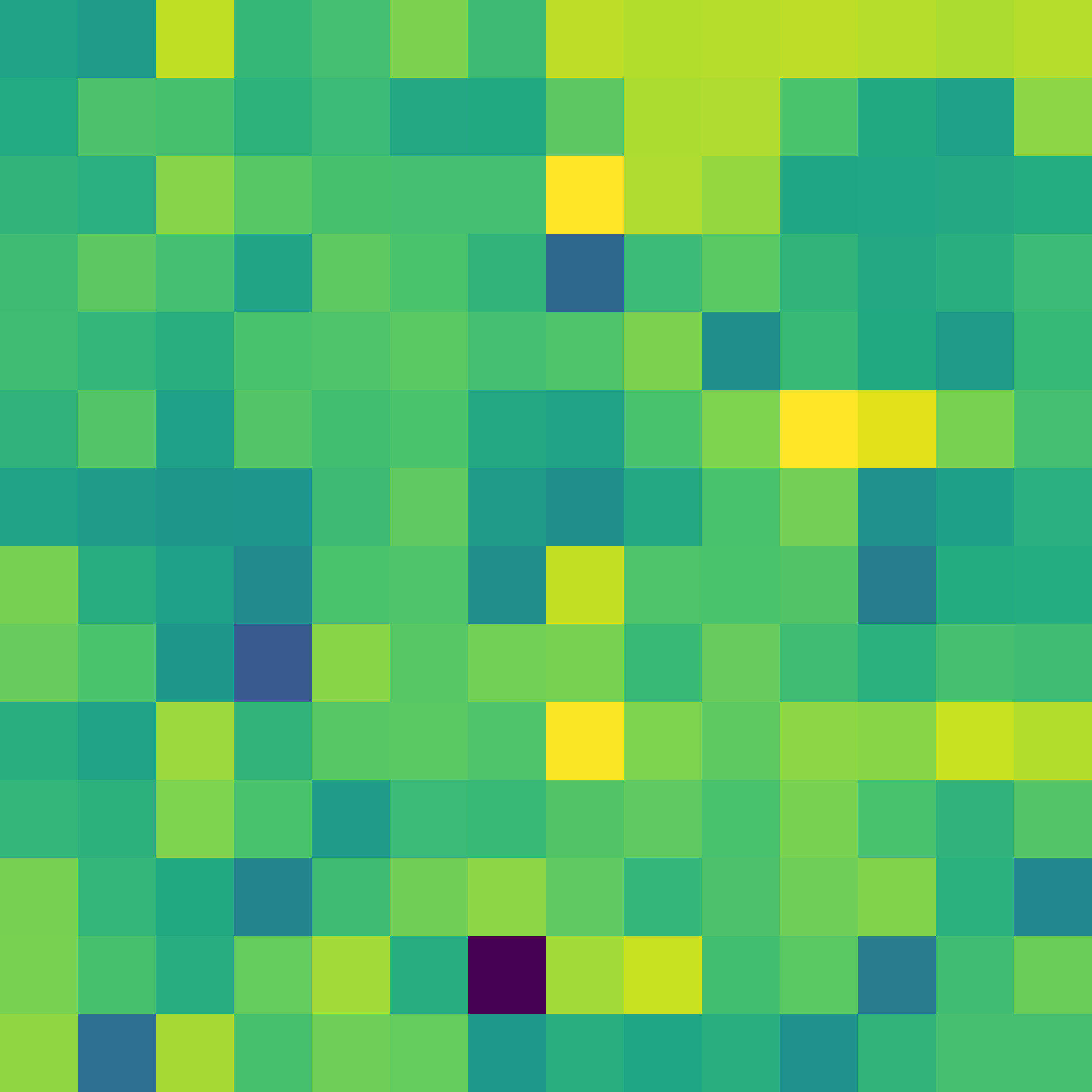}{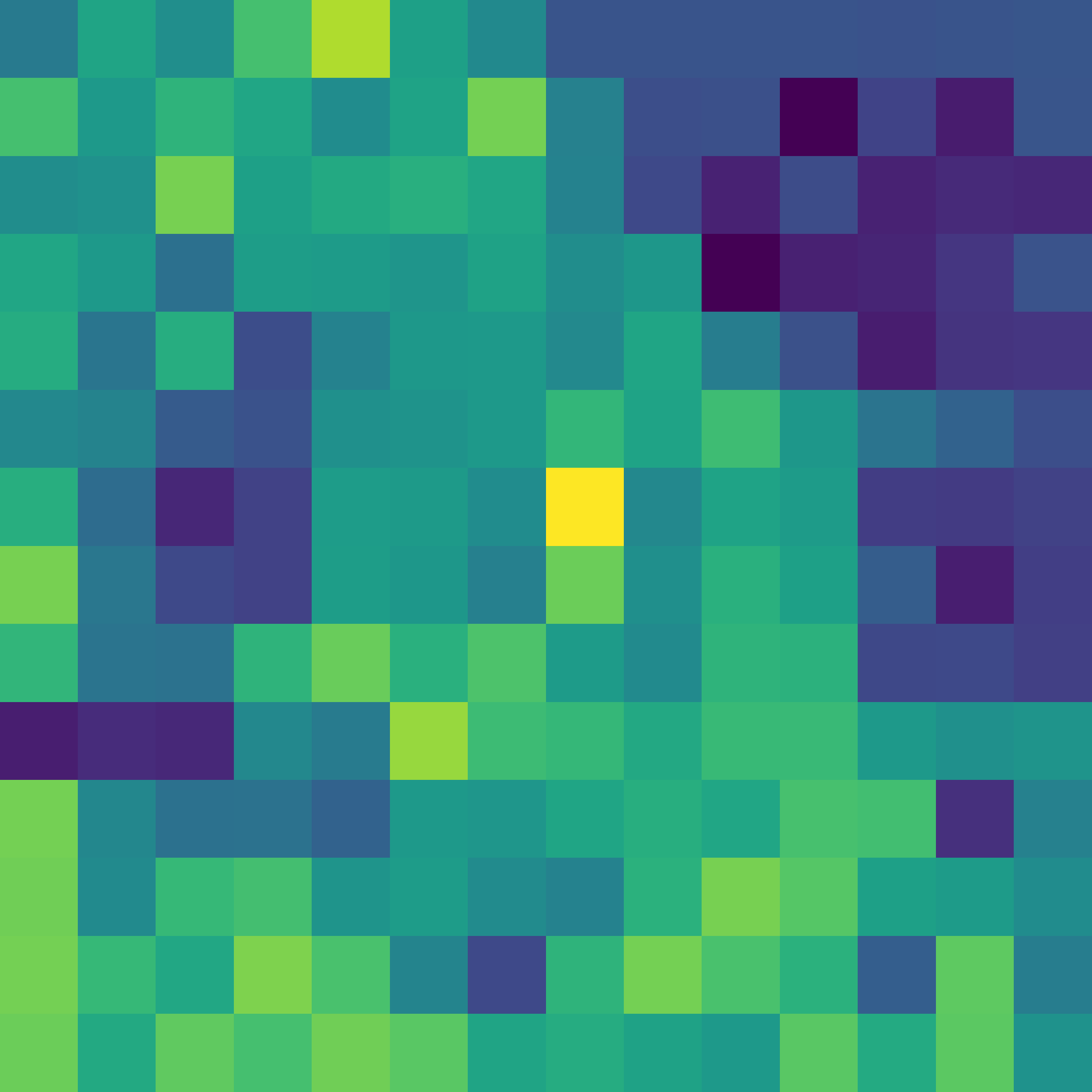}{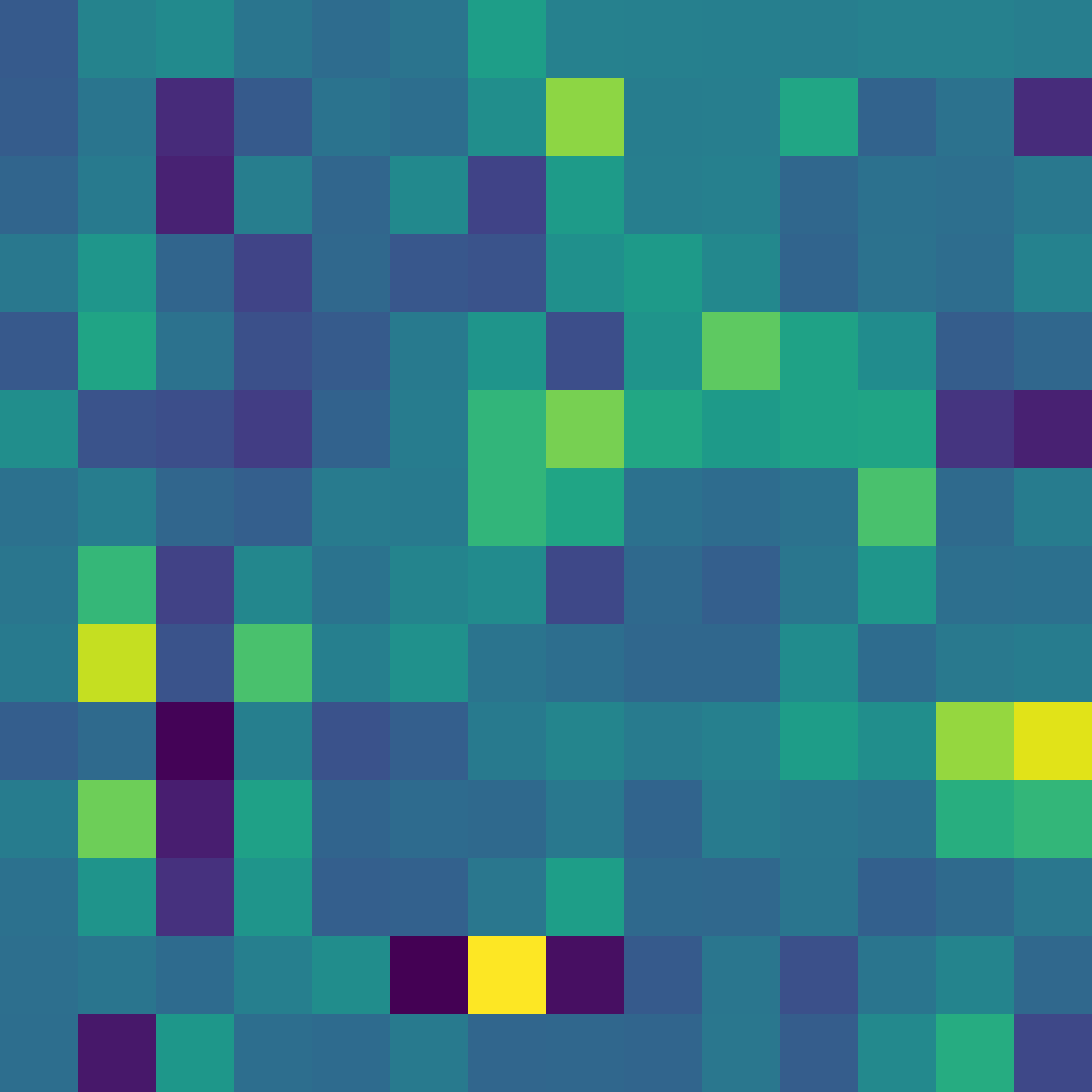}{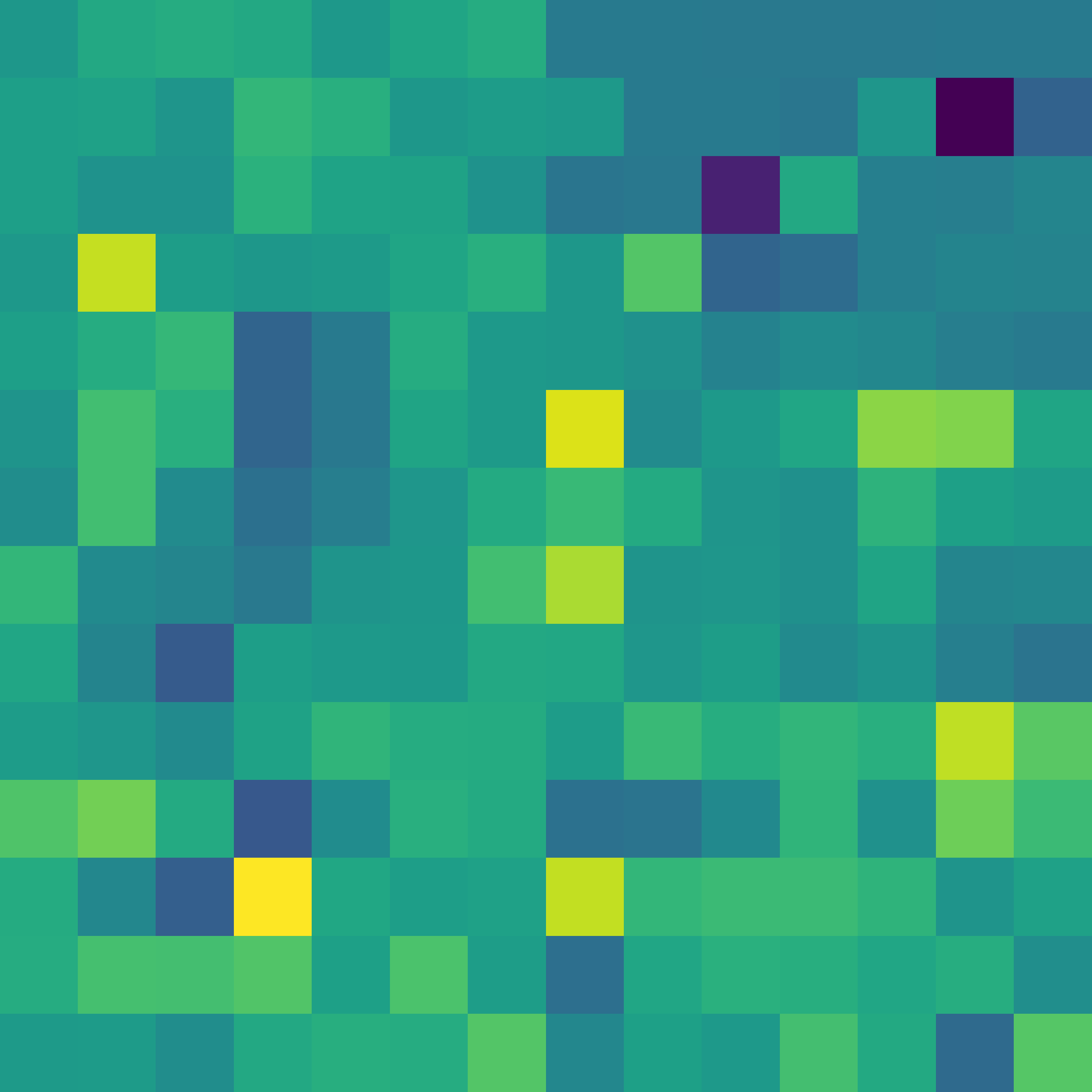} &
    \onebyfour{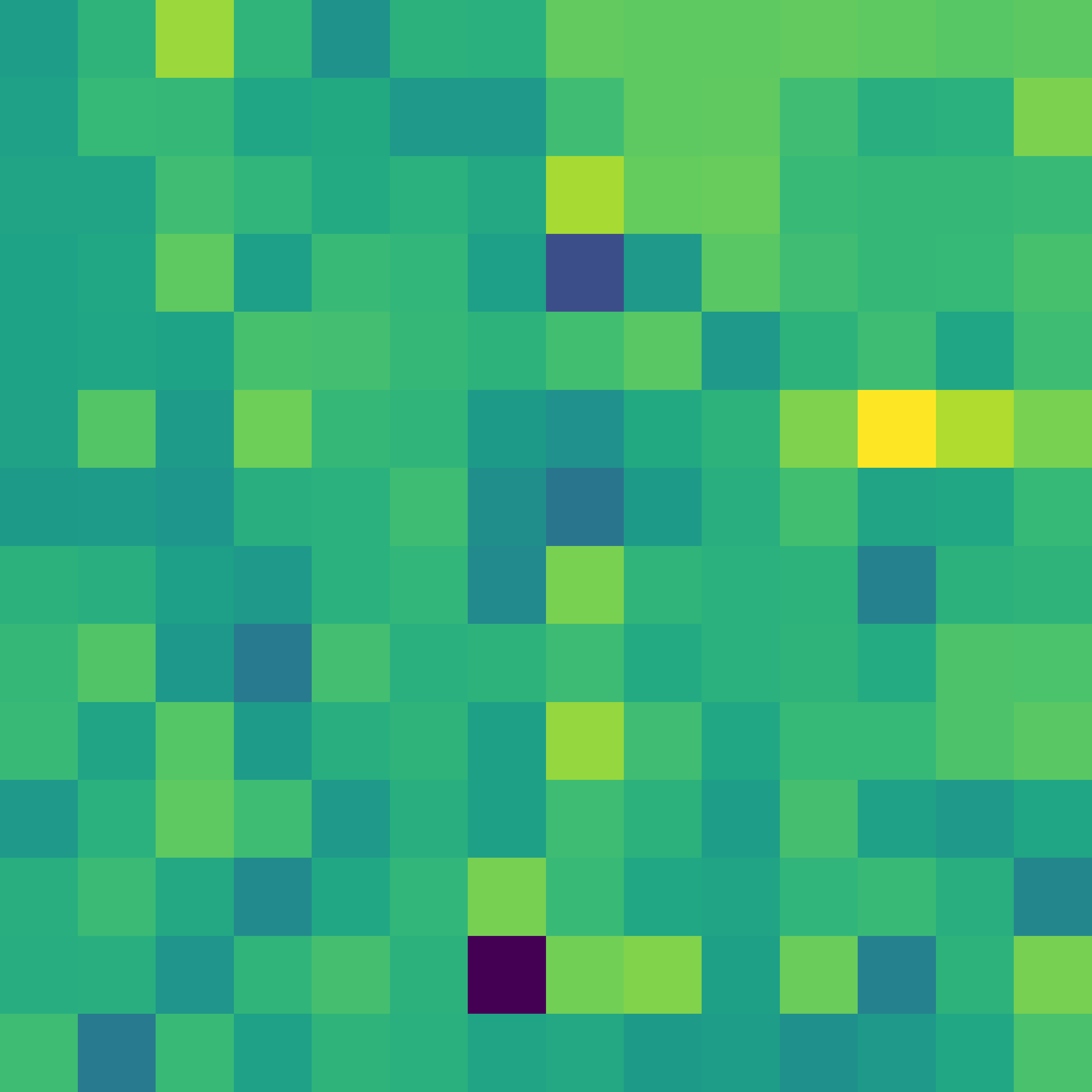}{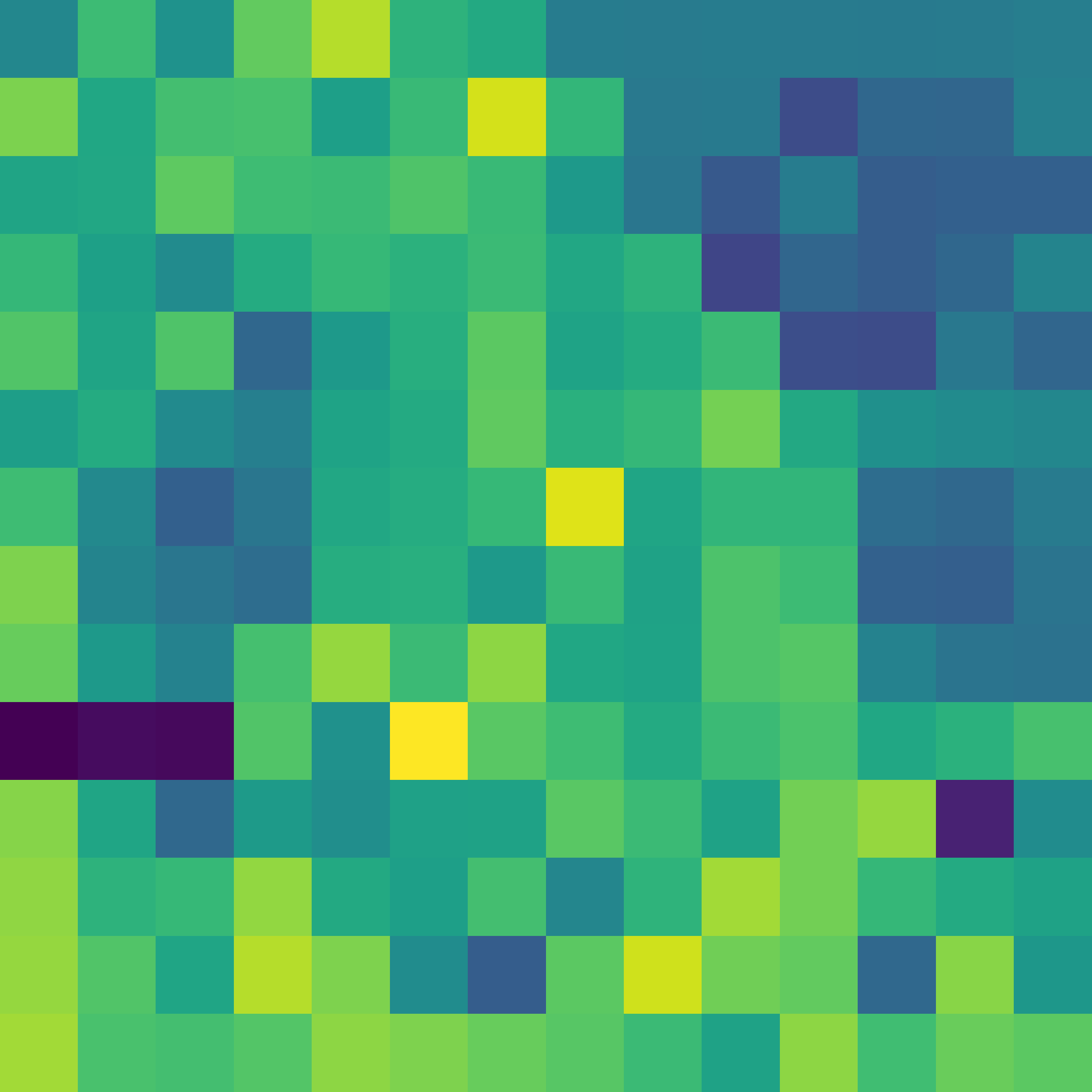}{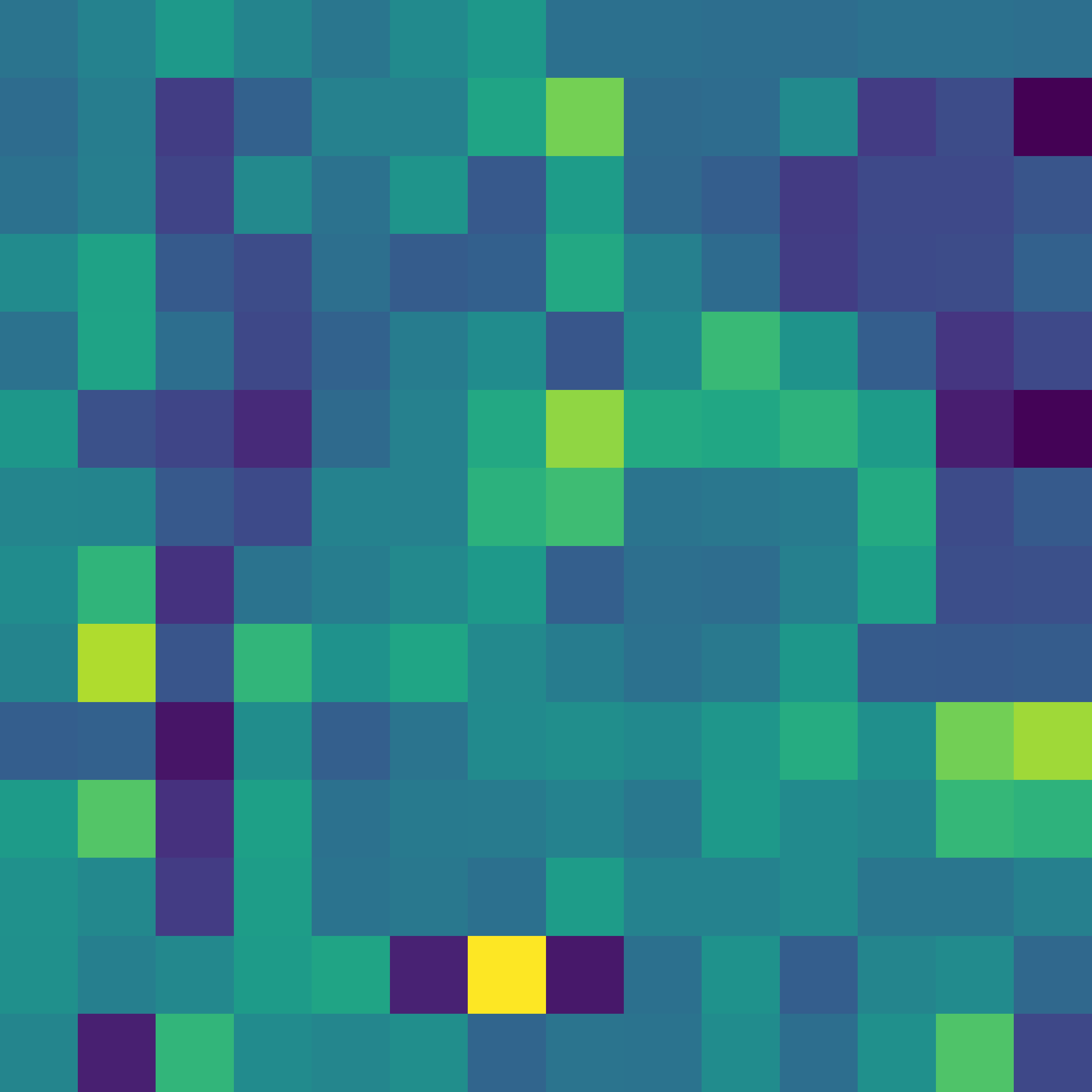}{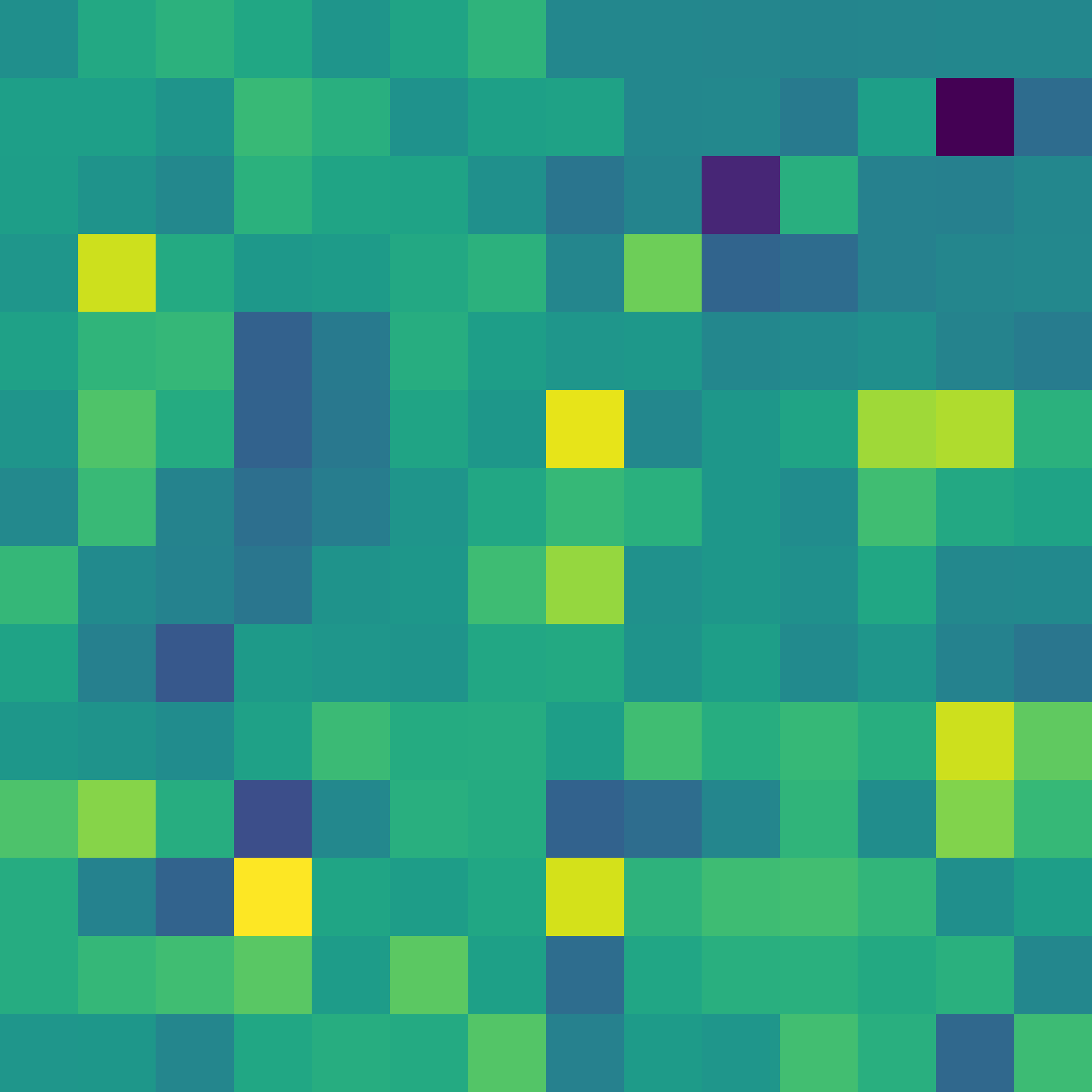} &
    \onebyfour{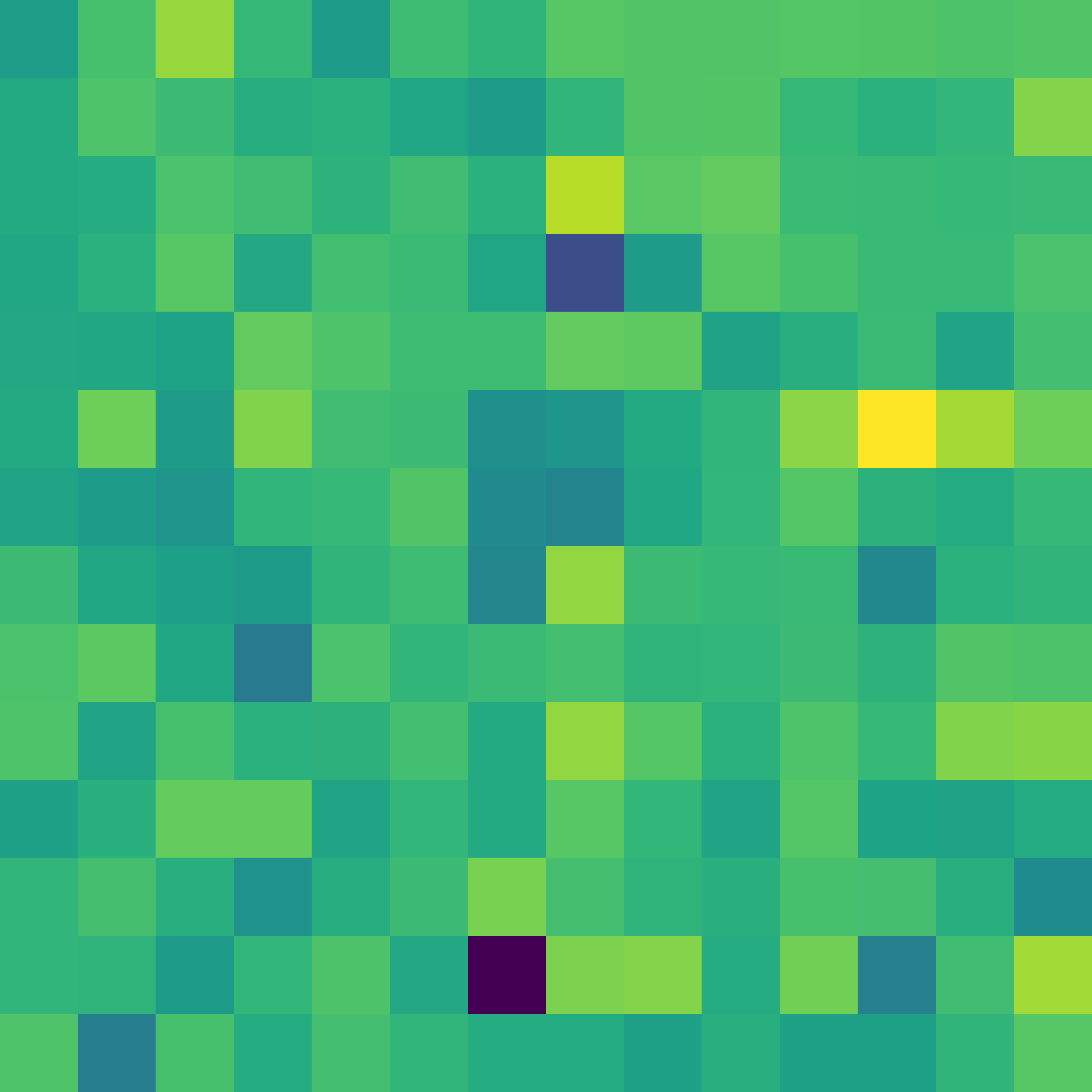}{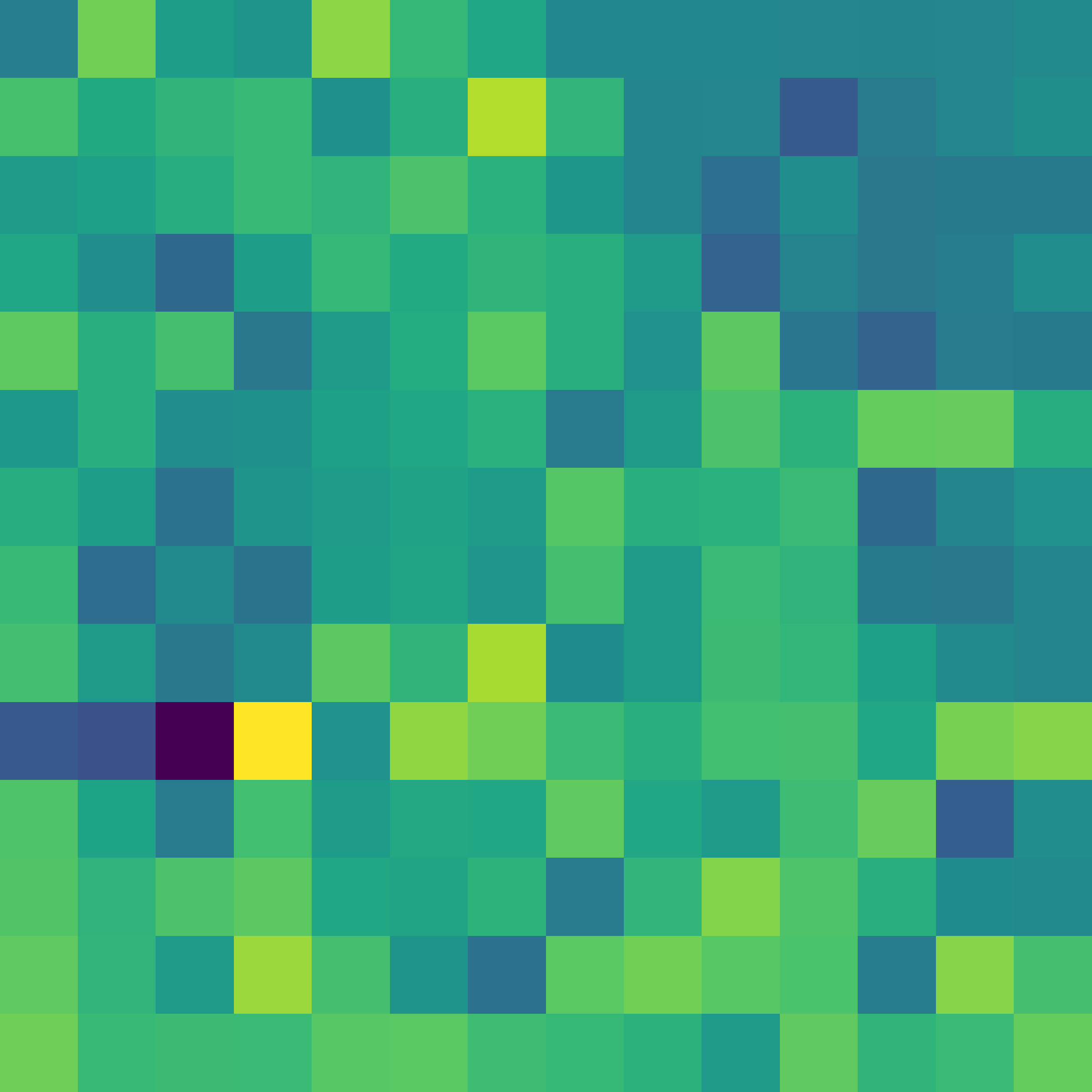}{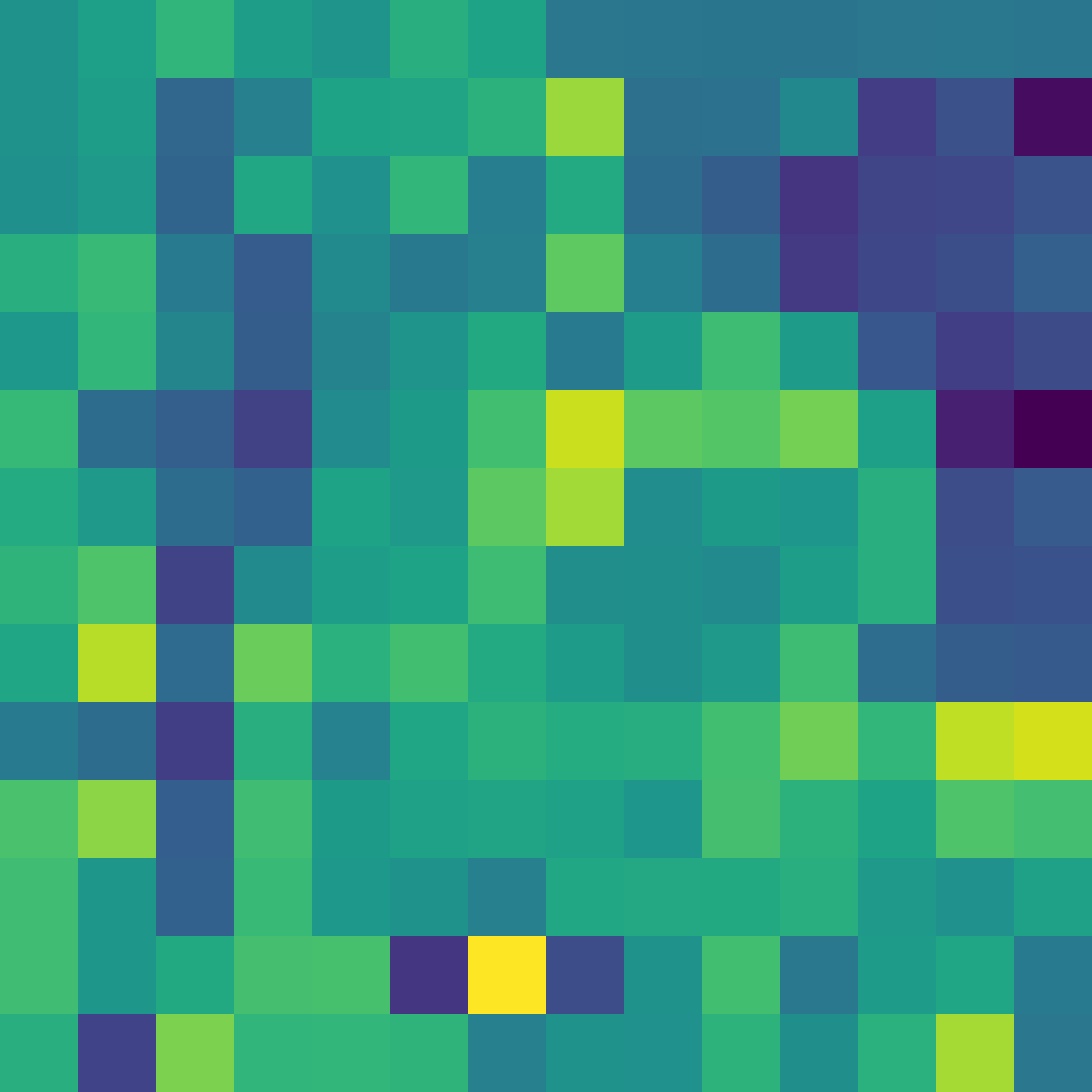}{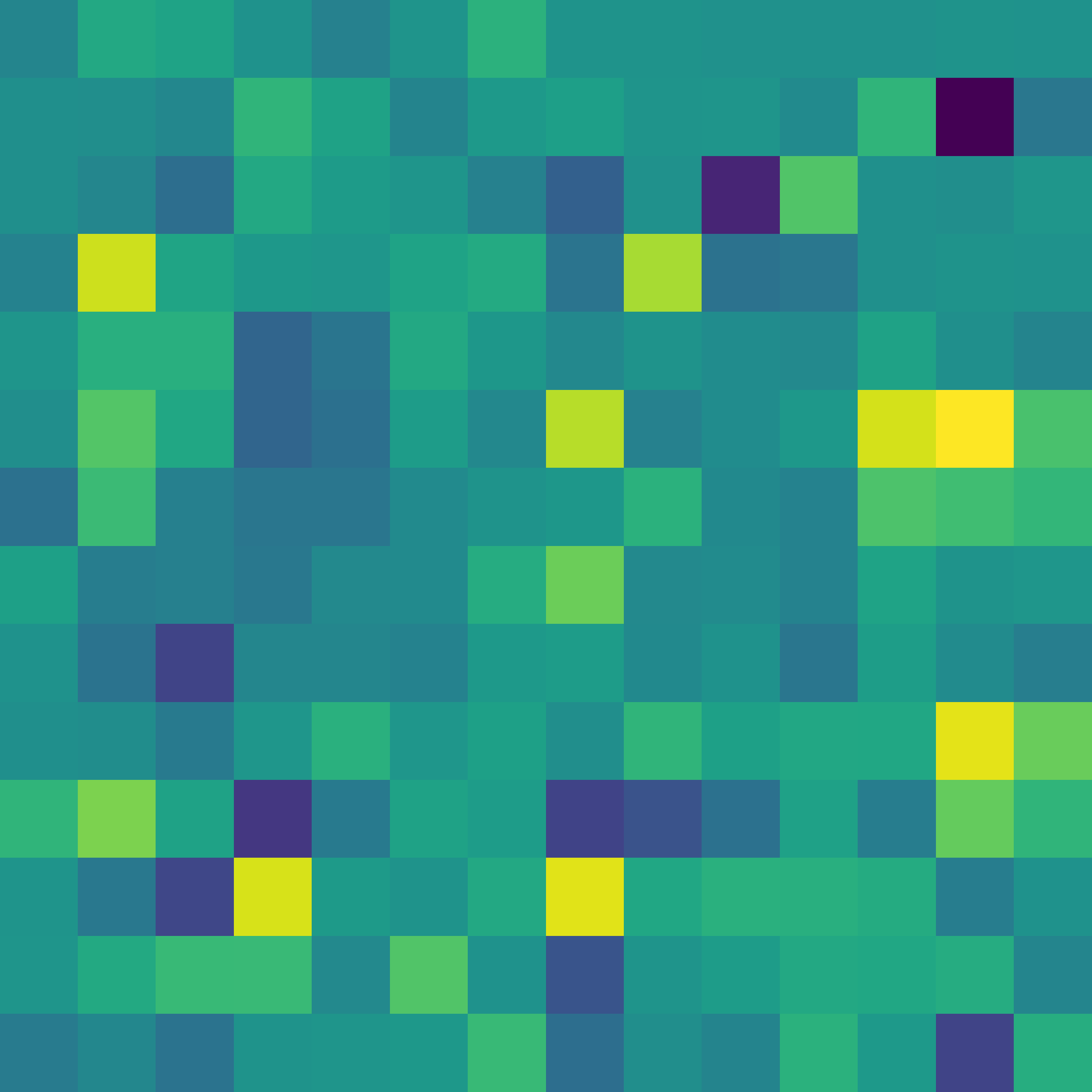} &
    \onebyfour{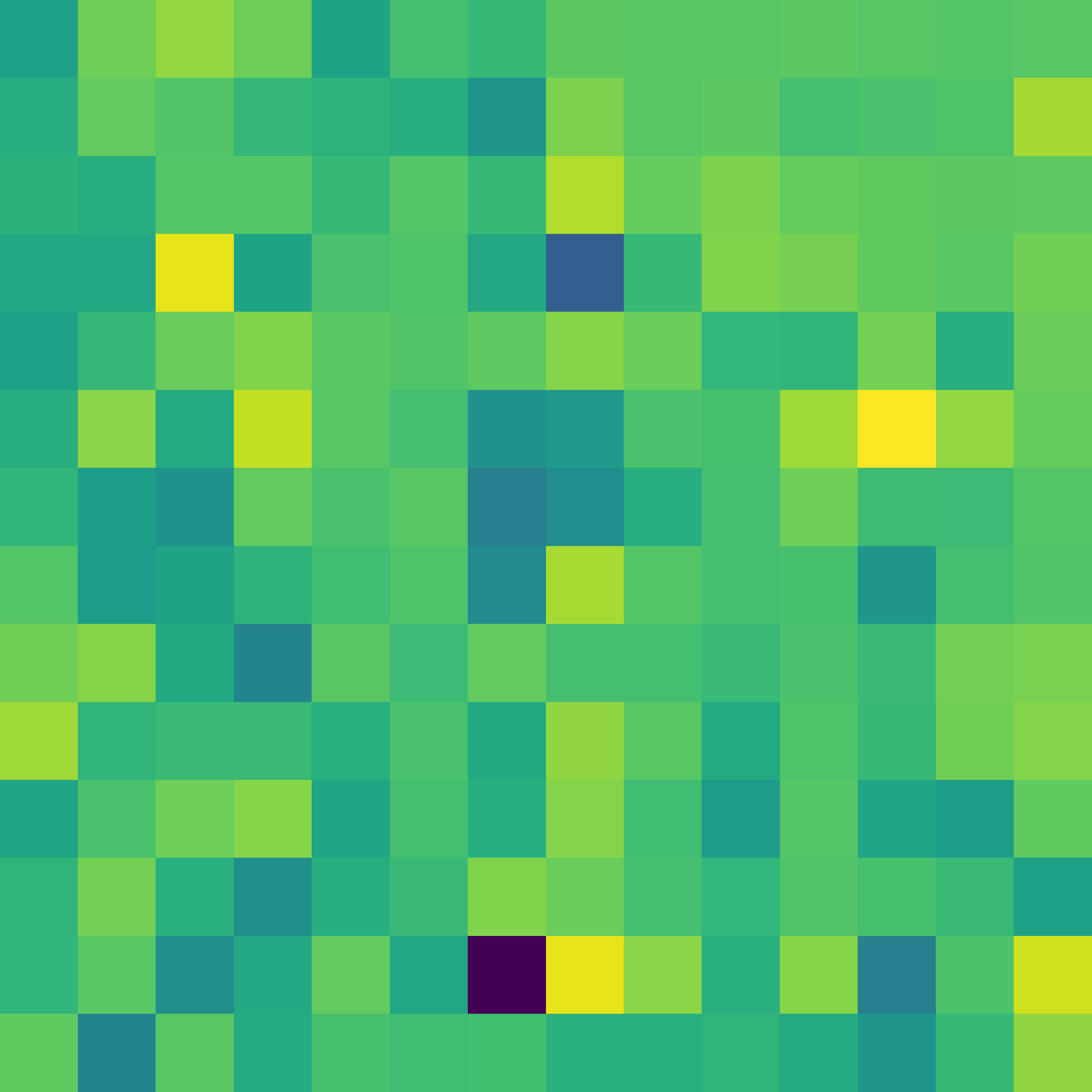}{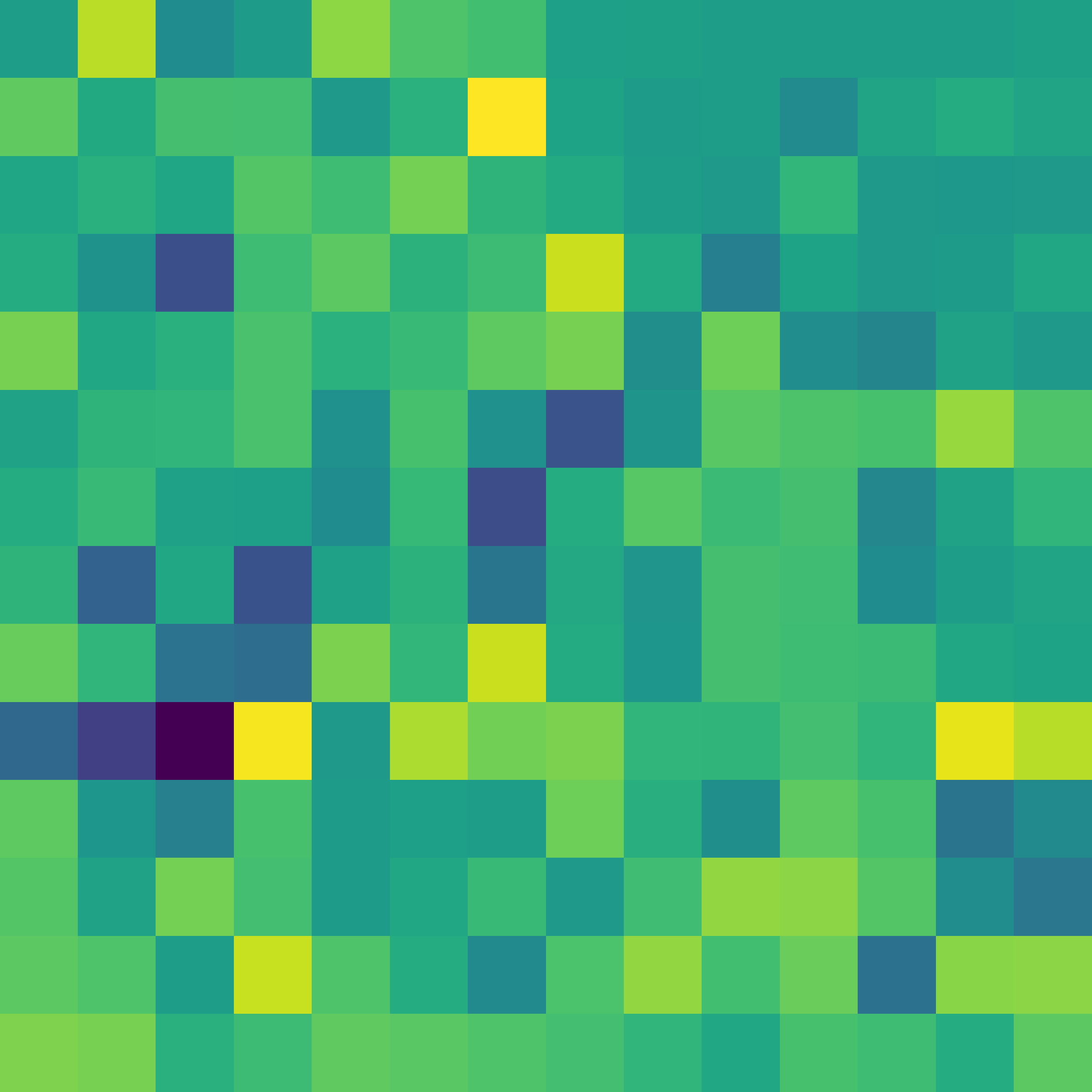}{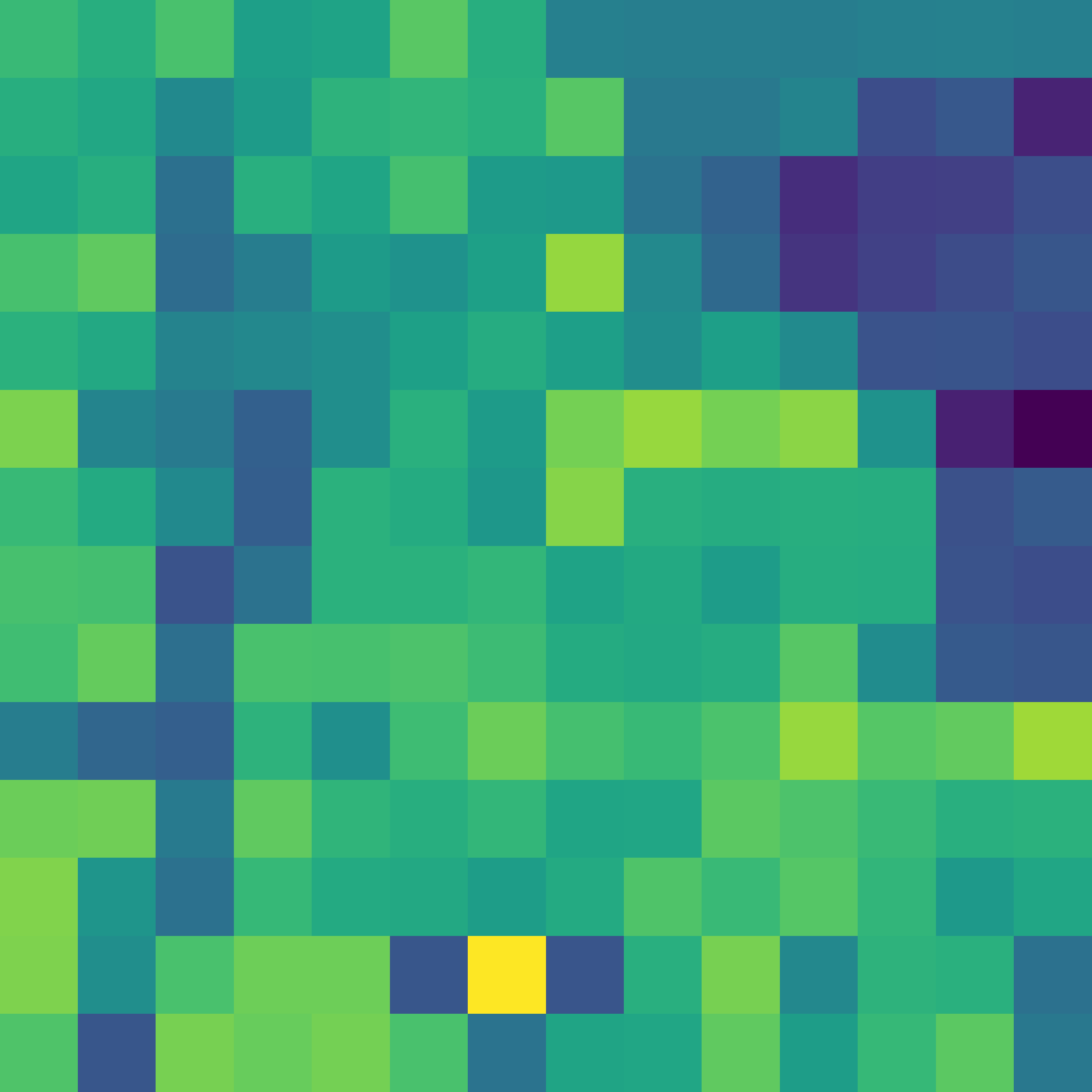}{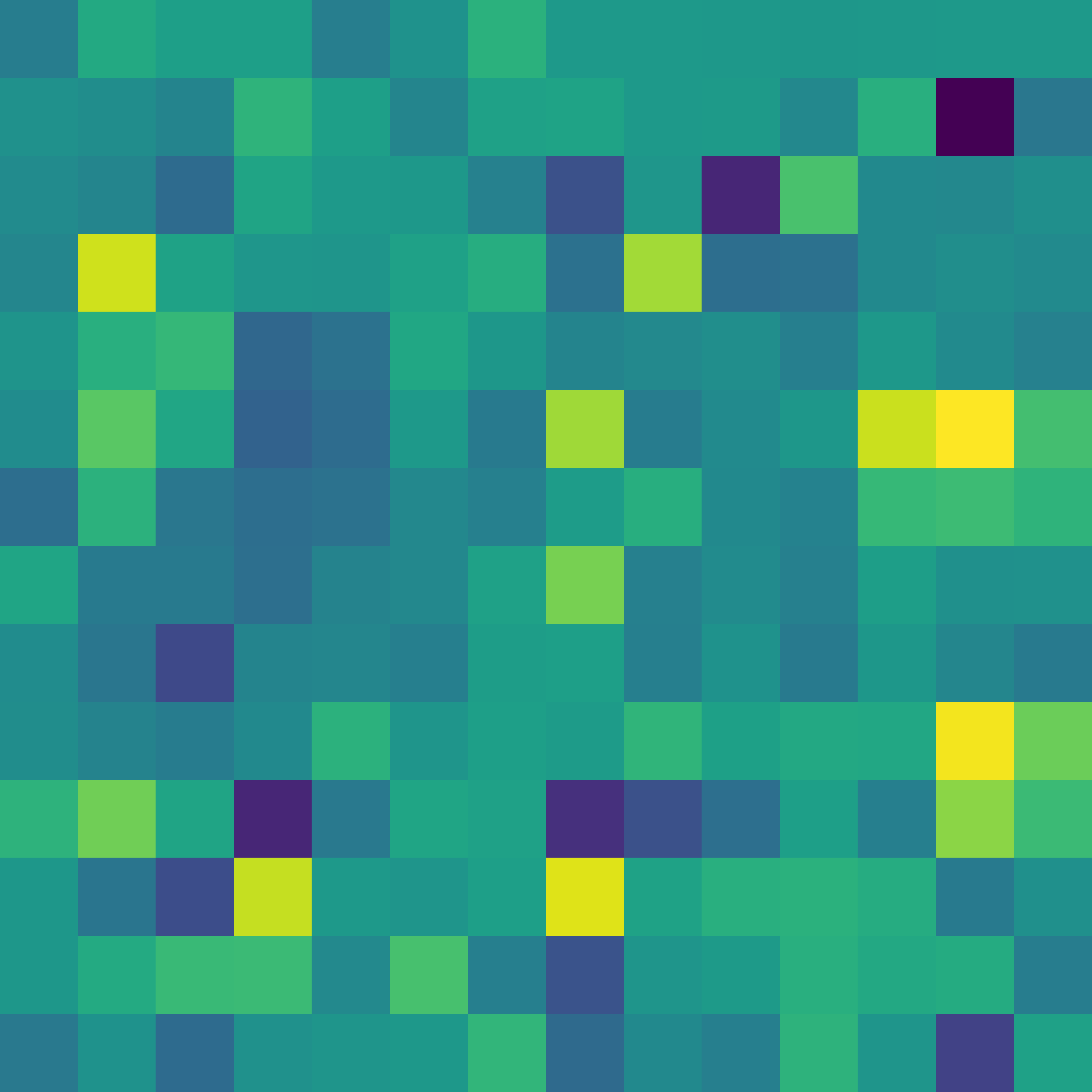} \\[0.45cm]
    Eq. Error & \onebyfour{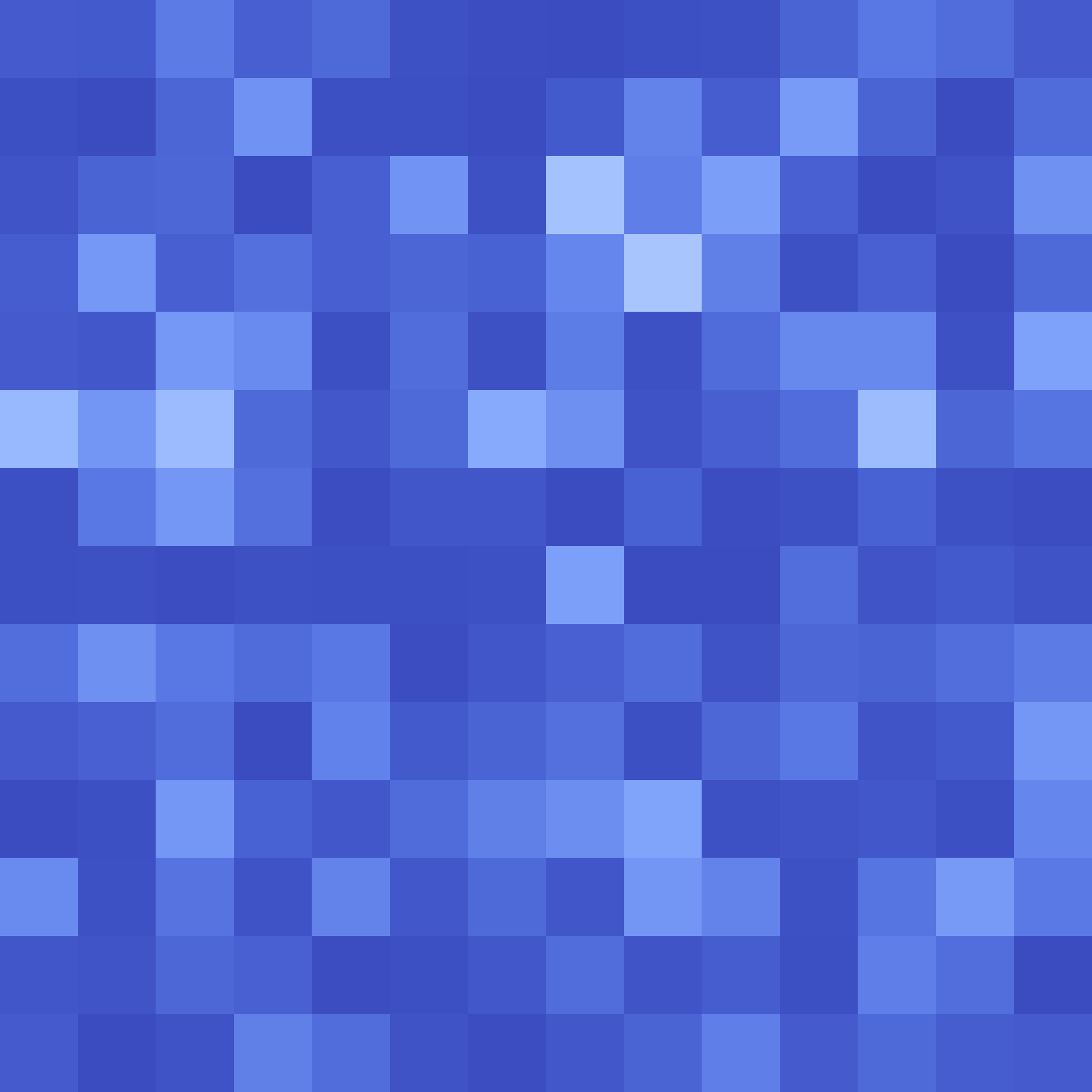}{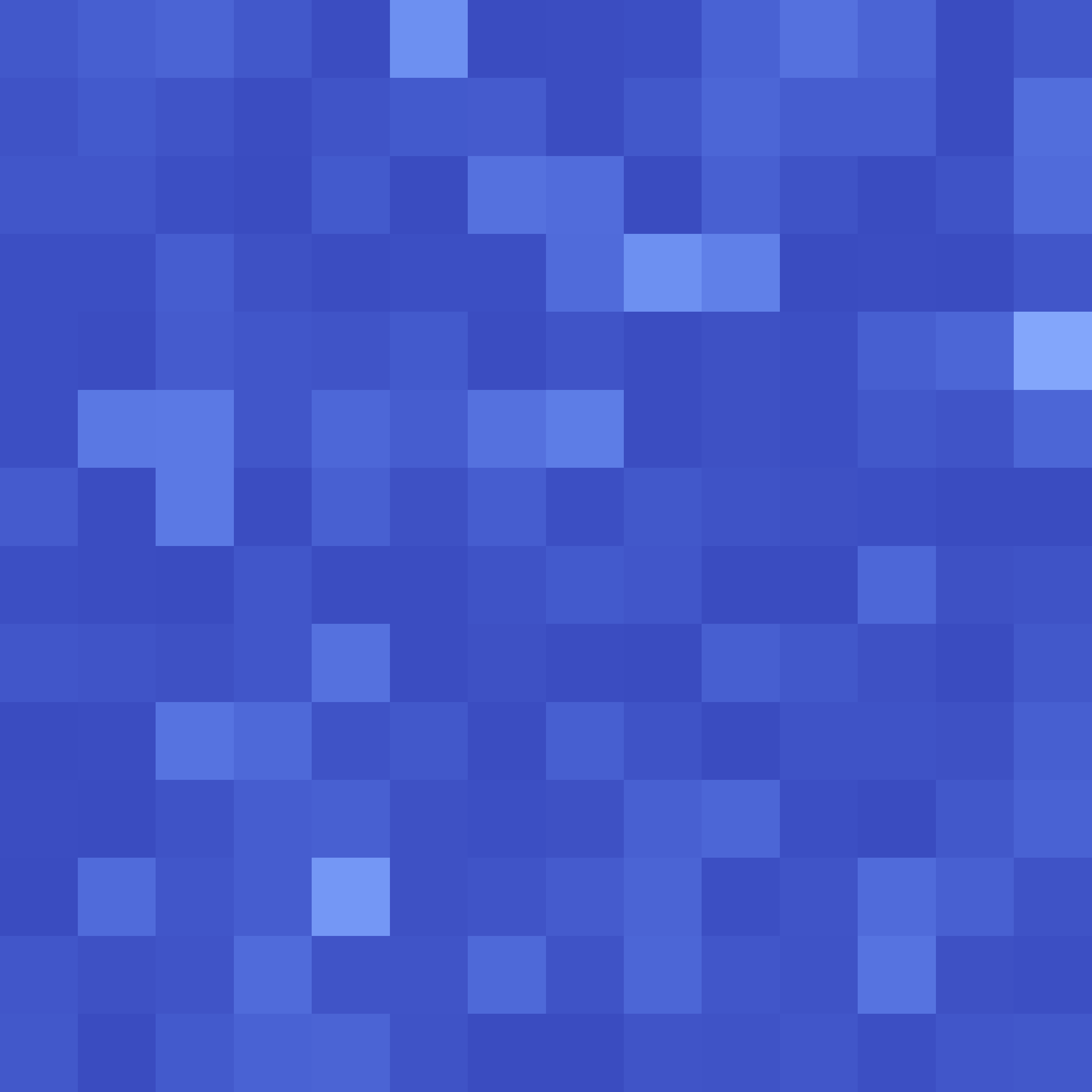}{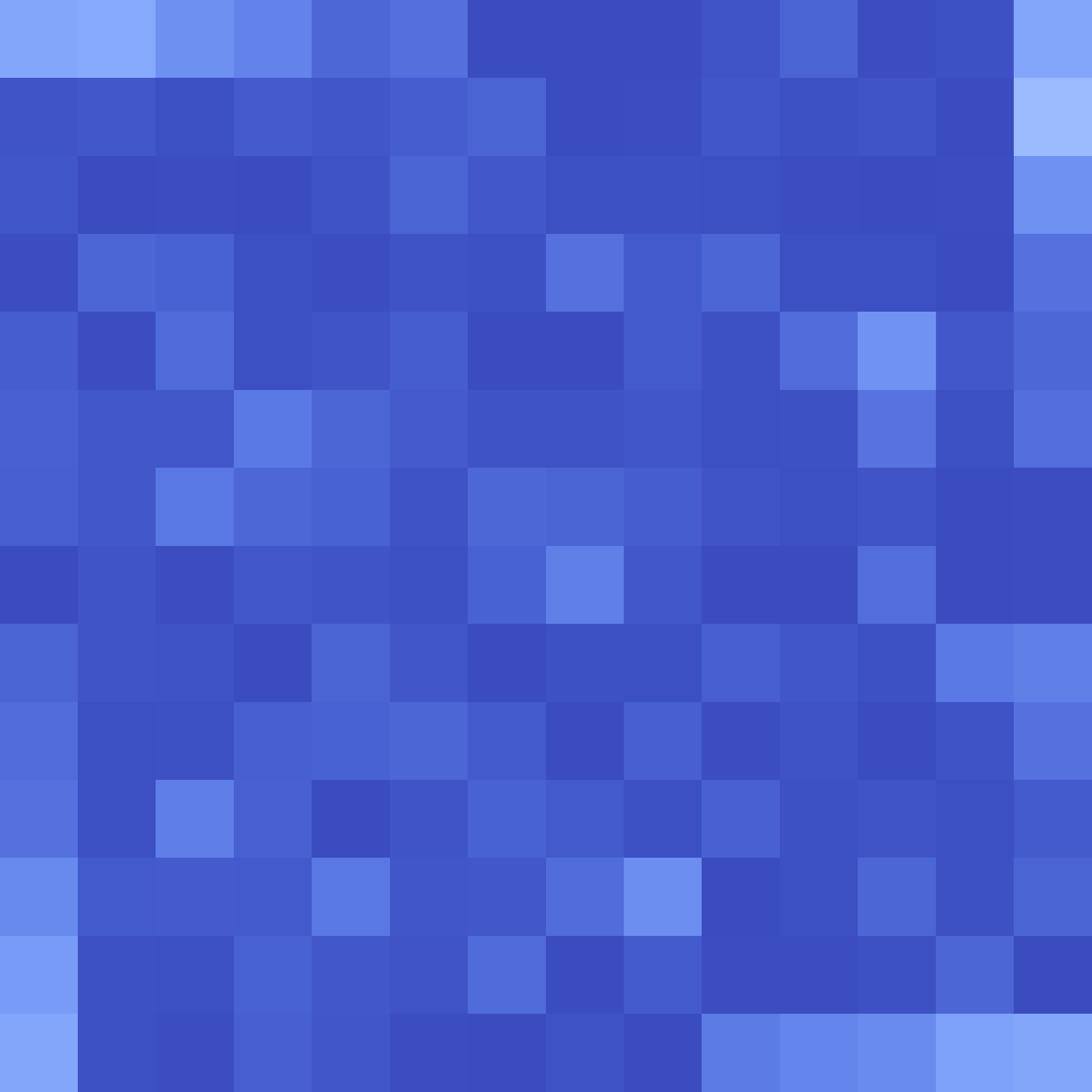}{figs/nrot36/soft_0.01_nrot_36/2736665098_0b0870f51f_z/ch7_error_10.png} &
    \onebyfour{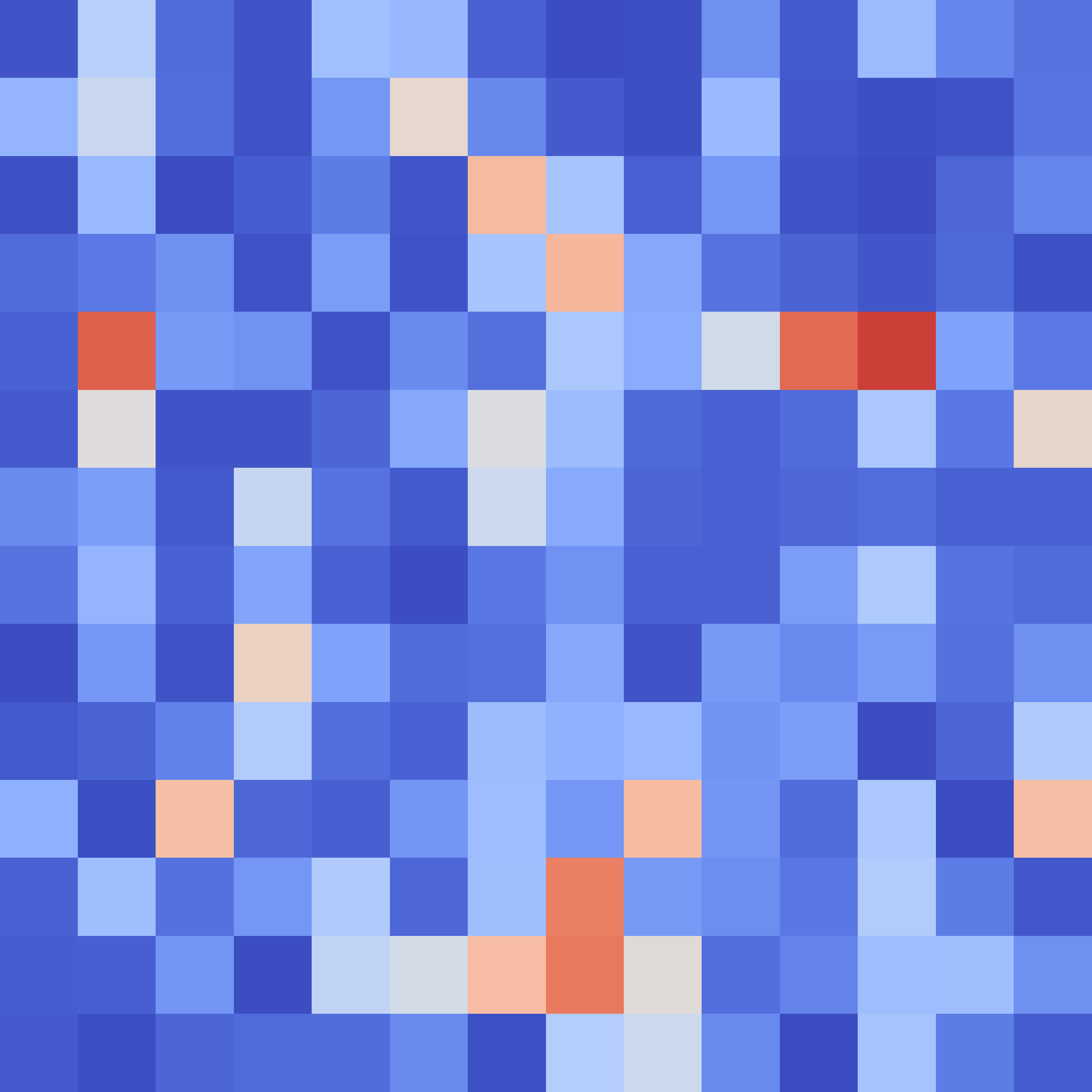}{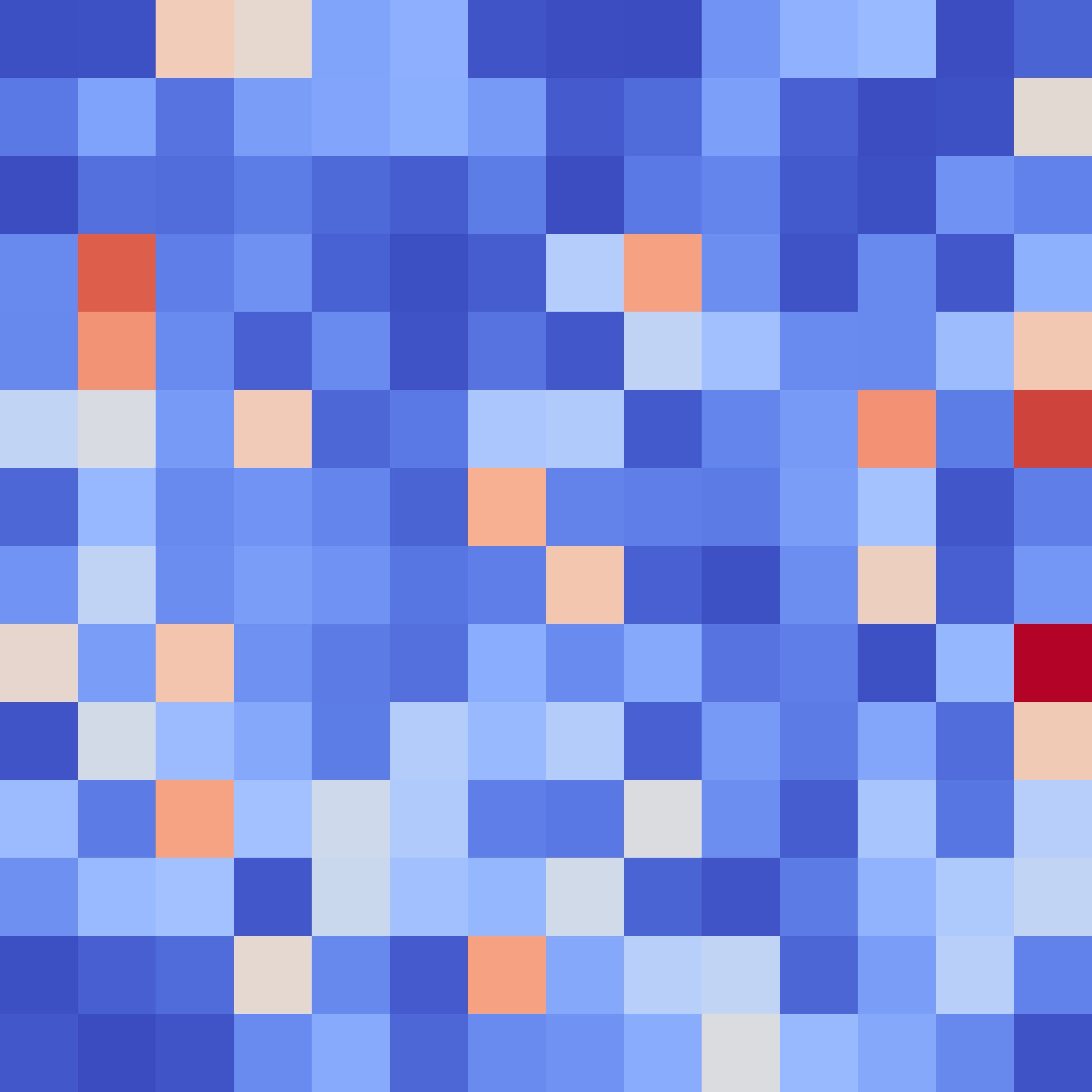}{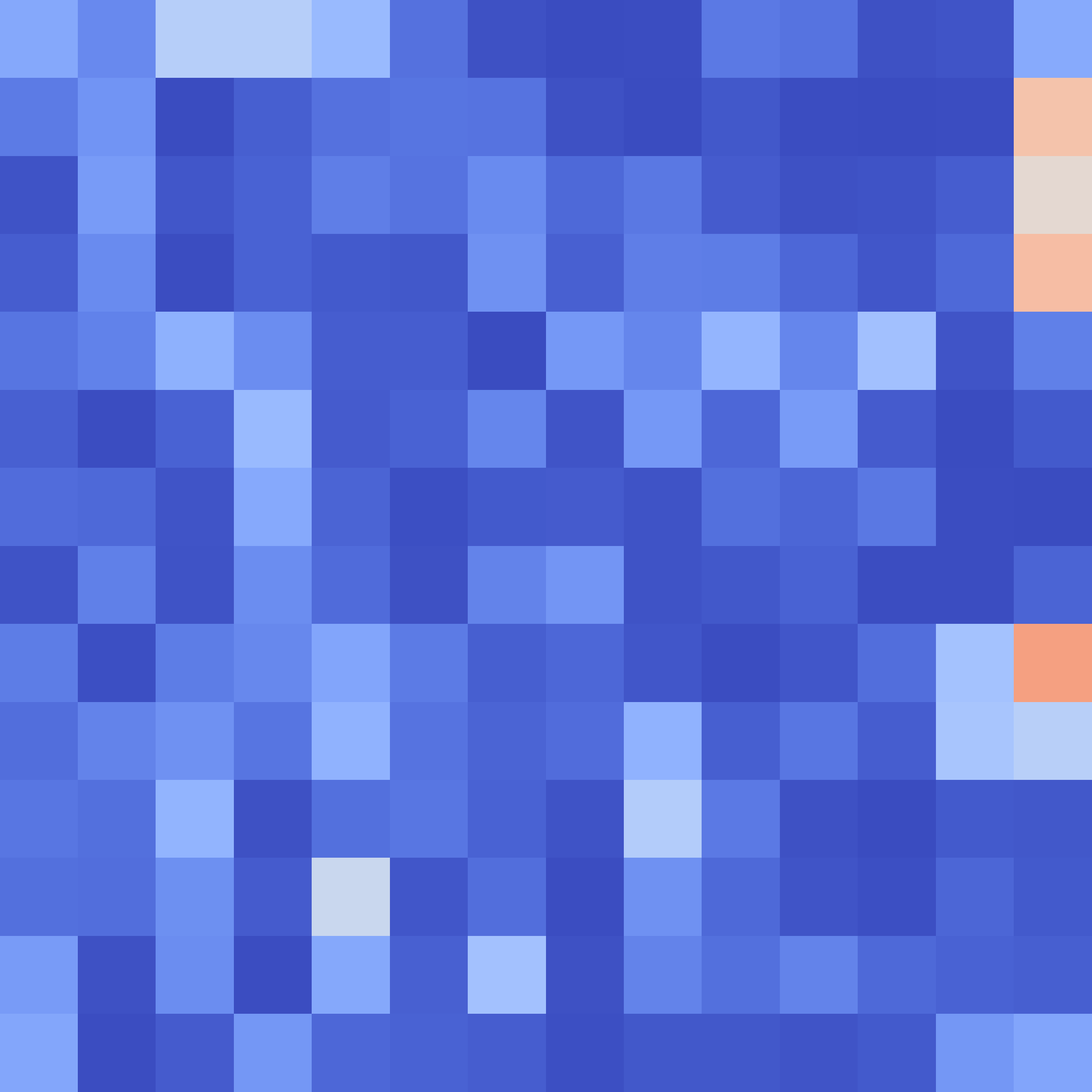}{figs/nrot36/soft_0.25_nrot_36/2736665098_0b0870f51f_z/ch7_error_10.png} &
    \onebyfour{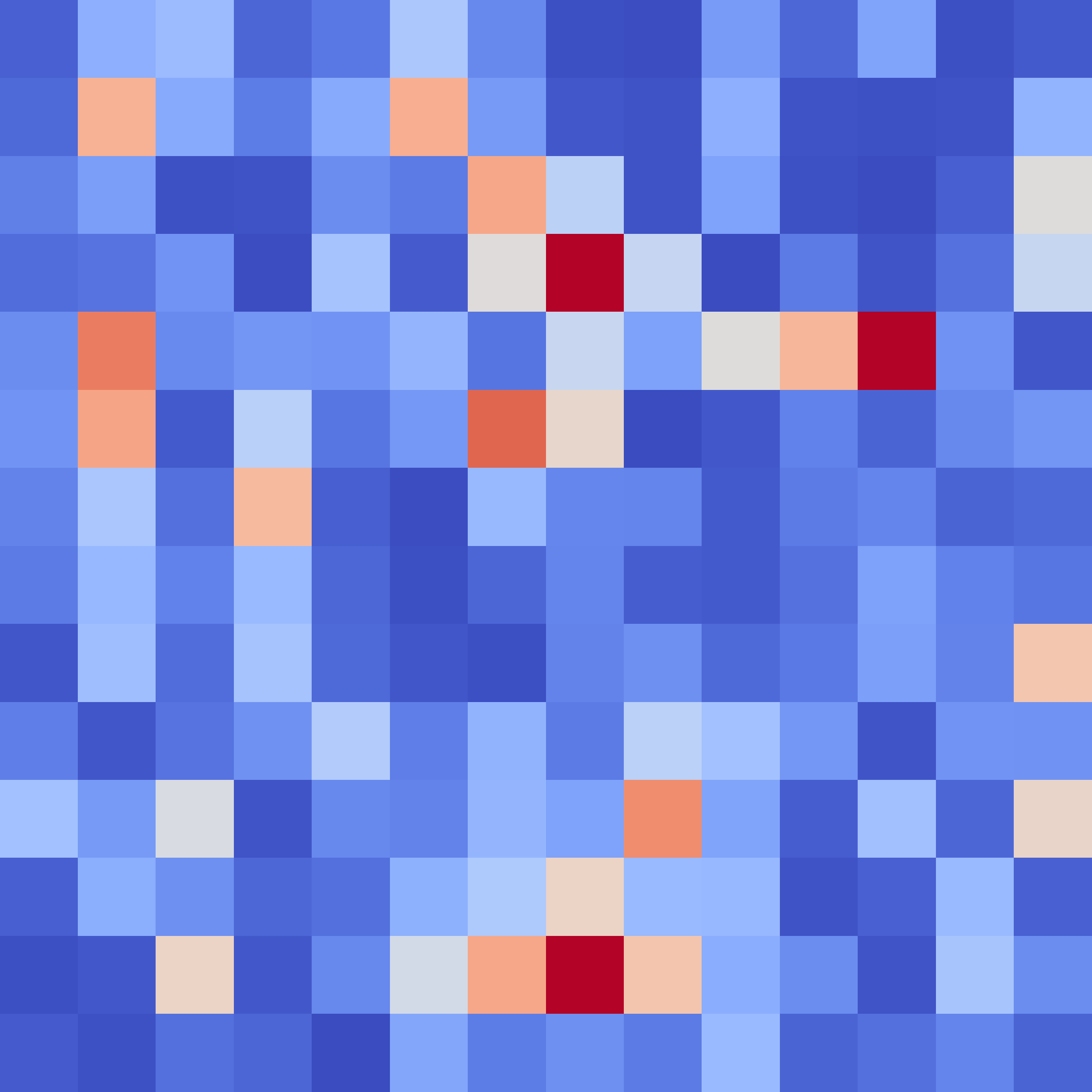}{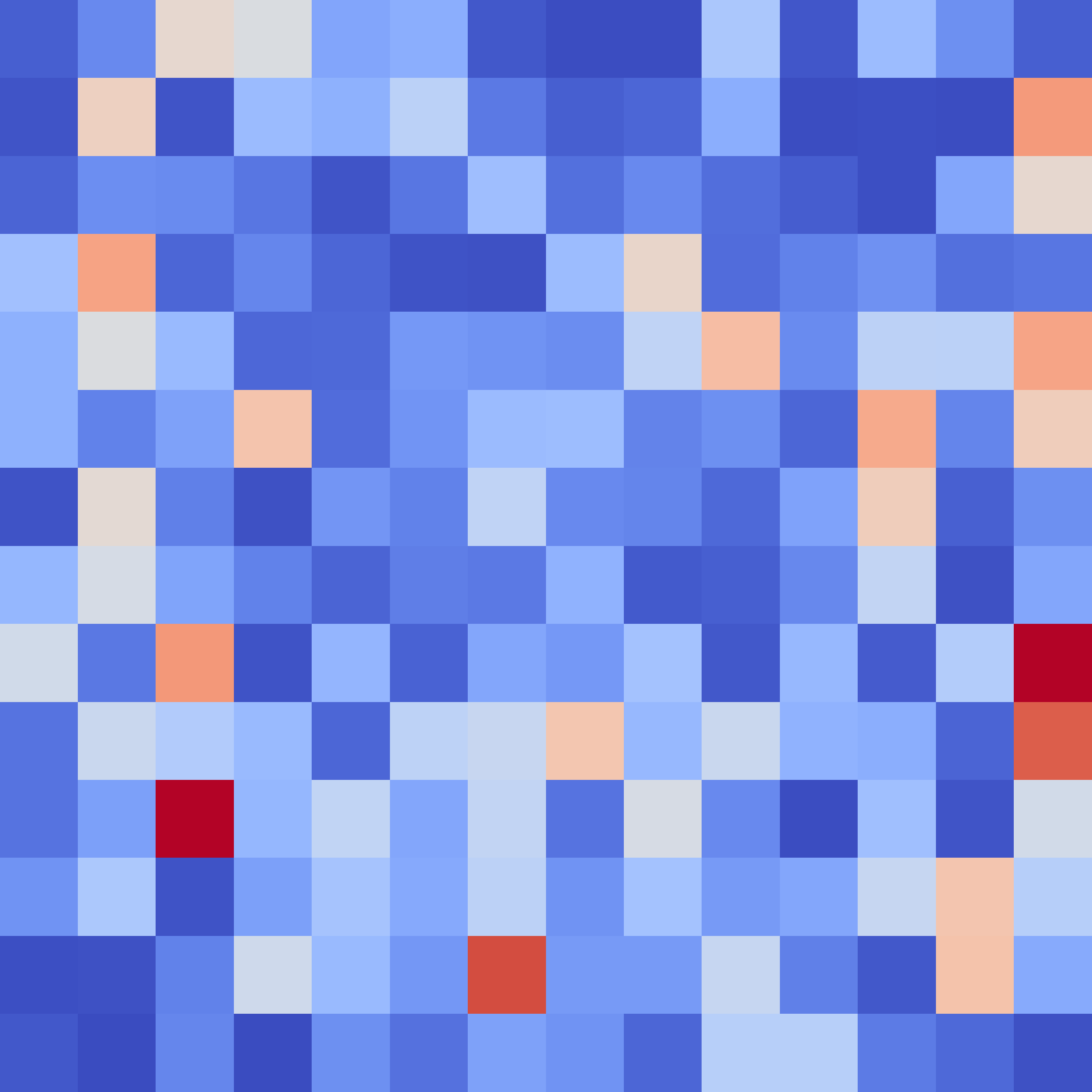}{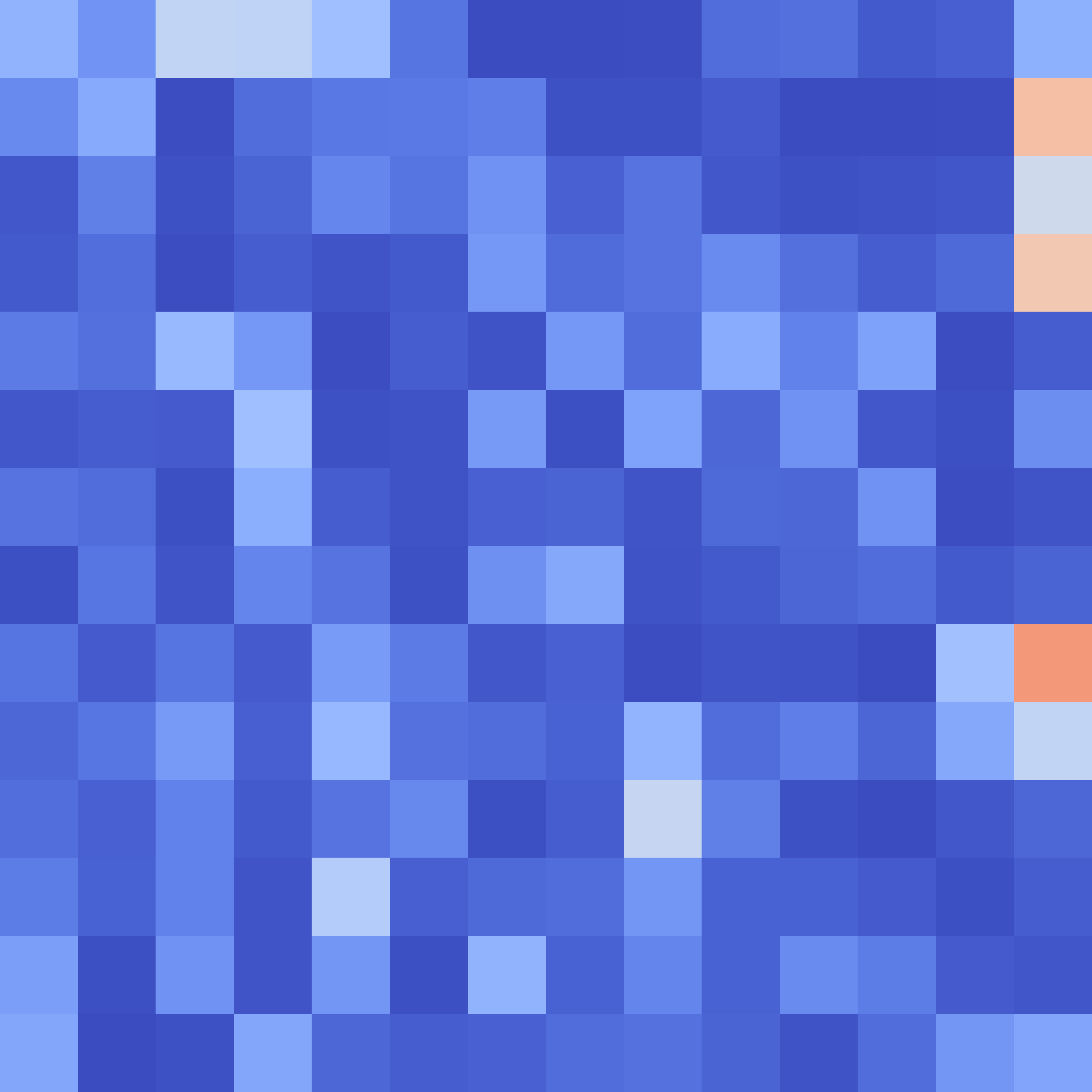}{figs/nrot36/soft_0.35_nrot_36/2736665098_0b0870f51f_z/ch7_error_10.png} &
    \onebyfour{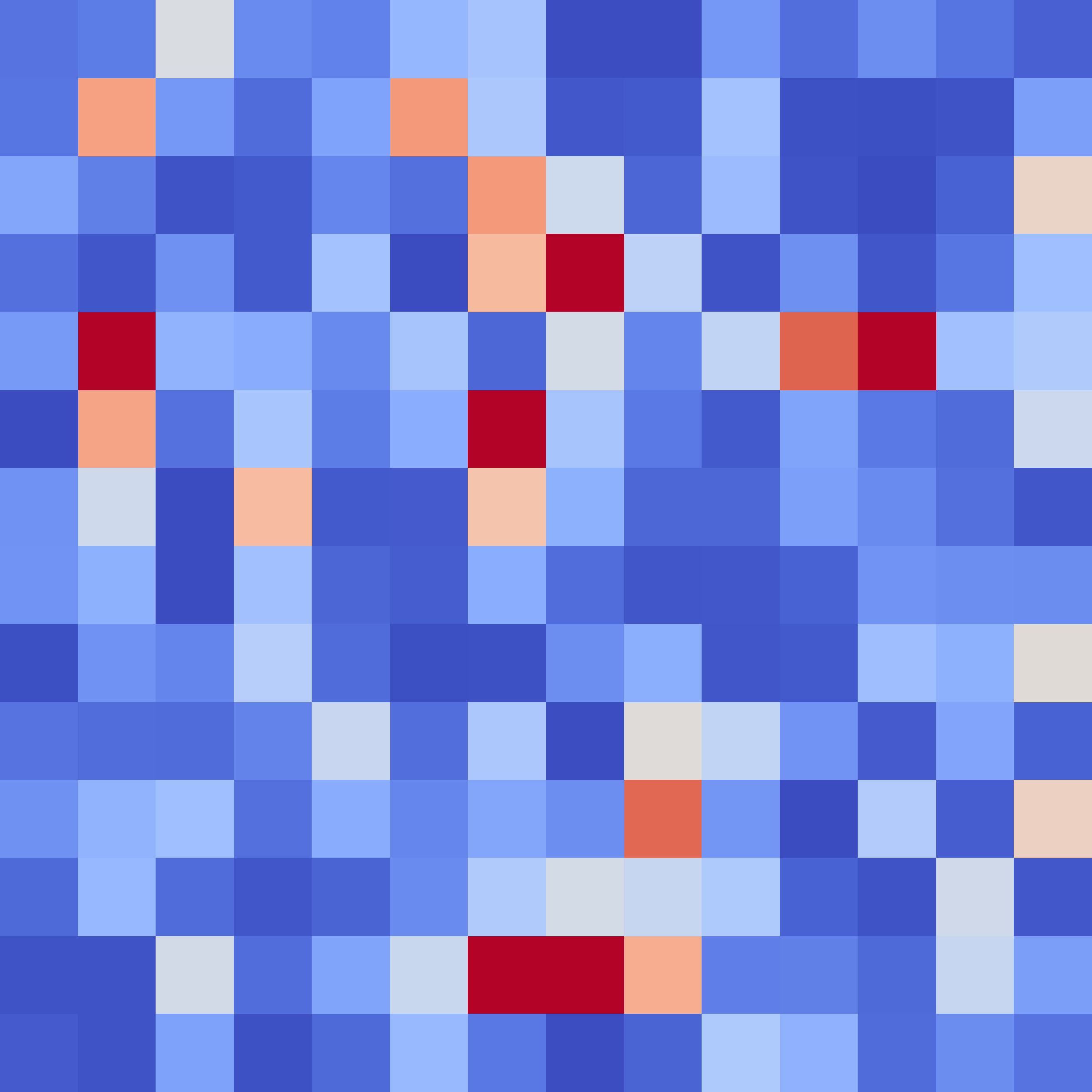}{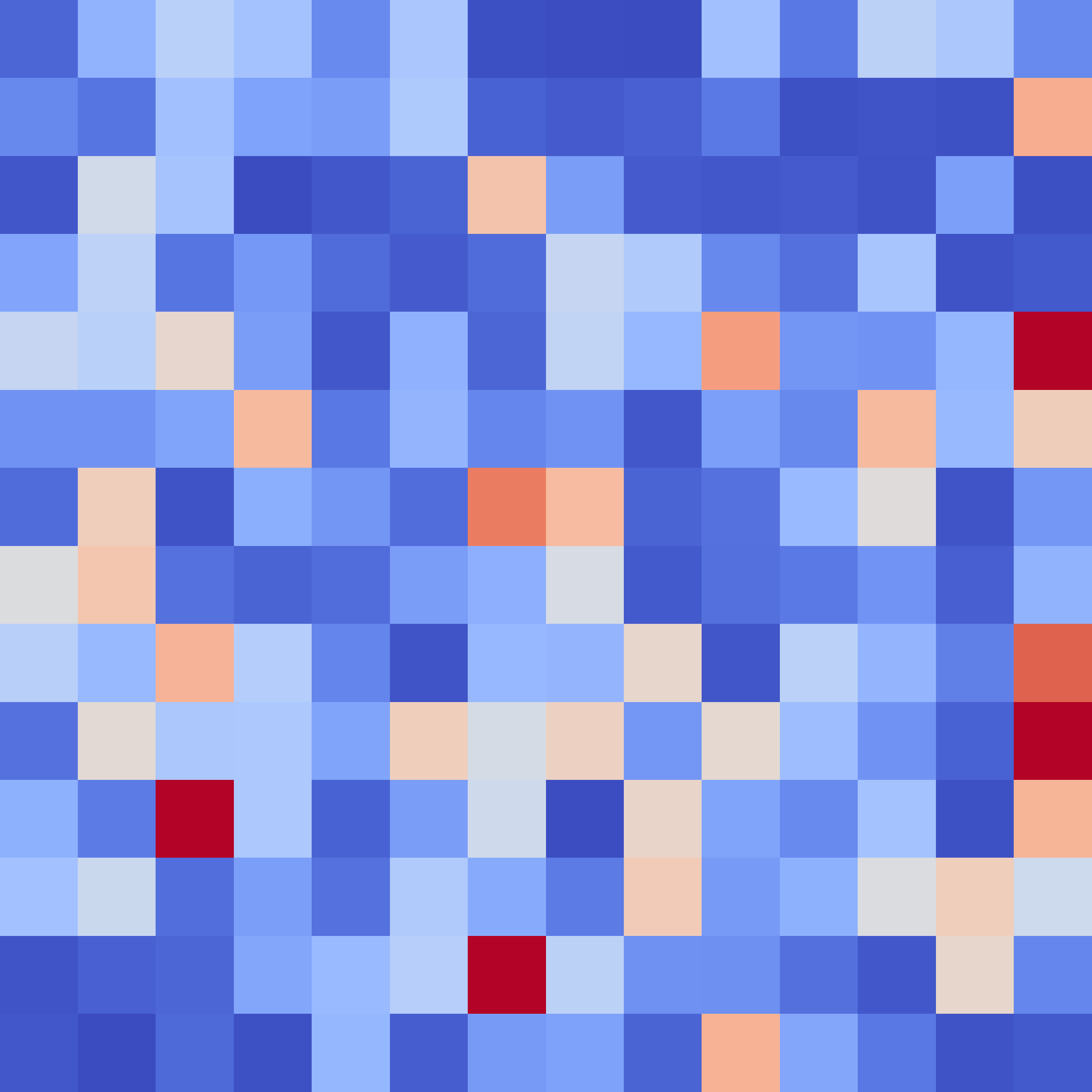}{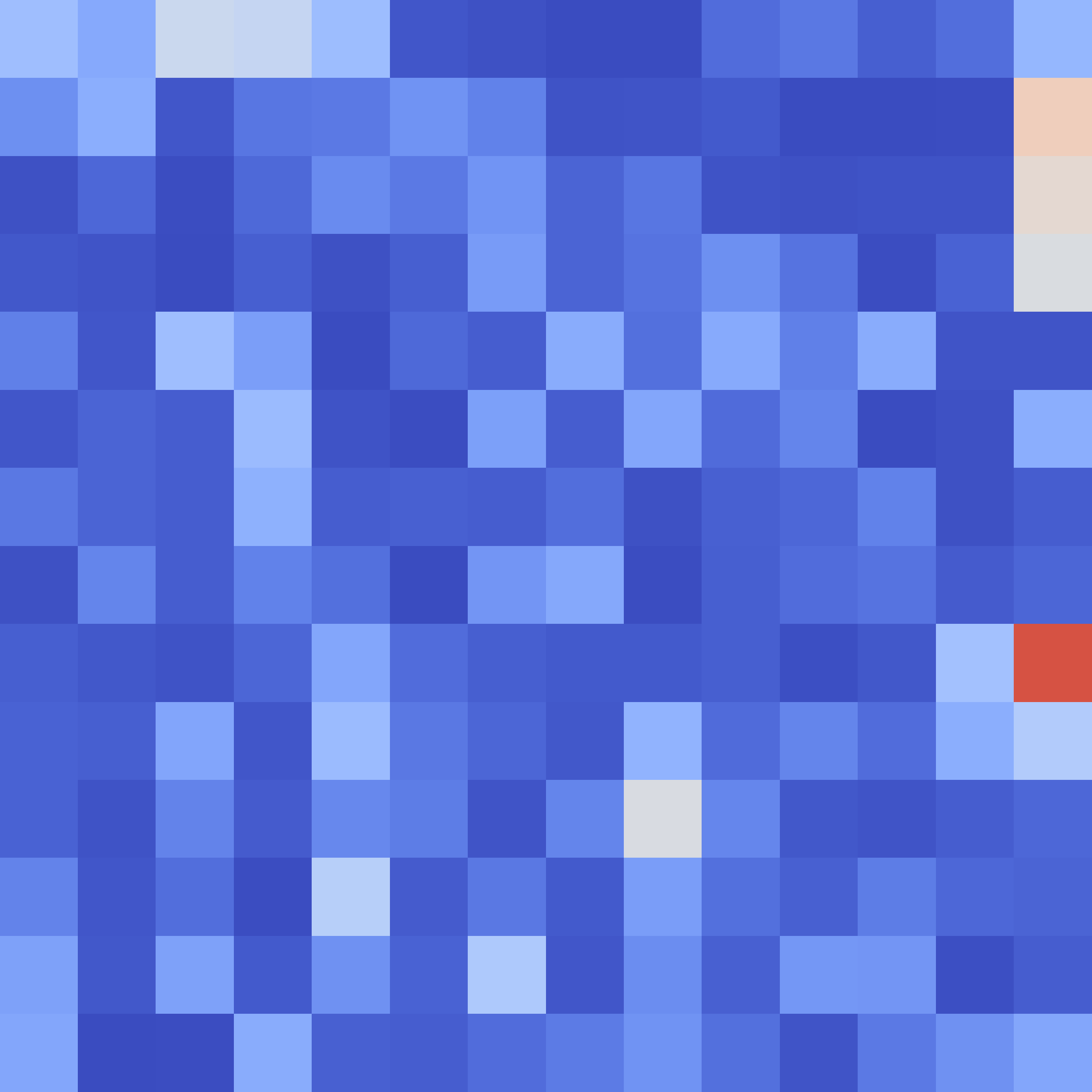}{figs/nrot36/soft_0.5_nrot_36/2736665098_0b0870f51f_z/ch7_error_10.png} &
    \onebyfour{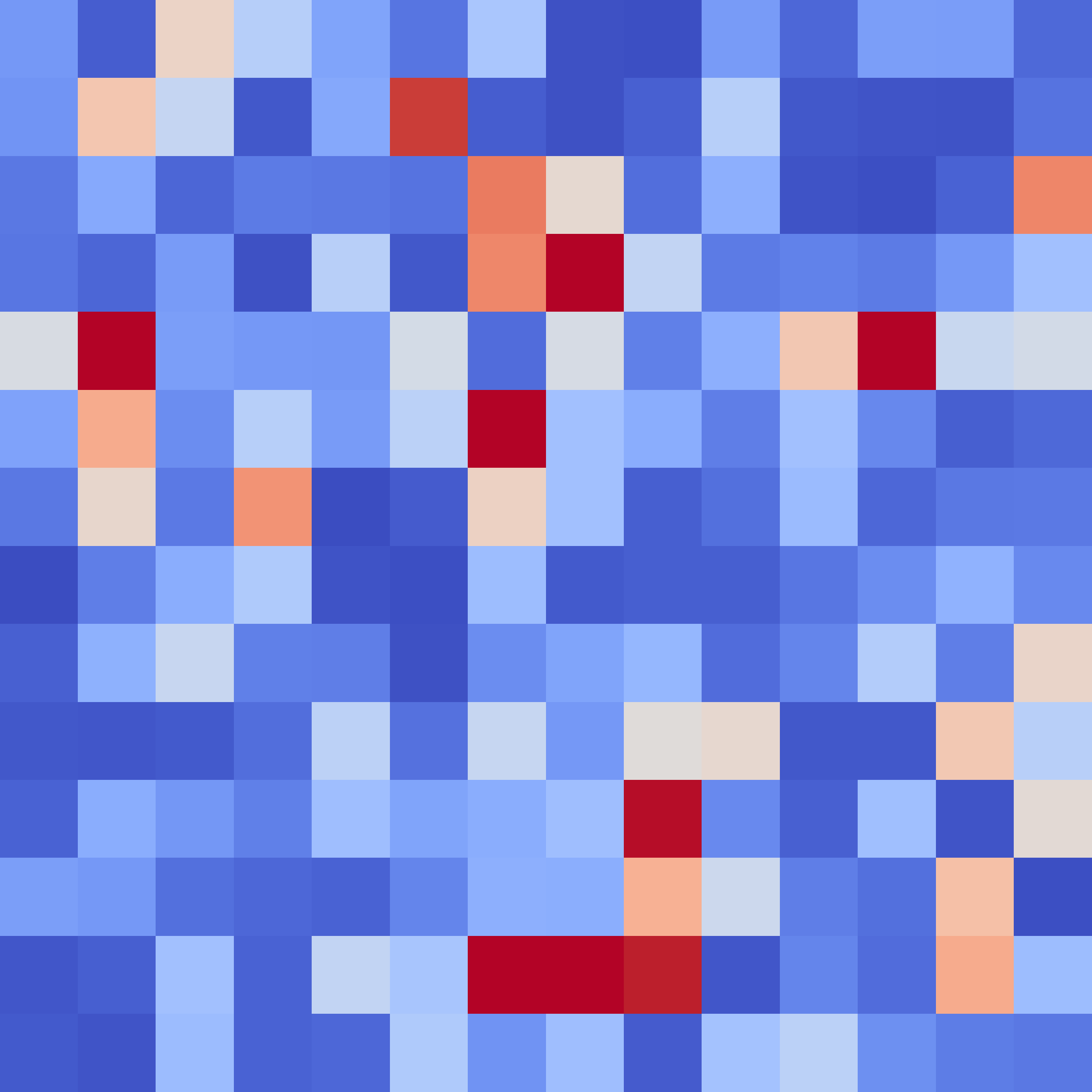}{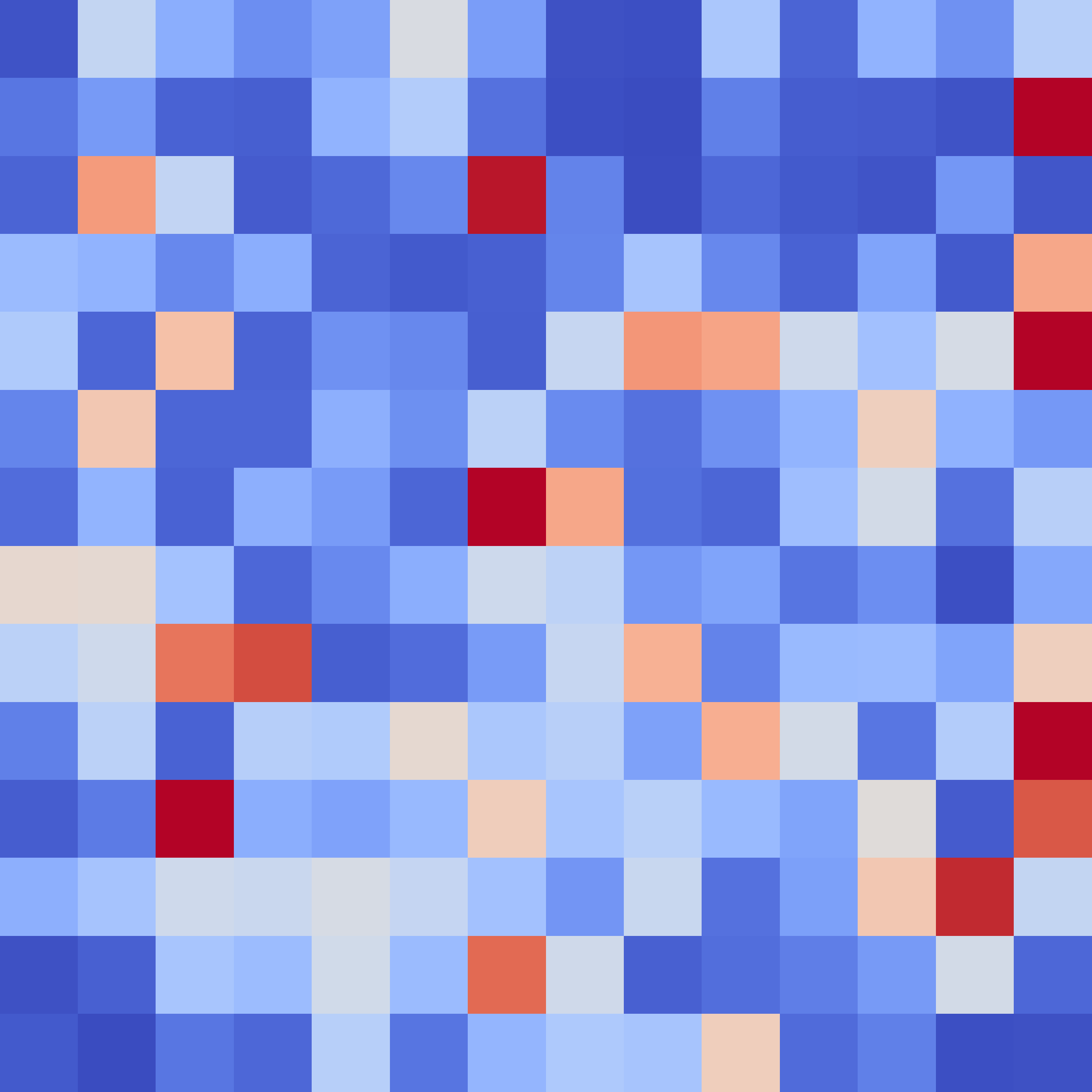}{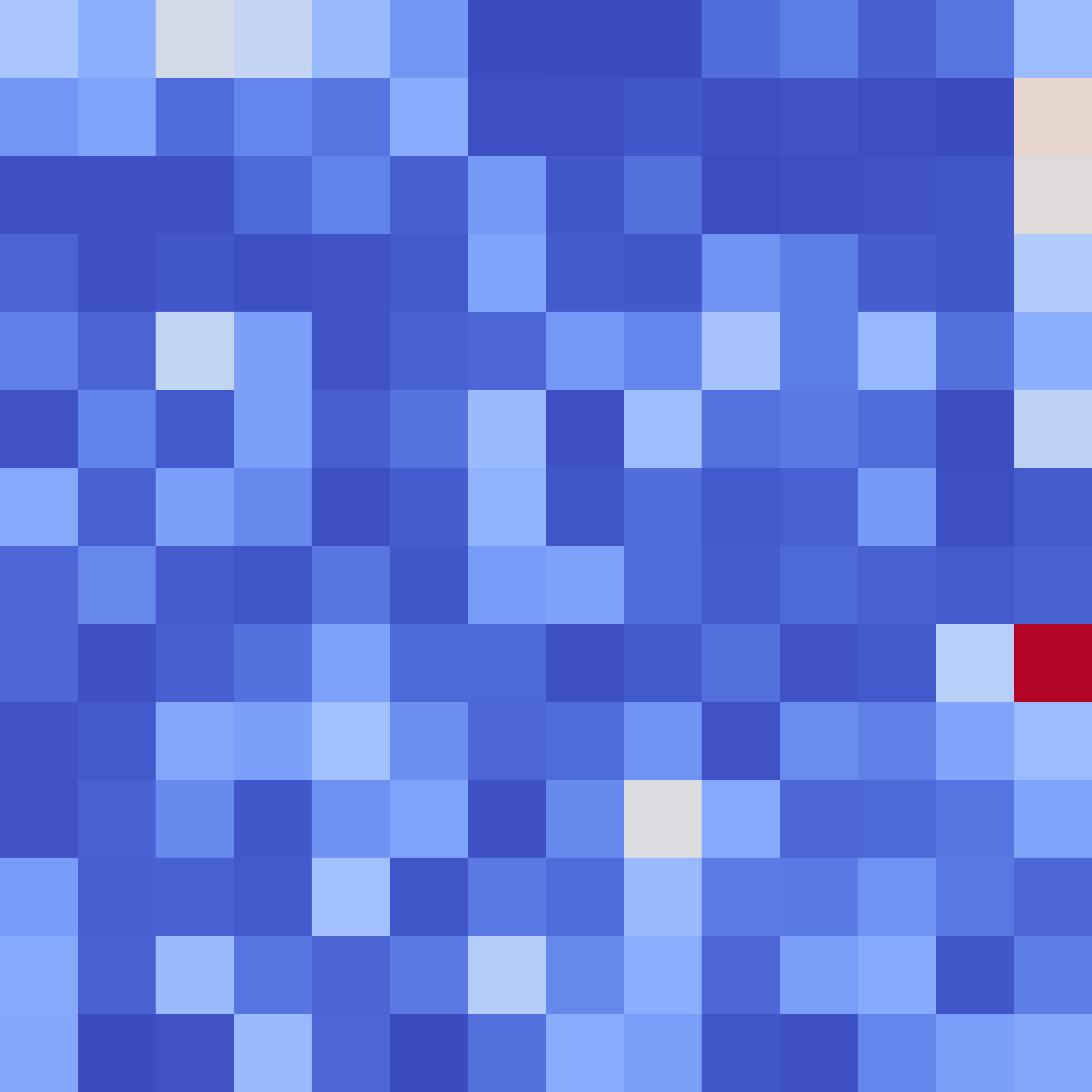}{figs/nrot36/unmod/2736665098_0b0870f51f_z/ch7_error_10.png}
\end{tabular}
}%
\vspace{2pt}

\end{minipage}
\hfill
\begin{minipage}{0.079\textwidth}
\hspace{1.5cm}
\end{minipage}
\begin{minipage}{0.93\textwidth}
    \vspace{2pt}
    \noindent\resizebox{\textwidth}{!}{%
    \begin{tikzpicture}[x=1\textwidth/13]
        \shade[left color={rgb,255:red,59;green,76;blue,192},
               right color=white]
               (0.61,-0.10) rectangle (7.00,-0.00);
        \shade[left color=white,
               right color={rgb,255:red,180;green,4;blue,38}]
               (7.20,-0.10) rectangle (13.3,-0.00);
        \node[anchor=south] at (1.12,-0.6) {\small Low error};
        \node[anchor=south] at (12.95,-0.645) {\small High error};
    \end{tikzpicture}
    }%
\end{minipage}
\vspace{-0.38cm}
    \captionof{figure}{Visualization of the ViT~\cite{dosovitskiy2020image} weights with our soft equivariance layer (\wrt $10^\circ$ rotation) under different softness levels, along with the corresponding extracted features and the equivariance errors. Our tunable design allows the layers' weights to transition smoothly from perfectly equivariant to fully non-equivariant behavior in a controlled manner.
    }
    \label{fig:teaser3}
\begin{textblock*}{5cm}(1.8cm,17.1cm) %
  {\rotatebox{90}{\small Features}}
\end{textblock*}

%% file: supp/math_prelim.tex
\section{Mathematical Background}
\label{supp:math}
In this section, we review the mathematical preliminaries needed to construct the soft-invariant and soft-equivariant projection operator.
\subsection{Groups}
\label{supp:groups}
\myparagraph{Group.} A group is a set $G$ together with a binary operation $\cdot : G \times G \to G$ that combines any two elements $a, b \in G$ to form another element denoted $a \cdot b$. And it satisfies the following four requirements:
\begin{itemize}
    \item {\it Closure:} For all $a, b \in G$, the result of the operation $a \cdot b$ is also in $G$.
    \item {\it  Associativity:} For all $a, b, c \in G$, $(a \cdot b) \cdot c = a \cdot (b \cdot c)$.
    \item {\it  Identity element:} There exists an element $e \in G$ such that for every element $a \in G$, the equation $e \cdot a = a \cdot e = a$ holds.
    \item {\it Inverse element:} For each $a \in G$, there exists an element $b \in G$ such that $a \cdot b = b \cdot a = e$, where $e$ is the identity element.
\end{itemize}
The binary operation is often referred to as the group product.

\myparagraph{Generating Set.} A subset $\sS \subseteq G$ is called a generating set of the group $G$ if every element of $G$ can be expressed as product of the elements of the $\sS$. So for any element $g \in G$, there exists a sequence of elements $s_1, s_2, \ldots, s_m$ from $\sS$ such that $g = s_1^{n_1} \cdot s_2^{n_2} \cdot \ldots \cdot s_m^{n_m}$.

\myparagraph{Word Metric.} Given a generating set $\sS$ of $G$, any element $g \in G$ 
can be written as a product $g = s_1^{n_1} \cdot s_2^{n_2} \cdots s_m^{n_m}$ with 
$s_i \in \sS$. The \emph{word length} of $g$ with respect to $\sS$ is defined as the 
minimum number of generator applications required to express $g$:
\begin{equation}
    \ell_{\sS}(g) \triangleq \min \left\{ \sum_{i=1}^m n_i \;\middle|\; 
    g = s_1^{n_1} \cdots s_m^{n_m},\; s_i \in \sS \right\}.
\end{equation}
The \emph{word metric} $d_{\sS}: G \times G \to \mathbb{Z}_{\geq 0}$ is then defined as:
\begin{equation}
    d_{\sS}(g, g') \triangleq \ell_{\sS}(g^{-1} g'),
\end{equation}
which measures the minimum number of generator steps needed to reach $g'$ from $g$.
When $\sS$ is clear from context, we write $d(\cdot, \cdot)$ and $\ell(\cdot)$ to denote word metric and word length, respectively.

{\bf Note:} In general, the decomposition $g = s_1^{n_1} \cdot s_2^{n_2} \cdots s_m^{n_m}$ is not unique. For this, a fixed minimal-length word representation for each element $g\in G$ is chosen, which is called the `canonical' word representation.

\myparagraph{Group Action.} The group action or the representation of a group $G$ on space $\gX$ is a map $\rho : G \rightarrow GL(\gX)$, where $GL(\gX)$ is the set of invertible linear transformations on $\gX$, such that for any $g, h \in G$ and any $\vx \in \gX$, we have $\rho(e)\vx = \vx$ and $\rho(g h)\vx = \rho(g)(\rho(h)\vx)$, where $e$ is the identity element of the group $G$.

\myparagraph{Discrete Rotation Group (Cyclic group $C_n$).}
Let $r$ denote the planar rotation by angle $\tfrac{2\pi}{n}$. The set
\[
C_n \triangleq \{\, r^k \mid k=0,1,\dots,n-1 \,\}
\]
With composition as the group product, it is a discrete (finite) group.

The generating set is $\{r\}$ since every element is $r^k$.
The $2D$ linear representation $\rho$ maps $r^k$ to $2D$ rotation matrix $ R(\tfrac{2\pi k}{n})$ where
\[
R(\theta) \triangleq 
\begin{pmatrix}
\cos\theta & -\sin\theta\\
\sin\theta & \ \cos\theta
\end{pmatrix}.
\]
The action on $\vx\in\R^2$ is $\rho(r^k)\vx = R\!\big(\tfrac{2\pi k}{n}\big)\vx$.

\myparagraph{Lie Group.}
A Lie group $G$ is a smooth manifold equipped with a group structure such that the group operations—multiplication $(g,h)\mapsto gh$ and inversion $g\mapsto g^{-1}$ are smooth maps.

\myparagraph{Lie Algebra.}
The Lie algebra $\mathfrak{g}$ associated with a Lie group $G$ is the tangent space at the identity element $e$, denoted as $\mathfrak{g} = T_e G$. It is a vector space equipped with a bilinear operation $[\cdot, \cdot]: \mathfrak{g} \times \mathfrak{g} \to \mathfrak{g}$ called the Lie bracket. For matrix Lie groups (where $G \subseteq \mathrm{GL}(n, \mathbb{R})$), the bracket is defined as the matrix commutator:
\begin{equation}
    [A, B] = AB - BA, \quad \forall A, B \in \mathfrak{g}.
\end{equation}
\paragraph{Lie Algebra Basis.}
A basis for a Lie algebra $\mathfrak{g}$ of dimension $d$ is a set of linearly independent elements $\{A_1, \dots, A_d\}$ such that every element $A \in \mathfrak{g}$ can be uniquely represented as a linear combination $A = \sum_{i=1}^d c_i A_i$, where $c_i$ are scalar coefficients.
\myparagraph{Exponential Map.}
The exponential map $\exp: \mathfrak{g} \to G$ creates a mapping from the tangent space to the manifold. For matrix Lie groups, it is given by the matrix exponential:
\begin{equation}
    \exp(A) = \sum_{k=0}^{\infty} \frac{A^k}{k!}.
\end{equation}
The map satisfies $\exp(0) = I$ and $\frac{d}{dt}\exp(tA)|_{t=0} = A$. Crucially, for commutative algebras, $\exp(A+B) = \exp(A)\exp(B)$, though this does not hold generally for non-abelian groups like $\mathrm{SO}(3)$.

\myparagraph{Group and Algebra Representations.}
A group representation is a homomorphism $\rho: G \to \mathrm{GL}(V)$ acting on a vector space $V$. This induces a Lie algebra representation $d\rho: \mathfrak{g} \to \mathfrak{gl}(V)$, derived by differentiating the group action at the identity:
\begin{equation}
    d\rho(A) = \left. \frac{d}{dt} \rho(\exp(tA)) \right|_{t=0}.
\end{equation}
This map preserves the bracket structure: $d\rho([A, B]) = [d\rho(A), d\rho(B)]$.

\myparagraph{Case Studies: Rotations in $\mathbb{R}^2$ and $\mathbb{R}^3$.}

We explicitly detail the geometry of 2D and 3D rotations used in the main paper.

\myparagraph{The $\mathrm{SO}(2)$ Group.}
The $\mathrm{SO}(2)$ or the special orthogonal group in 2D consists of planar rotation matrices parameterized by an angle $\theta \in [0, 2\pi)$:
\begin{equation}
    R(\theta) = \begin{pmatrix} \cos\theta & -\sin\theta \\ \sin\theta & \cos\theta \end{pmatrix}.
\end{equation}
The corresponding Lie algebra, $\mathfrak{so}(2)$, consists of $2\times2$ skew-symmetric matrices. It is 1-dimensional, spanned by the generator $A$:
\begin{equation}
    \mathfrak{so}(2) = \{ t A \mid t \in \mathbb{R} \}, \quad \text{where } A = \begin{pmatrix} 0 & -1 \\ 1 & 0 \end{pmatrix}.
\end{equation}
Since $\mathrm{SO}(2)$ is abelian, the exponential map is straightforward: $\exp(t A) = R(t)$.
The standard representation acting on $\mathbf{x} \in \mathbb{R}^2$ yields the generator action:
\begin{equation}
    d\rho(A)\mathbf{x} = A\mathbf{x} = \begin{pmatrix} -x_2 \\ x_1 \end{pmatrix}.
\end{equation}

\myparagraph{The $\mathrm{SO}(3)$ Group .}
The group of 3D rotations is non-abelian. Its Lie algebra, $\mathfrak{so}(3)$, consists of all $3\times3$ skew-symmetric matrices. A general element $A \in \mathfrak{so}(3)$ has the form:
\begin{equation}
    A = \begin{pmatrix} 
    0 & -c & b \\ 
    c & 0 & -a \\ 
    -b & a & 0 
    \end{pmatrix}, \quad a,b,c \in \mathbb{R}.
\end{equation}
Any such matrix can be written as $A = \theta K$, where $\theta$ is the rotation angle and $K$ is a skew-symmetric matrix representing the unit rotation axis ($K^T = -K$ and $\|K\|_{\tt F} = \sqrt{2}$). The exponential map is given by Rodrigues' rotation formula:
\begin{equation}
    \exp(\theta K) = I + \sin\theta K + (1 - \cos\theta) K^2.
\end{equation}
For example, the generator for a rotation about the $z$-axis corresponds to $a=b=0, c=1$:
\begin{equation}
    K_z = \begin{pmatrix} 
    0 & -1 & 0 \\ 
    1 & 0 & 0 \\ 
    0 & 0 & 0 
    \end{pmatrix}, \quad 
    \exp(\theta K_z) = \begin{pmatrix} \cos\theta & -\sin\theta & 0 \\ \sin\theta & \cos\theta & 0 \\ 0 & 0 & 1 \end{pmatrix}.
\end{equation}

The Lie algebra action $d\rho(A)$ describes the infinitesimal velocity of a point $\mathbf{x}$ being rotated. It is computed via matrix-vector multiplication:
\begin{equation}
    d\rho(A)\mathbf{x} = A\mathbf{x} = 
    \begin{pmatrix} 
    0 & -c & b \\ 
    c & 0 & -a \\ 
    -b & a & 0 
    \end{pmatrix}
    \begin{pmatrix} x_1 \\ x_2 \\ x_3 \end{pmatrix}
    =
    \begin{pmatrix} -cx_2 + bx_3 \\ cx_1 - ax_3 \\ -bx_1 + ax_2 \end{pmatrix}.
\end{equation}
This linear action is geometrically equivalent to the cross product $\mathbf{v} \times \mathbf{x}$, where the vector $\mathbf{v}=(a,b,c)^\top$ corresponds to the axis of rotation scaled by the angular velocity.

{\bf Lie algebra basis.} Since $\mathfrak{so}(3)$ is a 3-dimensional vector space, any element $A$ can be uniquely expressed as a linear combination of three basis matrices, $A = a A_x + b A_y + c A_z$. These basis elements correspond to infinitesimal rotations about the standard axes:
\begin{equation}
    A_x = \begin{pmatrix} 
    0 & 0 & 0 \\ 
    0 & 0 & -1 \\ 
    0 & 1 & 0 
    \end{pmatrix}, \quad
    A_y = \begin{pmatrix} 
    0 & 0 & 1 \\ 
    0 & 0 & 0 \\ 
    -1 & 0 & 0 
    \end{pmatrix}, \quad
    A_z = \begin{pmatrix} 
    0 & -1 & 0 \\ 
    1 & 0 & 0 \\ 
    0 & 0 & 0 
    \end{pmatrix}.
\end{equation}

\paragraph{Real Schur Decomposition.}

For any real square matrix $\mM \in \R^{n \times n}$, there exists an orthogonal matrix $\mU$ such that $\mU \mSigma \mU^\top=\mM$, where $\mSigma$ is upper quasi-triangular.
When $\mM$ is normal, \ie, $\mM\mM^\top=\mM^\top\mM$, $\mSigma$ is block diagonal with $1 \times 1$ and $2 \times 2$ blocks where the $2 \times 2$ blocks correspond to pairs of complex conjugate eigenvalues. And the canonical form for a these $2 \times 2$ block is (see Theorem 2.5.8 by~\citet{horn2012matrix}):
\[
\begin{pmatrix}
        a & b \\
        -b & a
\end{pmatrix},
\]
where the associated eigenvalues are $a \pm ib$. The spectral norm of these $2 \times 2$ matrices models the scaling factor of the associated basis vectors in $\mU$.

%% file: supp/method_details.tex
\section{Method Details}
\label{sec:sup_method_details}
In this section, we provide additional explanations and technical details of our proposed technique.
\subsection{Multi-generator Equivariance}
\label{sup:multi_gen}
In practice, many Lie groups of interest are generated by multiple generators. For example, the special orthogonal group $SO(3)$ is generated by three generators corresponding to rotations about the x, y, and z axes. To design a soft equivariant layer for a group $G$ with multiple Lie algebra generators $\{A_i\}_{i=1}^k$, we use the right singular vectors of the combined  constraint (following \equref{eq:equiv_condition_mat}) as
\be
\bar\mL = \begin{bmatrix} \mL_i \\ \vdots \\ \mL_k \end{bmatrix} \in \R^{k\cdot d\cdot d' \times d \cdot d'},
\ee 
where, $\mL_i = (d\rho_{\gX}(A_i)^\top \otimes \rmI_{d'} - \rmI_d \otimes d\rho_{\gY}(A_i))$.
The right singular vector of $\bar\mL$ corresponding is then used to design $\mB_{\tt eq}$ following \equref{eq:soft_equiv_filter}.
\subsection{Details on Schur Equivariance prediction}
\label{sup:Schur_details}
We present two examples showing the implications of Lemma~\ref{clm:schur_equiv} and the constraints on the weights for exact equivariance. 
\myparagraph{Example: 4D to 3D Equivariant Map.}
Consider a map from a 4D input space $\gX$ to a 3D output space $\gY$. Let the Schur decomposition of their respective Lie algebra representations be:
\be 
d\rho_\gX = \mU_\gX \mSigma_\gX \mU_\gX^\top, \quad
d\rho_\gY = \mU_\gY \mSigma_\gY \mU_\gY^\top
\ee
The input and output Schur forms $\mSigma_\gX$ and $\mSigma_\gY$ are block diagonal matrices with the following blocks:
\bea
\mS_1 = \begin{pmatrix} \eva & \evb \\ -\evb & \eva \end{pmatrix}, \
\mS_2 = \begin{pmatrix} \evc & \evd \\ -\evd & \evc \end{pmatrix} \\ 
\mSigma_\gX = \operatorname{diag}(\mS_1, \mS_2) \\
\mT_1 = \begin{pmatrix} \evc & \evd \\ -\evd & \evc \end{pmatrix}, \
\mT_2 = \begin{pmatrix} \evlambda_1 \end{pmatrix} \\
\mSigma_\gY = \operatorname{diag}(\mT_1, \mT_2)
\eea
The transformed weight (in Schur basis) matrix $\Theta'$ is a $3 \times 4$ matrix. To be equivariant, non-zero blocks $\Theta'_{IJ}$ can only exist where the representation blocks share eigenvalues, \ie, $\mT_I \simeq \mS_J$. All other blocks must be zero \cite{horn2012matrix} and the non-zero block must take the form of \equref{eq:2x2_commute}.

The final sparse structure for $\Theta'$ is
\be
\Theta' =
\left(
\begin{array}{c|c}
 \vzero_{2 \times 2} & \begin{pmatrix} \evalpha & \evbeta \\ -\evbeta & \evalpha \end{pmatrix} \\
 \hline
 \vzero_{1 \times 2} & \vzero_{1 \times 2}
\end{array}
\right).
\ee
This equivariant design reduces the number of learnable parameters from $3 \times 4 = 12$ in a standard dense layer to just 2, while guaranteeing the desired symmetry.

\myparagraph{Example: 4D to 4D Equivariant Map.}
Consider a map from a 4D input space $\gX$ to a 4D output space $\gY$. Let the Schur decomposition of their respective Lie algebra representations be:
\be 
d\rho_\gX = \mU_\gX \mSigma_\gX \mU_\gX^\top, \quad
d\rho_\gY = \mU_\gY \mSigma_\gY \mU_\gY^\top
\ee

Let the block diagonal Schur forms $\mSigma_\gX$ and $\mSigma_\gY$ be written as:
\be 
\mSigma_\gX = \operatorname{diag}(\mS_1, \mS_2), \quad
\mSigma_\gY = \operatorname{diag}(\mT_1, \mT_2)
\ee
We define the specific structure of these blocks using scalars $a, b$ and $c, d$ as follows:
\bea
\mS_1 = \begin{pmatrix} a & b \\ -b & a \end{pmatrix}, \
\mS_2 = \begin{pmatrix} c & d \\ -d & c \end{pmatrix} \\
\mT_1 = \begin{pmatrix} a & b \\ -b & a \end{pmatrix}, \
\mT_2 = \begin{pmatrix} c & d \\ -d & c \end{pmatrix}
\eea
The transformed weight matrix $\Theta'$ is a $4 \times 4$ matrix. Acoording to Lemma \ref{clm:schur_equiv} $\rmW'_{IJ}$ relating input block $\mS_J$ to output block $\mT_I$ can be non-zero only if the representations are equivalent ($\mT_I \simeq \mS_J$).

Here, we observe that $\mT_1 \simeq \mS_1$ and $\mT_2 \simeq \mS_2$, while cross-pairings (e.g., $\mT_1$ vs $\mS_2$) possess distinct eigenvalues. Consequently, the weight matrix allows for learnable parameters (following \equref{eq:2x2_commute}) on the diagonal blocks but enforces zeros elsewhere:
\be
\Theta' =
\left(
\begin{array}{c|c}
 \begin{pmatrix} \evalpha_1 & \evbeta_1 \\ -\evbeta_1 & \evalpha_1 \end{pmatrix} & \vzero_{2 \times 2} \\
 \hline
 \vzero_{2 \times 2} & \begin{pmatrix} \evalpha_2 & \evbeta_2 \\ -\evbeta_2 & \evalpha_2 \end{pmatrix}
\end{array}
\right)
\ee
This constrains the layer to take a block-diagonal structure in Schur basis, reducing the number of learnable parameters from 16 to 4. We illustrate these examples in \figref{fig:shur_filter}.
\input{figs/shur_filter}

\myparagraph{Schur equivariance projection for multiple generators.} To design the Schur projection operator for a Lie algebra with multiple generators, we compose the projection operator for each generator, \ie, $\mB_{\tt Schur} = \mB_{\tt Schur}^1 \circ \mB_{\tt Schur}^2 \circ \dots \circ \mB_{\tt Schur}^{n_G}$, where $\mB_{\tt Schur}^i$ is the projection operator design from $d\rho_\gX(A_i)$ and  $d\rho_\gY(A_i)$. For commuting generators, this yields the exact projection operator. However, for non-commutative generators, this is an approximation of the true projection operator.

\subsection{From hard thresholding to smooth cut-off }
\label{sec:smooth_cutoff}
Instead of applying a sharp cut-off (threshold) to the singular values as in~\equref{eq:inv_blur}, an alternative design is to use a smooth transition, \ie,
\bea 
\label{eq:exp_blur_inv}
\mB_{\tt inv} \triangleq \sum_{i} \gamma_i \vu_i \vu_i^\top,
\eea 
where $\gamma_i \triangleq  1 \text{ if~} \sigma_i < b$, else $\gamma_i \triangleq  \exp(-\sigma_i^2 / s^2)$. Here, $s \in \R^+$ is a hyperparameter controlling the decay rate.

\myparagraph{\color{cvprblue}\textit{Remarks.}} 
For the special case of shift invariance, the smooth cut-off in~\equref{eq:exp_blur_inv} is analogous to the concept of “transition band’’ in filter design~\cite{manolakis2011applied,vetterli2014foundations}. A sharp cut-off in the Fourier domain, \ie, an ideal filter, produces ripples in the time domain, which is undesirable in practice.

\myparagraph{Soft Schur equivariance projection.} Instead of following a strict cut-off when designing the projection operator, we follow a smooth cut-off defined as:
\bea
\mW'_{lk} = 
\begin{cases}
\gamma_{lk} \Theta'_{lk} & \text{if~} \mT_l \not \simeq \mS_k,  \lambda_{\mS_k}+\lambda_{\mT_l} > b, \\
{\tt Sym}(\Theta'_{lk}) + \gamma_{lk} (\Theta'_{lk} - {\tt Sym}(\Theta'_{lk}) ) & \text{if }  \mT_l \simeq  \mS_k, \lambda_{\mS_k}+\lambda_{\mT_l} > b, \\
\Theta'_{lk}  & \text{otherwise}
\end{cases}
\eea 
Where $\gamma_{lk} = \exp(-\frac{(\lambda_{\mS_k}+\lambda_{\mT_l})}{s^2})$ is a decay factor based on the eigenvalues of the corresponding representation blocks, with hyperparameters $s$.

\section{Additional Clarifications}
\label{sec:additional_clarifications}
\subsection{Properties of $\eta$-Soft Equivariance Metric}
\label{sup:stability}
The relative equivariance error in \equref{eq:soft_eq} provides a metric that is unaffected by local rescaling of the function and the input. For linear layers, it is a scale-invariant measure: rescaling the weight  {($\mW \rightarrow  \alpha \mW$)} or the input {($\vx \rightarrow \beta \vx$)}
does not change the relative equivariance error as
\bea 
\frac{\| \alpha \mW \rho_\gX(g) \beta \vx - \rho_\gY(g) \alpha \mW \beta \vx\|}{\| \alpha \mW \| \| \beta \vx\|} = \frac{\| \mW \rho_\gX(g)\vx - \rho_\gY(g) \mW \vx\|}{\| \mW \| \| \vx\|}.,
\eea
This property ensures that the metric captures the intrinsic equivariance error without being influenced by the scale of the weights or inputs. 

\begin{figure}[ht]
        \centering
        \includegraphics[width=0.9\linewidth]{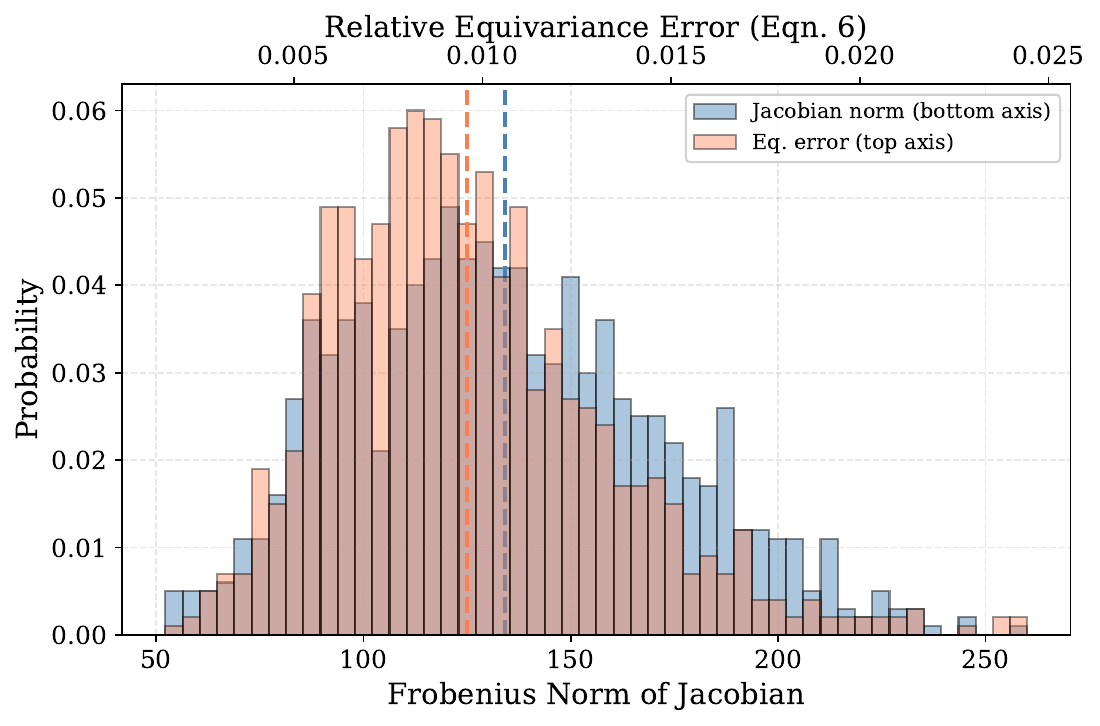}
        \caption{Distribution of the norm of the Jacobian and the relative equivariance metric (\equref{eq:soft_eq}) from a pre-trained ResNet-18 model on ImageNet. The relative equivariance error does not collapse or explode.}
        \label{fig:stability_relative_error}
\end{figure}

Furthermore, we empirically validate the stability of this metric by analyzing a pre-trained ResNet-18 model. As illustrated in \figref{fig:stability_relative_error}, the norm of the Jacobian of the logits with respect to the input does not vanish, and the relative equivariance error neither explodes nor collapses. This confirms the robustness and reliability of our proposed metric.

\subsection{Example}
\input{supp/example.tex}

%% file: figs/shur_filter.tex
\begin{figure}[h]
    \centering
    \includegraphics[width=0.9\linewidth]{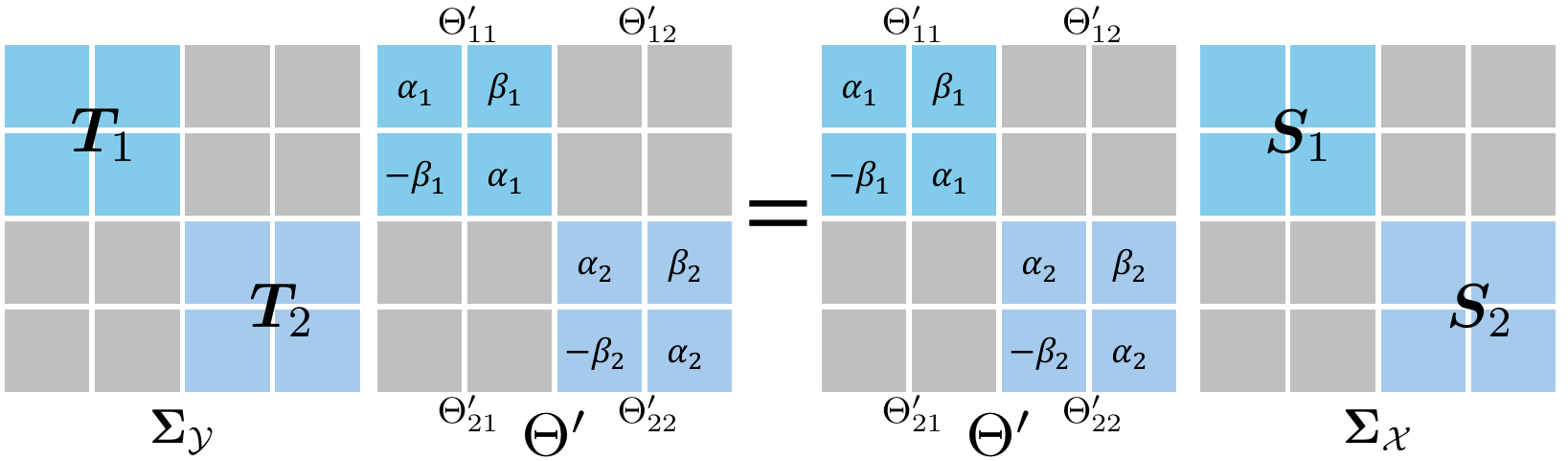}
    \caption{Visualization of condition of strict equivariance condition on the weight $\Theta'$ which is described in Lemma \ref{clm:schur_equiv}. }
    \label{fig:shur_filter}
\end{figure}

%% file: supp/example.tex
Here we will show a worked-out example for constructing a projection operator for weights $\mW \in \R^{3 \times 3}$ working on 3D vectors. We aim to achieve approximate equivariance with respect to rotation about the $z$-axis, corresponding to the following Lie algebra generator:
\begin{equation}\label{eq:Az}
    A_z = \begin{pmatrix} 0 & -1 & 0 \\ 1 & 0 & 0 \\ 0 & 0 & 0 \end{pmatrix}.
\end{equation}
Here we directly used the matrix form of $A_z$, \ie, $d\rho(A_z)=A_z$

\subsubsection{Using SVD}
Following \ref{eq:equiv_condition_mat}, we form the Kronecker product
\begin{equation}\label{eq:L}
    \mL = A_z^\top \otimes I_3 - I_3 \otimes A_z \;\in\; \R^{9 \times 9}.
\end{equation}
When written explicitly, we obtain:
\begin{equation}\label{eq:L-explicit}
    \mL =
    \begin{pmatrix}
        0 & 1 & 0 & 1 & 0 & 0 & 0 & 0 & 0 \\
        -1 & 0 & 0 & 0 & 1 & 0 & 0 & 0 & 0 \\
        0 & 0 & 0 & 0 & 0 & 1 & 0 & 0 & 0 \\
        -1 & 0 & 0 & 0 & 1 & 0 & 0 & 0 & 0 \\
        0 & -1 & 0 & -1 & 0 & 0 & 0 & 0 & 0 \\
        0 & 0 & -1 & 0 & 0 & 0 & 0 & 0 & 0 \\
        0 & 0 & 0 & 0 & 0 & 0 & 0 & 1 & 0 \\
        0 & 0 & 0 & 0 & 0 & 0 & -1 & 0 & 0 \\
        0 & 0 & 0 & 0 & 0 & 0 & 0 & 0 & 0
    \end{pmatrix}.
\end{equation}
The singular values of $\mL$ are $\{0^{(\times 3)},\; 1^{(\times 4)},\; 2^{(\times 2)}\}$.  

The 3-dimensional null space corresponds to singular values of $0$, and yields the following 3-parameter family of exactly equivariant weights, i.e., exact equivariance:
\begin{equation}\label{eq:W-equiv}
    W_{\mathrm{eq}} = \begin{pmatrix} \alpha & \beta & 0 \\ -\beta & \alpha & 0 \\ 0 & 0 & \gamma \end{pmatrix}, \quad \alpha, \beta, \gamma \in \R.
\end{equation}

With SVD $\mL = \mU^L \Sigma^L (\mV^L)^\top$, the projection $\mB_{\tt eq} = \sum_{i:\,\sigma_i < b} v^L_i\,(v^L_i)^\top$ keeps right singular vectors below cutoff~$b$.  For $b = 1.5$, the 7 vectors with $\sigma \in \{0, 1\}$ are retained, yielding a 7-dimensional projected subspace. Reshaping each of the 7 right singular vectors $v^L_i \in \R^9$ into $3 \times 3$ matrices $\mV_i = \mathrm{reshape}(v^L_i)$, the projected weight can be written as:
\begin{equation}\label{eq:W-basis}
    \mW = \sum_{i=1}^{7} a_i \mV_i = a_1 \mV_1 + a_2 \mV_2 + a_3 \mV_3 + a_4 \mV_4 + a_5 \mV_5 + a_6 \mV_6 + a_7 \mV_7,
\end{equation}
where the 7 basis matrices are (3 from $\sigma=0$, 4 from $\sigma=1$):
\begin{align*}
    \mV_1 &= \begin{pmatrix} 1 & 0 & 0 \\ 0 & 1 & 0 \\ 0 & 0 & 0 \end{pmatrix}, \quad
    \mV_2 = \begin{pmatrix} 0 & 1 & 0 \\ -1 & 0 & 0 \\ 0 & 0 & 0 \end{pmatrix}, \quad
    \mV_3 = \begin{pmatrix} 0 & 0 & 0 \\ 0 & 0 & 0 \\ 0 & 0 & 1 \end{pmatrix}, \\[4pt]
    \mV_4 &= \begin{pmatrix} 0 & 0 & 1 \\ 0 & 0 & 0 \\ 0 & 0 & 0 \end{pmatrix}, \quad
    \mV_5 = \begin{pmatrix} 0 & 0 & 0 \\ 0 & 0 & 1 \\ 0 & 0 & 0 \end{pmatrix}, \quad
    \mV_6 = \begin{pmatrix} 0 & 0 & 0 \\ 0 & 0 & 0 \\ 1 & 0 & 0 \end{pmatrix}, \quad
    \mV_7 = \begin{pmatrix} 0 & 0 & 0 \\ 0 & 0 & 0 \\ 0 & 1 & 0 \end{pmatrix}.
\end{align*}
Here $\mV_1, \mV_2, \mV_3$ span the exactly equivariant subspace ($\sigma=0$), while $\mV_4, \mV_5, \mV_6, \mV_7$ correspond to $\sigma=1$ directions that mildly break equivariance by coupling the $xy$-plane to the $z$-axis.

\subsubsection{Schur decomposition approach}
For this example $A_z$ is already in real Schur form, $\mU_\gX = \mU_\gY = I_3$. Furthermore, $\mSigma_\gX = \diag(\{\mS_k\}_{k=1}^2) = \mSigma_\gY = \diag(\{\mT_l\}_{l=1}^2)= A_z$ with $A_z$ having the following block diagonal structure:
\begin{equation}\label{eq:schur-form}
    A_z = \diag\!\left(
        \underbrace{\begin{pmatrix} 0 & -1 \\ 1 & 0 \end{pmatrix}}_{S_1,\;\lambda_{\mS_1}=1},\;
        \underbrace{(0)}_{\mS_2,\;\lambda_{S_2}=0}
    \right).
\end{equation}

The weight in Schur basis $\Theta' = \mU_\gY^\top \Theta \mU_\gX = \Theta$ is partitioned into blocks. Following Lemma \ref{clm:schur_equiv}, each block $\Theta'_{lk}$ is governed by the Sylvester equation $T_l\,\Theta'_{lk} = \Theta'_{lk}\,S_k$ which can be summarized as follows:

\begin{center}
\begin{tabular}{@{}ccccc@{}}
    \toprule
    Block & Size & $\lambda_{\mT_l} + \lambda_{\mS_k}$ & $\mT_l \simeq \mS_k$? & Equivariant solution \\
    \midrule
    $\Theta'_{11}$ & $2\!\times\!2$ & $2$ & Yes & $\begin{psmallmatrix} \alpha & \beta \\ -\beta & \alpha \end{psmallmatrix}$ \\[4pt]
    $\Theta'_{12}$ & $2\!\times\!1$ & $1$ & No & $\mathbf{0}$ \\
    $\Theta'_{21}$ & $1\!\times\!2$ & $1$ & No & $\mathbf{0}$ \\
    $\Theta'_{22}$ & $1\!\times\!1$ & $0$ & Yes & $\gamma$ \\
    \bottomrule
\end{tabular}
\end{center}

\myparagraph{Schur projection with $b = 1.5$:}
Following \equref{eq:schur_conditions}, blocks with $\lambda_{\mT_l} + \lambda_{\mS_k} > 1.5$ are projected; blocks below this threshold are unconstrained.  At $b = 1.5$:
\begin{itemize}
    \item $\Theta'_{11}$ ($\lambda = 2 \geq 1.5$, $T_1 \simeq S_1$): symmetrized to $\begin{psmallmatrix} \alpha & \beta \\ -\beta & \alpha \end{psmallmatrix}$ form (2 degree of freedom) after projection.
    \item $\Theta'_{12}$, $\Theta'_{21}$ ($\lambda = 1 < 1.5$): unconstrained (4 degree of freedom) ,\ie, remain unchanged after projection.
    \item $\Theta'_{22}$ ($\lambda = 0 < 1.5$): unconstrained (1 degree of freedom).
\end{itemize}

\myparagraph{Numerical Example: Schur projection in action}
Let the weights before projection be
$$\Theta = \begin{pmatrix} 2 & 3 & 1 \\ -1 & 4 & 2 \\ 3 & -1 & 5 \end{pmatrix}$$.

Applying the Schur projection with cutoff $b = 1.5$:
\begin{itemize}
    \item $\Theta'_{11} = \begin{pmatrix} 2 & 3 \\ -1 & 4 \end{pmatrix} \;\xrightarrow{\Sym}\; \begin{pmatrix} 3 & 2 \\ -2 & 3 \end{pmatrix}$
    \item $\Theta'_{12} = \begin{pmatrix} 1 \\ 2 \end{pmatrix}$, \;$\Theta'_{21} = \begin{pmatrix} 3 & -1 \end{pmatrix}$ : unchanged.
    \item \;$\Theta'_{22} = (5)$: unchanged.
\end{itemize}

The final projected weight matrix is 
$$
    \mW = \begin{pmatrix} 3 & 2 & 1 \\ -2 & 3 & 2 \\ 3 & -1 & 5 \end{pmatrix}.
$$

\myparagraph{\color{cvprblue}\textit{ Remarks.}} We note that both the SVD and the Schur decomposition approach give rise to the same projection operator. However, computing the projection operator via Schur decomposition is much more efficient.

%% file: supp/math_proofs.tex
\section{Complete Proofs of the Claims and Lemmas}
\label{supp:proofs}

{\bf \noindent Reformulation of equivariance constraint.} To study deviation of a function $F$ from perfect equivariance at an input $\vx$, we define the equivariance residual/error $R(g)$ under the action of group element $g \in G$:
\begin{equation}
R(g) = F(\rho_\gX(g)\vx) - \rho_\gY(g)F(\vx).
\end{equation}

For perfect equivariance, $\|R(g)\| = 0$ for all $g \in G$ and for soft equivariance the $\|R(g)\|$ is bounded for all $g \in G$. Applying the first order Taylor approximation from \equref{eq:lie_taylor} around the identity element $e \in G$, we can rewrite the error as:
\begin{align}
\label{eq:equivariance_residual_error}
R(g) &= R(e) + \sum_i t_i \left.\frac{\partial R(\exp(\sum_i t_i A_i))}{\partial t_i}\right|_{t=0} + O(\|A\|_{\mathfrak{g}}^2)\\ 
     & = \sum_i t_i \left.\frac{\partial R( \exp(\sum_i t_i A_i))}{\partial t_i}\right|_{t=0} + O(\|A\|_{\mathfrak{g}}^2) , \quad \quad \quad\text{(as~} R(e) = 0)  
\end{align}
where $g = \exp(A)$ with $A=\sum_i t_i A_i$ and $\|A\|_{\mathfrak g} \le r_G$, where $r_G$ is the injective radius (or injectivity radius) of $G$. Hence, the Taylor remainder is bounded by
\[
O(\|A\|_{\mathfrak g}^2) \le O(r_G^2) \le \epsilon_G,
\]
for some group-dependent constant $\epsilon_G$. In the following proofs, we first bound the linear term and then add the residual term $\epsilon_G$ in the final estimate.

\subsection{Proof of~\clmref{clm:soft_inv}}
\label{supp:soft_inv}
\begin{myclaime}
\claimone*
\end{myclaime}

\begin{proof}
We apply the Taylor expansion of the equivariance error (\equref{eq:equivariance_residual_error}) on the function $F_{\tt FC}=\vw^\top \vx$ with $\vw = \mB_{\tt inv} \theta$. Following the convention established in 
Eq.~\eqref{eq:equivariance_residual_error}, we work with the 
first-order terms and absorb the Taylor residual error 
$\epsilon_G$ into the final bound and obtain:

\begin{align}
R(g) &= \sum_i t_i \left.\frac{\partial \left(\vw^\top \rho_\gX(\exp(\sum_i t_i A_i)) \vx - \vw^\top \vx\right)}{\partial t_i}\right|_{   t=0} \\
&= \sum_i t_i \vw^\top \left. \frac{\partial \rho_\gX(\exp(\sum_i t_i A_i)) \vx}{\partial t_i}\right|_{t=0} \\
&= \sum_i t_i \vw^\top d\rho_\gX(A_i)\vx \quad \text{(using Lie algebra representation)}.
\end{align}
For $F_{\tt FC}$ the exact invariance condition becomes:
\begin{equation}
R(g) = 0 \implies \sum_i t_i \vw^\top d\rho_\gX(A_i)\vx = 0 \quad \forall t_i, \vx, g.
\end{equation}
This implies that if $\vw$ is in the left null space $d\rho_\gX(A_i)$ for all $i$, then the exact invariance condition is satisfied. We now derive the bound on the equivariance error when $\vw$ is not in the null space of $d\rho_\gX(A_i)$.

We denote the singular value decomposition (SVD) of each Lie algebra representation $d\rho_\gX(A_i)$ as:
\be
d\rho_\gX(A_i) = \mU^i \Sigma^i {\mV^i}^\top,
\ee
where $U^i$ and $V^i$ are orthonormal matrices and $\Sigma^i$ is a diagonal matrix with singular values $0 \le \sigma^i_1 \le \sigma^i_2 \le \ldots$ arranged in increasing order.

We can express the invariance error norm, $\|R(g)\|$, as:
\begin{align}
\|R(g)\| &= \left\|\sum_i t_i \vw^\top d\rho_\gX(A_i)\vx\right\| \\
&\leq \sum_i |t_i|  |\vw^\top d\rho_\gX(A_i)\vx| \\
& =  \sum_i  |t_i|  |\vw^\top U^i \Sigma^i {V^i}^\top \vx | \\
&=  \sum_i   |t_i|  |({\mU^i}^\top \vw)^\top \Sigma^i ({\mV^i}^\top \vx)| \\
&=  \sum_i   |t_i|  \left|{W^i}^\top \Sigma^i X^i\right|
\end{align}
where $W^i = {\mU^i}^\top \vw$ and $X^i = {\mV^i}^\top \vx$ are the coordinates of $\vw$ and $\vx$ in the left and right singular vector bases, respectively.

As $b$ is the cut-off value of the projection of operator $\mB_{\tt inv}$, $W^i_j = 0$ for all $\sigma^i_j > b$ (true for all generators by definition). 

Therefore, we can write:

\begin{align}
\|R(g)\| &\leq \sum_i |t_i| \sum_{j:\sigma^i_j < b} \sigma^i_j |W^i_j| |X^i_j| \quad (\text{as $\Sigma^i$ is a diagonal matrix})\\
    &\leq \sum_i b |t_i| \sum_{j:\sigma^i_j < b} |W^i_j| |X^i_j|  \\
    &\leq \sum_i b |t_i| \|W^i\| \|X^i\| \quad \text{(by Cauchy-Schwarz)}  \\
    &\leq \sum_i b |t_i| \|\vw\| \|\vx\| \quad (\text{as $U^i$ and $V^i$ are orthonormal } \|W^i\| = \|\vw\|)\\
    &= b \|\vw\| \|\vx\| \sum_i |t_i|
\end{align}
For compact and connected Lie groups $\sqrt{\sum_i t_i^2} \le r_G $ (assuming an orthonormal basis for the Lie algebra), where $r_G$ is the injective radius of the group $G$ \cite{klingenberg1995riemannian}. Applying Cauchy-Schwarz inequality on $\sum_i |t_i|$ gives:
\begin{align}
\sum_i |t_i| \leq \sqrt{n_G} \sqrt{\sum_i t_i^2} \leq \sqrt{n_G} r_G,
\end{align}
where $n_G$ is the number of Lie algebra basis elements of the group $G$.

For the linear function $F_{\tt FC}$, the norm of the Jacobian at $\vx$ is $\|\rmJ_{F_{\tt FC}(;\vw)}(\vx)\|_{\tt F} = \|\vw\|$. Lastly, substituting the definition of $R(g)$ and $\vw$, and reintroducing the Taylor residual error $\epsilon_G$,  we obtain the final bound on the equivariance error:
\begin{align}
\| (\mathbf{B}\theta)^\top \mathbf{x} - (\mathbf{B}\theta)^\top \rho_\gX(g)\mathbf{x} \|
\leq b \|\vw\| \|\mathbf{x}\| \sqrt{n_G} r_G + \epsilon_G
\implies 
\frac{\| (\mathbf{B}\theta)^\top \mathbf{x} - (\mathbf{B}\theta)^\top \rho_\gX(g)\mathbf{x} \|}
     {\|\mathbf{J}_{F_{\tt FC}(; \vw)}(\mathbf{x})\|_{\tt F} \|\mathbf{x}\|}
\leq b\, \sqrt{n_G} r_G +\varepsilon_G.
\end{align}
Here, $\varepsilon_G$ is the Taylor residual normalized by $\|\vw\| \|\mathbf{x}\|$.  
\end{proof}

\subsection{Proof for Multiple Generators.}
We now prove that the Claim~\ref{clm:soft_inv} also holds for the projection operator $\mB_{\tt inv}$ designed using the combined SVD of multiple Lie algebra generators.  
\label{supp:multi_generator}
\begin{proof}
We denote 
\bea 
\mA \triangleq [ d\rho_\gX(A_1) \mid d\rho_\gX(A_2) \mid \dots \mid  d\rho_\gX(A_{n_G})],
\eea 
where, $A_i, \dots, A_{n_G}$ are the basis elements of the Lie algebra. We also denote the SVD of $\mA$ as $\mA = \mU \Sigma \mV^\top$.

We now define an extended input $\vz$ as
\bea 
\vz \triangleq  \begin{bmatrix} t_1 \vx \\ t_2 \vx \\ \vdots \\ t_{n_G} \vx \end{bmatrix},
\eea 
\end{proof}
\ie, $\vz$ is $n_G$ times repetition of the input $\vx$ scaled by parameter associated with the Lie algebra generator.

Now following \secref{supp:soft_inv} we express the invariance error as
\bea 
\|R(g)\| &=& \|\sum_i t_i \vw^\top d\rho_\gX(A_i)\vx \| \\
&=& \| \vw^\top \sum_i t_i  d\rho_\gX(A_i)\vx \| \\
&=& \| \vw^\top \mA \vz \| \\
&=& \| \vw^\top \mU \Sigma \mV^\top \vz \| \\
&=& \| (\mU^\top\vw)^\top  \Sigma (\mV^\top \vz) \| \\
&=& \| W^\top  \Sigma Z \|.
\eea 
Following the condition of $\mB_{\tt inv}$, $W_j = 0$ for $\sigma_j > b$. We can rewrite the bound as 
\bea
\|R(g)\| &\leq & \| \sum_{\sigma_j < b } b W_j   Z_j\| \\
&\leq & b \|\vw\| \|\vz\| \quad \text{by Cauchy-Schwarz and } \|W\| = \|\vw\|, \|Z\| = \|\vz\| \\
&\leq & b \|\vw\| \sqrt{\sum_i t_i^2} \|\vx\| \quad \text{following the construction of } \vz\\
&\leq & b r_G \|\vw\|  \|\vx\| \quad \text{using the definition of injective radius}\\
&\leq & b \sqrt{n_G} r_G  \|\vw\|  \|\vx\|. \quad \text{as } n_G \ge 1
\eea 
Finally, following the definition of $R(g)$, and Jacobian of the linear function $F_{\tt FC}$, we obtain
\bea
\frac{\| (\mathbf{B}\theta)^\top \mathbf{x} - (\mathbf{B}\theta)^\top \rho_\gX(g)\mathbf{x} \|}
     {\|\mathbf{J}_{F_{\tt FC}(;\vw)}(\mathbf{x})\|_{\tt F} \|\mathbf{x}\|}
 \leq b\, \sqrt{n_G} r_G + \varepsilon_G.
\eea 
Thus, the designed projection operator via combined decomposition satisfies Claim \ref{clm:soft_inv}.

{\bf Note:} An alternative approach for designing a projection operator for groups with multiple generators is to design a projection operator $B_{\tt inv, i}$ for each generator $A_i$ and the final projection operator can be approximated via compositions $B_{\tt inv} \approx B_{\tt inv, 1} \circ B_{\tt inv, 2} \circ \dots B_{\tt inv, n_G}$. For commutative generators, this approximation is exact. For non-commutative generators, better approximations can be obtained by recursive application of these projection operators.

\subsection{Proof of~\clmref{clm:soft_eqv}}
\label{supp:soft_eqv}
\begin{myclaime}
\claimtwo*
\end{myclaime}
\begin{proof}

Applying the Taylor expansion of the equivariance error \equref{eq:equivariance_residual_error} in the function $F_{\tt FC} = \mW \vx$ with ${\tt vec}(\mW) = \mB'_{\tt eq} \theta \in \R^{d' d}$ and retaining only the linear terms, we obtain:
\begin{align}
R(g) &= \sum_i t_i \left. \frac{\partial \left(\mW \rho_\gX(\exp(\sum_i t_i A_i)) \vx - \rho_\gY(\exp(\sum_i t_i A_i)) \mW \vx \right)}{\partial t_i}\right|_{t=0} \\ 
&= \sum_i t_i \left( \mW d\rho_{\gX}(A_i)\vx - d\rho_{\gY}(A_i) \mW \vx \right)  \\ 
&= \sum_i t_i \left( \mW d\rho_{\gX}(A_i)- d\rho_{\gY}(A_i) \mW \right) \vx  \label{eq:basic_eq_condition}\\ 
&= \sum_i t_i (\vx^\top \otimes \mI_{d'}) \text{\tt vec}(\mW d\rho_{\gX}(A_i) - d\rho_{\gY}(A_i) \mW) \quad \left(\text{using the fact~} \mM \vz = (\vz^\top \otimes \mI_{d'}) {\tt vec}(\mM) \right)\\
&= \sum_i t_i (\vx^\top \otimes \mI_{d'}) \text{\tt vec}(\mI_{d'}\mW d\rho_{\gX}(A_i) - d\rho_{\gY}(A_i) \mW \mI_d) \\
&= \sum_i t_i (\vx^\top \otimes \mI_{d'}) (d\rho_{\gX}(A_i)^\top \otimes \mI_{d'} - \mI_d \otimes d\rho_{\gY}(A_i)) \text{\tt vec}(\mW) \quad \left(\text{as } {\tt vec}(\mA\mB\mC) =  \left(\mC^\top \otimes \mA\right) {\tt vec}(\mB)\right).\label{eq:expanded_equivariance}
\end{align}
We denote $\mL_i = d\rho_{\gX}(A_i)^\top \otimes \rmI_{d'} - \rmI_d \otimes d\rho_{\gY}(A_i)$ and $\overline{\vx} = \vx^\top \otimes \rmI_{d'}$ and define the SVD of $\mL_i$ as
\begin{equation}
    \mL_i = \mU^{\tt L_i} \Sigma^{\tt L_i} {\mV^{\tt L_i}}^\top,
\end{equation}
where $\rmU^{L_i}$ and $\rmV^{L_i}$ are orthogonal matrices and $\Sigma^{L_i}$ is a diagonal matrix with singular values $0 \leq \sigma_1 \leq  \sigma_2 \le \ldots$ arranged in increasing order.
We can express the equivariance error norm as:
\begin{align}
\|R(g)\| &= \left\|\sum_i t_i \overline{\vx} \mL_i \text{\tt vec}(\mW)\right\| \\
&\leq \sum_i |t_i| \cdot \|\overline{\vx} \mL_i \text{\tt vec}(\mW)\| \\
&= \sum_i |t_i| \|\overline{\vx} \mU^{\tt L_i} \Sigma^{\tt L_i} {\mV^{\tt L_i}}^\top \text{\tt vec}(\mW)\| \\
&= \sum_i |t_i| \|({\mU^{\tt L_i}}^\top \overline{\vx}^\top)^\top \Sigma^{\tt L_i} ({\mV^{\tt L_i}}^\top \text{\tt vec}(\mW))\| \\
&= \sum_i |t_i| \|\overline{X}_i^\top \Sigma^{\tt L_i} \overline{W}_i\|,
\end{align}
where $\overline{X}^i = ({\mU^{\tt L_i}}^\top \overline{\vx}^\top) \in \mathbb{R}^{d'd \times d'}$ and $\overline{W}^i = {\mV^{\tt L_i}}^\top \text{\tt vec}(\mW) \in \mathbb{R}^{d'd}$ are the coordinates of $\overline{\vx}$ and $\text{\tt vec}(\mW)$ in the left and right singular vector bases, respectively. Here $\overline{X}^i$ is a matrix with $d'$ columns, where $\overline{X}^i_c$ denotes its $c$-th column, and $\overline{X}^i_{c,j}$ denotes the $j$-th component of that column.

As $b$ is the cut-off value of the projection of operator $\mB_{\tt eq}$, $\overline{W}^i_j = 0$ for all $\sigma^i_j > b$ (follows from the definition of the cut-off values and holds for all generators).

Therefore, we can write:
\begin{align}
\|R(g)\| &\leq \sum_i |t_i| \|\overline{X}^{i\top} \Sigma^{\tt L_i} \overline{W}^i\| \\
&\leq \sum_i |t_i| \sum_{c=1}^{d'} \left\| {\overline{X}^i_{c}}^\top \Sigma^{\tt L_i} \overline{W}^i\right\| \\ 
&\leq \sum_i |t_i| \sum_{c=1}^{d'} \left\|\sum_{j:\sigma^i_j < b} \sigma^i_j \overline{X}^i_{c,j} \overline{W}^i_j\right\| \\ 
&\leq \sum_i |t_i| \sum_{c=1}^{d'} \sum_{j:\sigma^i_j < b} \sigma^i_j |\overline{X}^i_{c,j}| |\overline{W}^i_j| \\
&\leq \sum_i |t_i| b \sum_{c=1}^{d'} \sum_{j:\sigma^i_j < b} |\overline{X}^i_{c,j}| |\overline{W}^i_j| \\
&\leq \sum_i |t_i| b \left(\sum_{c=1}^{d'} \sum_{j:\sigma^i_j < b} |\overline{X}^i_{c,j}|^2\right)^{1/2} \left(\sum_{j:\sigma^i_j \le b} |\overline{W}^i_j|^2\right)^{1/2}\\
&\leq \sum_i |t_i| b \|\overline{X}^i\|_{\tt F} \|\overline{W}^i\| \\
&\leq \sum_i |t_i| b \|\overline{\vx}\|_{\tt F} \|\text{\tt vec}(\mW)\| \quad \text{(as $\rmU^{L_i}$ and $\rmV^{L_i}$ are orthonormal)}.
\end{align}

Now we compute $\|\overline{\vx}\|_{\tt F}$ where $\overline{\vx} = \vx^\top \otimes \rmI_{d'} \in \mathbb{R}^{d' \times d'd}$:
\begin{align}
\|\overline{\vx}\|_{\tt F}^2 &= \|\vx^\top \otimes \rmI_{d'}\|_{\tt F}^2 \\
&= \text{tr}\left((\vx^\top \otimes \rmI_{d'})(\vx^\top \otimes \rmI_{d'})^\top\right) \\
&= \text{tr}\left((\vx^\top \otimes \rmI_{d'})(\vx \otimes \rmI_{d'})\right) \\
&= \text{tr}\left((\vx^\top \vx) \otimes (\rmI_{d'} \rmI_{d'})\right) \\
&= \text{tr}\left(\|\vx\|^2 \otimes \rmI_{d'}\right) \\
&= \|\vx\|^2 \cdot \text{tr}(\rmI_{d'}) \\
&= d' \|\vx\|^2.
\end{align}
Therefore, $\|\overline{\vx}\|_{\tt F} = \sqrt{d'} \|\vx\|$. 
Substituting the norms, we can rewrite the inequality as:
\begin{align}
\|R(g)\| &\leq \sum_i |t_i| b \sqrt{d'} \|\vx\| \|\mW\|_{\tt F}
\end{align}
as $\|\text{\tt vec}(\mW)\| = \|\mW\|_{\tt F}$.

For compact and connected Lie groups $\sqrt{\sum_i t_i^2} \le r_G$ (assuming an orthonormal basis for the Lie algebra), where $r_G$ is the injective radius of the group $G$. Applying Cauchy-Schwarz inequality on $\sum_i |t_i|$ gives:
\begin{align}
\sum_i |t_i| \leq \sqrt{n_G} \sqrt{\sum_i t_i^2} \leq r_G \sqrt{n_G},
\end{align}
where $n_G$ is the number of Lie algebra basis elements of the group $G$. 

As the function $F_{\tt FC}$ is linear, the norm of the Jacobian at $\vx$ is $\|\rmJ_{F(;\mW)_{\tt FC}}(\vx)\|_{\tt F} = \|\mW\|_{\tt F}$. Substituting the definition of $R(g)$ and $\|\rmJ_{F(;\mW)_{\tt FC}}(\vx)\|$, and reintroducing the Taylor residual error $\epsilon_G$, we obtain the final bound on the equivariance error:
\begin{align}
&\| \mW \rho_\gX(g) \vx - \rho_\gY(g)\mW \vx \| \leq b  r_G \sqrt{n_G d'}  \|\rmJ_{F(;\mW)_{\tt FC}}(\vx)\|_{\tt F} \|\vx\| + \epsilon_G \\
&\implies \frac{\| \rmB' \Theta \rho_\gX(g) \vx - \rho_\gY(g) \rmB' \Theta \vx \|}{\|\rmJ_{F(;\mW)_{\tt FC}}(\vx)\|_{\tt F} \|\vx\|} \leq b r_G \sqrt{n_G d'} + \varepsilon_G.
\end{align}
Here, $\varepsilon_G$ is the normalized Taylor residual.
\end{proof}

\subsection{Proof of Lemma~\ref{clm:schur_equiv}}
\label{supp:schur_equiv}
\begin{myclaime}
\lemmaone*
\end{myclaime}
\begin{proof}
Following \equref{eq:basic_eq_condition}, we can write the condition for exact equivariance as for weight $\Theta \in \R^{d' \times d}$ as:
\begin{equation}
\|R(g)\| = 0 \implies \norm{\sum_i t_i \left( \Theta d\rho_{\gX}(A_i)- d\rho_{\gY}(A_i) \Theta \right) \vx}= 0.
\end{equation}
The above condition holds if 
\be
\|\Theta d\rho_{\gX}(A_i)- d\rho_{\gY}(A_i) \Theta\|_{\tt F} = 0 \quad \forall i.
\ee
As the condition is the same for all Lie algebra generators $A_i$, we drop the index $A_i$ for readability. %
We define the Schur decomposition of the Lie algebra representations as:
\bea
d\rho_{\gX} = \mU_\gX \mSigma_\gX \mU_\gX^\top, \quad d\rho_{\gY} = \mU_\gY \mSigma_\gY \mU_\gY^\top,
\eea
where $\mU_\gX$ and $\mU_\gY$ are orthonormal.
And we denote $\Theta' = \mU_\gY^\top \Theta \mU_\gX$
Substituting the Schur decomposition in the equivariance condition, we obtain:
\begin{align}
&\|\Theta d\rho_{\gX} - d\rho_{\gY} \Theta \|_{\tt F}= 0 \\
\implies &\|\Theta \mU_\gX \mSigma_\gX \mU_\gX^\top - \mU_\gY \mSigma_\gY \mU_\gY^\top \Theta \|_{\tt F}= 0 \\
\implies &\|\mU_\gY^\top (\Theta \mU_\gX \mSigma_\gX \mU_\gX^\top - \mU_\gY \mSigma_\gY \mU_\gY^\top \Theta)\mU_\gX \|_{\tt F} = 0 \quad (\text{orthonormal matrix preserves norm})\\
\implies &\|\mU_\gY^\top \Theta \mU_\gX \mSigma_\gX \mU_\gX^\top \mU_\gX - \mU_\gY^\top \mU_\gY \mSigma_\gY \mU_\gY^\top \Theta \mU_\gX \|_{\tt F} = 0 \\
\implies &\|\Theta' \mSigma_\gX - \mSigma_\gY \Theta'\|_{\tt F} = 0
\end{align}
This condition holds when 
\be 
\mSigma_\gY \Theta' - \Theta' \mSigma_\gX = \vzero. \label{eq:equiv_condition_in_schur}
\ee
Now as $d\rho_{\gX}$ and $d\rho_{\gY}$ are normal matrices, $\mSigma_\gX$ and $\mSigma_\gY$ are block diagonal matrices \ie, $\mSigma_\gX=\operatorname{diag}(\{\mS_k\}_{k=1}^p)$ and $\mSigma_\gY=\operatorname{diag}(\{\mT_l\}_{l=1}^q)$, where $\{\mS_k\}_{k=1}^p$ and $\{\mT_l\}_{l =1}^q$ are sets of $1\times1$ or $2\times2$ Schur form blocks. We decompose $\Theta'$ into blocks $\Theta'_{lk}$ corresponding to the blocks of $\mSigma_\gX$ and $\mSigma_\gY$  of size $\dim(\mT_l) \times \dim(\mS_k)$ \ie, $\Theta'_{lk}$ is a submatrix of $\Theta'$ corresponding to the rows associated with block he $\mT_l$ in $\mSigma_\gY$ and columns associated with the block $\mS_k$ in $\mSigma_\gX$.

As the $0$ entries of $\mSigma_\gX$ and $\mSigma_\gY$ do not contribute to the matrix multiplication, we can rewrite \equref{eq:equiv_condition_in_schur} as a blockwise condition:
\be
\mT_l \Theta'_{lk} - \Theta'_{lk} \mS_k = \vzero \quad \forall l,k.
\ee
\end{proof}

\subsection{Proof of~\clmref{clm:schur_filter}}
\label{supp:schur_filter}
\begin{myclaime}
\claimfour*
\end{myclaime}
\begin{proof}
Applying the Taylor expansion of the equivariance error \equref{eq:equivariance_residual_error} for the function $F_{\tt FC}(x, \mW) = \mW \vx$ and retaining only the linear terms, we obtain:
\begin{align}
R(g) &= \sum_i t_i \left. \frac{\partial \left(\rmW \rho_\gX(\exp(t_i A_i)) \vx - \rho_\gY(\exp(t_i A_i)) \rmW \vx \right)}{\partial t_i}\right|_{t=0} \\ 
&= \sum_i t_i \left( \rmW d\rho_{\gX}(A_i)\vx - d\rho_{\gY}(A_i) \rmW \vx \right)  \\ 
&= \sum_i t_i \left( \rmW d\rho_{\gX}(A_i)- d\rho_{\gY}(A_i) \rmW \right) \vx.
\end{align}
Now, we denote ${\mW^i}' = {\mU_\gY^i}^\top \mW \mU_\gX^i$ where $\mU_\gX^i$ and $\mU_\gY^i$ are the orthonormal matrices from the Schur decomposition of $d\rho_{\gX}(A_i)$ and $d\rho_{\gY}(A_i)$ respectively.
We denote $\mSigma_\gX=\operatorname{diag}(\{\mS_k^i\}_{k=1}^p)$ and $\mSigma_\gY=\operatorname{diag}(\{\mT_l^i\}_{l=1}^q)$, where $\{\mS_k^i\}_{k=1}^p$ and $\{\mT_l^i\}_{l =1}^q$ are sets of $1\times1$ or $2\times2$ Schur form blocks corresponding to the Schur decomposition of $d\rho_{\gX}(A_i)$ and $d\rho_{\gY}(A_i)$. By $\lambda_{\mS^i_k}$ and $\lambda_{\mT_l^l}$ we denote the norm of the highest eigenvalue of the blocks $\mS^i_k$ and $\mT_l^l$ respectively. We can express the equivariance error norm $|R(g)|$ to be upper-bounded by:
{
\allowdisplaybreaks
\bea
& & \sum_i |t_i| \|\mW d\rho_{\gX}(A_i)- d\rho_{\gY}(A_i) \mW\|_2 \|\vx\| \\
&&= \sum_i |t_i| \|(\mU_\gY^i {\mW^i}' {\mU_\gX^i}^\top) (\mU_\gX^i \mSigma_\gX^i {\mU_\gX^i}^\top) - (\mU_\gY^i \mSigma_\gY^i {\mU_\gY^i}^\top) (\mU_\gY^i {\mW^i}' {\mU_\gX^i}^\top)\|_2 \|\vx\| \\
&&= \sum_i |t_i| \| {\mW^i}' \mSigma_\gX^i - \mSigma_\gY^i {\mW^i}'\|_2 \|\vx\| \quad (\text{spectral norm of orthonormal matrices is } 1)\\
&&\leq  \sum_i |t_i| \| {\mW^i}' \mSigma_\gX^i - \mSigma_\gY^i {\mW^i}'\|_{\tt F} \|\vx\| \\
&&=  \sum_i |t_i| \left(\sum_{l,k}\| {\mW^i}'_{lk} \mS_k^i - \mT_l^i {\mW^i}'_{lk}\|_{\tt F}^2\right)^{1/2} \|\vx\|\\
&&=  \sum_i |t_i| \left(\sum_{l,k: \lambda_{\mS_k^i}+\lambda_{\mT_l^i} \leq b} \| {\mW^i}'_{lk} \mS_k^i - \mT_l^i {\mW^i}'_{lk}\|_{\tt F}^2\right)^{1/2} \|\vx\| \quad {\scriptstyle (\text{as }\mB_{\tt Schur}, {\mW^i}'_{lk} \mS_k^i - \mT_l^i {\mW^i}'_{lk} = \vzero \text{ when }\lambda_{\mS_k^i}+\lambda_{\mT_l^i} > b)} \\
&&\leq  \sum_i |t_i| \left(\sum_{l,k: \lambda_{\mS_k^i}+\lambda_{\mT_l^i} \leq b} (\lambda_{\mS_k^i}+\lambda_{\mT_l^i})^2\|{\mW^i}'_{lk}\|_{\tt F}^2\right)^{1/2} \|\vx\| \quad {\scriptstyle (\text{ maximum scaling is bounded by the highest eigenvalues})}\\
&&\leq  \sum_i |t_i| \left(\sum_{l,k: \lambda_{\mS_k^i}+\lambda_{\mT_l^i} \leq b} b^2 \|{\mW^i}'_{lk}\|_F^2\right)^{1/2} \|\vx\|\\
&&\leq  \sum_i |t_i| b \left(\sum_{l,k} \|{\mW^i}'_{lk}\|_{\tt F}^2\right)^{1/2} \|\vx\|\\
&&=  \sum_i |t_i| b \|{\mW^i}'\|_{\tt F} \|\vx\|\\
&&=  \sum_i |t_i| b \|\mW\|_{\tt F} \|\vx\| \quad \text{(as $\mU_\gX^i$ and $\mU_\gY^i$ are orthonormal)}.
\eea
}

For compact and connected Lie groups, applying the Cauchy-Schwarz inequality and using the injective radius property:
\begin{align}
\sum_i |t_i| \leq \sqrt{n_G} \sqrt{\sum_i t_i^2} \leq \sqrt{n_G} r_G.
\end{align}

Substituting the definition of $R(g)$ and using $\|\rmJ_{F_{\tt FC}(;\mW)}(\vx)\|_F = \|\mW\|_F$, and reintroducing the Taylor residual error $\epsilon_G$, we can state

\begin{align}
\| \mB_s \Theta \rho_\gX(g) \vx - \rho_\gY(g) \mB_s \Theta \vx \| &\leq b r_G \sqrt{n_G} \|\mW\|_F \|\vx\| + \epsilon_G
\implies \frac{\| \mB_s \Theta \rho_\gX(g) \vx - \rho_\gY(g) \mB_s \Theta \vx \|}{\|\rmJ_{F_{\tt FC}(;\mW)}(\vx)\|_F\|\vx\|} &\leq b  \sqrt{n_G} r_G + \varepsilon_G.
\end{align}
Here, $\varepsilon_G$ is the normalized Taylor residual.
\end{proof}

\subsection{Proof of Lemma~\ref{lemma:forward_diff}}
\label{supp:forward_diff}
\begin{mylemmae}
\lemmatwo*
\end{mylemmae}
\begin{proof}
By definition of $\hat{f}$, the approximation error is:
\begin{equation}
    \left| f(g) - \hat{f}(g) \right| 
    = \left| f(g) - f(e) - \sum_{i=1}^k n_{s_i}\,\Delta_{s_i} f(e) \right|.
\end{equation}
Applying the triangle inequality:
\begin{equation}
    \left| f(g) - \hat{f}(g) \right| 
    \leq \underbrace{|f(g) - f(e)|}_{\text{(I)}} 
    + \underbrace{\sum_{i=1}^k n_i\,|\Delta_{s_i} f(e)|}_{\text{(II)}}.
\end{equation}
Recall that the word metric $d_{\sS}(g, g')$ counts the minimum number of generator from $\sS$ needed to reach $g'$ from $g$; in particular, for $g = s_1^{n_1} \cdots s_m^{n_m}$, the word metric can be expressed as 
$d_{\sS}(e, g) = \sum_{i=1}^m n_i = \sum_{i=1}^k n_{s_i} $, where $n_{s_i}$ is the total number of occurrence of $s_i$ in $s_1^{n_1} \cdots s_m^{n_m}$ the canonical  word representation of $g$  .

{\bf \noindent Bounding (I).} Since $d_{\sS}(e, g) = \sum_{i=1}^k n_i$ 
and $f$ is $h$-Lipschitz:
\begin{equation}
    |f(g) - f(e)| \leq h \cdot d_{\sS}(e, g).
\end{equation}

{\bf \noindent Bounding (II).} Each generator $s_i$ satisfies $d_{\sS}(e, s_i) = 1$, 
so by $h$-Lipschitz:
\begin{equation}
    |\Delta_{s_i} f(e)| = |f(s_i \cdot e) - f(e)| \leq h \cdot d_{\sS}(e, s_i) = h.
\end{equation}
Summing over all generators:
\begin{equation}
    \sum_{i=1}^k n_i\,|\Delta_{s_i} f(e)| \leq h \sum_{i=1}^k n_i = h \cdot d_{\sS}(e, g).
\end{equation}

Combining the two bounds:
\begin{equation}
    \left| f(g) - \hat{f}(g) \right| 
    \leq h \cdot d_{\sS}(e, g) + h \cdot d_{\sS}(e, g) 
    = 2h \cdot d_{\sS}(e, g),
\end{equation}
which completes the proof.

\end{proof}

%% file: supp/additional_results.tex
\newpage
\input{tables/human_traj_full.tex}
\section{Additional Experiments and Results}
\label{sec:additional_res}
\subsection{Synthetic $O(5)$ benchmark}
We follow the setup of \cite{finzi2021practical} to create a synthetic benchmark for $O(5)$ invariant learning.
 
\myparagraph{Task Description}
Given two vectors $x_1, x_2 \in \mathbb{R}^5$, the task is to predict the following $O(5)$ invariant quantity $f(x_1,x_2) = \sin(\|x_1\|) - \|x_2\|^3/2 + \frac{x_1^\top x_2}{\|x_1\| \|x_2\|}$. To induce a more challenging learning problem, we deviate from the strict invariant target function $f$ and reformulate the target function as 
\be
\tilde{f}(x_1,x_2) = \sin(\|x_1\|) - \|x_2\|^3/2 + \frac{x_1^\top x_2}{\|x_1\| \|x_2\|} + \underbrace{\textcolor{MidnightBlue}{\gamma \frac{|x_1^1| +|x_1^2|+|x_2^1|+|x_2^2| }{|x_1^3| +|x_1^4|+|x_2^3|+|x_2^4| + \varepsilon}}}_{\text{non-invariant term}},
\ee 
where $\gamma$ is a hyperparameter that controls the degree of deviation from strict invariance and $x_j^k$ denotes the $k$th component of the vector $x_j$. When $\gamma=0$, the target function is strictly invariant, and as $\gamma$ increases, the target function becomes less invariant.

We present the results in \tabref{tab:o5}, which show that our method outperforms the baselines, with EMLP-RPP \cite{finzi2021residual} being the second-best.

\input{tables/lie_o5.tex}

\subsection{Additional ablation studies and extended results}
We evaluate the benefits of soft equivariant models we also report the performance of full equivariant models. However, fully equivariant pretrained models with architectures and parameter counts similar to ViT or DINOv2 are not publicly available. So we use our projection operator to create a full equivariant model from a pretrained non-equivariant model. We present the results in \tabref{tab:full_eq} on the classification and segmentation tasks following the setup of \cref{sec:exp_classification} and \cref{sec:exp_segmentation}. We observe that the full equivariance constraint degrades the performance (see \tabref{tab:cifar_results} and \tabref{tab:voc_seg_results}).

\begin{minipage}{.58\textwidth}
     \input{tables/full_eq.tex}

\end{minipage}
\begin{minipage}{.40\textwidth}
        \vspace{-0.5cm}
     \input{tables/wang_vs_soft.tex}

\end{minipage}

We also compare our method with the group convolution-based soft equivariant model proposed by \citet{wang2022approximately} on the MNIST digits classification task. We follow the setup of \cref{sec:exp_tunability} and report the results in \tabref{tab:wang_vs_soft}. We observe that across different levels of softness, which is controlled by the number of filter banks in \citet{wang2022approximately}, our method maintains stable performance, outperforming the baseline.
\begin{table}[t]
\begin{minipage}[t]{.49\textwidth}
\input{tables/seg_multi_seed.tex}
\end{minipage}
\hfill
\begin{minipage}[t]{.49\textwidth}
\input{tables/cifar_multi_seed}
\end{minipage}
\end{table}

We provide the full results on the human trajectory prediction task in \tabref{tab:human_traj_full}. We observe that, across most datasets, our method outperforms both strict-equivariant and non-equivariant baselines on both standard metrics (ADE, FDE) and augmented metrics (aADE, aFDE). While \ref{tab:cifar_results} and \ref{tab:voc_seg_results} reports results for a single run, \tabref{tab:cifar_results_ms} and \tabref{tab:voc_seg_results_ms} report mean\,$\pm$\, std over 5 random seeds following the exact setup of \cref{sec:exp_classification} and \cref{sec:exp_segmentation}. We observe that our method consistently outperforms the baselines.

%% file: tables/human_traj_full.tex
\begin{table*}[t]
\centering
\setlength{\tabcolsep}{2pt}
\caption{\textbf{Human Trajectory Prediction Results (ETH/UCY).} All metrics are lower is better.}
\label{tab:human_traj_full}
\vspace{-0.1cm}
\resizebox{\textwidth}{!}{
\begin{tabular}{@{}l>{\hskip 3pt} cccccc c cccccc c cccccc c cccccc c cccccc @{}}
\specialrule{.15em}{.05em}{.05em}
\textbf{Method}
& \multicolumn{6}{c}{\textbf{ETH}} & &
  \multicolumn{6}{c}{\textbf{UNIV}} & &
  \multicolumn{6}{c}{\textbf{ZARA1}} & &
  \multicolumn{6}{c}{\textbf{ZARA2}} & &
  \multicolumn{6}{c}{\textbf{HOTEL}} \\
& {\small ADE}{\small$\downarrow$} & {\small aADE}{\small$\downarrow$} & {\small cADE}{\small$\downarrow$} & {\small FDE}{\small$\downarrow$} & {\small aFDE}{\small$\downarrow$} & {\small cFDE}{\small$\downarrow$} & &
  {\small ADE}{\small$\downarrow$} & {\small aADE}{\small$\downarrow$} & {\small cADE}{\small$\downarrow$} & {\small FDE}{\small$\downarrow$} & {\small aFDE}{\small$\downarrow$} & {\small cFDE}{\small$\downarrow$} & &
  {\small ADE}{\small$\downarrow$} & {\small aADE}{\small$\downarrow$} & {\small cADE}{\small$\downarrow$} & {\small FDE}{\small$\downarrow$} & {\small aFDE}{\small$\downarrow$} & {\small cFDE}{\small$\downarrow$} & &
  {\small ADE}{\small$\downarrow$} & {\small aADE}{\small$\downarrow$} & {\small cADE}{\small$\downarrow$} & {\small FDE}{\small$\downarrow$} & {\small aFDE}{\small$\downarrow$} & {\small cFDE}{\small$\downarrow$} & &
  {\small ADE}{\small$\downarrow$} & {\small aADE}{\small$\downarrow$} & {\small cADE}{\small$\downarrow$} & {\small FDE}{\small$\downarrow$} & {\small aFDE}{\small$\downarrow$} & {\small cFDE}{\small$\downarrow$} \\
\hline
Base~\cite{giuliari2021transformer}
  & 4.67 & 4.80 & 4.73 & \bf 6.04 & \bf 6.27 & \bf 6.15 & &
    7.90 & 7.92 & 7.91 & 8.16 & 8.17 & 8.16 & &
    3.57 & 3.65 & 3.61 & \bf 4.63 & 4.73 & 4.68 & &
    3.18 & 3.15 & 3.17 & 3.83 & 3.76 & 3.79 & &
    5.83 & 5.85 & 5.84 & 6.64 & 6.71 & 6.67 \\
EqAuto~\cite{deng2021vector}
  & 5.40 & 5.40 & 5.40 & 7.33 & 7.33 & 7.33 & &
    8.16 & 8.16 & 8.16 & 8.33 & 8.33 & 8.33 & &
    3.66 & 3.66 & 3.66 & 4.98 & 4.98 & 4.98 & &
    2.94 & 2.94 & 2.94 & 3.63 & 3.63 & 3.63 & &
    6.16 & 6.16 & 6.16 & 6.78 & 6.78 & 6.78 \\
\rowcolor{myblue!15!White}
Ours
  & \bf 4.50 & \bf 4.66 & \bf 4.58 & 6.10 & 6.35 & 6.23 & &
    \bf 7.82 & \bf 7.88 & \bf 7.85 & \bf 8.04 & \bf 8.09 & \bf 8.07 & &
    \bf 3.39 & \bf 3.41 & \bf 3.40 & 4.65 & \bf 4.68 & \bf 4.67 & &
    \bf 2.91 & \bf 2.91 & \bf 2.91 & \bf 3.60 & \bf 3.60 & \bf 3.60 & &
    \bf 5.69 & \bf 5.69 & \bf 5.69 & \bf 6.26 & \bf 6.25 & \bf 6.26 \\
\specialrule{.15em}{.05em}{.05em}
\end{tabular}
}
\vspace{-0.15cm}
\end{table*}

%% file: tables/lie_o5.tex
\begin{table}[h]
\centering
\small
\caption{\bf Relative MSE ($10^{-1}$) for varying $\gamma$ on Synthetic $O(5)$ benchmark.}
\label{tab:o5}
\begin{tabular}{lccc}
\toprule
\textbf{Model} & \textbf{$\gamma = 0.3$} & \textbf{$\gamma = 0.4$} & \textbf{$\gamma = 0.5$} \\
\midrule
EMLP \cite{finzi2021practical}        & \underline{0.78} & 0.88 & 1.00 \\
EMLP-RPP \cite{finzi2021residual}    & 0.79 & \underline{0.81} & \underline{0.84} \\
MLP         & 1.21 & 1.20 & 1.29 \\
\rowcolor{myblue!15!White} Ours        & \bf{0.72} & \bf{0.74} & \bf{0.81} \\
\bottomrule
\end{tabular}
\end{table}

%% file: tables/full_eq.tex
\captionof{table}{Performance of full equivariant models}
\label{tab:full_eq}
\resizebox{\linewidth}{!}{
\small
\begin{tabular}{@{}llccc@{}}
\toprule
\textbf{Task} & \textbf{Dataset} & \textbf{ViT} & \textbf{DINOv2} & \textbf{ResN/Seg} \\
\midrule
\multirow{2}{*}{Class. (cAcc)} & CIFAR-10 & 84.10 & 93.07 & 73.21$^{\dagger}$ \\
& CIFAR-100 & 66.98 & 81.99 & 58.85$^{\dagger}$ \\
Segmen. (cIoU ) & Pascal VOC & 63.45 & 68.87 & 29.20$^{\ddagger}$ \\
\bottomrule
\multicolumn{5}{@{}l@{}}{\footnotesize $^{\dagger}$ResNet-50, $^{\ddagger}$SegFormer} \\
\end{tabular}
}

%% file: tables/wang_vs_soft.tex
\captionof{table}{\citet{wang2022approximately} vs Ours.}
\label{tab:wang_vs_soft}
\resizebox{\linewidth}{!}{
\small
\begin{tabular}{@{}cc cc@{}}
\toprule
\multicolumn{2}{c}{\textbf{\citet{wang2022approximately}}} & \multicolumn{2}{c}{\textbf{Ours}} \\
\textbf{\# Filter Bank} & \textbf{cAcc (\%)} & \textbf{Softness} & \textbf{cAcc (\%)} \\
\midrule
2 & 83.08 & \cellcolor{myblue!15!White}0.7 & \cellcolor{myblue!15!White}85.52 \\
3 & 81.25 & \cellcolor{myblue!15!White}0.8 & \cellcolor{myblue!15!White} 85.22 \\
4 & 82.74 & \cellcolor{myblue!15!White}0.9 & \cellcolor{myblue!15!White} \textbf{85.11} \\
\bottomrule
\end{tabular}
}

%% file: tables/seg_multi_seed.tex
\centering
\setlength{\tabcolsep}{4pt}
\captionof{table}{\textbf{Segmentation Performance on PASCAL VOC~\cite{everingham2010pascal}.}Results reported as mean\,$\pm$\,std over multiple runs.
eErr is in $\times 10^{-2}$.
}
\vspace{-0.1cm}
\label{tab:voc_seg_results_ms}
\resizebox{\linewidth}{!}{
\small
\begin{tabular}{@{}l >{\hskip 4pt} cc S cc S cc @{}}
\specialrule{.15em}{.05em}{.05em}
\textbf{Arch.} &
\multicolumn{2}{c}{\textbf{ViT}~\cite{dosovitskiy2020image}} & &
\multicolumn{2}{c}{\textbf{DINOv2}~\cite{oquab2023dinov2}} & &
\multicolumn{2}{c}{\textbf{Segformer}~\cite{xie2021segformer}} \\
& cIoU{\small$\uparrow$} & eErr{\small$\downarrow$} & &
  cIoU{\small$\uparrow$} & eErr{\small$\downarrow$} & &
  cIoU{\small$\uparrow$} & eErr{\small$\downarrow$} \\
\hline
Base
& $72.99{\scriptstyle\,\pm\,0.95}$ & $11{\scriptstyle\,\pm\,01}$ & &
  $87.97{\scriptstyle\,\pm\,1.02}$ & $04{\scriptstyle\,\pm\,00}$ & &
  $63.67{\scriptstyle\,\pm\,0.98}$ & $11{\scriptstyle\,\pm\,00}$ \\
Canon.
& $64.37{\scriptstyle\,\pm\,0.75}$ & $19{\scriptstyle\,\pm\,00}$ & &
  $82.83{\scriptstyle\,\pm\,0.18}$ & $38{\scriptstyle\,\pm\,07}$ & &
  $55.90{\scriptstyle\,\pm\,1.26}$ & $24{\scriptstyle\,\pm\,02}$ \\
\rowcolor{myblue!15!White}
Ours
& $\mathbf{73.25}{\scriptstyle\,\pm\,0.38}$ & $\mathbf{10}{\scriptstyle\,\pm\,00}$ & &
  $\mathbf{88.61}{\scriptstyle\,\pm\,0.18}$ & $\mathbf{03}{\scriptstyle\,\pm\,00}$ & &
  $\mathbf{63.87}{\scriptstyle\,\pm\,0.67}$ & $\mathbf{10}{\scriptstyle\,\pm\,00}$ \\
\specialrule{.15em}{.05em}{.05em}
\end{tabular}
\vspace{-0.1cm}
}

%% file: tables/cifar_multi_seed.tex
\setlength{\tabcolsep}{2.5pt}
\captionof{table}{\textbf{Performance on CIFAR10/100 across various backbones.} Results reported as mean\,$\pm$\,std over multiple runs. iErr is in $\times 10^{-2}$.
}
\vspace{-0.15cm}
\label{tab:cifar_results_ms}
\resizebox{\linewidth}{!}{
\small
\begin{tabular}{@{}ll >{\hskip 4pt} cc S cc S cc @{}}
\specialrule{.15em}{.05em}{.05em}
& \textbf{Arch.}
& \multicolumn{2}{c}{\textbf{ViT}~\cite{dosovitskiy2020image}} & &
  \multicolumn{2}{c}{\textbf{DINOv2}~\cite{oquab2023dinov2}} & &
  \multicolumn{2}{c}{\textbf{ResNet-50}~\cite{he2016deep}} \\
& &
cAcc{\small$\uparrow$} & iErr{\small$\downarrow$} & &
cAcc{\small$\uparrow$} & iErr{\small$\downarrow$} & &
cAcc{\small$\uparrow$} & iErr{\small$\downarrow$} \\
\hline
\multirow{3}{*}{\rotatebox{90}{\scriptsize{CIFAR10\;}}}
& Base
& $97.47{\scriptstyle\,\pm\,0.05}$ & $06{\scriptstyle\,\pm\,00}$ & &
  $98.63{\scriptstyle\,\pm\,0.06}$ & $04{\scriptstyle\,\pm\,00}$ & &
  $94.88{\scriptstyle\,\pm\,0.17}$ & $21{\scriptstyle\,\pm\,01}$ \\
& Canon.
& $93.42{\scriptstyle\,\pm\,0.22}$ & $13{\scriptstyle\,\pm\,02}$ & &
  $97.78{\scriptstyle\,\pm\,0.22}$ & $\mathbf{03}{\scriptstyle\,\pm\,00}$ & &
  $91.00{\scriptstyle\,\pm\,0.26}$ & $11{\scriptstyle\,\pm\,01}$ \\
& \cellcolor{myblue!15!White} Ours
& \cellcolor{myblue!15!White} $\mathbf{97.49}{\scriptstyle\,\pm\,0.14}$
& \cellcolor{myblue!15!White} $\mathbf{06}{\scriptstyle\,\pm\,00}$
& \cellcolor{myblue!15!White}
& \cellcolor{myblue!15!White} $\mathbf{98.70}{\scriptstyle\,\pm\,0.05}$
& \cellcolor{myblue!15!White} $04{\scriptstyle\,\pm\,00}$
& \cellcolor{myblue!15!White}
& \cellcolor{myblue!15!White} $\mathbf{95.05}{\scriptstyle\,\pm\,0.27}$
& \cellcolor{myblue!15!White} $\mathbf{19}{\scriptstyle\,\pm\,03}$ \\
\hline
\multirow{3}{*}{\rotatebox{90}{\scriptsize{CIFAR100\;}}}
& Base
& $85.63{\scriptstyle\,\pm\,0.22}$ & $24{\scriptstyle\,\pm\,00}$ & &
  $91.62{\scriptstyle\,\pm\,0.20}$ & $25{\scriptstyle\,\pm\,01}$ & &
  $79.78{\scriptstyle\,\pm\,0.18}$ & $51{\scriptstyle\,\pm\,02}$ \\
& Canon.
& $78.41{\scriptstyle\,\pm\,0.44}$ & $25{\scriptstyle\,\pm\,02}$ & &
  $88.64{\scriptstyle\,\pm\,0.65}$ & $\mathbf{16}{\scriptstyle\,\pm\,01}$ & &
  $74.18{\scriptstyle\,\pm\,1.02}$ & $\mathbf{27}{\scriptstyle\,\pm\,03}$ \\
& \cellcolor{myblue!15!White} Ours
& \cellcolor{myblue!15!White} $\mathbf{85.67}{\scriptstyle\,\pm\,0.28}$
& \cellcolor{myblue!15!White} $\mathbf{23}{\scriptstyle\,\pm\,01}$
& \cellcolor{myblue!15!White}
& \cellcolor{myblue!15!White} $\mathbf{91.90}{\scriptstyle\,\pm\,0.12}$
& \cellcolor{myblue!15!White} $23{\scriptstyle\,\pm\,00}$
& \cellcolor{myblue!15!White}
& \cellcolor{myblue!15!White} $\mathbf{80.13}{\scriptstyle\,\pm\,0.24}$
& \cellcolor{myblue!15!White} $45{\scriptstyle\,\pm\,00}$ \\
\specialrule{.15em}{.05em}{.05em}
\end{tabular}
\vspace{-0.1cm}
}

%% file: supp/training_detail.tex
\section{Experiment Details}
\label{sup:exp_details}

\subsection{Validating tunable softness level}
We use a two-layer MLP trained with a batch size of $512$ for $30$ epochs. The MLPs take flattened images as input. The models are trained on the standard MNIST~\cite{lecun2002gradient} training set with $10\%$ held out for validation. We choose the best-performing model on the validation set for final evaluation on the test set. We use Adam optimizer with a learning rate of $1e^{-3}$ and a weight decay of $1e^{-4}$. The input images are padded to prevent the boundary artifact during rotation. We use the Adam optimizer with linear learning rate decay. 

\subsection{Image Classification} 
\myparagraph{CIFAR-10 and CIFAR-100.} We use the pretrained ViT-B/16 model from \cite{dosovitskiy2020image} as the backbone which is pretrained in ImageNet-1k dataset for classification task. We replace the final classification layer to match the number of classes in the CIFAR-10 and CIFAR-100 datasets. The ResNet-50 \cite{he2016deep} backbone is also pretrained on ImageNet-1k, and only the final classification is replaced. The DINOv2 \cite{oquab2023dinov2} is trained on the large-scale LVD-142M dataset. We add a single-layer classification head with the DINOv2 backbone. 

For the Canon. \cite{mondal2023equivariant} baseline we use a $3$ layer $C_{18}$  equivariant steerable (a $20^\circ$ rotation equivariant) CNN architecture \cite{cohen2016steerable} as the canonicalization network. The model is trained from scratch with random initialization. We also use the prior loss as described in \cite{mondal2023equivariant} to stabilize the canonicalization network. We use the same rotation group for designing our projection operators. See \secref{sec:softness_selection} for details on the choice of the cutoff value for the projection operator.

All the models are finetuned for $30$ epochs. Images are interpolated to $224 \times 224$ resolution to match the input size of the pretrained models. We use a larger learning rate of $1e^{-3}$ for the classification head and  $1e^{-5}$ and $5e^{-5}$ for the ViT/DINOv2 and ResNet backbone, respectively. Models are trained with random rotation augmentation. We use the AdamW optimizer and a cosine learning rate scheduler. $20\%$ of the training data is held out for validation. The best model on the validation set is used for final evaluation on the test set.

\myparagraph{ImageNet-1k.} We use publicly available pretrained ViT-B/16 \cite{dosovitskiy2020image}, DINOv2-base \cite{oquab2023dinov2,facebook_dinov2}, and ResNet-50 \cite{he2016deep} models. We directly evaluate the models on the ImageNet-1k validation set. We use the same architecture for the canonicalization module for the Canon. \cite{mondal2023equivariant} baseline as described above. We use the training script from Pytorch Image Models (timm) library \cite{wightman2019pytorch} for finetuning and evaluation. We train all the models on a single GPU with a batch size of $128$ for $15$ epochs with $4$ warmup epochs. We use a cosine learning rate scheduler with a learning rate of $1e^{-4}$. We use standard data augmentation techniques from ImageNet training \cite{he2022masked,dosovitskiy2020image}, with mixup, cutmix, random erasing, and RandAugment. The hyperparameters are summarized in~\tabref{tab:training_hyperparams}. 
\input{tables/imnet_hyper.tex}

\subsection{Semantic segmentation}
We use the PASCAL VOC 2012 \cite{everingham2010pascal} dataset for the semantic segmentation task. Following the DINOv2 \cite{oquab2023dinov2}, we use a linear classifier over the pre-trained DINOv2-base and ViT-B/16 backbones. For SegFormer \cite{xie2021segformer}, we change the final classification layer of the decoder head to match the number of classes in the PASCAL VOC dataset. We use the same canonicalization network architecture as described above for the Canon. baseline. The SegFormer and DINOv2 models are trained with images of size $512 \times 512$ following their original setup. For ViT-B/16, we use an image size of $224 \times 224$. All the models are trained for $100$ epochs with a batch size of $32$. We use a cosine annealing learning rate scheduler with $8$ warmup epochs. The learning rate is set to $5e^{-5}$ for ViT-B/16, $1e^{-5}$ for DINOv2, and $1e^{-4}$ for SegFormer baselines. We use standard data augmentation techniques, including random scaling, cropping, horizontal flipping, and rotation. 

\input{figs/trajectory_sample}

\subsection{Human trajectory prediction}
We use the human trajectory datasets from ETH~\cite{pellegrini2009you} and UCY~\cite{lerner2007crowds}, consisting of five subsets with the number of subjects (humans) varying from $5-57$ across different scenes and datasets. We visualize two samples from the dataset in \figref{fig:traj_sample}, illustrating the task and the dataset's complexity. We use an auto-regressive transformer model with $4$ transformer layers, each with $4$ attention heads. We follow the technique of Vector neurons \cite{deng2021vector} for designing an equivariant auto-regressive (EqAuto) transformer. Models are trained for $100$ epochs with a learning rate of $5e^{-4}$ with a weight decay of $1e^{-5}$. The models are trained with mean squared error. The best-performing model on the validation set is used for the final evaluation on the test set. We use the Lie algebra representation of the $2D$ rotation group to design the projection operators. The combined ADE (cADE) and combined FDE (cFDE) are calculated as 
\be
\text{cADE} = \frac{\text{ADE} + \text{aADE}}{2}, \quad \text{cFDE} = \frac{\text{FDE} + \text{aFDE}}{2},
\ee

\subsection{Softness level selection}
\label{sec:softness_selection}
In \cref{sec:exp_tunability}, we show that when training from scratch, the level of `softness' can be selected depending on the trade-off between performance and consistency. 

\begin{figure}
  \centering
  \begin{subfigure}{.48\textwidth}
    \centering
    \caption{cIoU vs softness level for SegFormer}
    \label{fig:segformer_0_1}
    \includegraphics[width=.98\textwidth]{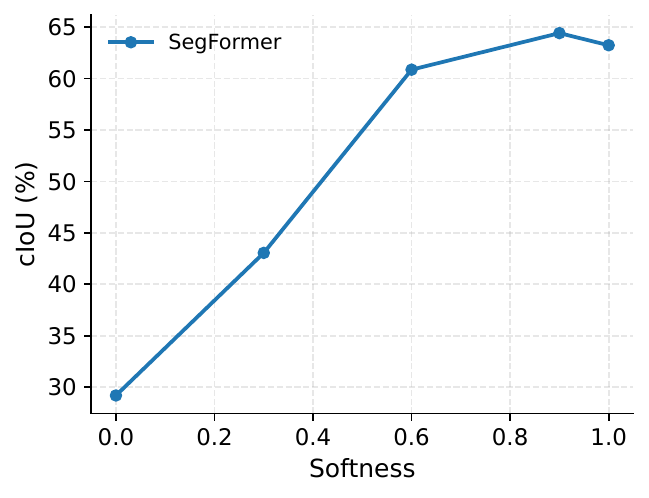}
  \end{subfigure}
  \begin{subfigure}{.48\textwidth}
    \centering
    \caption{cIoU vs high softness level for different pretrained models}
    \label{fig:softness_models}
    \includegraphics[width=.98\textwidth]{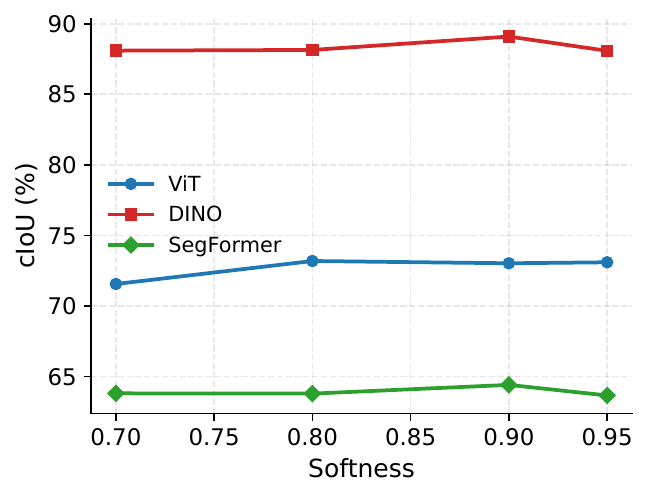}
  \end{subfigure}
\end{figure}

However, when using pretrained models, smaller softness levels (i.e., more aggressive projection) can lead to a significant drop in performance because they deviate further from the pretrained weights. As we see in \figref{fig:segformer_0_1}, the performance of SegFormer drops significantly when the softness level is smaller than $0.6$, and we do not see any benefit from pretraining. So, when adapting pretrained models, we use higher softness levels ($>0.7$) to retain the benefits of pretraining. As we see in \figref{fig:softness_models}, across all architectures, this range of softness levels leads to effective fine-tuning and better performance.

%% file: tables/imnet_hyper.tex
\begin{table}[h]
\centering
\caption{Training Hyperparameters for ImageNet-1k}
\label{tab:training_hyperparams}
\begin{tabular}{lc}
\toprule
\textbf{Hyperparameter} & \textbf{Value} \\
\midrule
\multicolumn{2}{l}{\textit{Optimization}} \\
Optimizer & AdamW \\
Learning Rate & $1e-4$ \\
Weight Decay & 0.05-0.04 \\
Epochs & 15 \\
Gradient Clipping & 1.0 \\
\midrule
\multicolumn{2}{l}{\textit{Learning Rate Schedule}} \\
Scheduler & Cosine Annealing \\
Warmup Epochs & 4 \\
Warmup LR & $1e-6$ \\
Layer-wise LR Decay & 0.65 \\
\midrule
\multicolumn{2}{l}{\textit{Augmentation}} \\
Mixup $\alpha$ & 0.8 \\
CutMix $\alpha$ & 1.0 \\
Random Erasing & 0.25 \\
RandAugment & rand-m9-mstd0.5-inc1 \\
\midrule
\multicolumn{2}{l}{\textit{Regularization}} \\
Label Smoothing & 0.1 \\
Drop Path Rate & 0.1 \\
Model EMA & 0.99 \\
\bottomrule
\end{tabular}
\end{table}

%% file: figs/trajectory_sample.tex
\begin{figure*}[t]
    \centering
    \setlength{\tabcolsep}{1pt}
    \begin{tabular}{cc}
    \includegraphics[width=0.49\columnwidth, trim=0.0cm 0.0cm 0.0cm 0.0cm, clip]{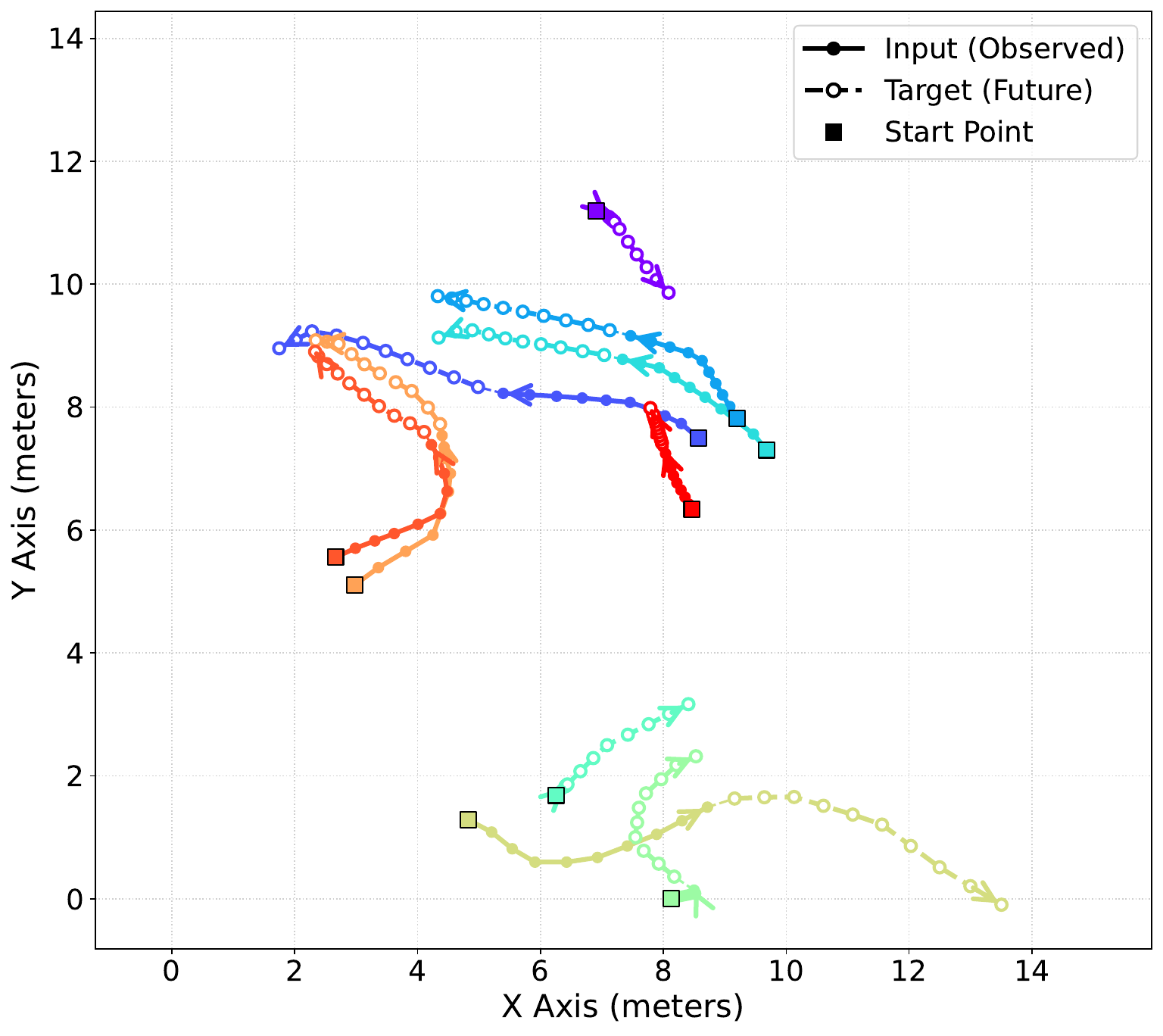} &
    \includegraphics[width=0.49\columnwidth, trim=0.0cm 0.0cm 0.0cm 0.0cm, clip]{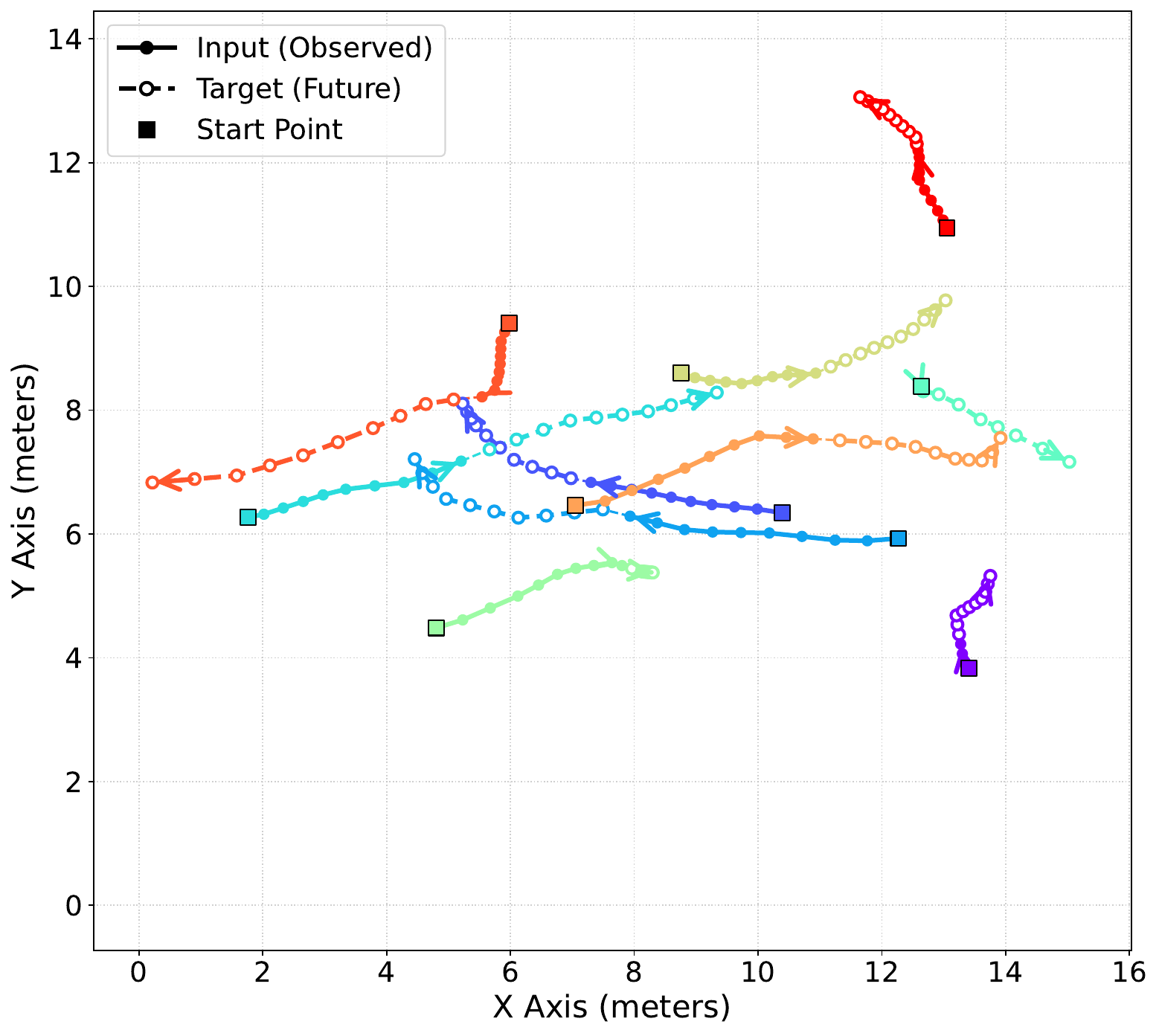} 
    \end{tabular}
    \vspace{-0.25cm}
    \caption{Sample trajectories from the ETH/UYC dataset. Solid circles indicate observed past positions, and hollow circles show the future. Square markers denote starting points. Each line with a distinct color corresponds to a different pedestrian in the scene.
    }
    \label{fig:traj_sample}
    \vspace{-0.25cm}
\end{figure*}